\documentclass[10pt,twocolumn,letterpaper]{article}

\usepackage{cvpr}
\usepackage{times}
\usepackage{epsfig}
\usepackage{graphicx}
\usepackage{amsmath}
\usepackage{amssymb}

\usepackage{color}
\usepackage{array}
\usepackage{caption}
\usepackage{subcaption}

\usepackage[pagebackref=true,breaklinks=true,letterpaper=true,colorlinks,bookmarks=false]{hyperref}

\definecolor{grey50}{rgb}{0.5,0.5,0.5}

\newcolumntype{C}[1]{>{\centering\let\newline\\\arraybackslash\hspace{0pt}}m{#1}}
\newcolumntype{L}[1]{>{\raggedright\let\newline\\\arraybackslash\hspace{0pt}}m{#1}}

\newcommand\Tstrut{\rule{0pt}{2.2ex}}       % "top" strut
\newcommand\Bstrut{\rule[-1.0ex]{0pt}{0pt}} % "bottom" strut
\newcommand{\TBstrut}{\Tstrut\Bstrut}       % top&bottom struts

\cvprfinalcopy % *** Uncomment this line for the final submission

\def\cvprPaperID{****} % *** Enter the CVPR Paper ID here

\begin{document}

%%%%%%%%% TITLE
% \title{\LaTeX\ Author Guidelines for CVPR Proceedings}
\title{Forecasting Human Dynamics from Static Images}

% \and
\author{Yu-Wei Chao$^1$, Jimei Yang$^2$, Brian Price$^2$, Scott Cohen$^2$, and Jia Deng$^1$ \vspace{3mm} \\
  \begin{minipage}{0.5\textwidth}
    \centering
    $^1$University of Michigan, Ann Arbor
    {\tt\small \{ywchao,jiadeng\}@umich.edu}\\
  \end{minipage}
  \begin{minipage}{0.5\textwidth}
    \centering
    $^2$Adobe Research
    {\tt\small \{jimyang,bprice,scohen\}@adobe.com}\\
  \end{minipage}
}

\maketitle

\begin{abstract}
This paper presents the first study on forecasting human dynamics from static
images. The problem is to input a single RGB image and generate a sequence of
upcoming human body poses in 3D. To address the problem, we propose the 3D Pose
Forecasting Network (3D-PFNet). Our 3D-PFNet integrates recent advances on
single-image human pose estimation and sequence prediction, and converts the 2D
predictions into 3D space. We train our 3D-PFNet using a three-step training
strategy to leverage a diverse source of training data, including image and
video based human pose datasets and 3D motion capture (MoCap) data. We
demonstrate competitive performance of our 3D-PFNet on 2D pose forecasting and
3D pose recovery through quantitative and qualitative results.
\end{abstract}

\section{Introduction}

\begin{figure}[t]
  % \vspace{-2mm}
  \centering
  \hspace{-36mm}
  \begin{subfigure}[c]{0.20\linewidth}
    \centering
    \includegraphics[height=1.8\textwidth]{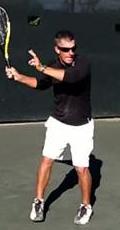}
  \end{subfigure}
  \begin{subfigure}[c]{0.07\linewidth}
    \centering
    \includegraphics[height=1\textwidth]{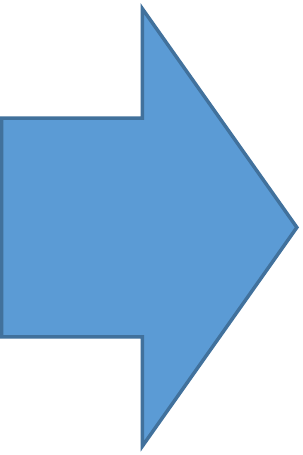}
  \end{subfigure}
  \begin{subfigure}[c]{0.20\linewidth}
    \centering
    \includegraphics[height=1.8\textwidth]{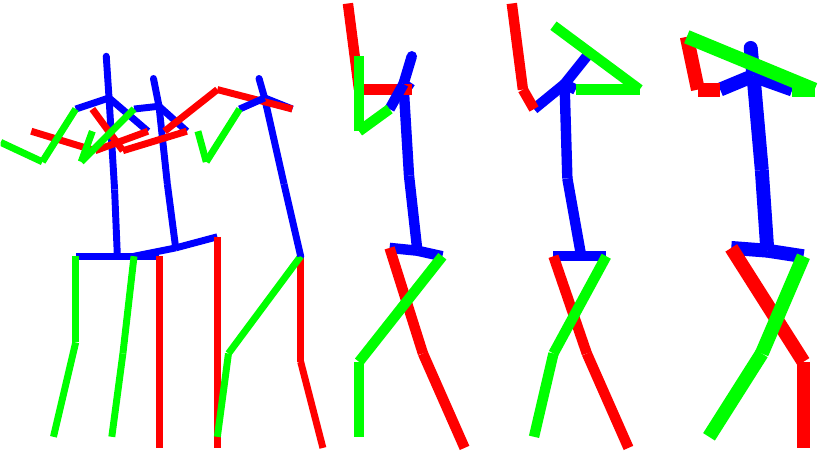}
  \end{subfigure}
  % \vspace{-2mm}
  \caption{\small Forecasting human dynamics from static images. Left: the
input image. Right: the sequence of upcoming poses.}
  \vspace{-2mm}
  \label{fig:intro}
\end{figure}

Human pose forecasting is the capability of predicting future human body
dynamics from visual observations. Human beings are endowed with this great
ability. For example, by looking at the left image of Fig.~\ref{fig:intro}, we
can effortlessly imagine the upcoming body dynamics of the target tennis
player, namely a forehand swing, as shown in the right image of
Fig.~\ref{fig:intro} Such prediction is made by reasoning on the scene context
(i.e. a tennis court), the current body pose of the target (i.e. standing and
holding a tennis racket), and our visual experience of a tennis forehand swing.

The ability of forecasting reflects a higher-level intelligence beyond
perception and recognition and plays an important role for agents to survive
from challenging natural and social environments. In the context of
human-robot interactions, such ability is particularly crucial for assistant
robots that need to interact with surrounding humans in an efficient and robust
manner. Apparently, the abilities of identifying and localizing the action
categories
\cite{simonyan:nips2014,donahue:cvpr2015,yeung:cvpr2016,shou:cvpr2016} after
observing an image or video are not sufficient to achieve this goal. For
example, when a person throws a ball at a robot, the robot needs to identify
the action and forecast the body pose trajectory even before the person
finishes so that it can response effectively (either by catching the ball or
dodging it).

This paper presents the first study on human pose forecasting from static
images. Our task is to take a single RGB image and output a sequence of future
human body poses. Our approach has two key features. First, as opposed to other
forecasting tasks that assume a multi-frame input (i.e. videos)
\cite{srivastava:icml2015,fragkiadaki:iccv2015,mathieu:iclr2016}, our work
assumes a single-frame input. Although this assumption increases the learning
challenge due to the lack of explicit motion cues, it encourages the algorithm
to learn high-level dynamics instead of low-level smoothness. Note that our
approach can be trivially extended to take multi-frame inputs as shown later in
the methodology section. Second, like most forecasting problems
\cite{yuen:eccv2010,walker:cvpr2014,pintea:eccv2014,walker:iccv2015,walker:eccv2016},
we first represent the forecasted poses in the 2D image space. However, we
include an extra component to our approach to further convert each forecasted
pose from 2D space to 3D space. Both forecasting and 3D conversion are
performed using a deep neural network (DNN). The two networks are integrated
into one single unified framework to afford end-to-end training. Since human
bodies feature a complex articulated structure, we believe the 3D output is
more actionable and useful for future applications (e.g. shape and texture
rendering) as we demonstrate in the supplementary materials.

The main contributions of this paper are three-fold: (1) We present the first
study on single-frame human pose forecasting. This extends the dimension of
current studies on human pose modeling from recognition (i.e. pose estimation
\cite{tompson:nips2014,newell:eccv2016}) to forecasting. The problem of pose
forecasting in fact generalizes pose estimation, since to forecast future poses
we need to first estimate the observed pose. (2) We propose a novel DNN-based
approach to address the problem. Our forecasting network integrates recent
advances on single-image human pose estimation and sequence prediction.
Experimental results show that our approach outperforms strong baselines on 2D
pose forecasting. (3) We propose an extra network to convert the forecasted 2D
poses into 3D skeletons. Our 3D recovery network is trained on a vast amount of
synthetic examples by leveraging motion capture (MoCap) data. Experimental
results show that our approach outperforms two state-of-the-art methods on 3D
pose recovery. In a nutshell, we propose a unified framework for 2D pose
forecasting and 3D pose recovery. Our \textit{3D Pose Forecasting Network}
(3D-PFNet) is trained by leveraging a diverse source of training data,
including image and video based human pose datasets and MoCap data. We
separately evaluate our 3D-PFNet on 2D pose forecasting and 3D pose recovery,
and show competitive results over baselines.

\section{Related Work}

\paragraph{Visual Scene Forecasting} Our work is in line with a series of
recent work on single-image visual scene forecasting. These works vary in the
predicted target and the output representation. \cite{lan:eccv2014} predicts
human actions in the form of semantic labels. Some others predict motions of
low level image features, such as the optical flow to the next frame
\cite{pintea:eccv2014,walker:iccv2015} or dense trajectories of pixels
\cite{yuen:eccv2010,walker:eccv2016}. A few others attempt to predict the
motion trajectories of middle-level image patches \cite{walker:cvpr2014} or
rigid objects \cite{mottaghi:cvpr2016}. However, these methods do not
explicitly output a human body model, thus cannot directly address human pose
forecasting. Notably, \cite{fragkiadaki:iccv2015} predicts the future dynamics
of a 3D human skeleton from its past motion. Despite its significance, their
method can be applied to only 3D skeleton data but not visual inputs. Our work
is the first attempt to predict 3D human dynamics from RGB images.

\vspace{-3mm}

\paragraph{Human Pose Estimation} Our work is closely related to the problem of
human pose estimation, which has long been attractive in computer vision. Human
bodies are commonly represented by tree-structured skeleton models, where each
node is a body joint and the edges capture articulation. The goal is to
estimate the 2D joint locations in the observed image
\cite{tompson:nips2014,newell:eccv2016} or video sequences
\cite{nie:cvpr2015,gkioxari:eccv2016}. Recent work has even taken one step
further to directly recover 3D joint locations
\cite{li:accv2014,yasin:cvpr2016,tekin:bmvc2016,du:eccv2016,chen:3dv2016,rogez:nips2016}
or body shapes \cite{bogo:eccv2016} from image observations. While promising,
these approaches can only estimate the pose of humans in the observed image or
video. Our approach not only estimate the human pose in the observed image, but
also forecasts the poses in the upcoming frames. Besides estimation from images
or videos, an orthogonal line of research addresses the recovery of 3D body
joint locations from their 2D projections
\cite{akhter:cvpr2015,zhou:cvpr2015,zhou:cvpr2016,wu:eccv2016}. Our work also
takes advantage of these approaches to transform the estimated 2D joint
locations into 3D space.

\vspace{-3mm}

\paragraph{Video Frame Synthesis} Two very recent works
\cite{xue:nips2016,vondrick:nips2016} attempt to synthesize videos from static
images by predicting pixels in future frames. This is a highly challenging
problem due to the extremely high dimensional output space and the massive
variations a scene can transform from a single image. Our work can provide
critical assistance to this task by using the predicted human poses as
intermediate representation to regularize frame synthesis, e.g. it is easier to
synthesize a baseball pitching video from a single photo of a player if we can
forecast his body dynamics. In addition to static images, there are also other
efforts addressing video prediction from video inputs
\cite{srivastava:icml2015,mathieu:iclr2016,finn:nips2016}, which can be
benefited by our work in the same way.

\section{Approach}

\subsection{Problem Statement}

\begin{figure}[t]
  % \vspace{-2mm}
  \centering
  \includegraphics[width=1.00\linewidth]{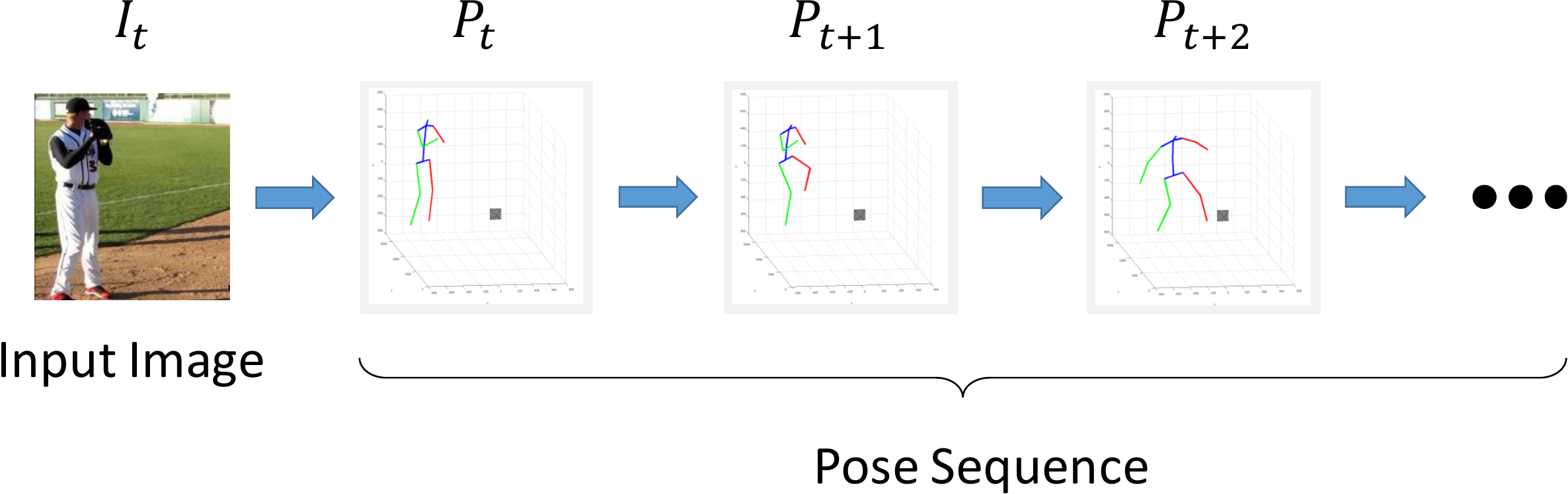}\\
  % \vspace{-2mm}
  \caption{\small The problem of human pose prediction. The input is a single
image, and the output is a 3D pose sequence.}
  \vspace{-2mm}
  \label{fig:problem}
\end{figure}

The problem studied in this paper assumes the input to be a single image
captured at time $t$. The output is a sequence of 3D human body skeletons
$P=\{P_t,\dots,P_{t+T}\}$, where $P_i\in\mathbb{R}^{3 \times N}$ denotes the
predicted skeleton at time $i$, represented by the 3D locations of $N$
keypoints. See Fig.~\ref{fig:problem} for an illustration of the problem. Note
that this formulation generalizes single-frame 3D human pose estimation, which
can be viewed as a special case when $T=0$.

\subsection{Network Architecture}

\begin{figure}[t]
  % \vspace{-2mm}
  \centering
  \includegraphics[width=1.00\linewidth]{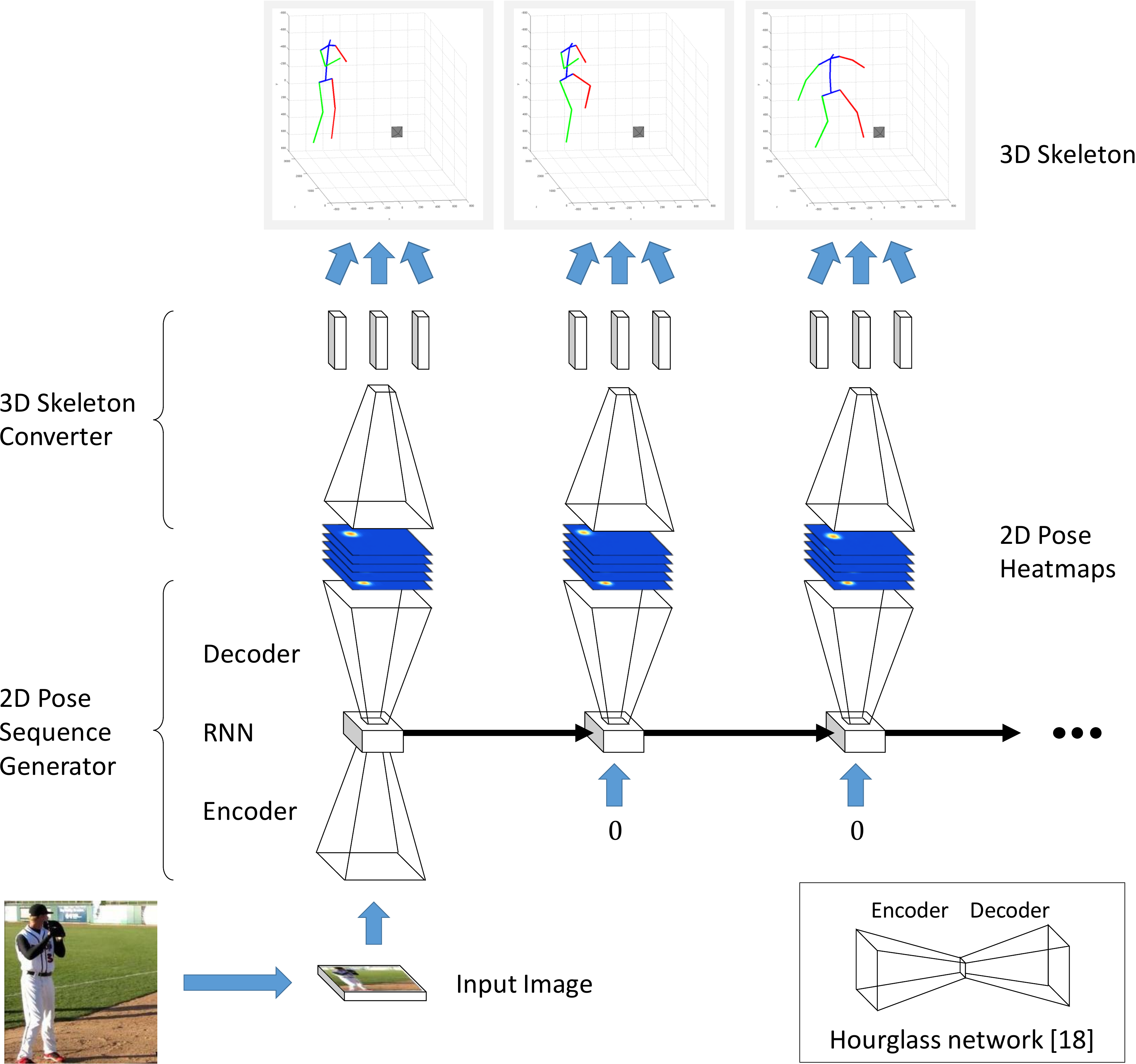}\\
  % \vspace{-2mm}
  \caption{\small A schematic view of the unrolled 3D-PFNet.}
  \vspace{-2mm}
  \label{fig:schematic}
\end{figure}

We propose a deep recurrent network to predict human skeleton sequences
(Fig.~\ref{fig:schematic}). The network is divided into two components: first,
a 2D pose sequence generator that takes an input image and sequentially
generates 2D body poses, where each pose is represented by heatmaps of
keypoints; second, a 3D skeleton converter that converts each 2D pose into a 3D
skeleton.

\vspace{-3mm}

\paragraph{2D Pose Sequence Generator} The first step is to generate a 2D body
pose sequence from the input image. The task can be decomposed into estimating
the body pose in the given frame and predicting the body poses in upcoming
frames. We thus leverage recent advances on single-frame human pose estimation
as well as sequence prediction. The recently introduced \textit{hourglass}
networks \cite{newell:eccv2016} have demonstrated state-of-the-art performance
on large-scale human pose datasets \cite{andriluka:cvpr2014}. We summarize the
hourglass architecture as follows: The first half of the hourglass processes
the input image with convolution and pooling layers to a set of low resolution
feature maps. This resembles conventional ConvNets (and is frequently referred
to as ``encoder'' in generative models). The second half (frequently referred
to as ``decoder'') then processes the low resolution feature maps with a
symmetric set of upsampling and convolution layers to generate dense detection
heatmaps for each keypoint at high resolution. A critical issue of this
architecture is the loss of high resolution information in the encoder output
due to pooling. Thus one key ingredient is to add a ``skip connection'' before
each pooling layer to create a direct path to the counterpart in the decoder.
As a result, the hourglass can consolidate features from multiple scales in
generating detection outputs.

While achieving promising results on single-frame pose estimation, the
hourglass network is incapable of predicting future poses. A straightforward
extension is to increase the channel size of its output to jointly generate
predictions for future frames \cite{walker:eccv2016,vondrick:nips2016}.
However, the drawback is that a trained network will always predict output for
a fixed number of frames. To bypass this constraint, we choose to formulate
pose forecasting as a sequence prediction problem by adopting recurrent neural
networks (RNNs).

RNNs extend conventional DNNs with feedback loops to enable sequential
prediction from internal states driven by both current and pass observations.
Our key idea is to introduce an RNN to the neck of the hourglass, i.e. between
the encoder and decoder. We hypothesize that the global pose features encoded
in the low resolution feature maps are sufficient to drive the future
predictions. We refer to the new network as the \textit{recurrent hourglass}
architecture. Fig.~\ref{fig:schematic} illustrates the process of generating
pose sequence from the unrolled version of the recurrent hourglass network.
First, the given image is passed into the encoder to generate low resolution
feature maps. These feature maps are then processed by an RNN to update its
internal states. Note that the internal states here can be viewed as the
``belief'' on the current pose. This ``belief'' is then passed to the decoder
to generate pose heatmaps for the input image. To generate pose for the next
timestep, this ``belief'' is fed back to the RNN and then updated to account
for the pose change. The updated ``belief'' is again passed to the decoder to
generate heatmaps for the second timestep. This process will repeat, and in the
end we will obtain a sequence of 2D pose heatmaps. Since we assume a
single-image input, the encoder is used only in the initial frame. Starting
from the second frame, the input to RNN is set to zeros. As mentioned earlier,
it is natural to extend our model to video inputs by adding an encoder at every
timestep.

\begin{figure*}[t]
  % \vspace{-2mm}
  \centering
  \includegraphics[width=1.00\linewidth]{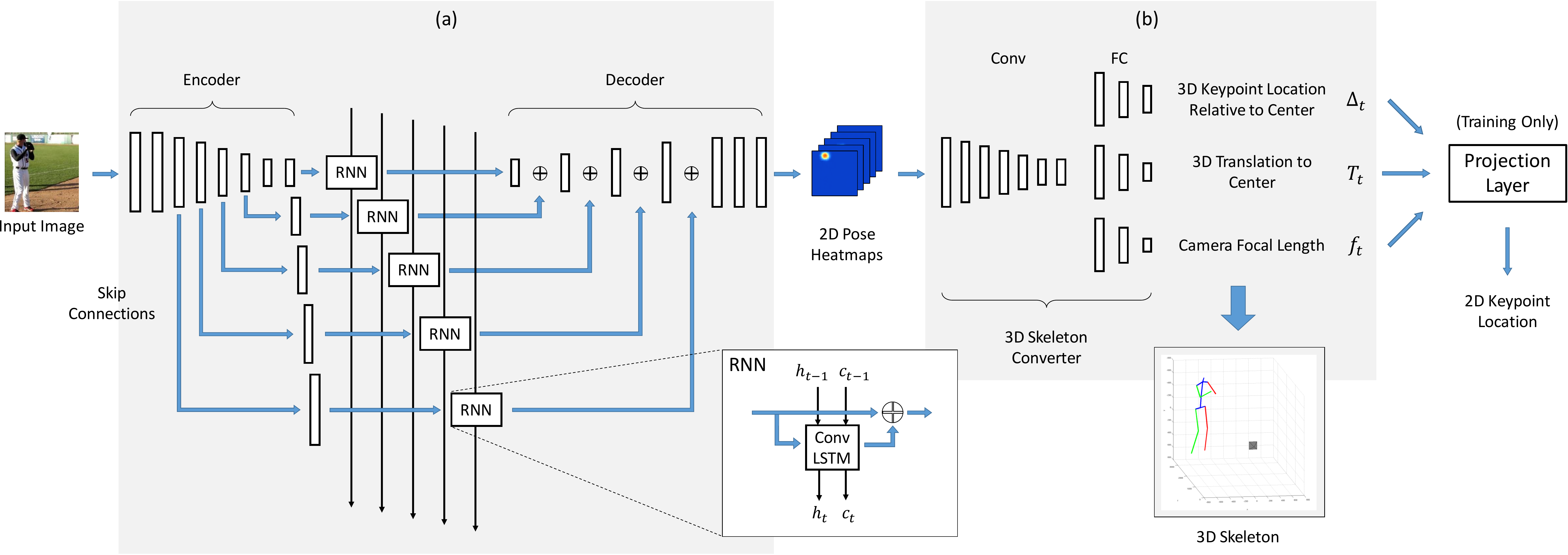}\\
  % \vspace{-2mm}
  \caption{\small Architecture of the 3D-PFNet. (a) The recurrent hourglass
architecture for 2D pose forecasting. (b) The 3D skeleton converter.}
  \vspace{-2mm}
  \label{fig:architecture}
\end{figure*}

For the RNN, we adopt the long short-term memory (LSTM) architecture
\cite{jozefowicz:icml2015} due to its strength in preserving long-term memory.
We apply two tricks: First, conventional LSTMs are used in fully-connected
architectures. Since the hourglass network is fully convolutional and the
encoder output is a feature map, we apply the LSTM convolutionally on each
pixel. This is equivalent to replacing the fully-connected layers in LSTM by
$1\times1$ convolution layers. Second, we apply the residual architecture
\cite{he:cvpr2016} in our RNN to retain a direct path from the encoder to the
decoder. As a result, we place less burden on the RNN as it only needs to learn
the ``changes'' in poses. Fig.~\ref{fig:architecture} (a) shows the detailed
architecture of our recurrent hourglass networks. Note that we also place an
RNN on the path of each skip connection of the hourglass.

\vspace{-3mm}

\paragraph{3D Skeleton Converter} The second step is to convert the heatmap
sequence into a sequence of 3D skeletons. Many recent works have addressed the
problem of recovering 3D skeleton structures from the 2D projection of their
keypoints \cite{zhou:cvpr2015,wu:eccv2016}. Zhou \emph{et
al.}\cite{zhou:cvpr2015} assumes the unknown 3D pose can be approximated by a
linear combination of a set of predefined basis poses, and propose to minimize
reprojection error with a convex relaxation approach. Wu \emph{et al.}
\cite{wu:eccv2016} adopts a similar assumption but instead uses a DNN to
estimate the linear coefficients and camera parameters. Both methods use a
top-down approach by leveraging a set of ``prior pose'' models. On the
contrary, we propose a bottom-up, data driven approach that directly predicts
the 3D keypoint locations from local 2D features. We hypothesize that the
bottom-up reconstruction can outperform top-down approaches given sufficiently
complex models and a vast amount of training data.

We model 3D skeletons and their 2D projection with a perspective projection
model. Recall that a 3D skeleton $P\in\mathbb{R}^{3 \times N}$ is represented
by $N$ keypoints in the camera coordinate system. We can decompose $P$ by $P =
\Delta + T1^{T}$, where $\Delta\in\mathbb{R}^{3 \times N}$ represents the
relative position of the $N$ keypoints to their center in 3D, and
$T\in\mathbb{R}^{3 \times 1}$ represents the translation to the center. Let $f$
be the camera focal length and assume the principal point is at the image
center. The goal of our 3D skeleton converter is to estimate $\{\Delta, T, f\}$
from the observed 2D heatmaps. Fig.~\ref{fig:architecture} (b) details the
architecture. The heatmaps generated at each timestep are first processed by
another encoder. Now instead of connecting to a decoder, the encoder output is
forwarded to three different branches. Each branch consists of three
fully-connected layers, and the three branches will output $\Delta$, $T$, and
$f$, respectively. Note that estimating camera parameters is unnecessary if we
have ground-truth 3D keypoint annotations to train our network. However, 3D
pose data is hard to collect and thus are often unavailable in in-the-wild
human pose datasets. With the estimated camera parameters, we can apply a
projection layer \cite{wu:eccv2016} at the output of the network to project 3D
keypoints to 2D, and measure the loss on reprojection error for training.

\subsection{Training Strategy}
\label{sub:training}

Our 3D-PFNet is composed of multiple sub-networks. Different sub-networks serve
different sub-tasks and thus can exploit different sources of training data. We
therefore adopt a three-step, task-specific training strategy.

\vspace{-3mm}

\paragraph{1) Hourglass} The hourglass network (i.e. encoder and decoder)
serves the task of single-frame 2D pose estimation. We therefore pre-train the
hourglass network by leveraging large human pose datasets that provide 2D body
joint annotations. We follow the training setup in \cite{newell:eccv2016} and
apply a Mean Squared Error (MSE) loss for the predicted and ground-truth
heatmaps.

\vspace{-3mm}

\paragraph{2) 3D Skeleton Converter} Training the 3D skeleton converter
requires correspondences between 2D heatmaps and 3D ground truth of $\{\Delta,
T, f\}$. We exploit the ground-truth 3D human poses from motion capture (MoCap)
data. We synthesize training samples using a technique similar to
\cite{wu:eccv2016}: First, we randomly sample a 3D pose and camera parameters
(i.e. focal length, rotation, and translation). We then project the 3D
keypoints to 2D coordinates using the sampled camera parameters, followed by
constructing the corresponding heatmaps. This provides us with a training set
that is diverse in both human poses and camera viewpoints. We apply an MSE loss
for each output of $\Delta$, $T$, and $f$, and an equal weighting to compute
the total loss.

\vspace{-3mm}

\paragraph{3) Full Network} Finally, we train the full network (i.e. hourglass
+ RNNs + 3D skeleton converter) using static images and their corresponding
pose sequences. To ease the training of LSTM, we apply curriculum learning
similar to \cite{yang:nips2015}: We start training the full network with pose
sequences of length 2. Once the training converges, we increase the sequence
length to 4 and resume the training. We repeat doubling the sequence length
whenever training converges. We train the network with two sources of losses:
The first source is the heatmap loss used for training the hourglass. Since we
assume the 3D ground truths are unavailable in image and video datasets, we
cannot apply loss directly on $\Delta$, $T$, and $f$. We instead apply a
projection layer as mentioned earlier and adopt an MSE loss on 2D keypoint
locations. Note that replacing 3D loss with projection loss might diverge the
training and output implausible 3D body poses, since a particular 2D pose can
be mapped from multiple possible 3D configurations. We therefore initialize the
3D converter network with weights learned from the synthetic data, and keep the
weights fixed during the training of the full network.

\section{Experiments}

We evaluate our 3D-PFNet on two tasks: (1) \textit{2D pose forecasting} and (2)
\textit{3D pose recovery}.

\subsection{2D Pose Forecasting}

% which datasets? how are they used?
\paragraph{Dataset} We evaluate pose forecasting in 2D using the Penn Action
dataset \cite{zhang:iccv2013}. Penn Action contains 2326 video sequences (1258
for training and 1068 for test) covering 15 sports action categories. Each
video frame is annotated with a human bounding box along with the locations and
visibility of 13 body joints. Note that we do not evaluate our forecasted 3D
poses due to the lack of 3D annotations in Penn Action. During training, we
also leverage two other datasets: MPII Human Pose (MPII)
\cite{andriluka:cvpr2014} and Human3.6M \cite{ionescu:pami2014}. MPII is a
large-scale benchmark for single-frame human pose estimation. Human3.6M
consists of videos of acting individuals captured in a controlled environment.
Each frame is provided with the calibrated camera parameters and the 3D human
pose acquired from MoCap devices.

\vspace{-3mm}

\begin{figure}[t]
  % \vspace{-2mm}
  \centering
  \captionsetup[subfigure]{labelformat=empty}
  \begin{subfigure}[c]{0.49\linewidth}
    \centering
    \hspace{-1.6mm}
    \includegraphics[height=0.31\textwidth]{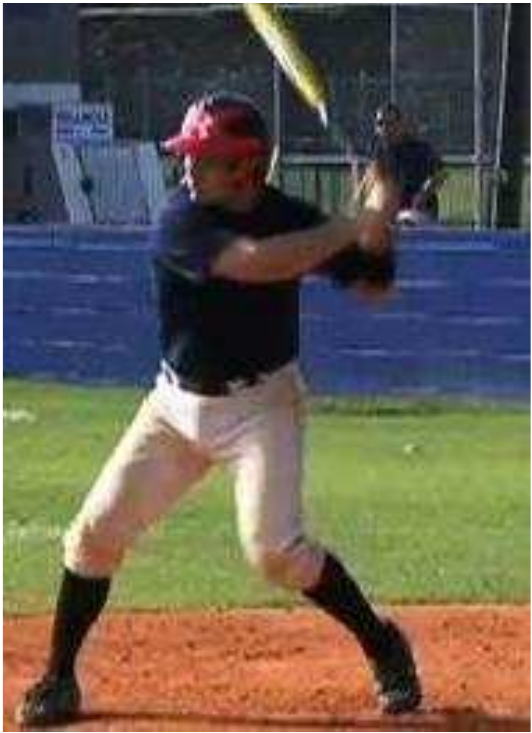}
    \includegraphics[height=0.31\textwidth]{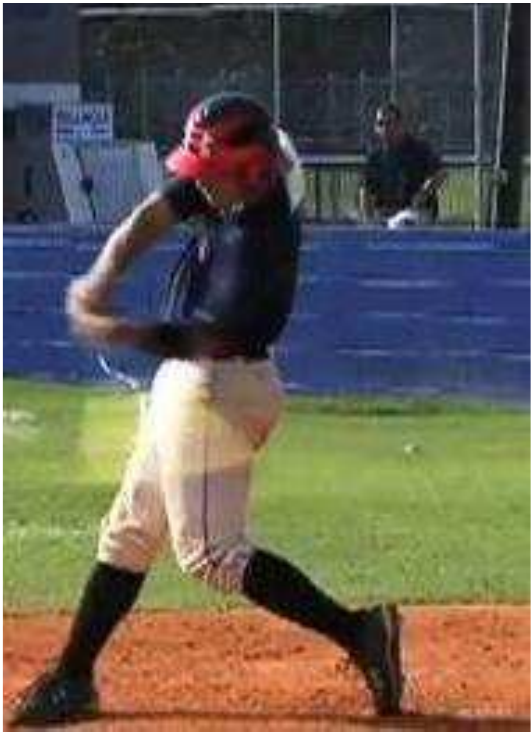}
    \includegraphics[height=0.31\textwidth]{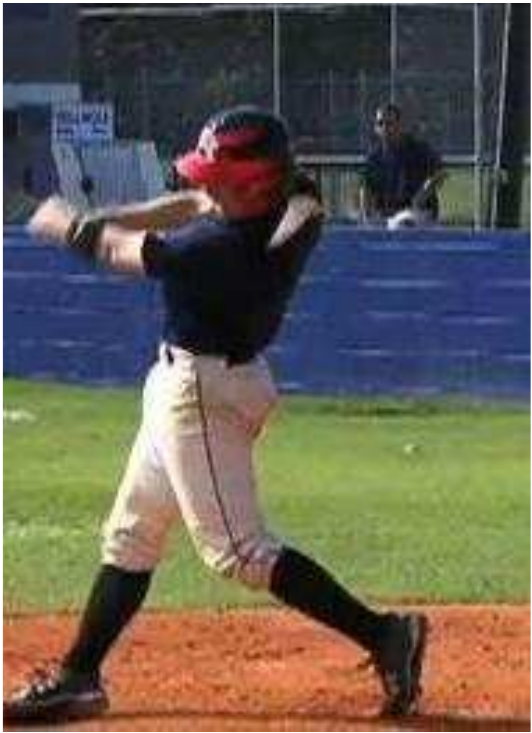}
    \includegraphics[height=0.31\textwidth]{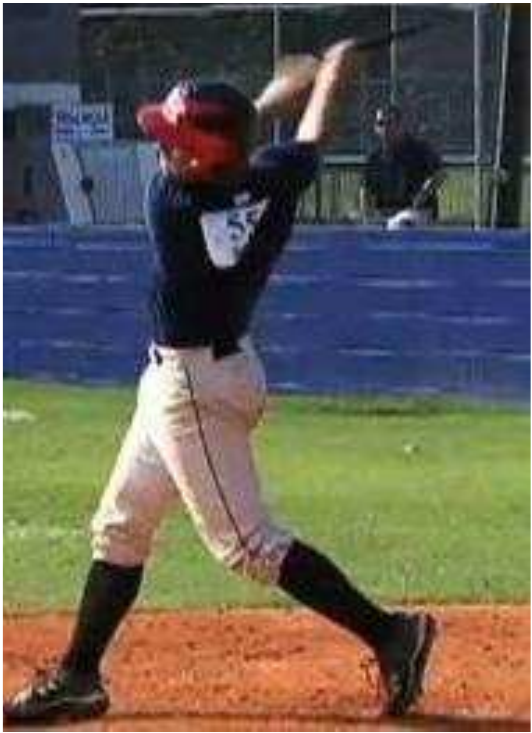}
    % \caption{\small Action}
  \end{subfigure}
  \begin{subfigure}[c]{0.49\linewidth}
    \centering
    \includegraphics[height=0.31\textwidth]{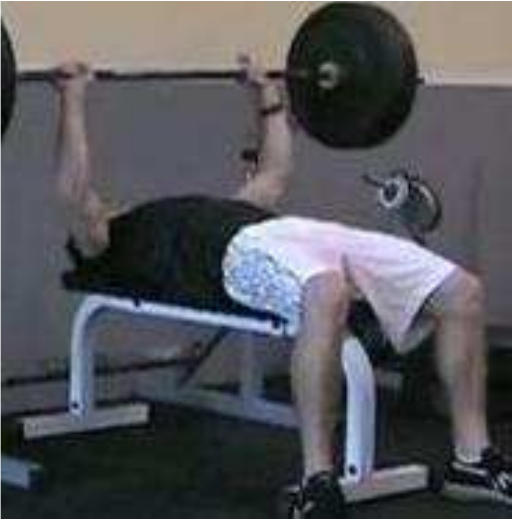}
    \includegraphics[height=0.31\textwidth]{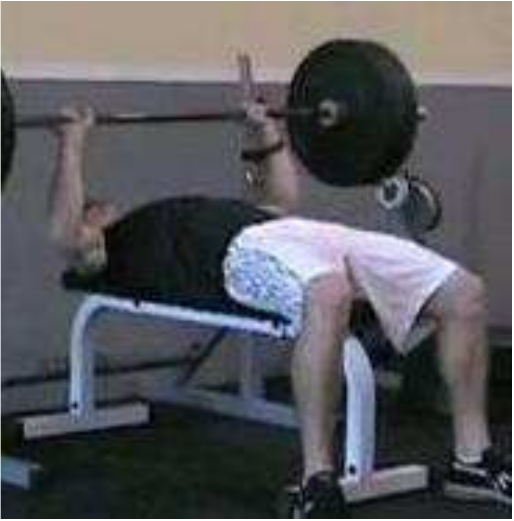}
    \includegraphics[height=0.31\textwidth]{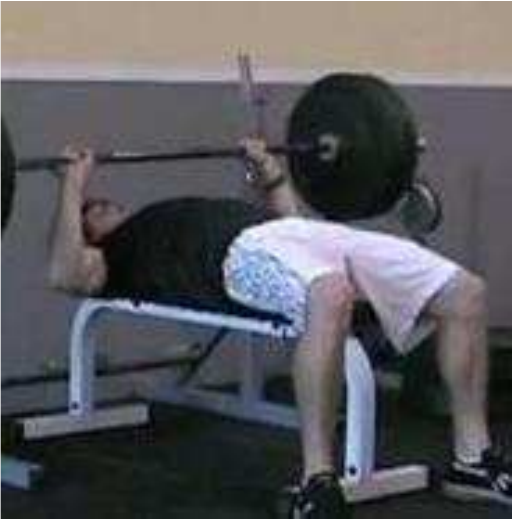}
    % \caption{\small Action}
  \end{subfigure}
  \\~\vspace{0mm}
  \\
  \begin{subfigure}[c]{0.49\linewidth}
    \centering
    \hspace{-1.6mm}
    \includegraphics[height=0.31\textwidth]{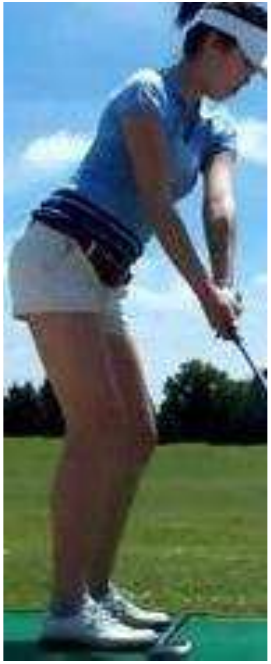}
    \includegraphics[height=0.31\textwidth]{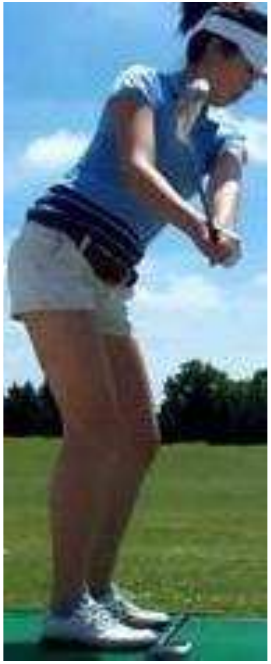}
    \includegraphics[height=0.31\textwidth]{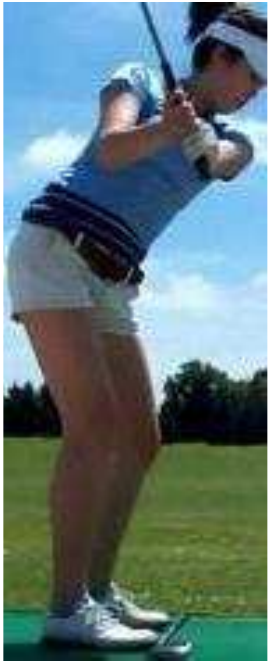}
    \includegraphics[height=0.31\textwidth]{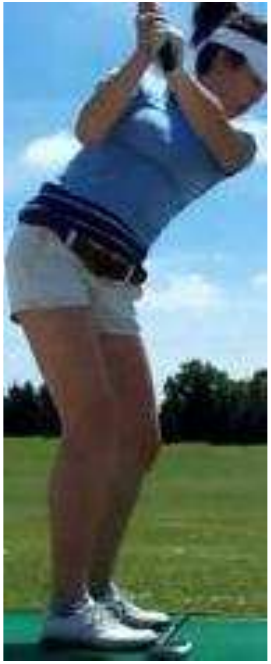}
    \includegraphics[height=0.31\textwidth]{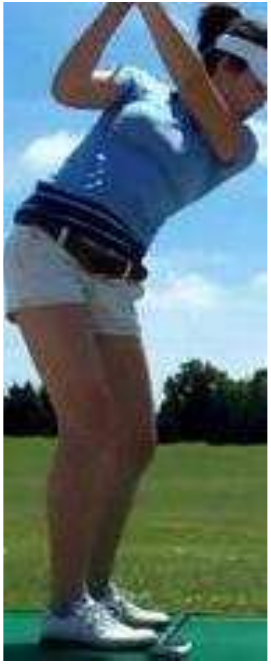}
    \includegraphics[height=0.31\textwidth]{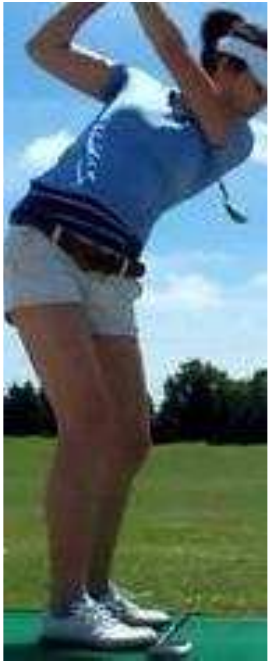}
    % \caption{\small Action}
  \end{subfigure}
  \begin{subfigure}[c]{0.49\linewidth}
    \centering
    \includegraphics[height=0.31\textwidth]{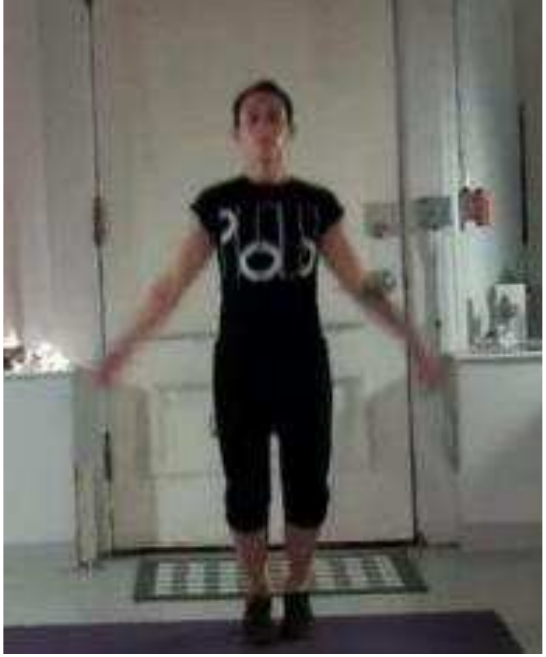}
    \includegraphics[height=0.31\textwidth]{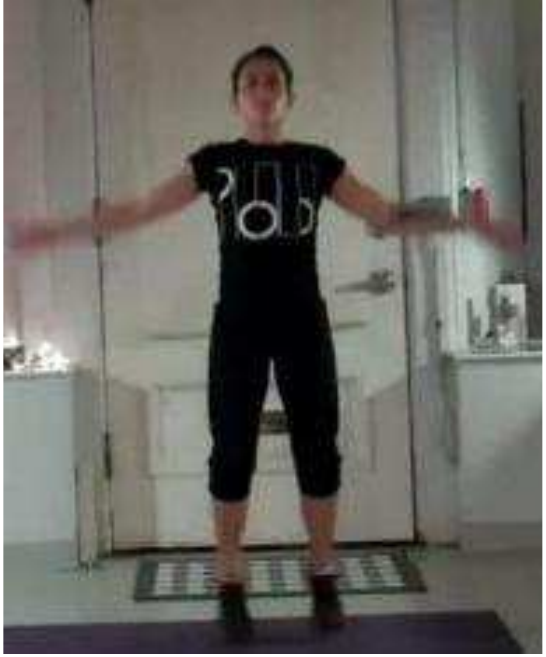}
    \includegraphics[height=0.31\textwidth]{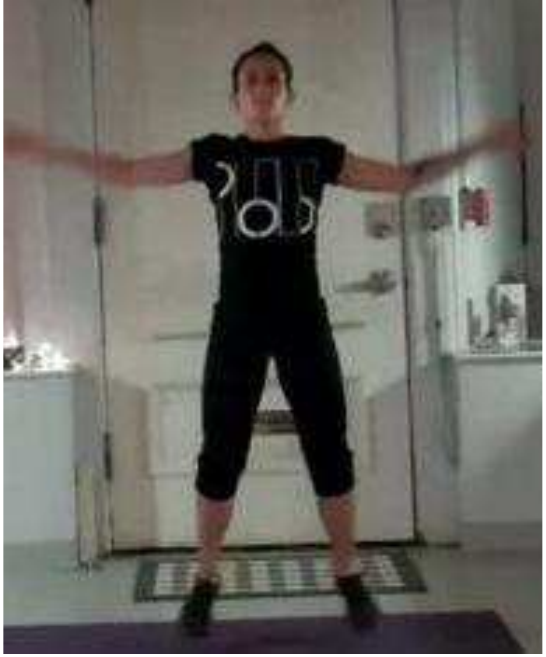}
    % \caption{\small Action}
  \end{subfigure}
  \\~\vspace{0mm}
  \\
  \begin{subfigure}[c]{0.49\linewidth}
    \centering
    \hspace{-1.6mm}
    \includegraphics[height=0.31\textwidth]{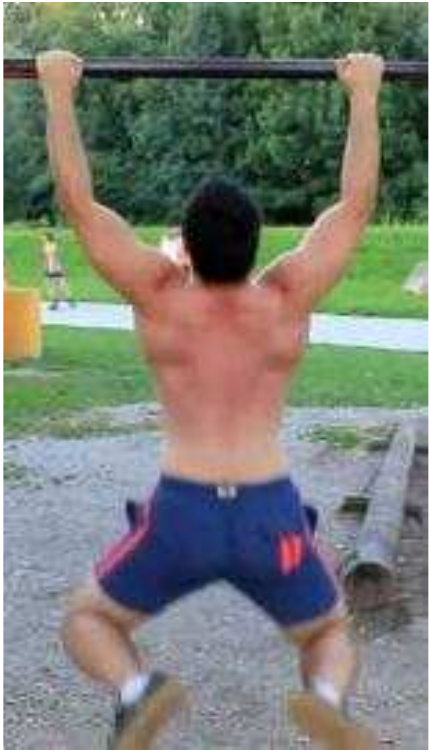}
    \includegraphics[height=0.31\textwidth]{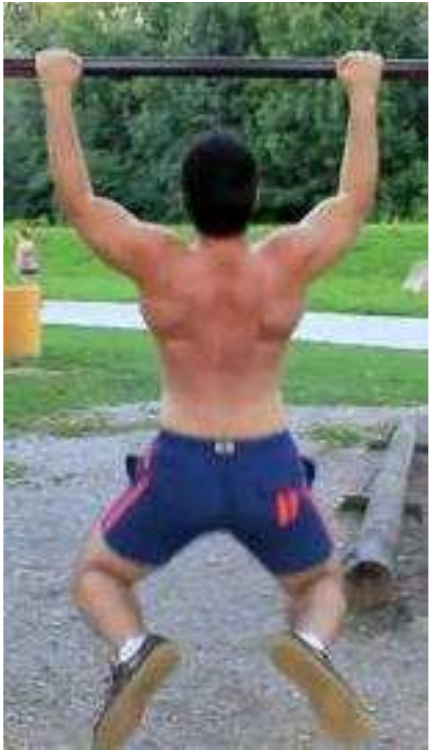}
    \includegraphics[height=0.31\textwidth]{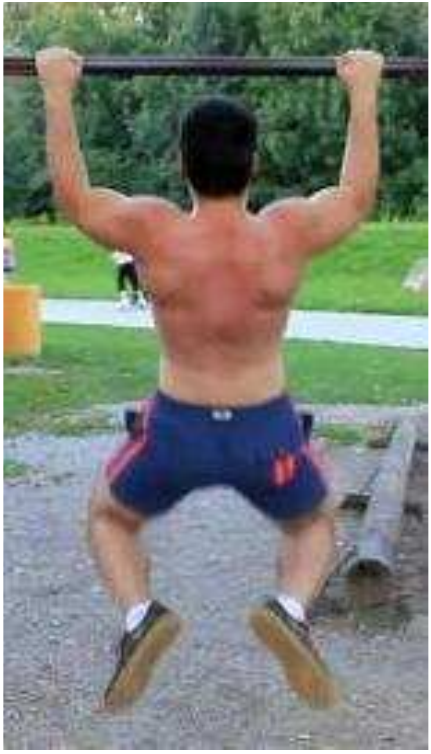}
    \includegraphics[height=0.31\textwidth]{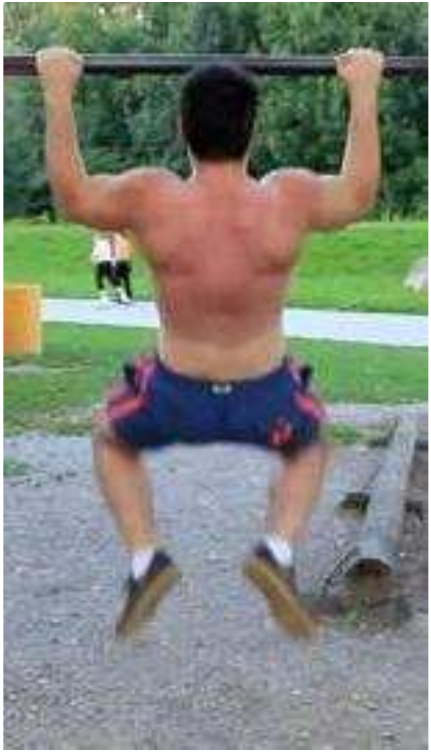}
    \includegraphics[height=0.31\textwidth]{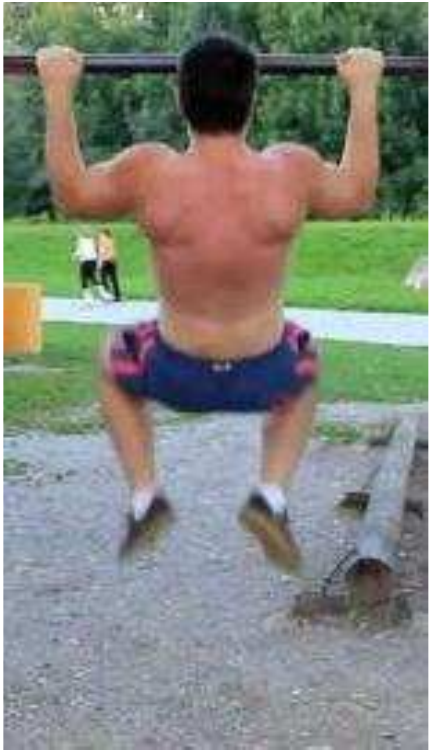}
    % \caption{\small Action}
  \end{subfigure}
  \begin{subfigure}[c]{0.49\linewidth}
    \centering
    \includegraphics[height=0.31\textwidth]{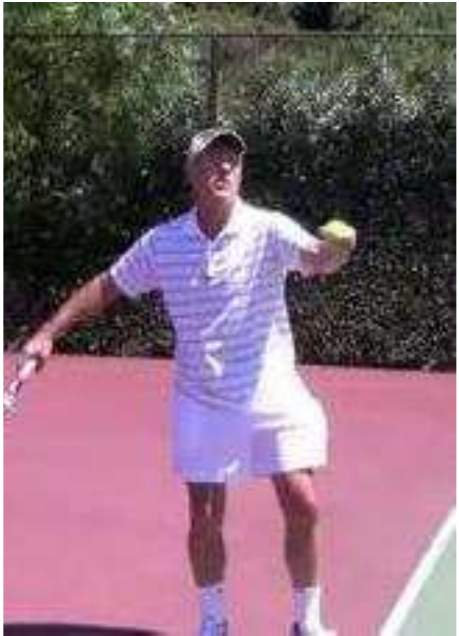}
    \includegraphics[height=0.31\textwidth]{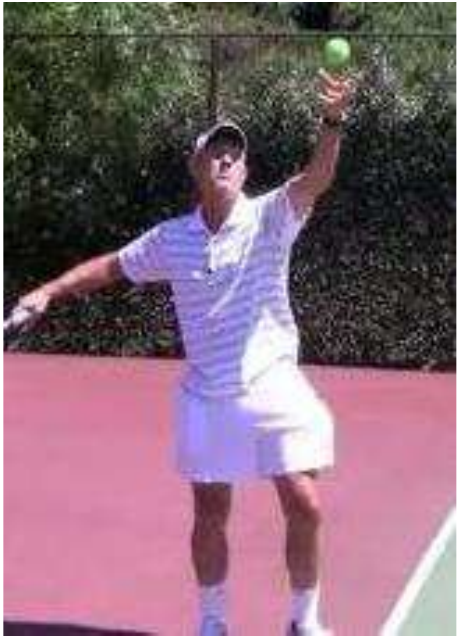}
    \includegraphics[height=0.31\textwidth]{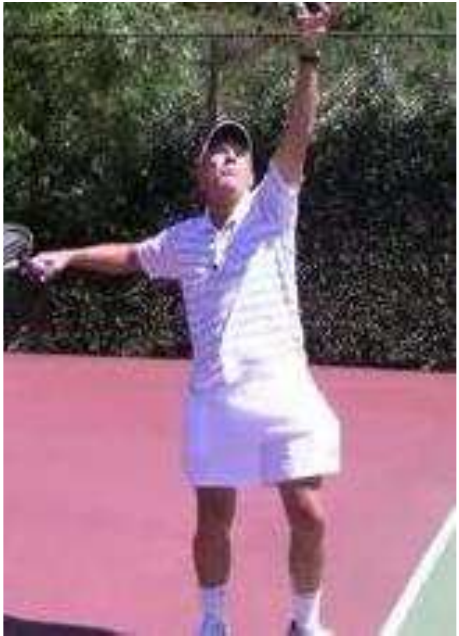}
    \includegraphics[height=0.31\textwidth]{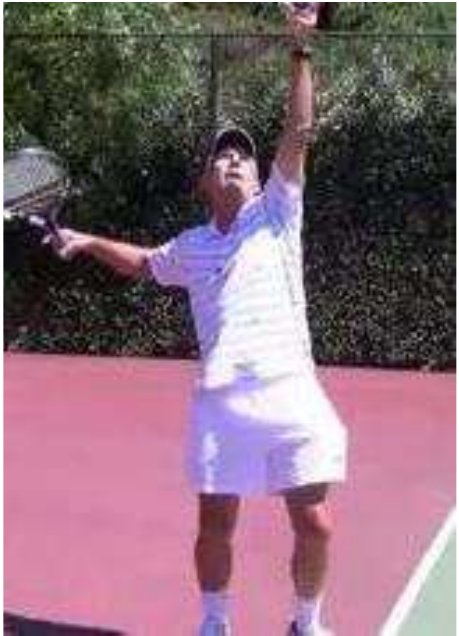}
    % \caption{\small Action}
  \end{subfigure}
  % \vspace{-2mm}
  \caption{\small Sample sequences of the processed Penn Action dataset. The
action classes are: baseball swing, bench press, golf swing, jumping jacks,
pull ups, and tennis serve.}
  % \vspace{-2mm}
  \label{fig:penn_crop}
\end{figure}

\paragraph{Evaluation Protocal} We preprocess Penn Action with two steps:
First, since our focus is not on human detection, we crop each video frame to
focus roughly around the human region: for each video sequence, we crop every
frame using the tight box that bounds the human bounding box across all frames.
Second, we do not assume the input image is always the starting frame of each
video (i.e. we should be able to forecast poses not only from the beginning of
a tennis forehand swing, but also from the middle or even shortly before the
action finishes). Thus for a video with $K$ frames, we generate $K$ sequences
by varying the starting frame. Besides, since adjacent frames contain similar
poses, we skip frames when generating sequences. The number of frames skipped
is video-dependent: Given a sampled starting frame, we always generate a
sequence of length 16, where we skip every $(K-1)/15$ frames in the raw video
sequence after the sampled starting frame. This is to ensure that our
forecasted output can ``finish'' each action in a predicted sequence of length
16. Note that once we surpass the end frame of a video, we will repeat the last
frame collected until we obtain 16 frames. This is to force the forecasting to
learn to ``stop'' and remain at the ending pose once an action has completed.
Fig.~\ref{fig:penn_crop} shows sample sequences of our processed Penn
Action.%~\footnote{Our experimental data and evaluation code are publicly
%available at
%\href{http://www.umich.edu/~ywchao/image-play/}{\texttt{http://www.umich.edu/$\sim$ywchao/image-play/}}.}

To evaluate the forecasted pose, we adopt the standard Percentage of Correct
Keypoints (PCK) metric \cite{andriluka:cvpr2014} from 2D pose estimation. PCK
measures the accuracy of keypoint localization by considering a predicted
keypoint correct if it falls within certain normalized distance of the ground
truth. This distance is normalized typically based on the size of the full body
bounding box \cite{yang:pami2013} or the head bounding box
\cite{andriluka:cvpr2014}. Since we have already cropped the frames based on
full body bounding boxes, we normalize the distance by $\max(h,w)$ pixels,
where $h$ and $w$ are the height and width of the cropped image. We ignore
invisible joints, and compute PCK separately for each of the 16 timesteps on
the test sequences.

\begin{figure}[t]
  % \vspace{-2mm}
  \centering
  \footnotesize
  \begin{tabular}{L{0.49\linewidth}@{\hspace{0.0mm}} L{0.49\linewidth}}
    \hspace{-2.5mm}
    \includegraphics[height=0.067\textwidth]{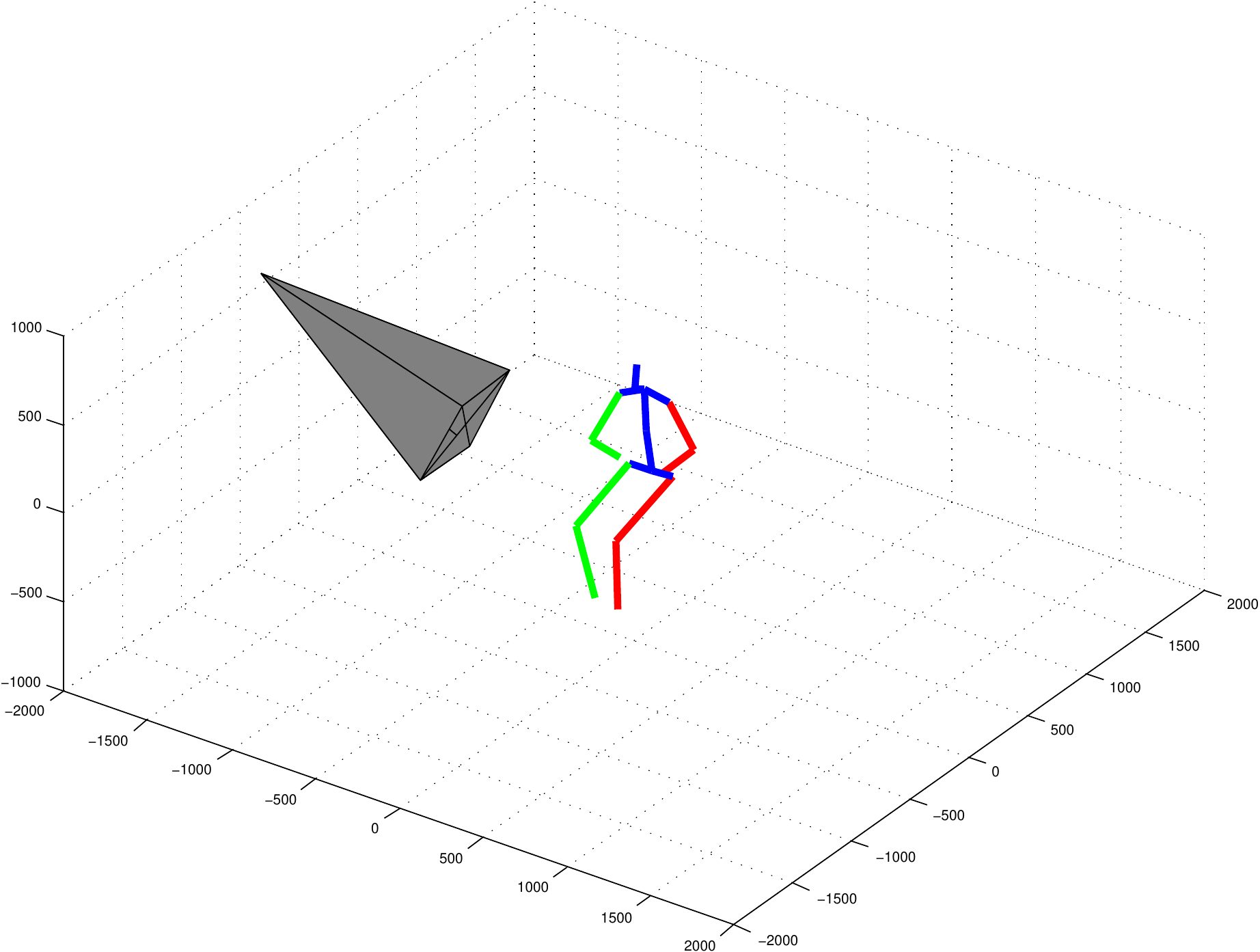}
    \includegraphics[height=0.067\textwidth]{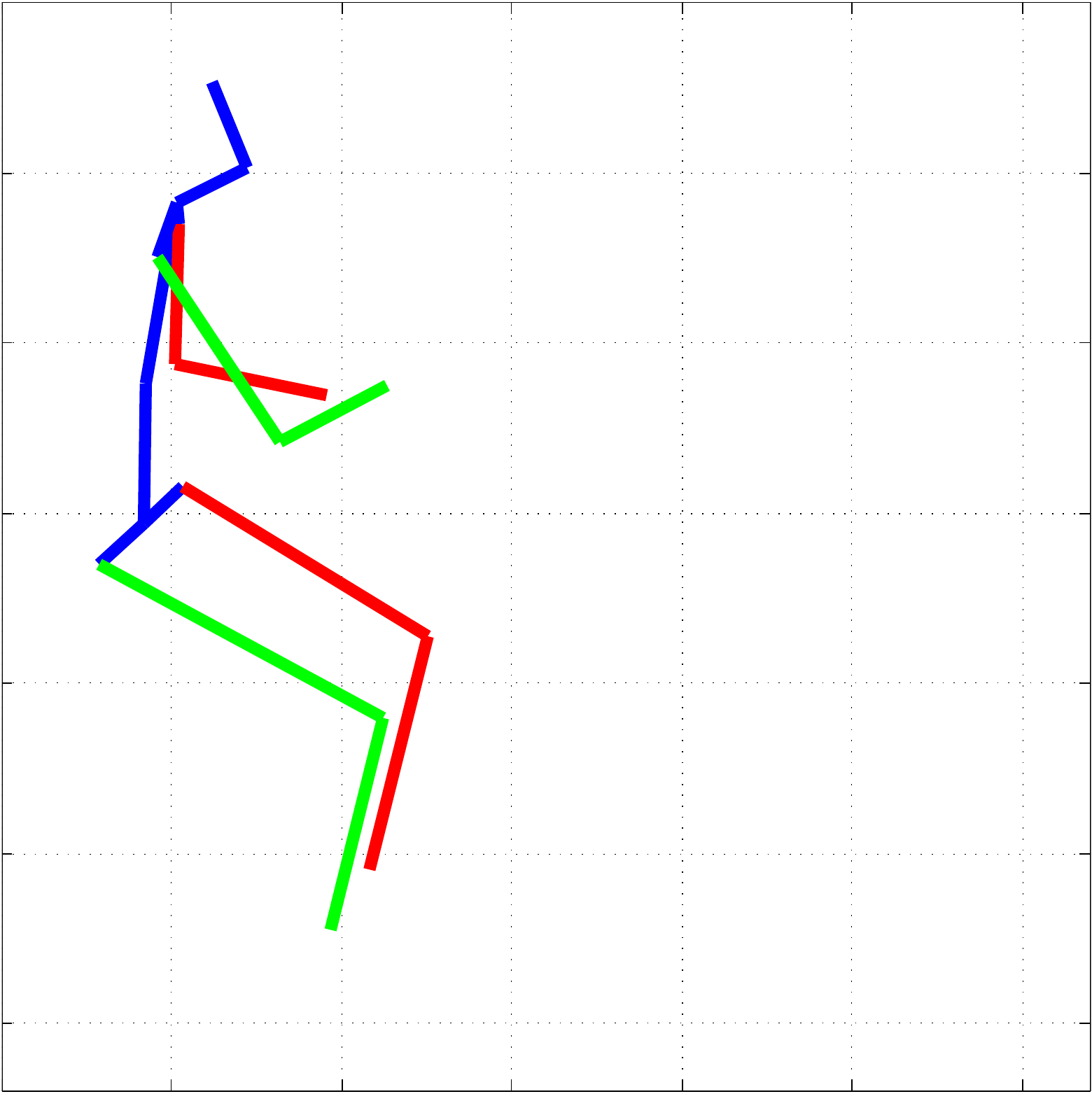}
    \includegraphics[height=0.067\textwidth]{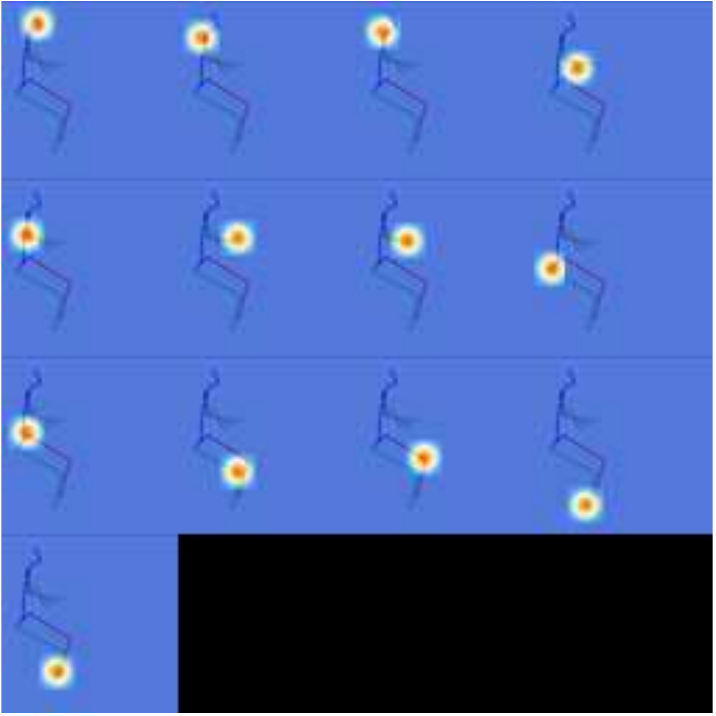}
    &
    \includegraphics[height=0.067\textwidth]{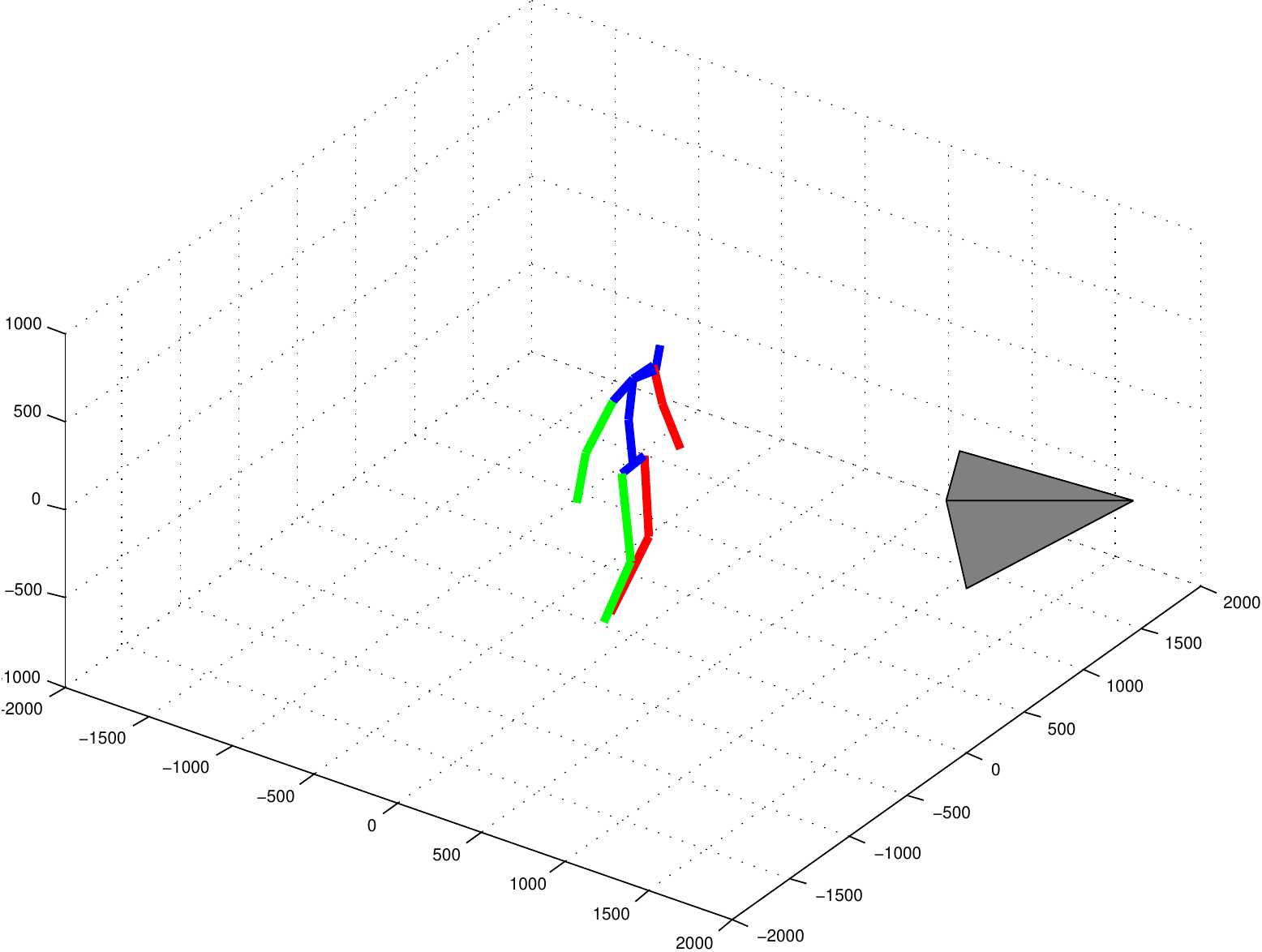}
    \includegraphics[height=0.067\textwidth]{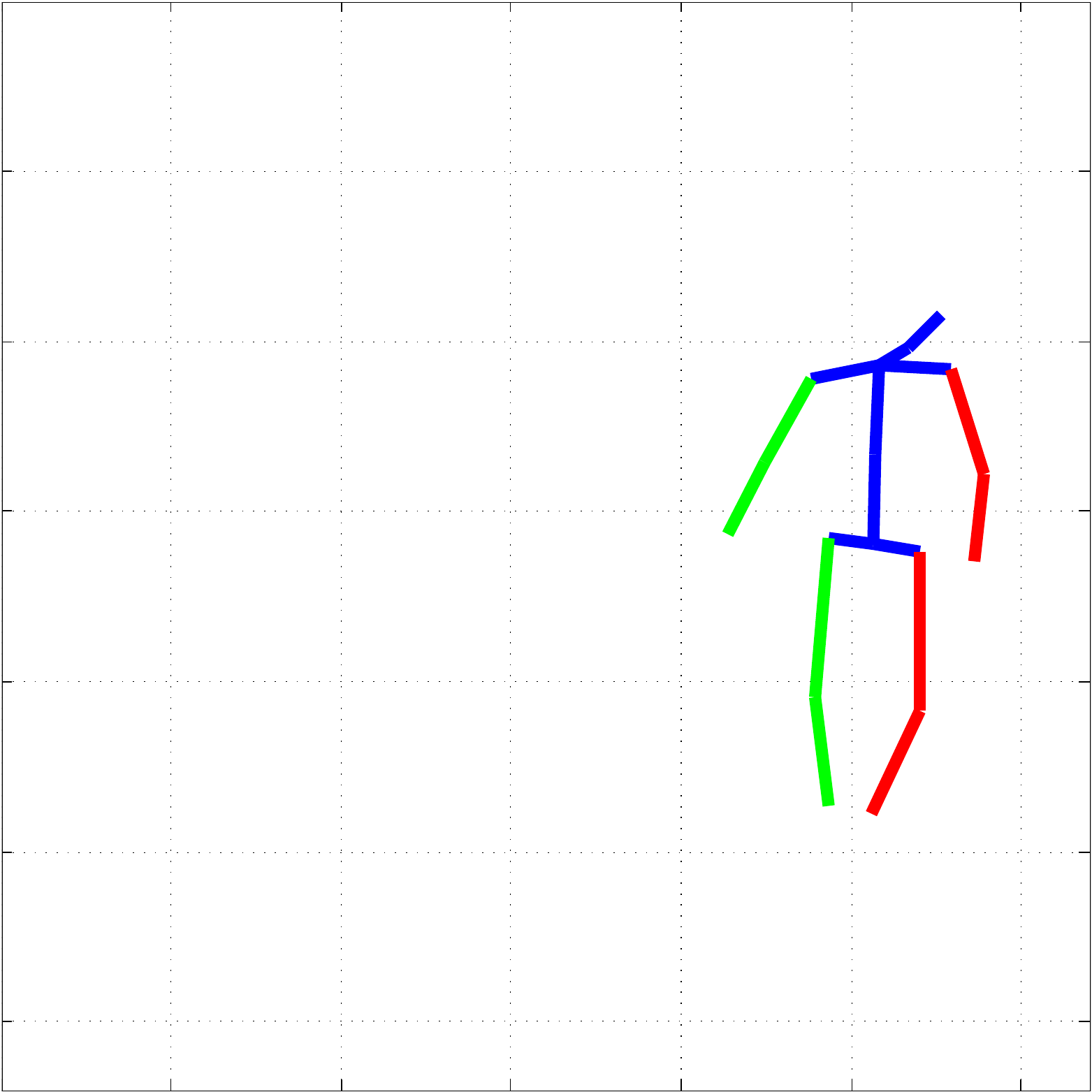}
    \includegraphics[height=0.067\textwidth]{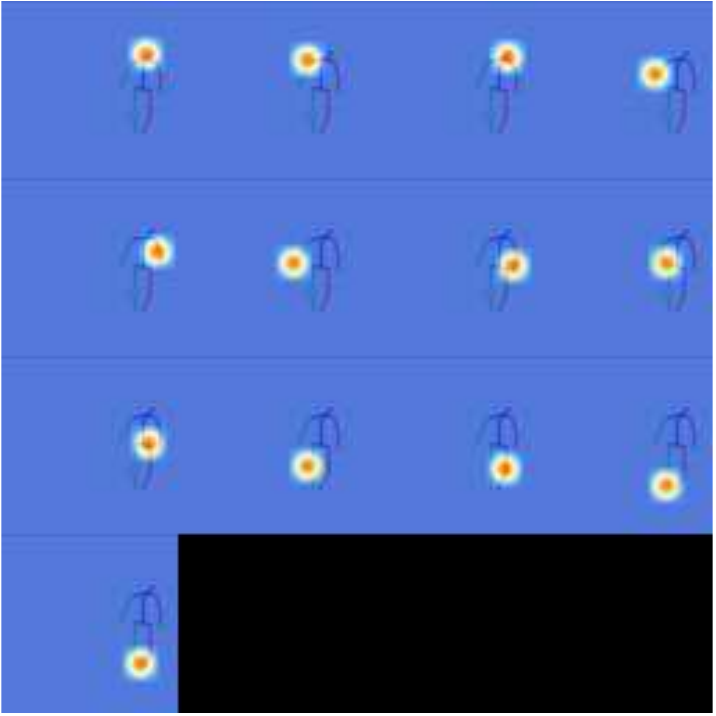}
    \\ [-1em] & \\
    \hspace{-2.5mm}
    \includegraphics[height=0.067\textwidth]{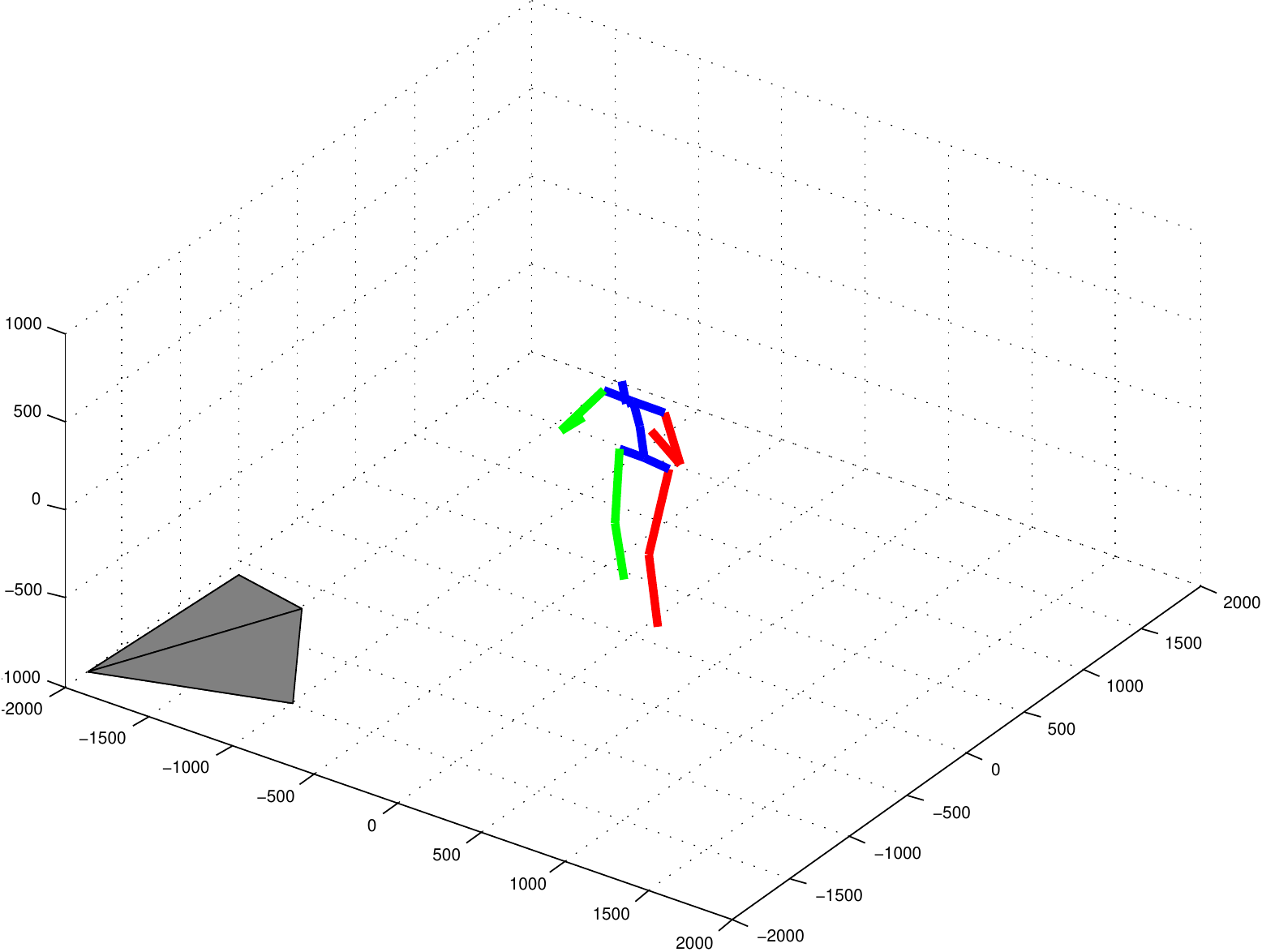}
    \includegraphics[height=0.067\textwidth]{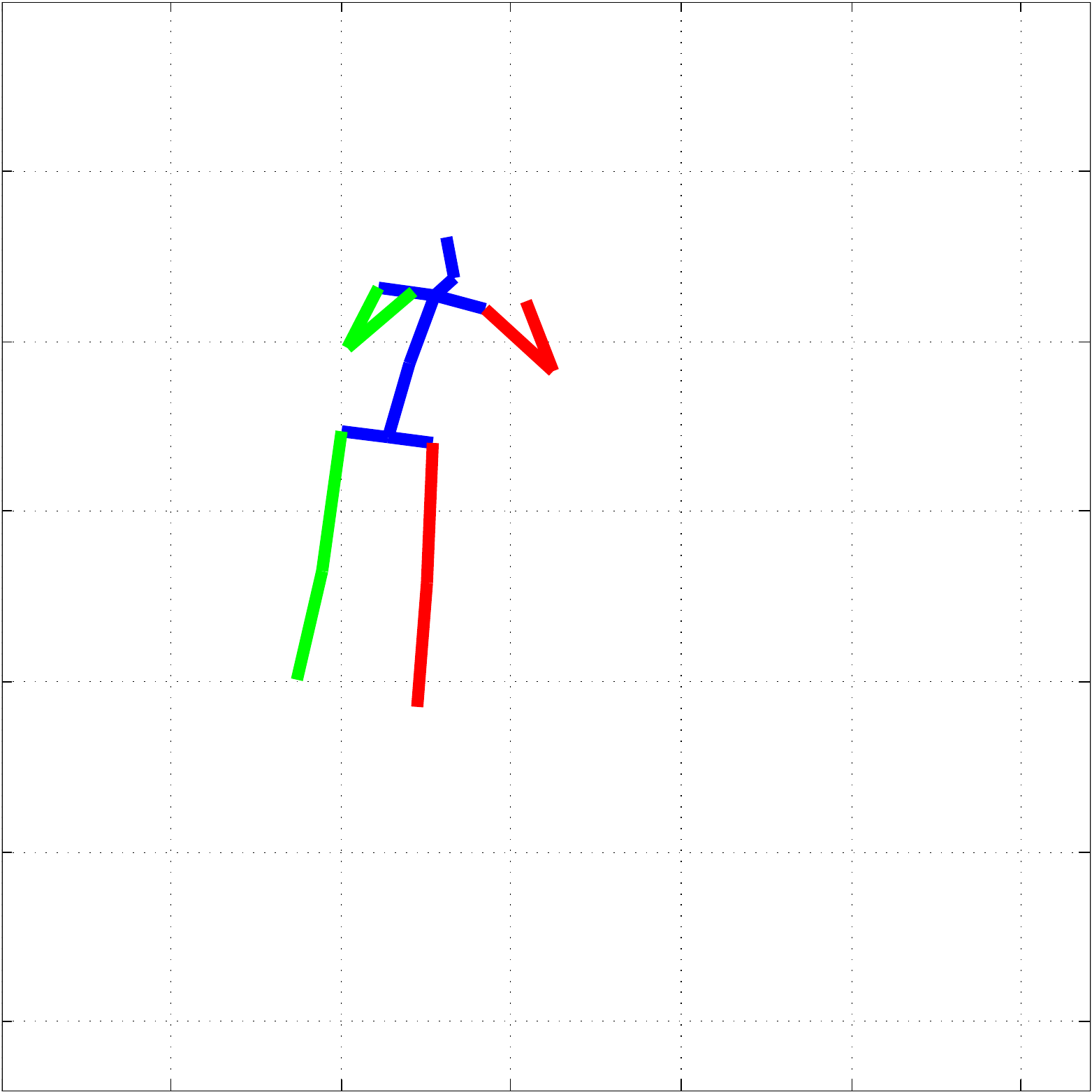}
    \includegraphics[height=0.067\textwidth]{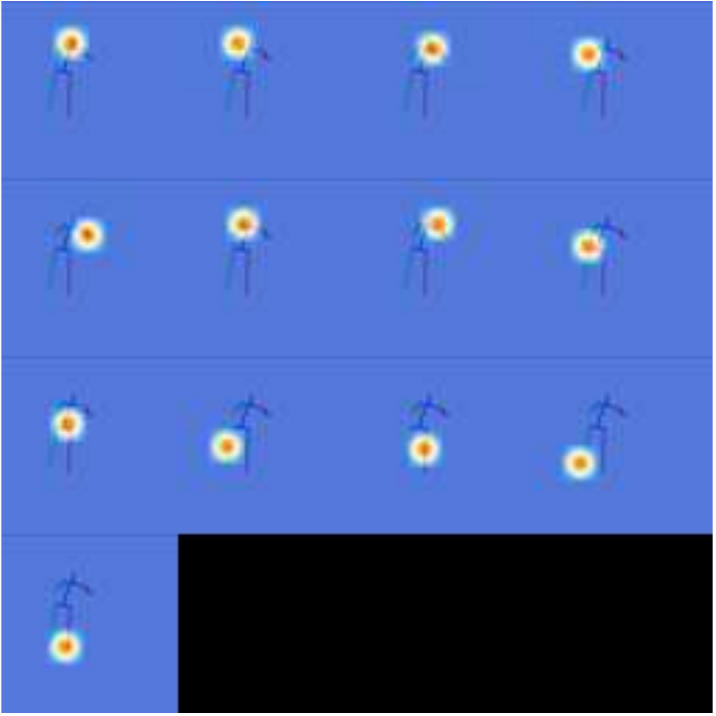}
    &
    \includegraphics[height=0.067\textwidth]{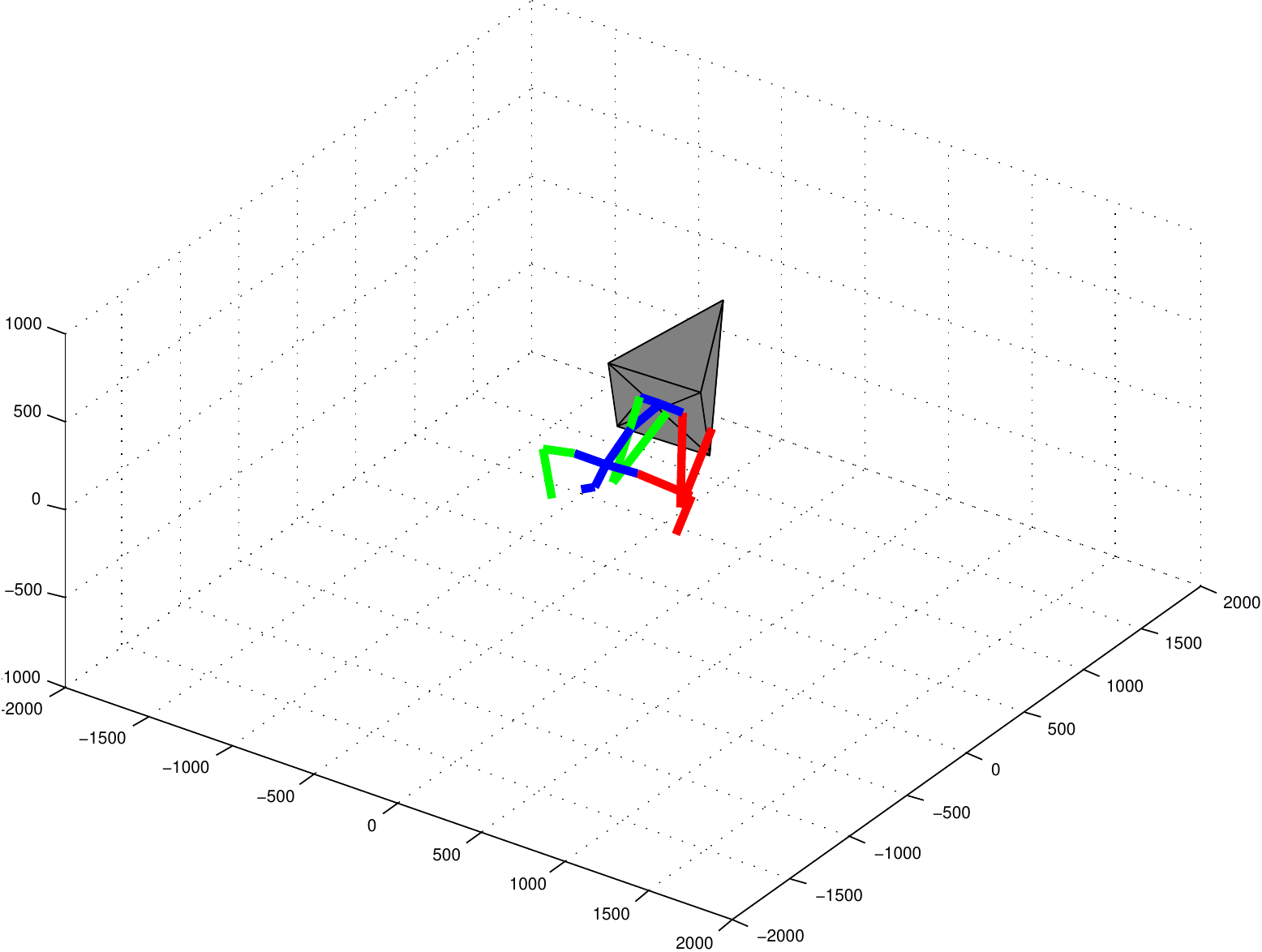}
    \includegraphics[height=0.067\textwidth]{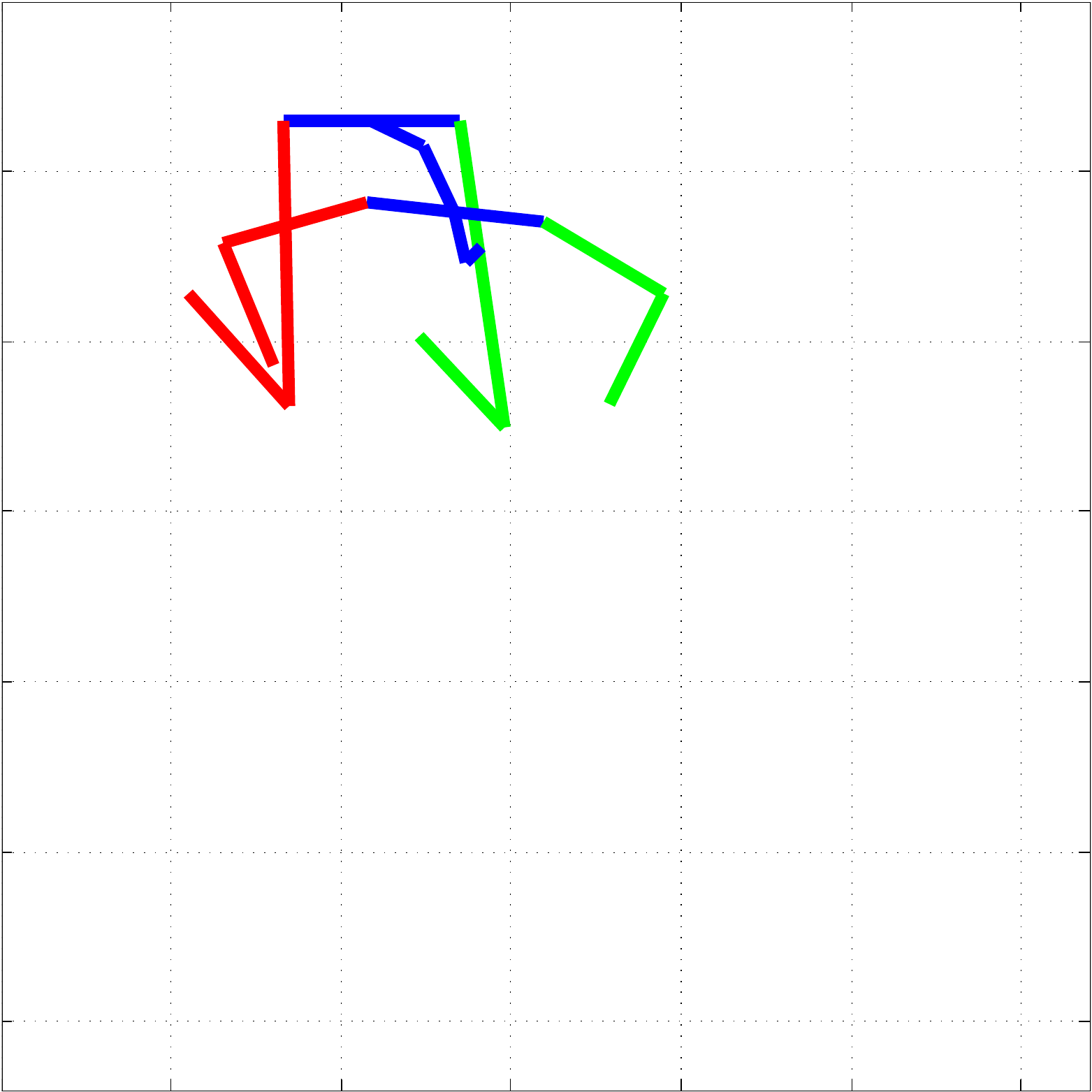}
    \includegraphics[height=0.067\textwidth]{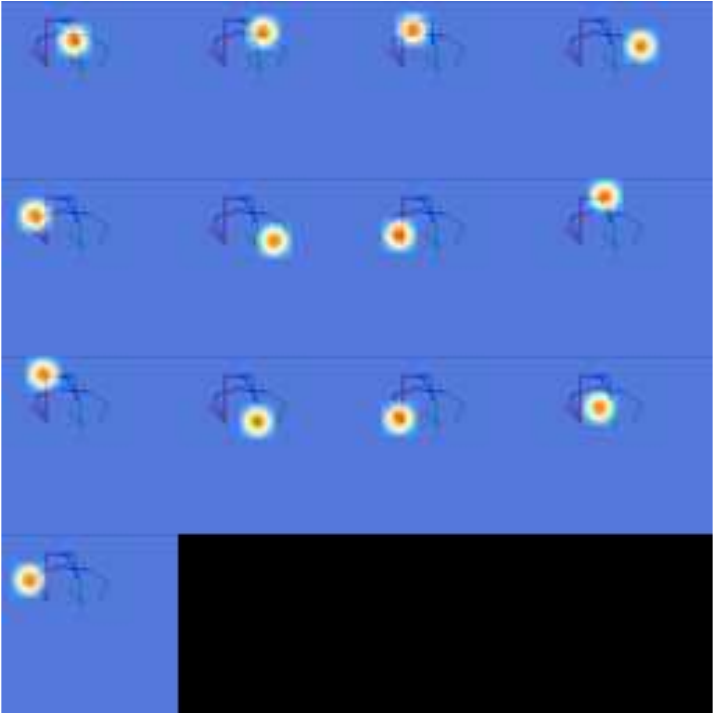}
  \end{tabular}
  % \vspace{-2mm}
  \caption{\small Samples of the synthetic data for training 3D skeleton
converter. Each triplet consists of (1) the sampled 3D pose and camera in world
coordinates, (2) the 2D projection, and (3) the converted heatmaps for 13
keypoints.}
  \vspace{-2mm}
  \label{fig:synthetic}
\end{figure}

\begin{figure*}[t]
  % \vspace{-2mm}
  \centering
  \captionsetup[subfigure]{labelformat=empty}
  \begin{subfigure}[c]{0.19\textwidth}
    \centering
    \includegraphics[width=1\textwidth]{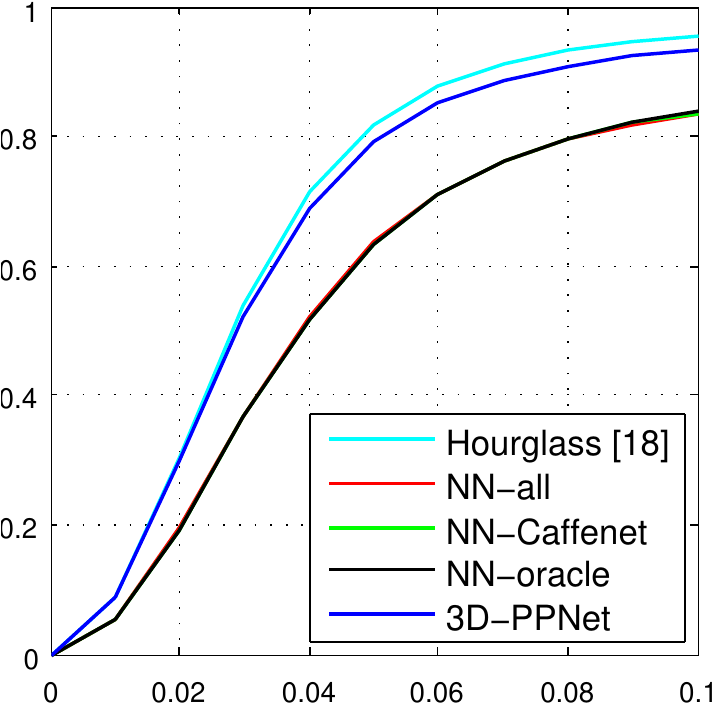}
    \caption{Timestep $1$}
  \end{subfigure}
  \begin{subfigure}[c]{0.19\textwidth}
    \centering
    \includegraphics[width=1\textwidth]{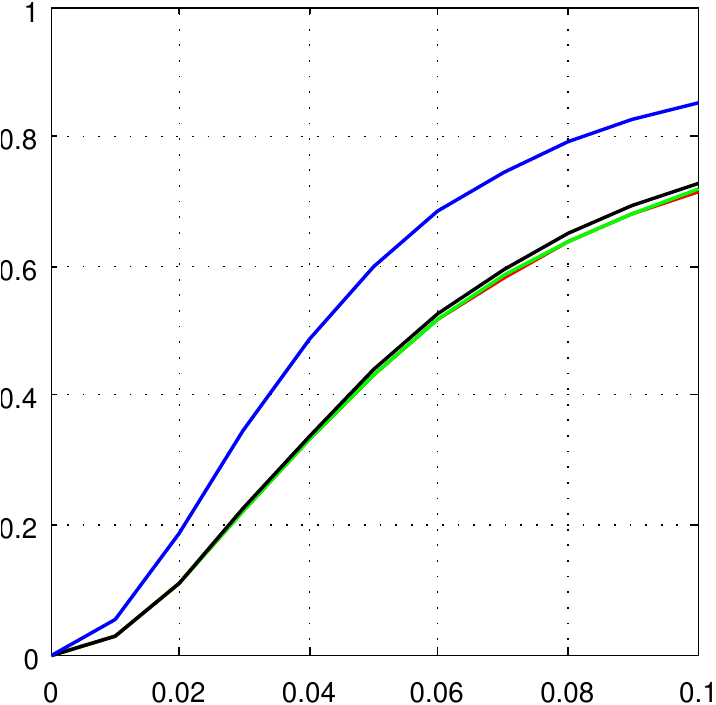}
    \caption{Timestep $2$}
  \end{subfigure}
  \begin{subfigure}[c]{0.19\textwidth}
    \centering
    \includegraphics[width=1\textwidth]{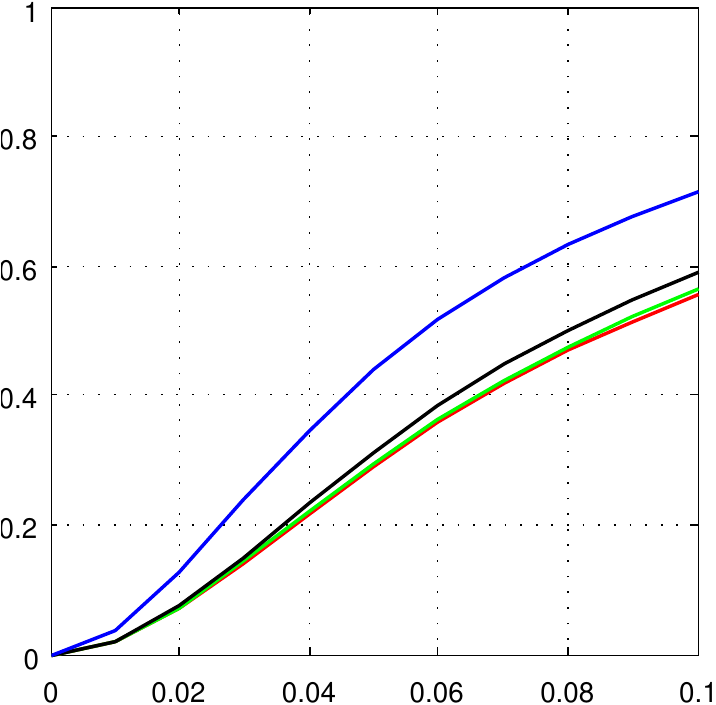}
    \caption{Timestep $4$}
  \end{subfigure}
  \begin{subfigure}[c]{0.19\textwidth}
    \centering
    \includegraphics[width=1\textwidth]{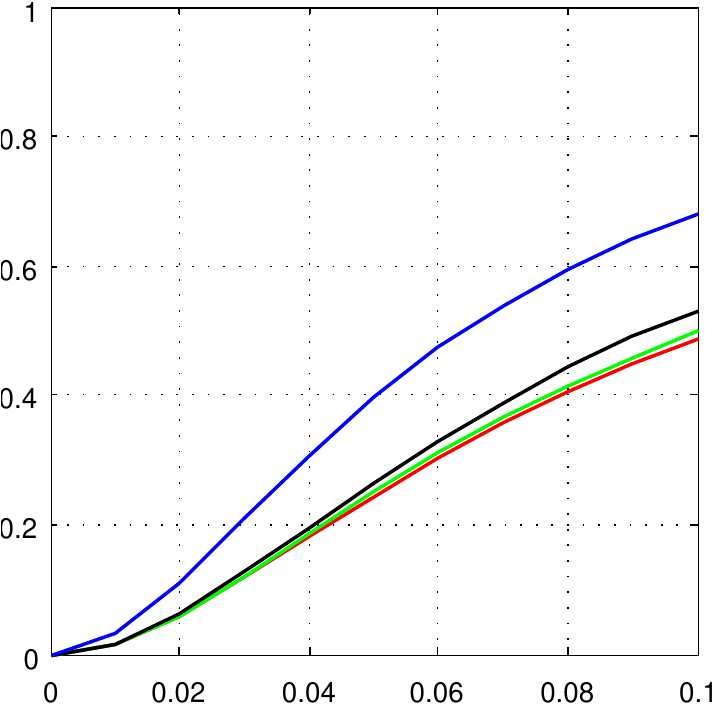}
    \caption{Timestep $8$}
  \end{subfigure}
  \begin{subfigure}[c]{0.19\textwidth}
    \centering
    \includegraphics[width=1\textwidth]{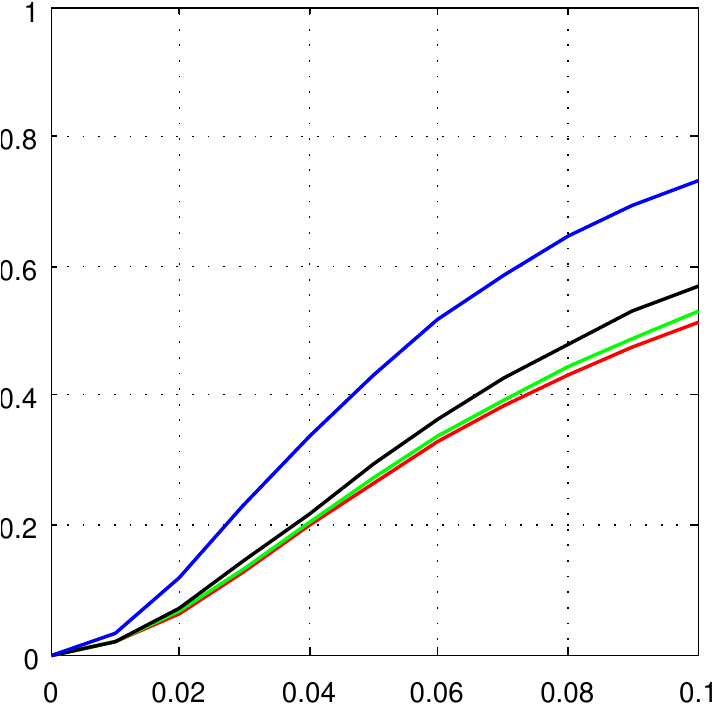}
    \caption{Timestep $16$}
  \end{subfigure}
  \vspace{-2mm}
  \caption{\small PCK curves at different timesteps. The x-axis is the distance
threshold and the y-axis is the PCK value. The hourglass network
\cite{newell:eccv2016} only estimates the current pose in timestep 1. Our
3D-PFNet outperforms all three NN baselines for all timesteps.}
  % \vspace{-2mm}
  \label{fig:pck}
\end{figure*}

\begin{table*}[t]
  \centering
  \small
  \setlength{\tabcolsep}{5.6pt}
  \begin{tabular}{l c c c c c c c c c c c c c c c c}
    \hline \TBstrut
    Timestep \#                      & 1    & 2    & 3    & 4    & 5    & 6    & 7    & 8    & 9    & 10   & 11   & 12   & 13   & 14   & 15   & 16   \\
    \hline\hline \TBstrut
    Hourglass \cite{newell:eccv2016} & \textbf{81.9} & --   & --   & --   & --   & --   & --   & --   & --   & --   & --   & --   & --   & --   & --   & --   \\ \hline \Tstrut
    NN-all                           & 63.5 & 43.2 & 33.8 & 29.1 & 26.9 & 25.8 & 24.8 & 24.5 & 24.4 & 24.5 & 24.7 & 25.0 & 25.5 & 26.0 & 26.5 & 26.5 \\
    NN-CaffeNet                      & 63.4 & 43.3 & 34.1 & 29.5 & 27.3 & 26.2 & 25.3 & 24.9 & 24.9 & 25.0 & 25.3 & 25.6 & 26.1 & 26.7 & 27.1 & 27.2 \\ \Bstrut
    NN-oracle                        & 63.4 & 44.1 & 35.5 & 31.2 & 29.1 & 28.0 & 27.0 & 26.5 & 26.5 & 26.8 & 27.3 & 27.6 & 28.1 & 28.8 & 29.2 & 29.3 \\ \hline \TBstrut
    3D-PFNet                         & 79.2          & \textbf{60.0} & \textbf{49.0} & \textbf{43.9} & \textbf{41.5} & \textbf{40.3} & \textbf{39.8} & \textbf{39.7}
                                     & \textbf{40.1} & \textbf{40.5} & \textbf{41.1} & \textbf{41.6} & \textbf{42.3} & \textbf{42.9} & \textbf{43.2} & \textbf{43.3} \\
    \hline
  \end{tabular}
  \vspace{-2mm}
  \caption{\small PCK values (\%) with threshold 0.5 (PCK@0.05) for timestep 1 to 16.}
  % \vspace{-2mm}
  \vspace{0mm}
  \label{tab:pck}
\end{table*}

\vspace{-3mm}

\paragraph{Implementation Details} We use Torch7 \cite{collobert:nipsw2011} for
our experiments. In all training, we use rmsprop for optimization. We train our
3D-PFNet in three steps as described in Sec.~\ref{sub:training}. First, we
train the hourglass for single-frame pose estimation by pre-training on MPII
and fine-tuning on the preprocessed Penn Action. For both datasets, we
partition a subset of the training set for validation. Second, we train the 3D
skeleton converter using Human3.6M. Note that the image data in Human3.6M are
unused here, since we only need 3D pose data for synthesizing camera parameters
and 2D heatmaps. Following the standard data split in \cite{ionescu:pami2014},
we use poses of 5 subjects (S1, S7, S8, S9, S11) for training and 2 subjects
(S5, S6) for validation. Fig.~\ref{fig:synthetic} shows samples of our
synthesized training data. We use mini-batches of size 64 and a learning rate
of 0.001. Finally, we train the full 3D-PFNet on the preprocessed Penn Action.
We apply the curriculum learning scheme until convergence at sequence length
16. At test time, we always generate pose sequences of length 16.

\vspace{-3mm}

\paragraph{Baselines} Since there are no prior approaches for pose forecasting,
we devise our own baselines for comparison. We consider three baselines based
on \textit{nearest neighbor (NN)}. (1) \textit{NN-all}: Given a test image, we
first estimate the current human pose with an hourglass network and find the NN
pose in the training images. We then transform the sequence of the NN pose to
the test image as output. To measure distance between two poses (each
represented by 13 2D keypoints), we first normalize the keypoints of each pose
to have zero mean and unit maximum length from the center. We define distance
by the MSE between two normalized poses. Since a ground-truth pose might
contain invisible keypoints, we compute MSE only on the visible keypoints.
Given the NN, we transform the associated sequence for the test image by
reversing the normalization. (2) \textit{NN-CaffeNet}: We hypothesize that the
NN results can be improved by leveraging scene contexts. We therefore
pre-filter the training set to keep only images with scene background similar
to the test image before applying NN-all. We compute the Euclidean distance on
the CaffeNet feature \cite{jia2014caffe}, and select the filtering threshold
using a validation set. (3) \textit{NN-oracle}: We exploit ground-truth action
labels to keep only the training images with the same action category as the
test image before applying NN-all. Note that this is a strong baseline since
our method does not use ground-truth action labels.

\begin{figure*}[t]
  % \vspace{-2mm}
  \centering
  \footnotesize
  \begin{tabular}{L{0.12\linewidth}@{\hspace{1.0mm}}|L{0.9\linewidth}@{\hspace{-0.0mm}}}
    \includegraphics[height=0.182\textwidth]{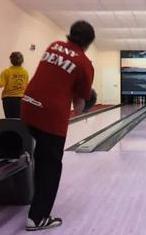}
    &
    \vspace{-2.7mm}
    \begin{tabular}{L{1.0\linewidth}}
      \hspace{-2.7mm}
      \includegraphics[height=0.10\textwidth]{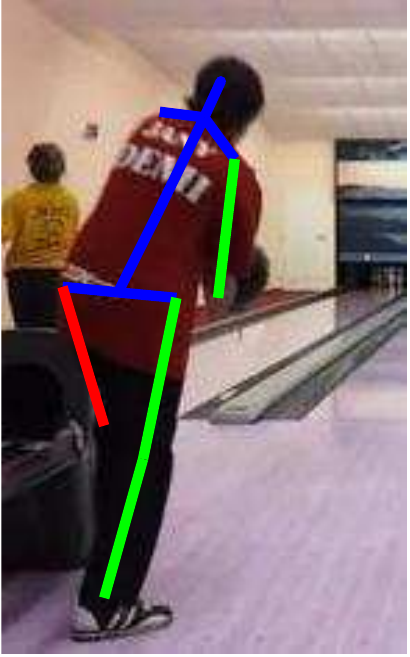}\hspace{0.9mm}
      \includegraphics[height=0.10\textwidth]{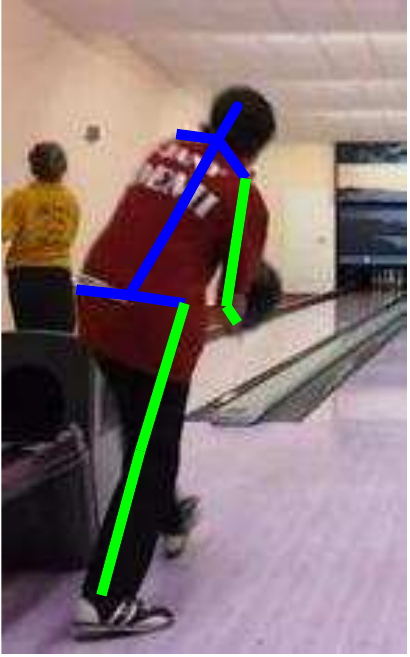}\hspace{0.9mm}
      \includegraphics[height=0.10\textwidth]{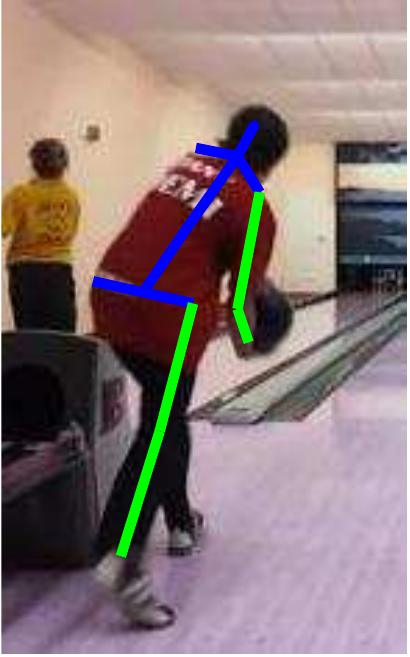}\hspace{0.9mm}
      \includegraphics[height=0.10\textwidth]{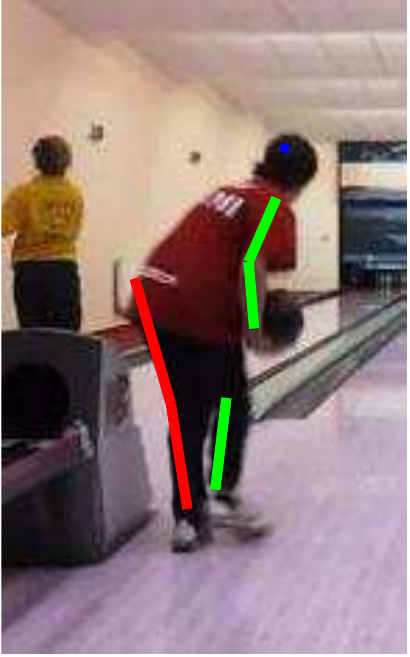}\hspace{0.9mm}
      \includegraphics[height=0.10\textwidth]{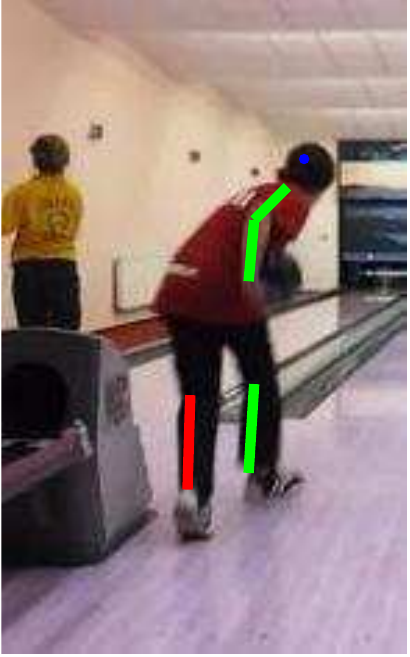}\hspace{0.9mm}
      \includegraphics[height=0.10\textwidth]{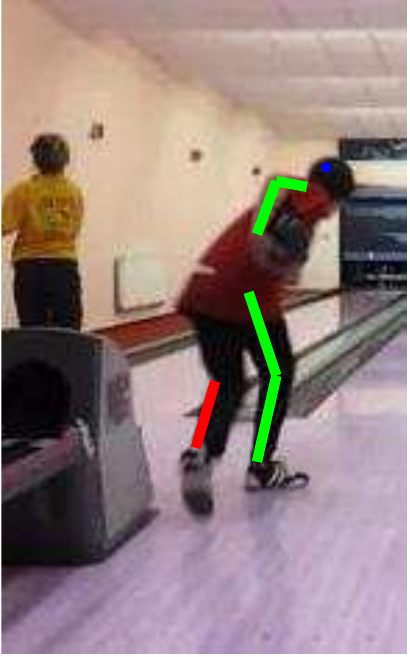}\hspace{0.9mm}
      \includegraphics[height=0.10\textwidth]{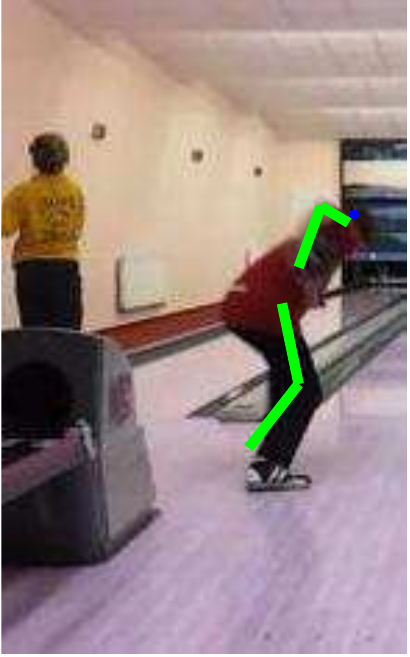}\hspace{0.9mm}
      \includegraphics[height=0.10\textwidth]{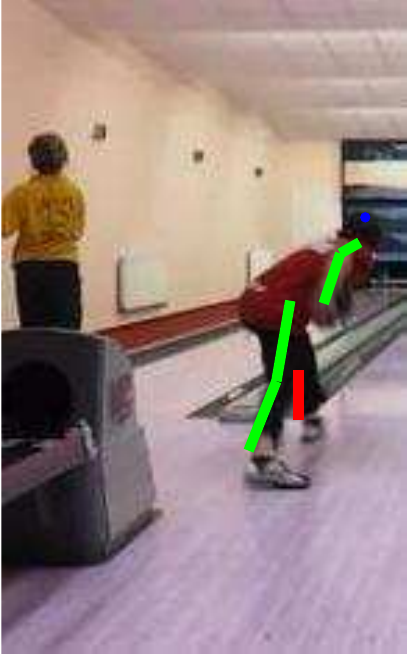}\hspace{0.9mm}
      \includegraphics[height=0.10\textwidth]{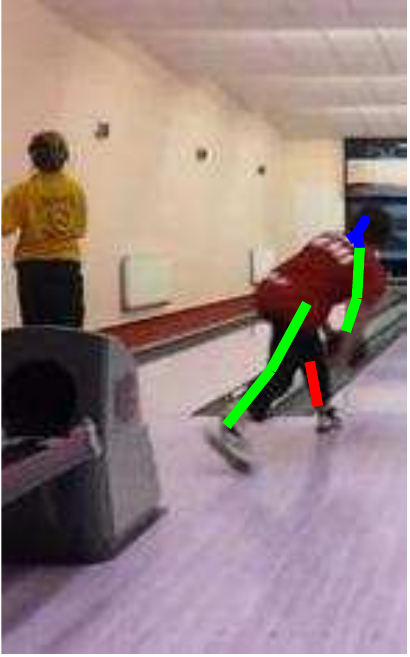}\hspace{0.9mm}
      \includegraphics[height=0.10\textwidth]{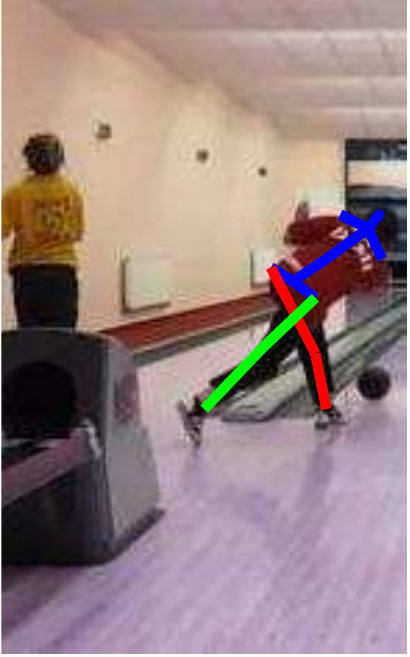}\hspace{0.9mm}
      \includegraphics[height=0.10\textwidth]{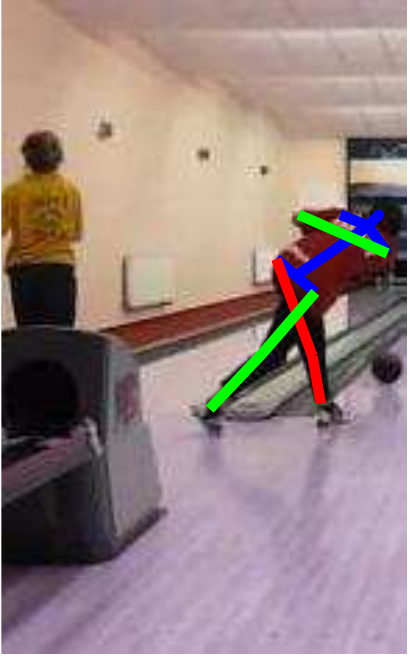}\hspace{0.9mm}
      \includegraphics[height=0.10\textwidth]{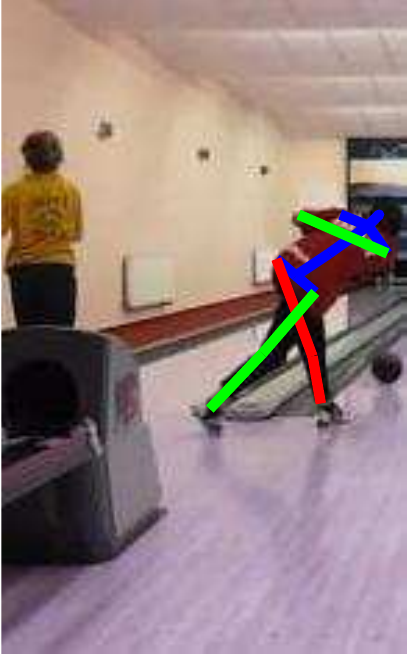}
      \\
      \hspace{-2.7mm}
      \includegraphics[height=0.10\textwidth]{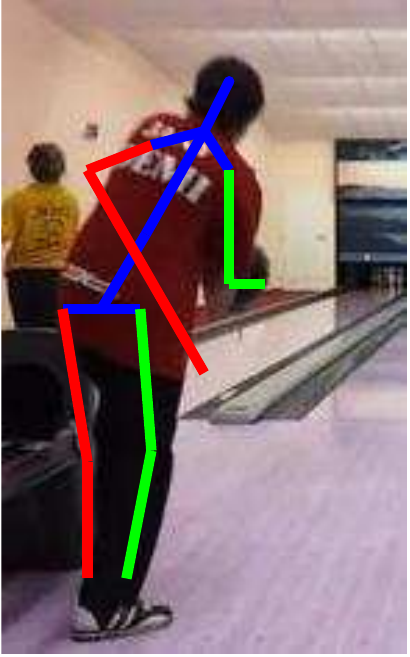}\hspace{0.9mm}
      \includegraphics[height=0.10\textwidth]{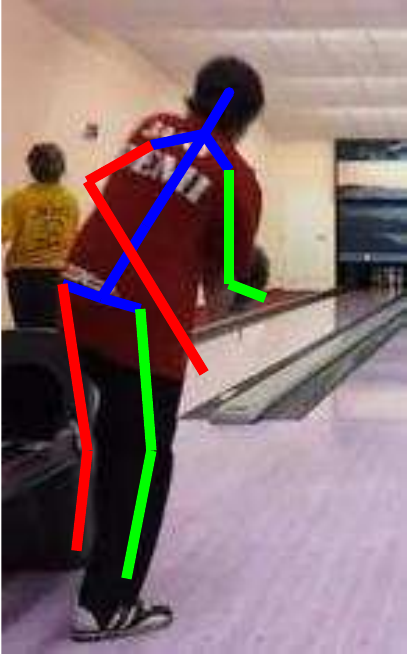}\hspace{0.9mm}
      \includegraphics[height=0.10\textwidth]{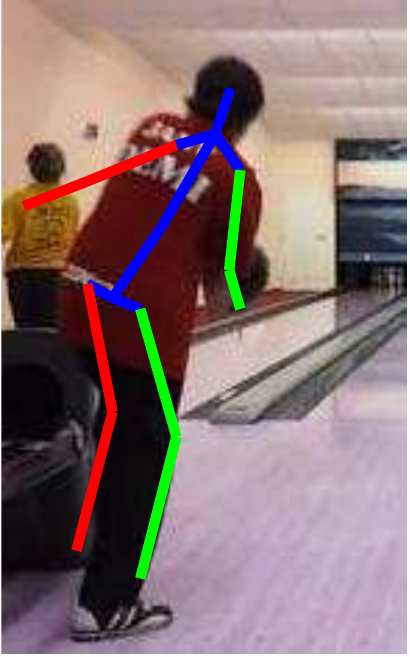}\hspace{0.9mm}
      \includegraphics[height=0.10\textwidth]{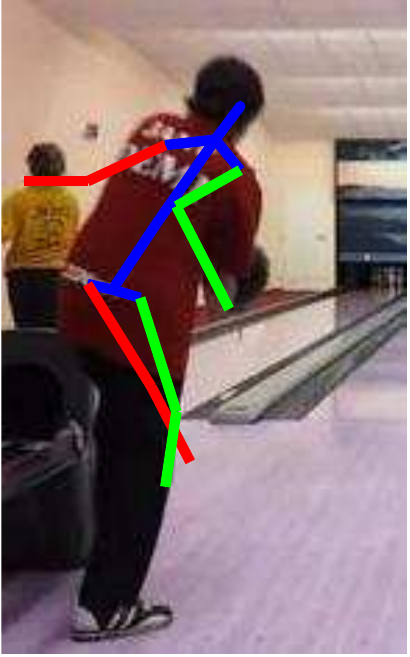}\hspace{0.9mm}
      \includegraphics[height=0.10\textwidth]{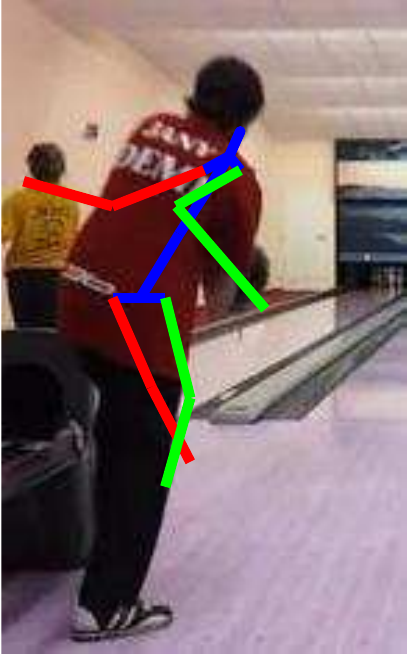}\hspace{0.9mm}
      \includegraphics[height=0.10\textwidth]{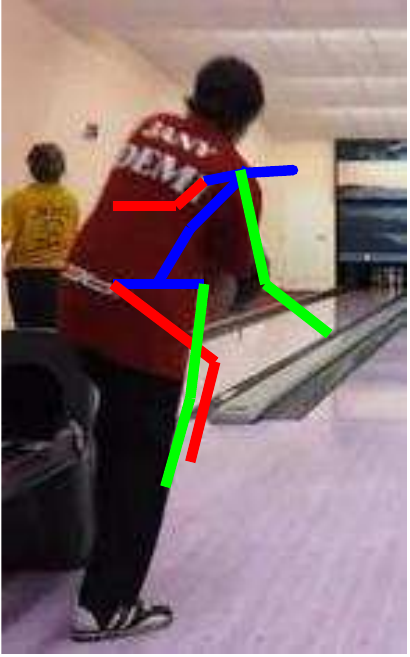}\hspace{0.9mm}
      \includegraphics[height=0.10\textwidth]{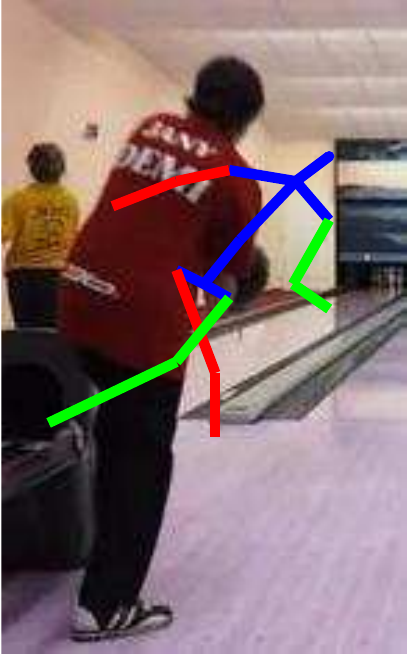}\hspace{0.9mm}
      \includegraphics[height=0.10\textwidth]{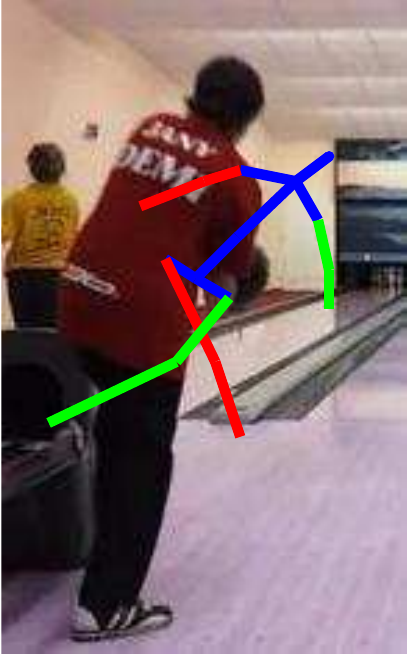}\hspace{0.9mm}
      \includegraphics[height=0.10\textwidth]{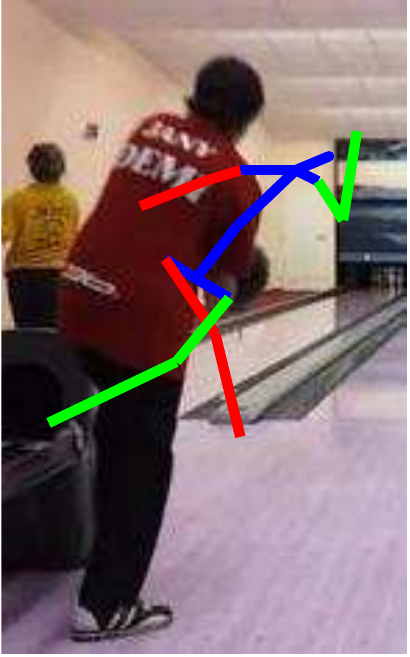}\hspace{0.9mm}
      \includegraphics[height=0.10\textwidth]{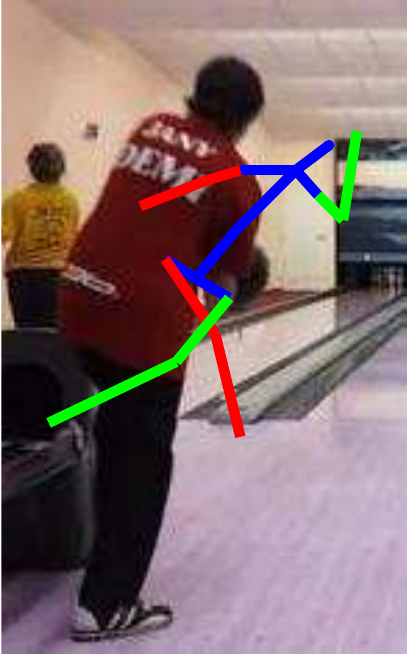}\hspace{0.9mm}
      \includegraphics[height=0.10\textwidth]{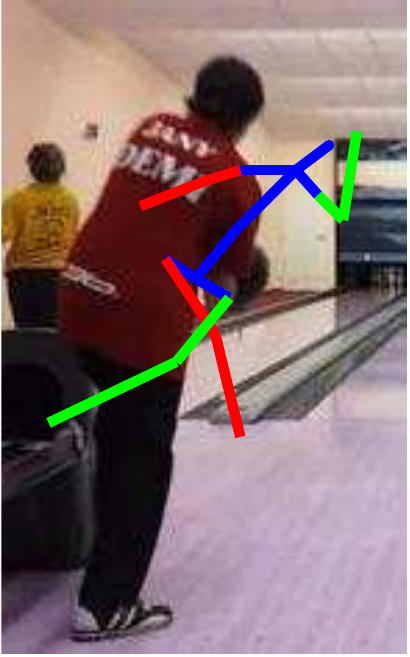}\hspace{0.9mm}
      \includegraphics[height=0.10\textwidth]{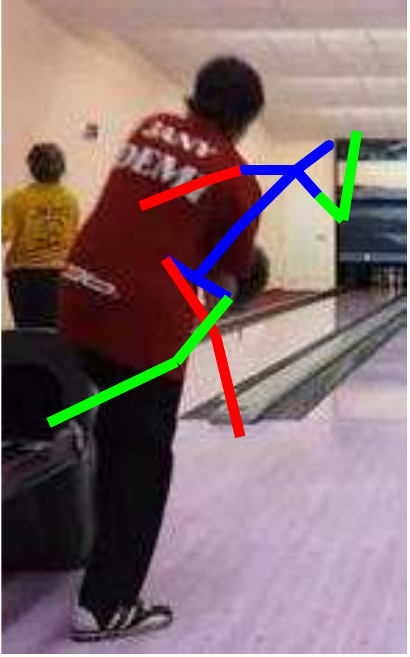}
      \\
      \includegraphics[height=0.10\textwidth]{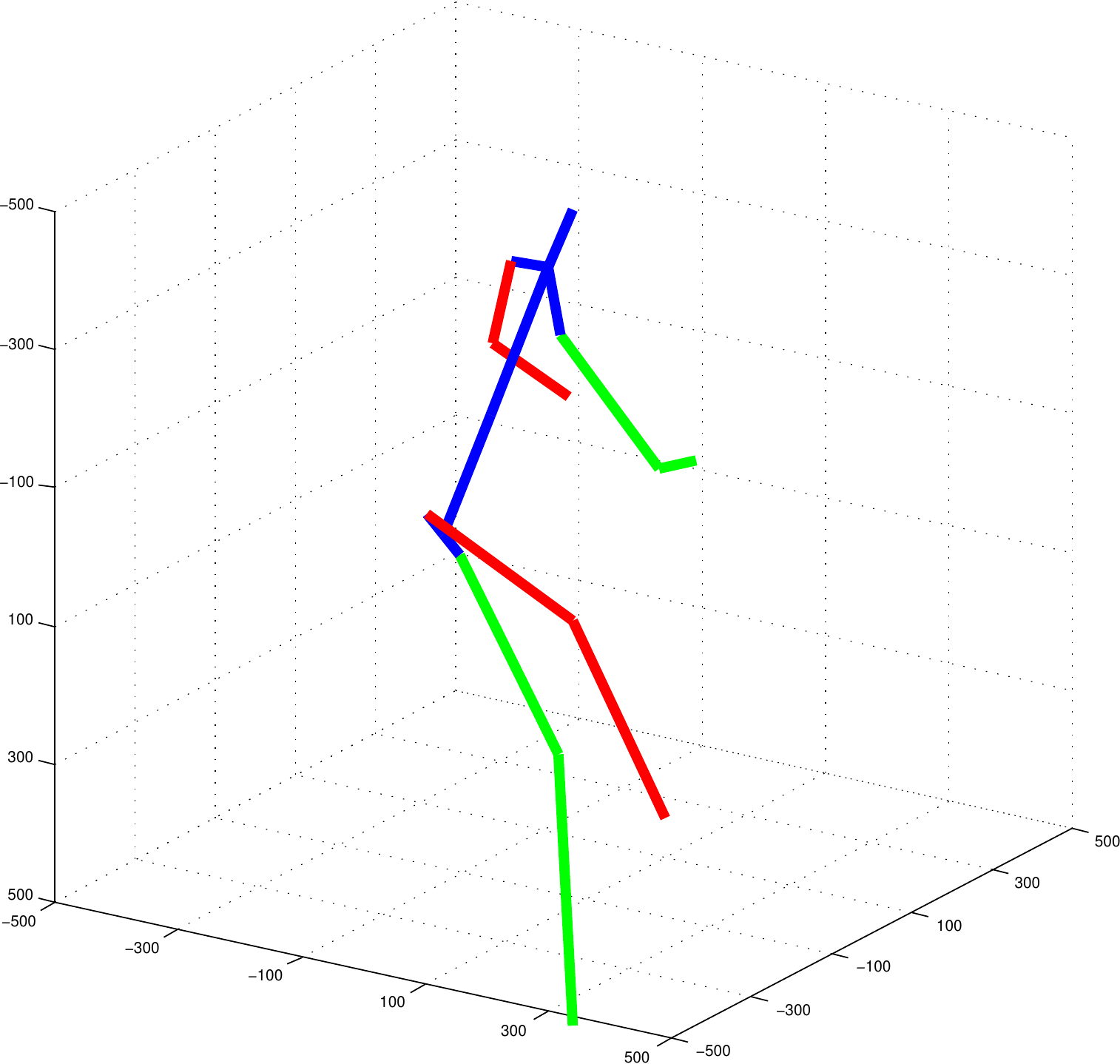}\hspace{1.9mm}
      \includegraphics[height=0.10\textwidth]{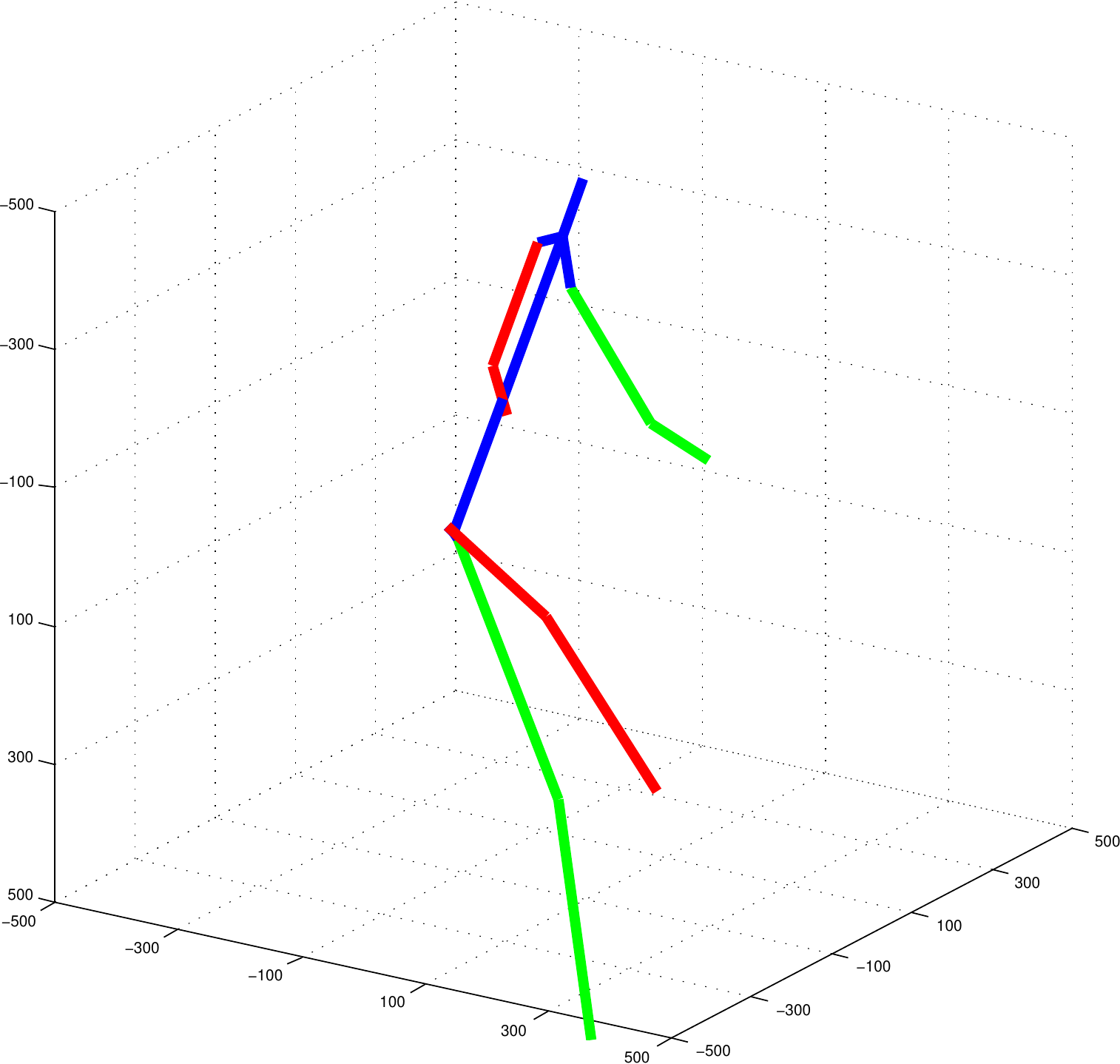}\hspace{1.9mm}
      \includegraphics[height=0.10\textwidth]{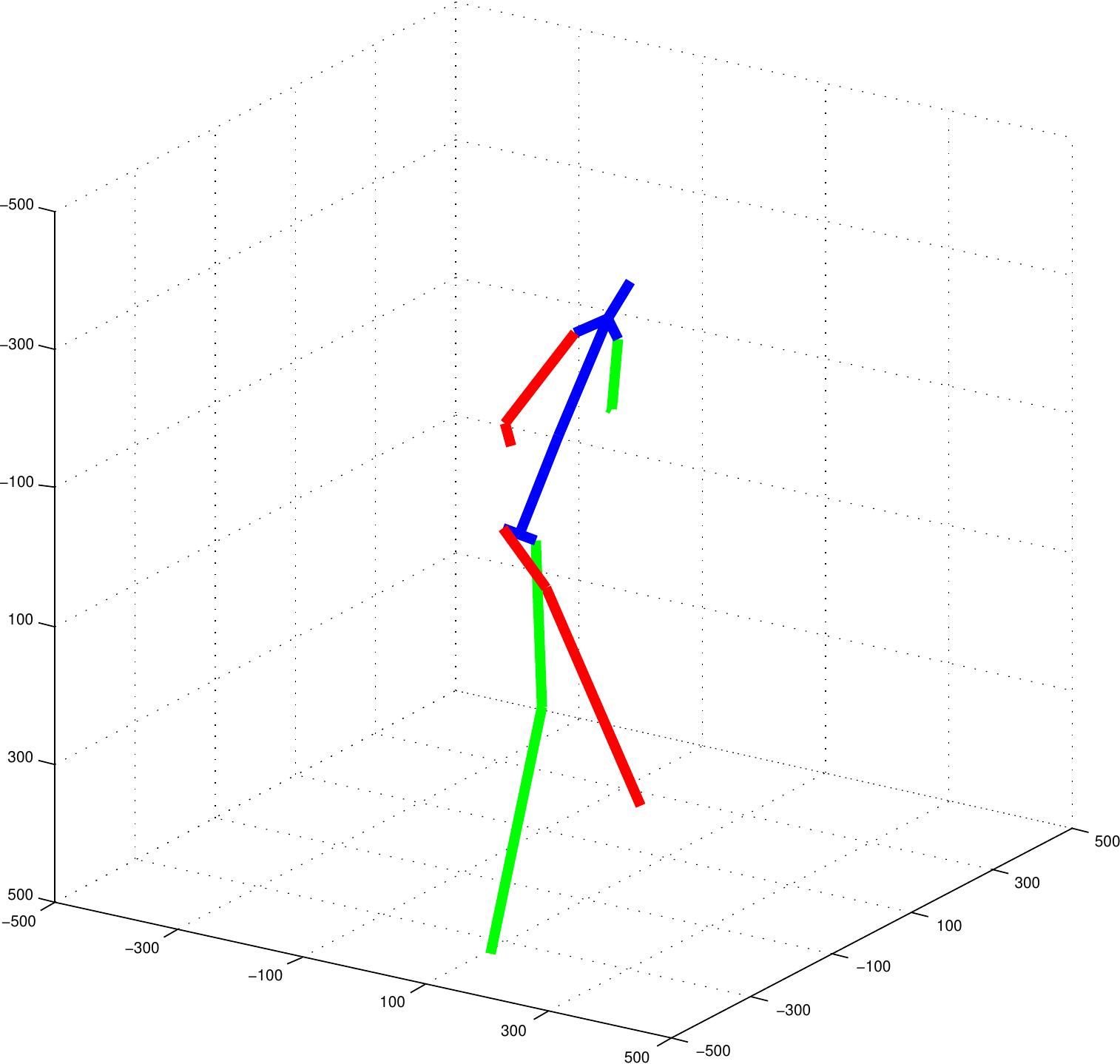}\hspace{1.9mm}
      \includegraphics[height=0.10\textwidth]{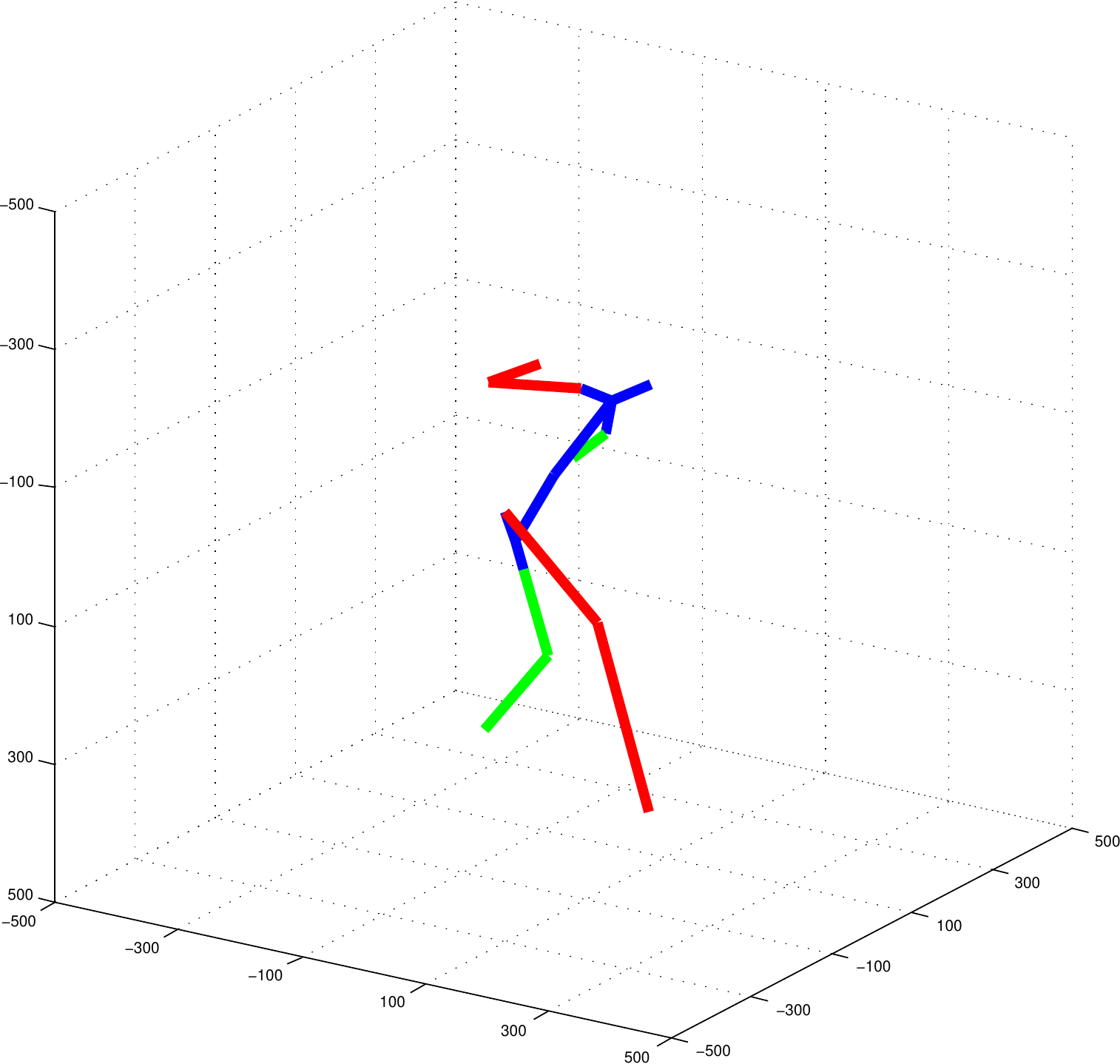}\hspace{1.9mm}
      \includegraphics[height=0.10\textwidth]{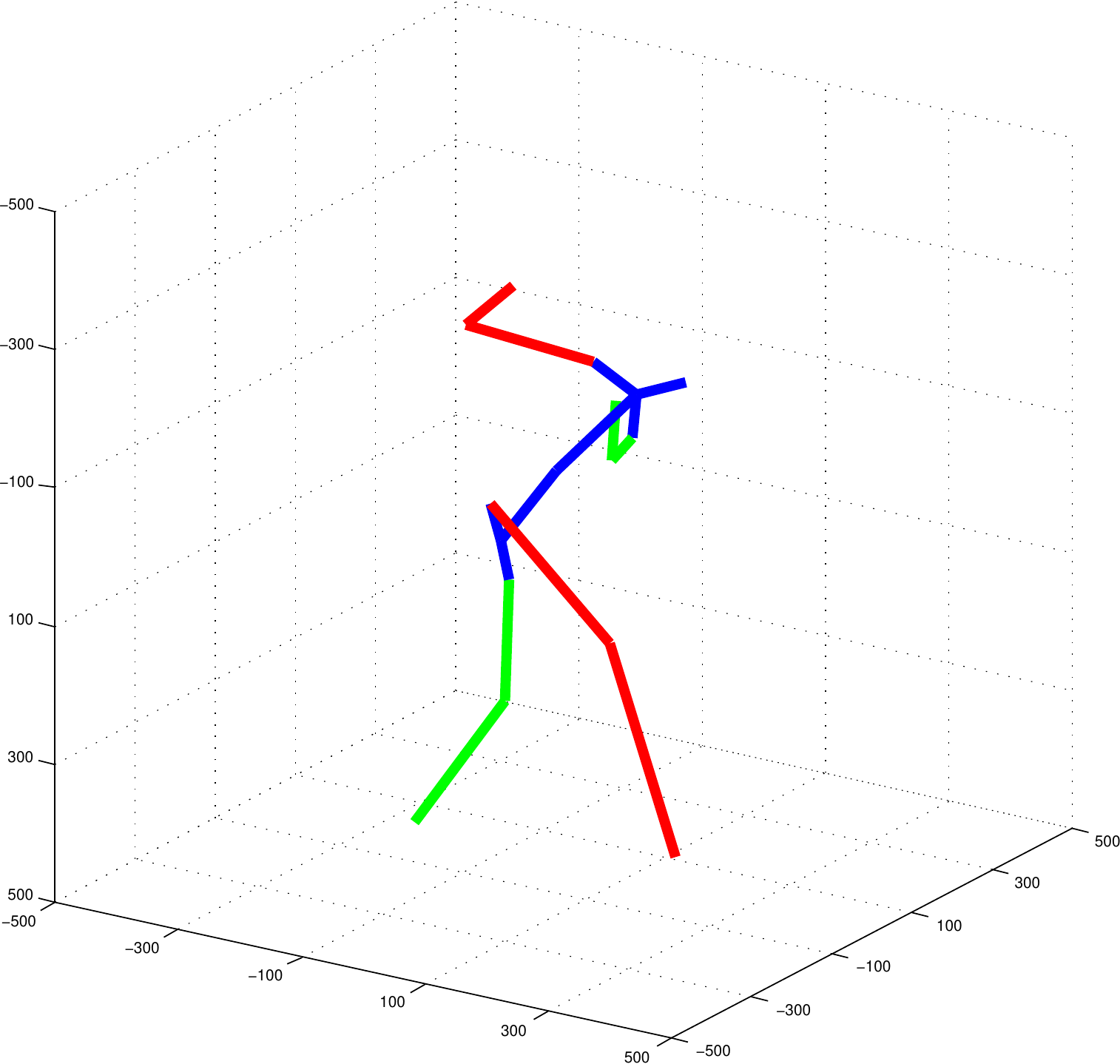}\hspace{1.9mm}
      \includegraphics[height=0.10\textwidth]{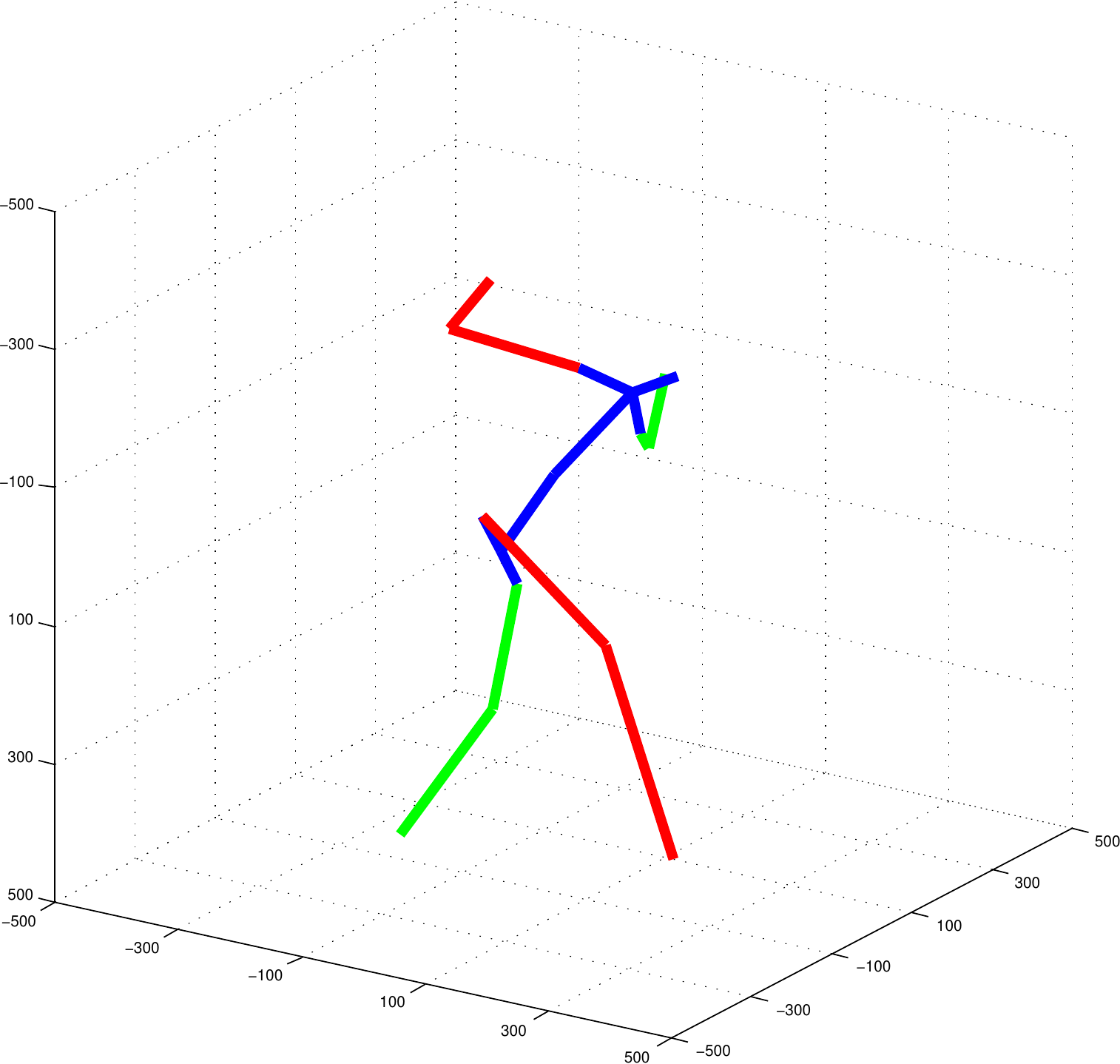}\hspace{1.9mm}
      \includegraphics[height=0.10\textwidth]{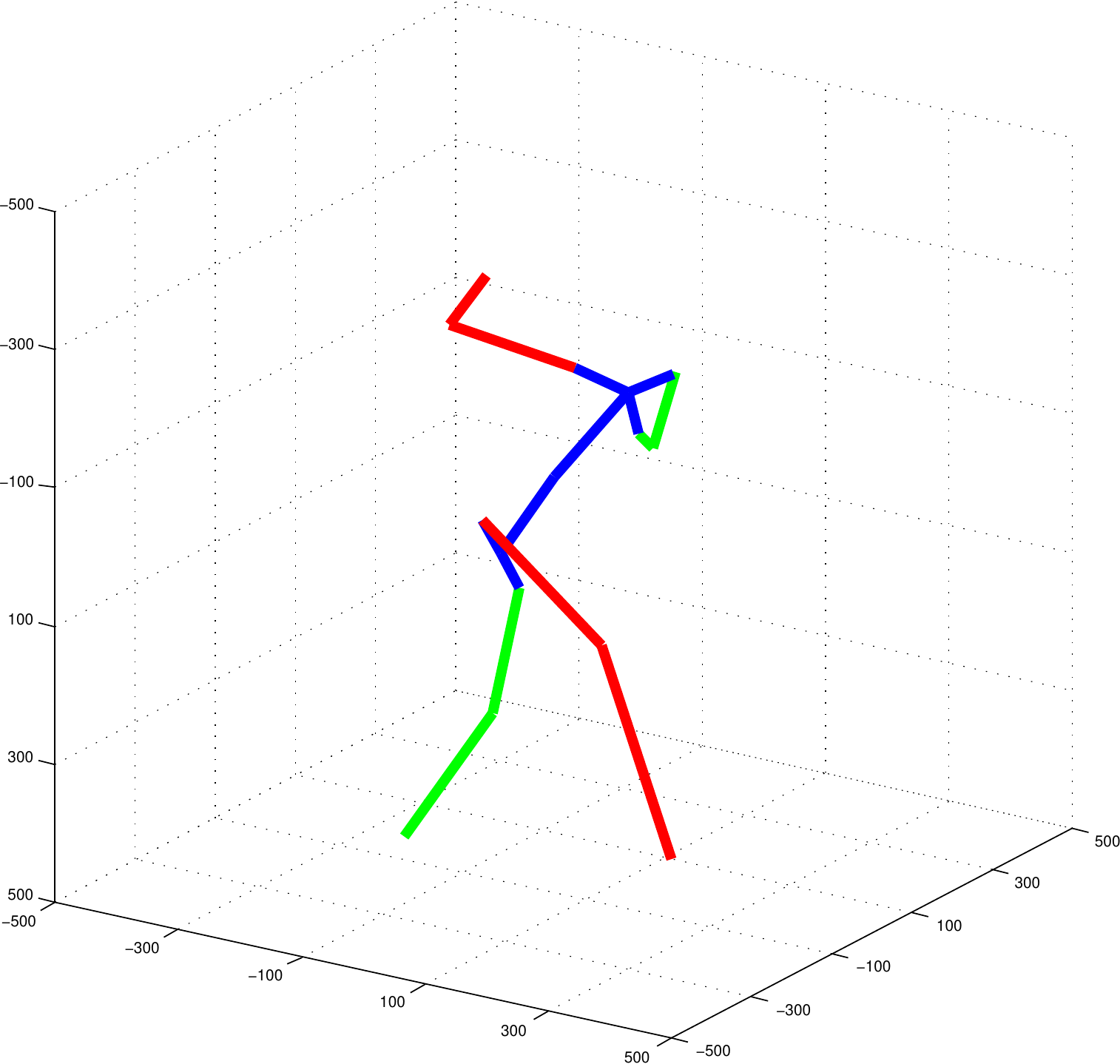}
    \end{tabular}
    \\ [-0.5em] & \\
    \includegraphics[height=0.232\textwidth]{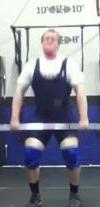}
    &
    \vspace{-2.7mm}
    \begin{tabular}{L{1.0\linewidth}}
      \hspace{-2.7mm}
      \includegraphics[height=0.10\textwidth]{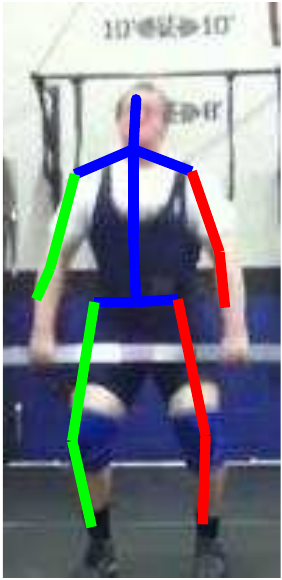}\hspace{0.82mm}
      \includegraphics[height=0.10\textwidth]{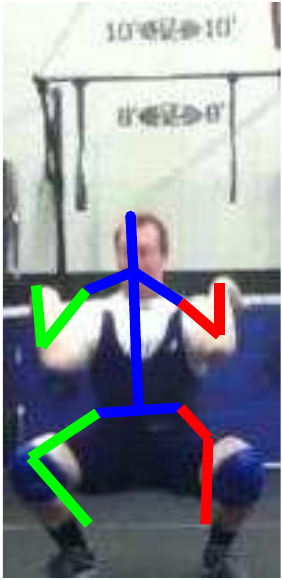}\hspace{0.82mm}
      \includegraphics[height=0.10\textwidth]{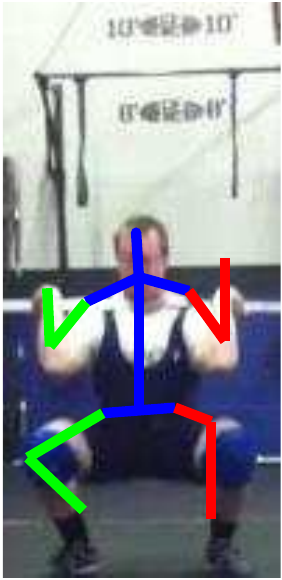}\hspace{0.82mm}
      \includegraphics[height=0.10\textwidth]{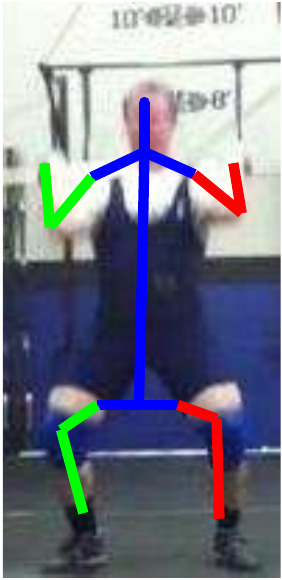}\hspace{0.82mm}
      \includegraphics[height=0.10\textwidth]{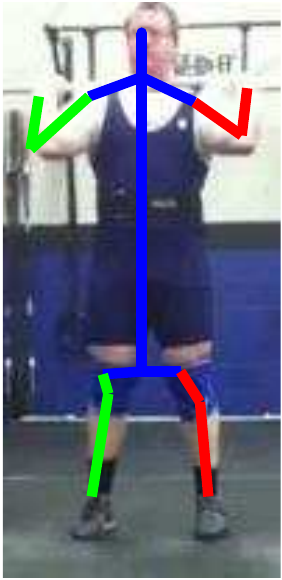}\hspace{0.82mm}
      \includegraphics[height=0.10\textwidth]{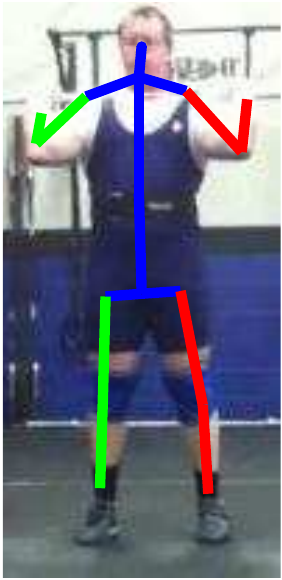}\hspace{0.82mm}
      \includegraphics[height=0.10\textwidth]{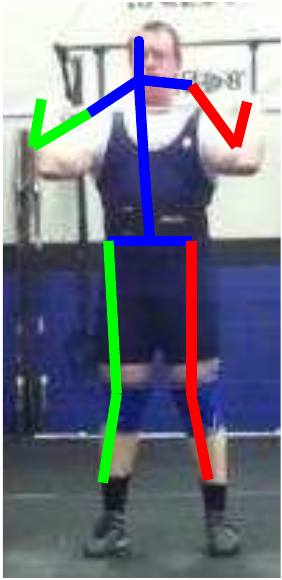}\hspace{0.82mm}
      \includegraphics[height=0.10\textwidth]{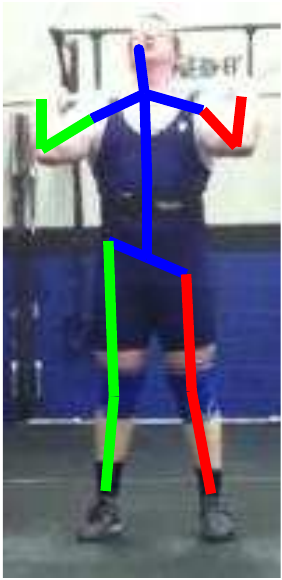}\hspace{0.82mm}
      \includegraphics[height=0.10\textwidth]{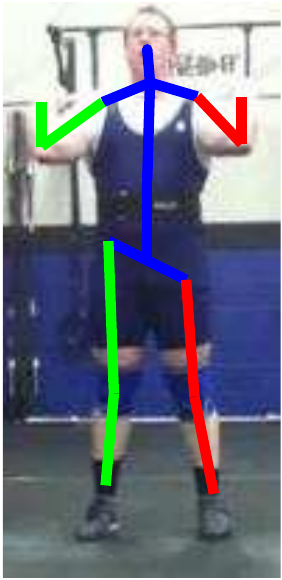}\hspace{0.82mm}
      \includegraphics[height=0.10\textwidth]{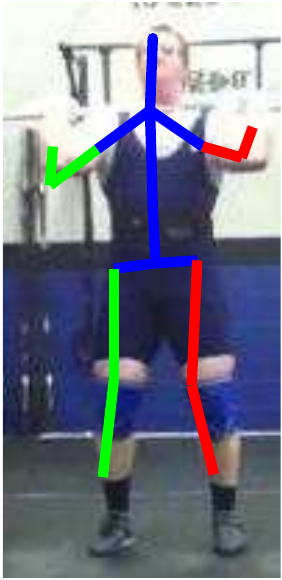}\hspace{0.82mm}
      \includegraphics[height=0.10\textwidth]{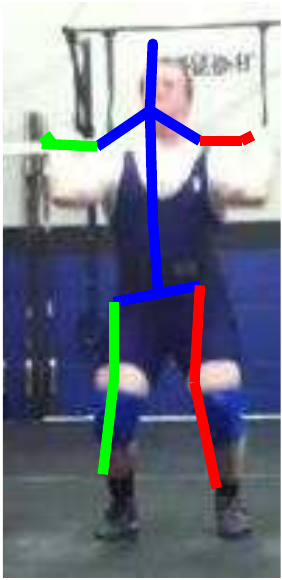}\hspace{0.82mm}
      \includegraphics[height=0.10\textwidth]{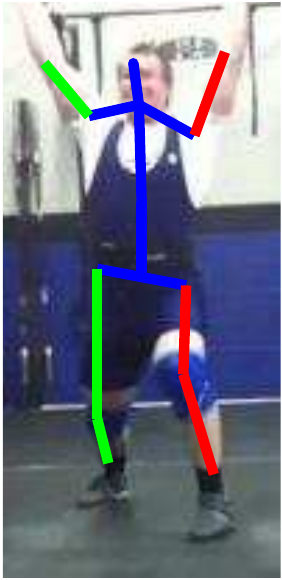}\hspace{0.82mm}
      \includegraphics[height=0.10\textwidth]{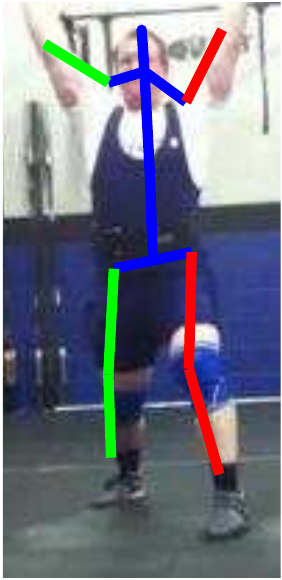}\hspace{0.82mm}
      \includegraphics[height=0.10\textwidth]{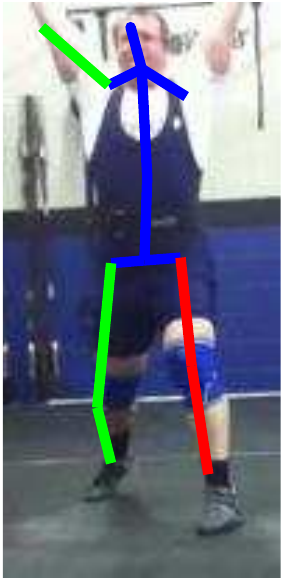}\hspace{0.82mm}
      \includegraphics[height=0.10\textwidth]{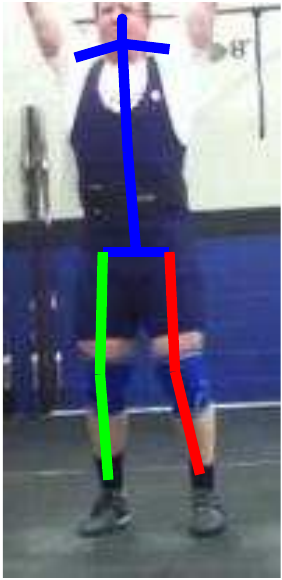}
      \\
      \hspace{-2.7mm}
      \includegraphics[height=0.10\textwidth]{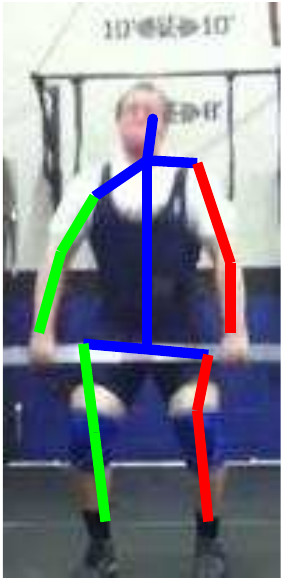}\hspace{0.82mm}
      \includegraphics[height=0.10\textwidth]{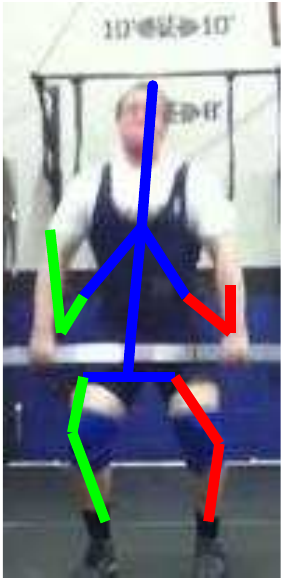}\hspace{0.82mm}
      \includegraphics[height=0.10\textwidth]{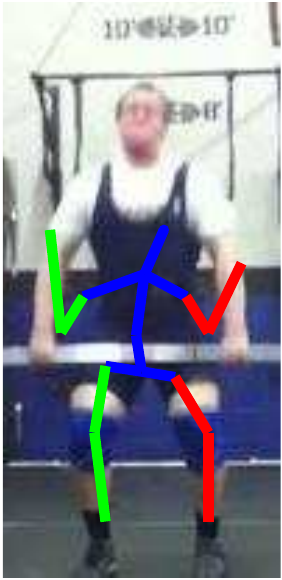}\hspace{0.82mm}
      \includegraphics[height=0.10\textwidth]{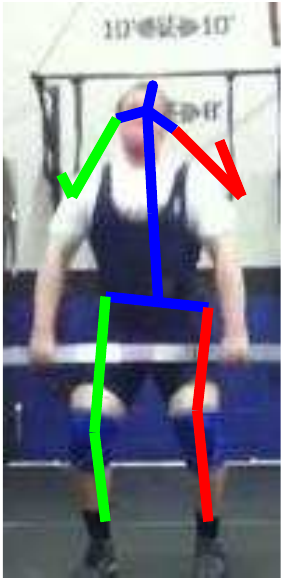}\hspace{0.82mm}
      \includegraphics[height=0.10\textwidth]{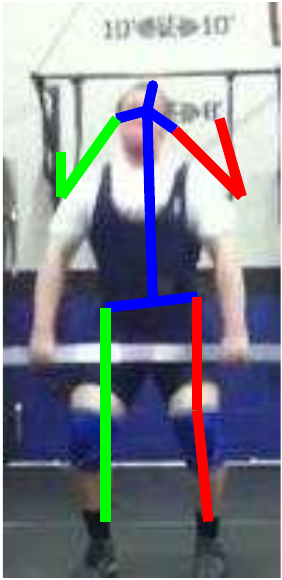}\hspace{0.82mm}
      \includegraphics[height=0.10\textwidth]{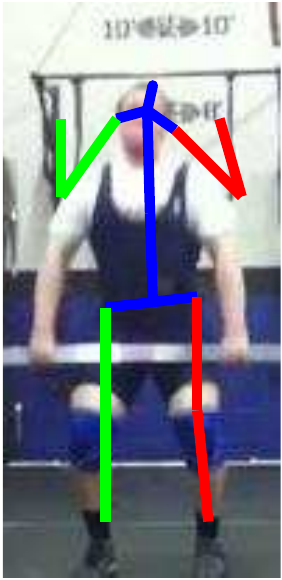}\hspace{0.82mm}
      \includegraphics[height=0.10\textwidth]{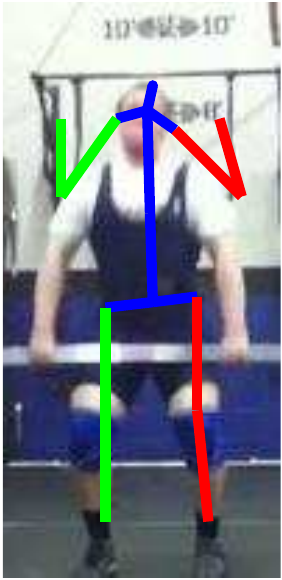}\hspace{0.82mm}
      \includegraphics[height=0.10\textwidth]{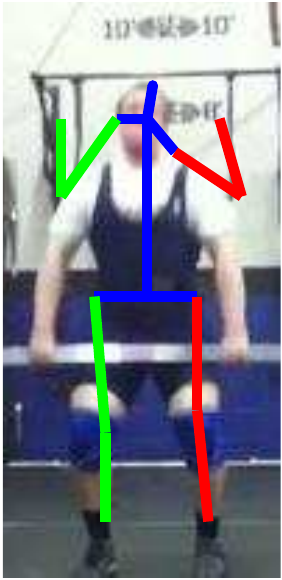}\hspace{0.82mm}
      \includegraphics[height=0.10\textwidth]{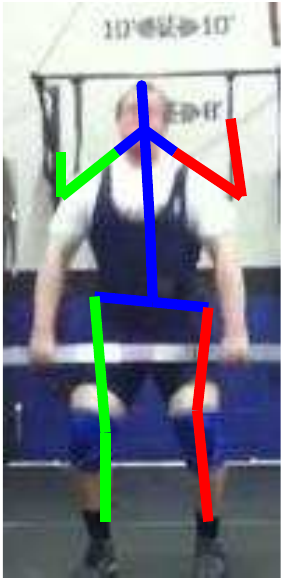}\hspace{0.82mm}
      \includegraphics[height=0.10\textwidth]{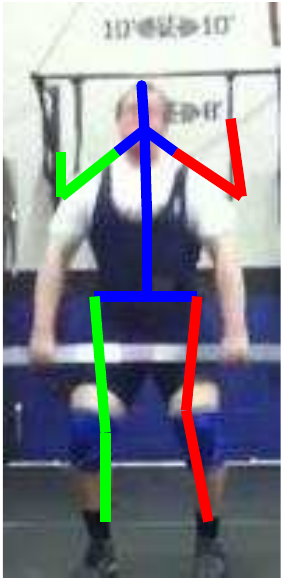}\hspace{0.82mm}
      \includegraphics[height=0.10\textwidth]{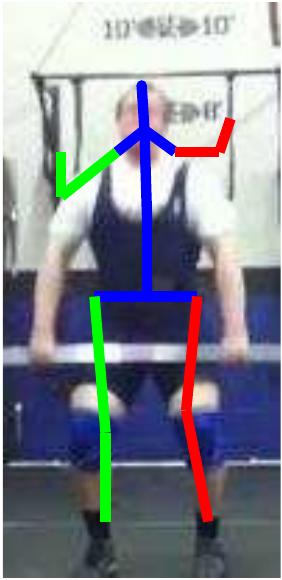}\hspace{0.82mm}
      \includegraphics[height=0.10\textwidth]{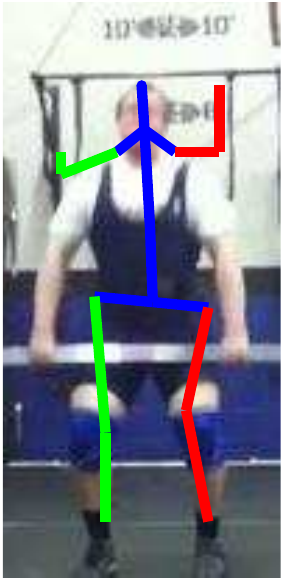}\hspace{0.82mm}
      \includegraphics[height=0.10\textwidth]{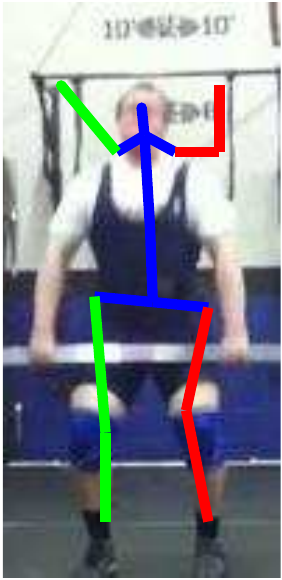}\hspace{0.82mm}
      \includegraphics[height=0.10\textwidth]{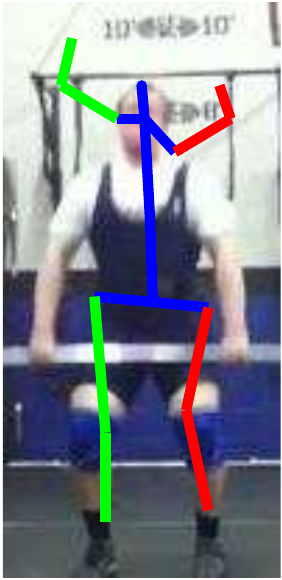}\hspace{0.82mm}
      \includegraphics[height=0.10\textwidth]{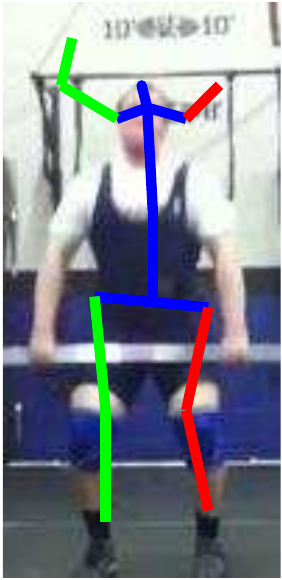}
      \\
      \includegraphics[height=0.10\textwidth]{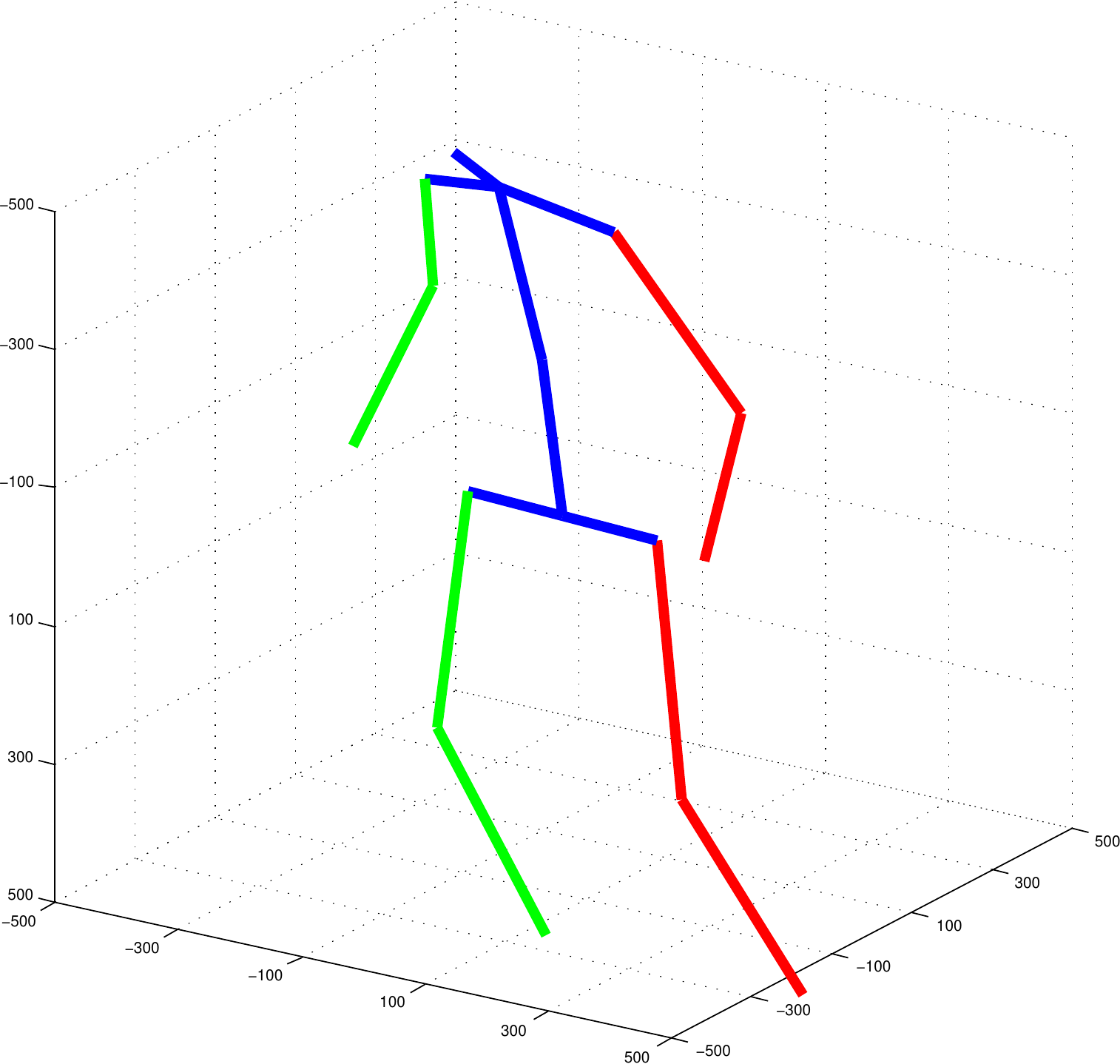}\hspace{1.9mm}
      \includegraphics[height=0.10\textwidth]{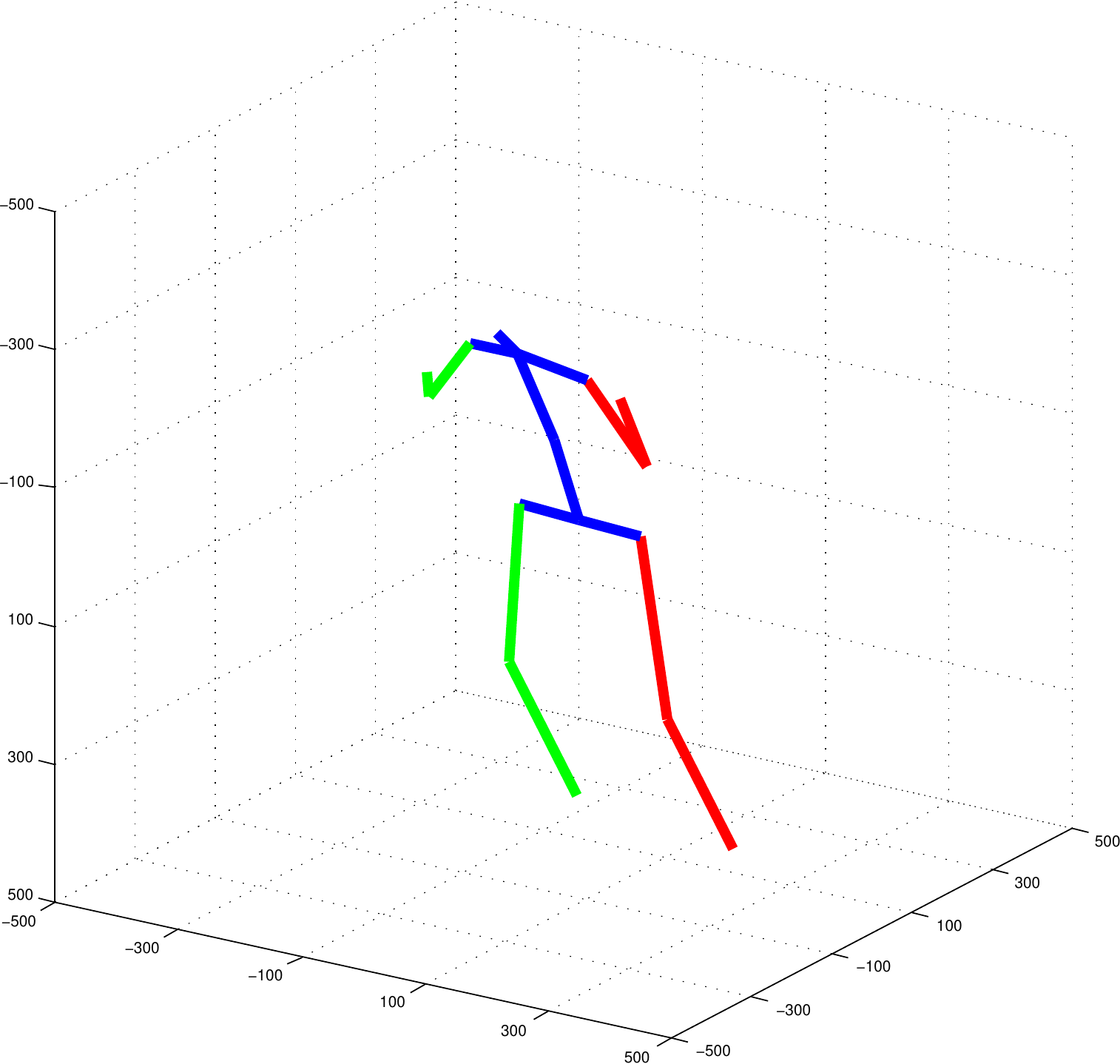}\hspace{1.9mm}
      \includegraphics[height=0.10\textwidth]{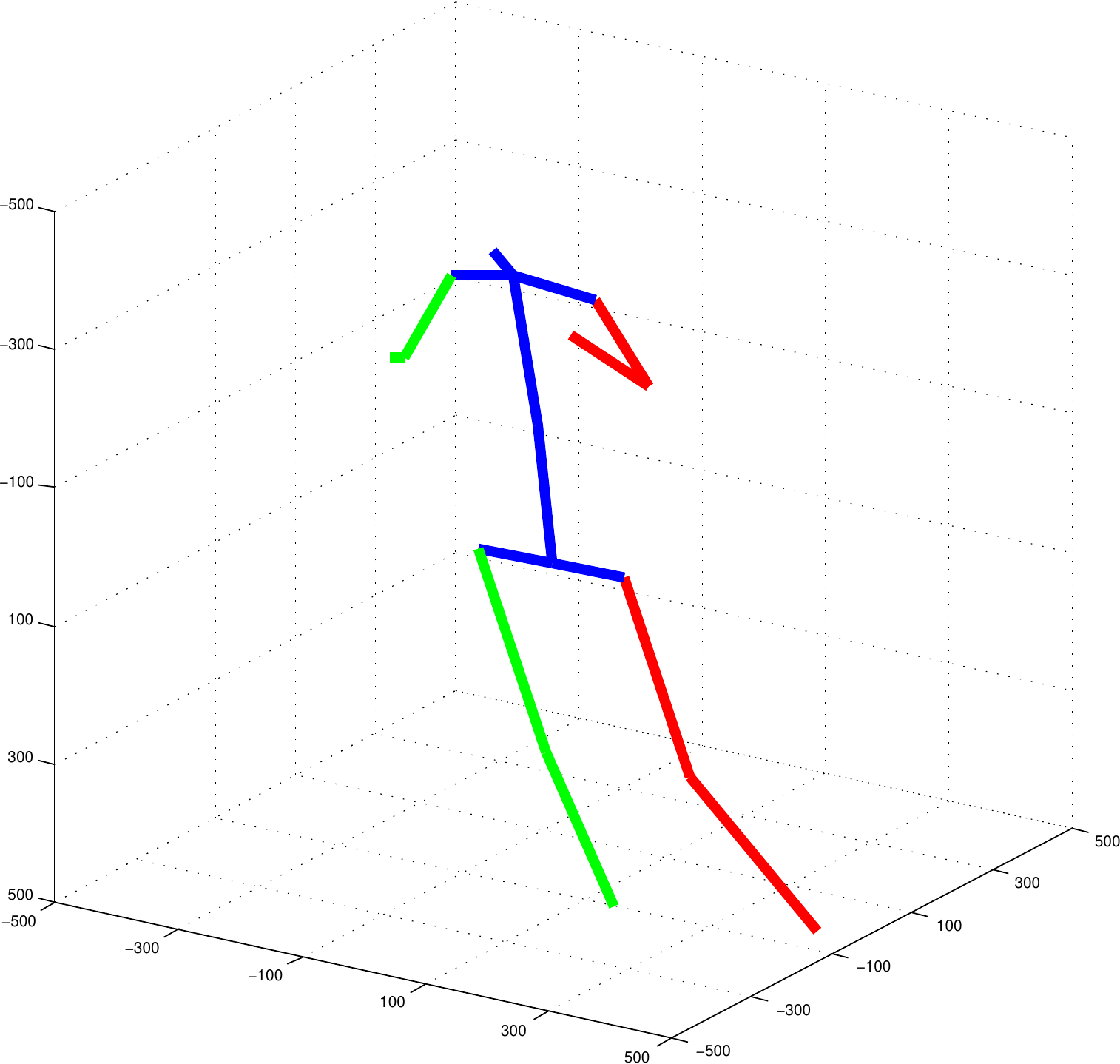}\hspace{1.9mm}
      \includegraphics[height=0.10\textwidth]{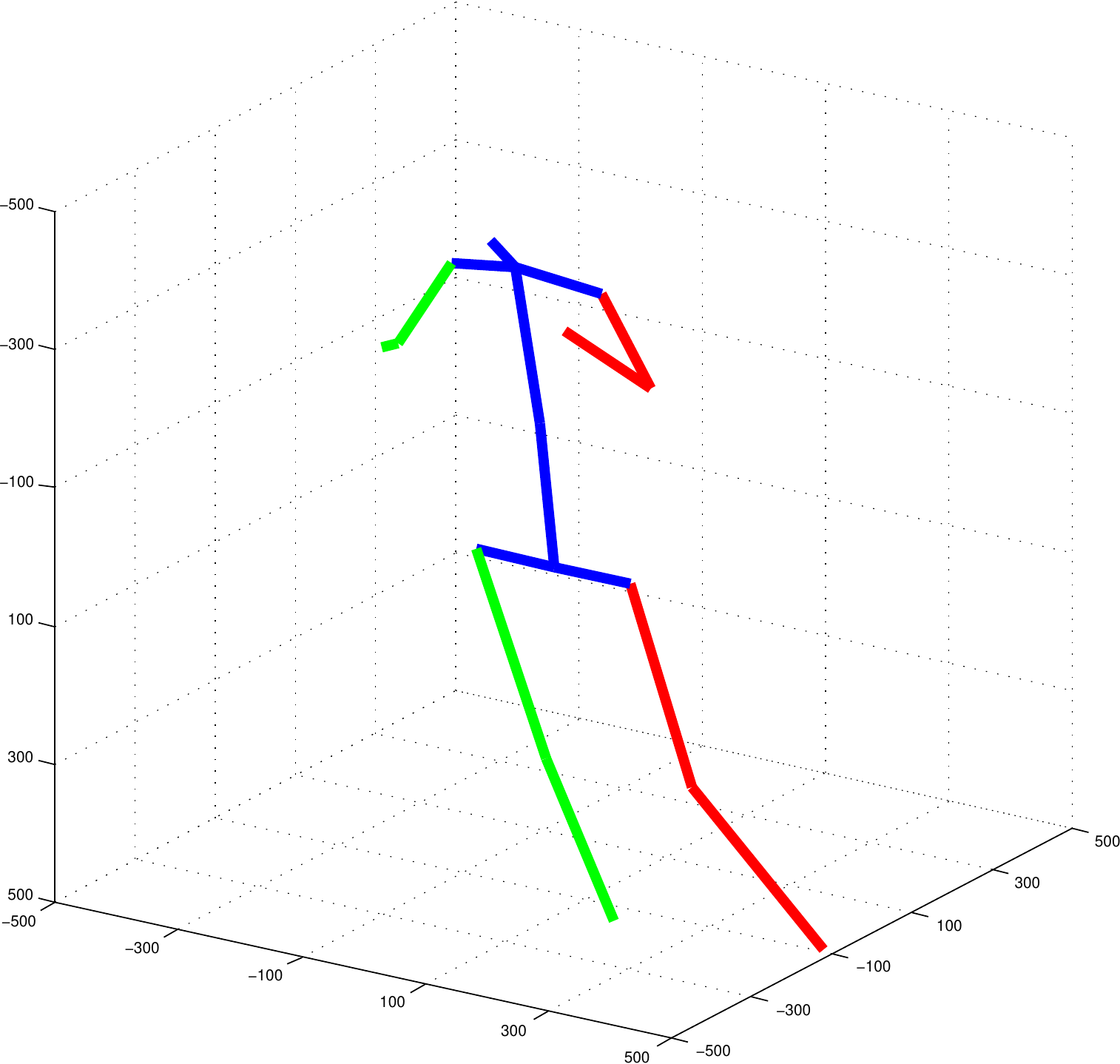}\hspace{1.9mm}
      \includegraphics[height=0.10\textwidth]{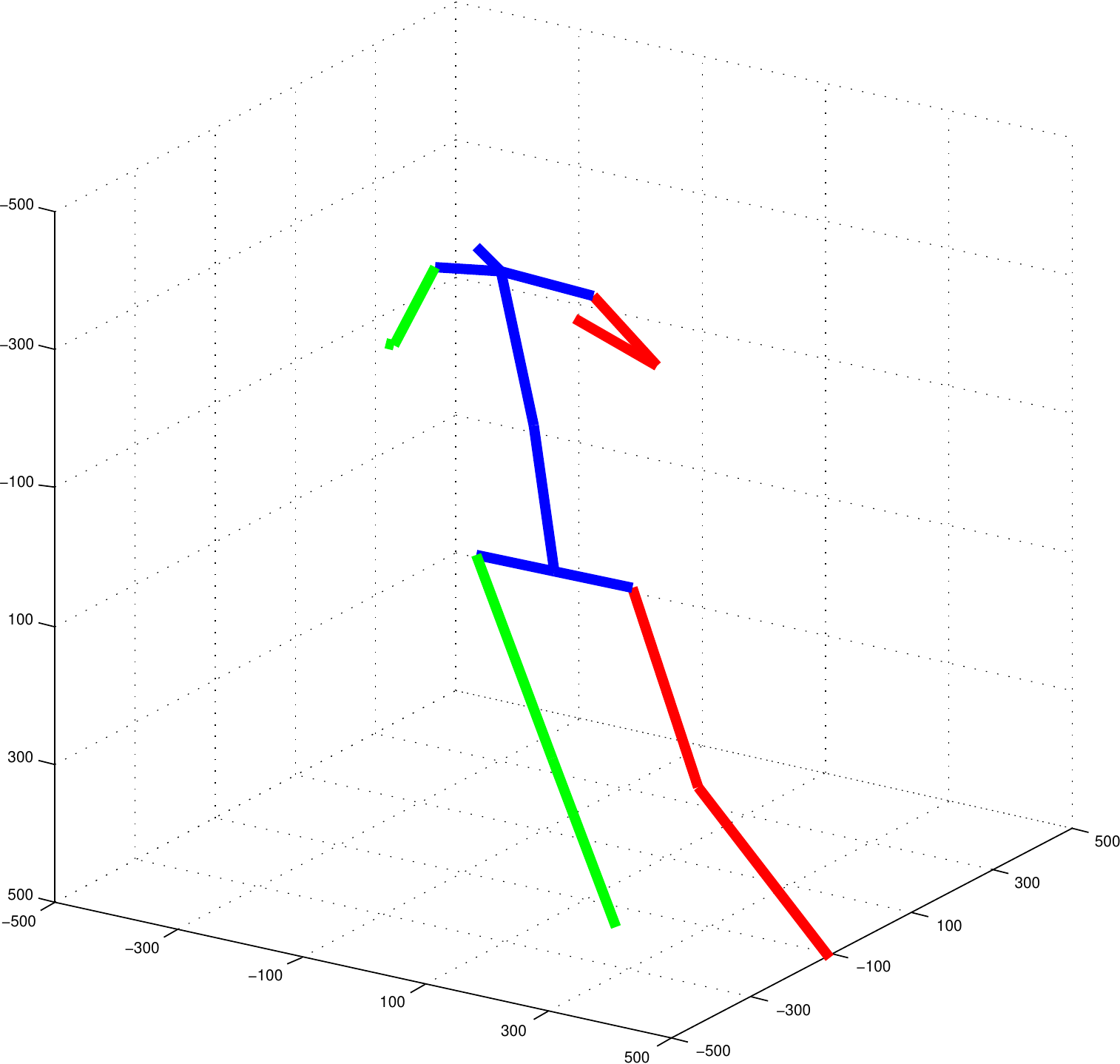}\hspace{1.9mm}
      \includegraphics[height=0.10\textwidth]{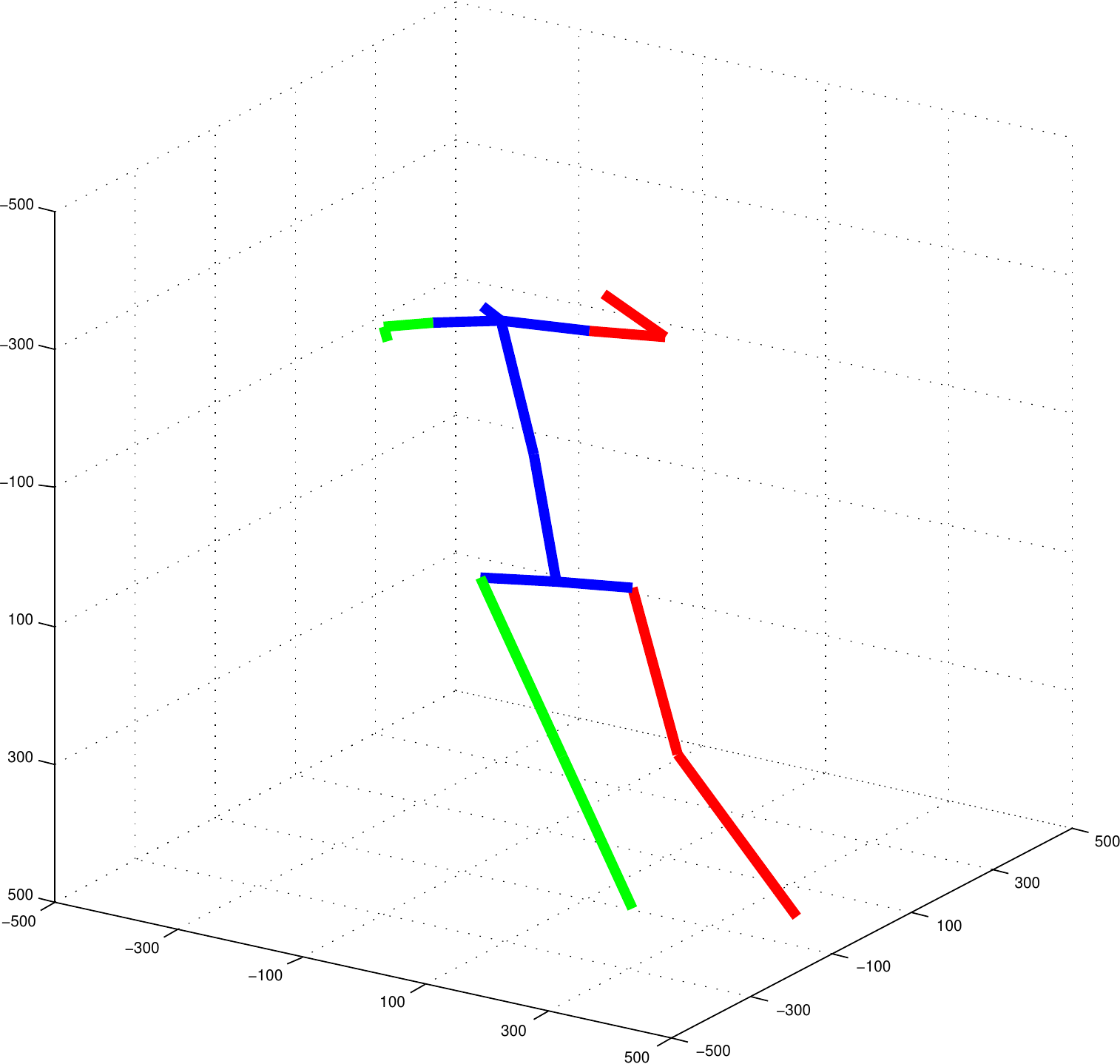}\hspace{1.9mm}
      \includegraphics[height=0.10\textwidth]{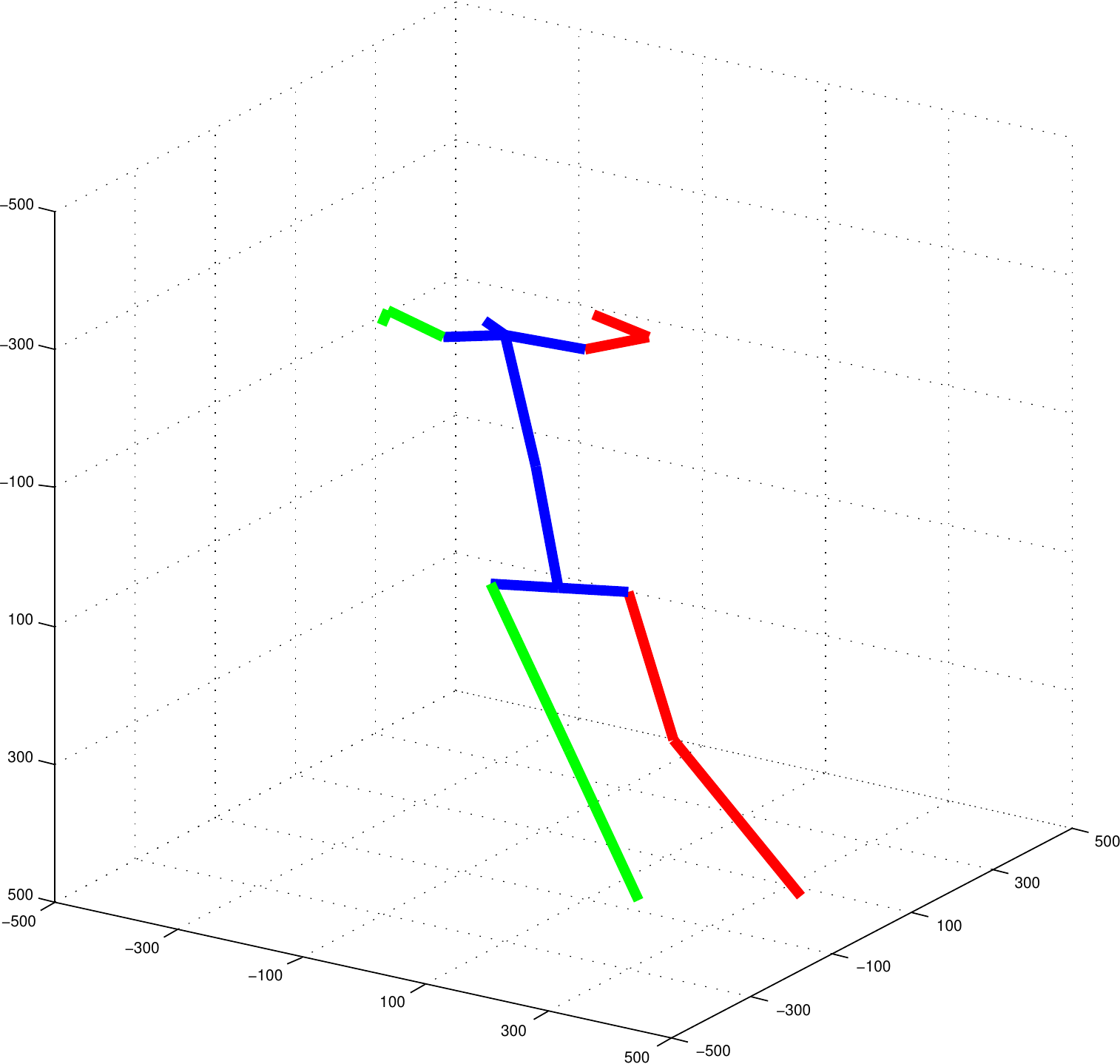}
    \end{tabular}
    \\ [-0.5em] & \\
    \includegraphics[height=0.057\textwidth]{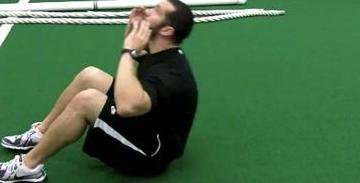}
    &
    \vspace{-2.7mm}
    \begin{tabular}{L{1.0\linewidth}}
       \hspace{-2.7mm}
      \includegraphics[height=0.07\textwidth]{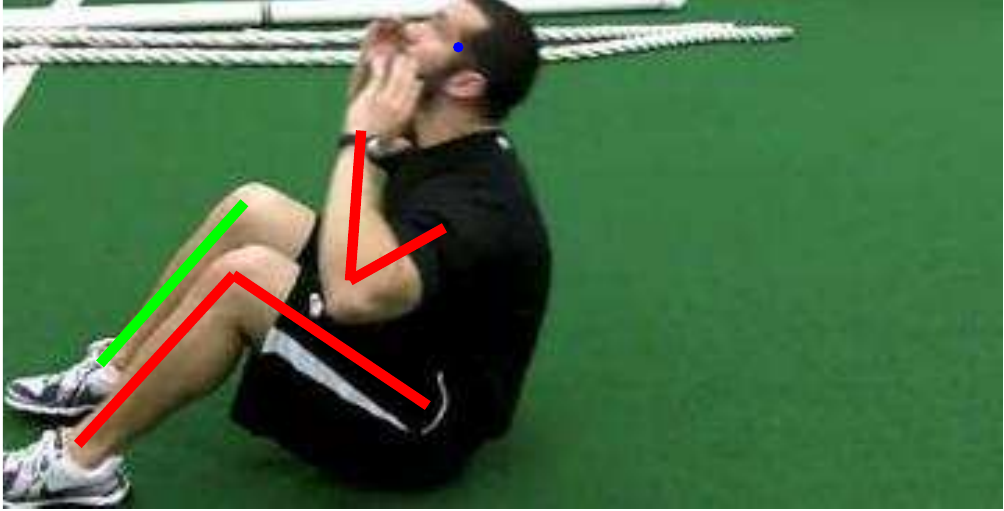}\hspace{0.15mm}
      \includegraphics[height=0.07\textwidth]{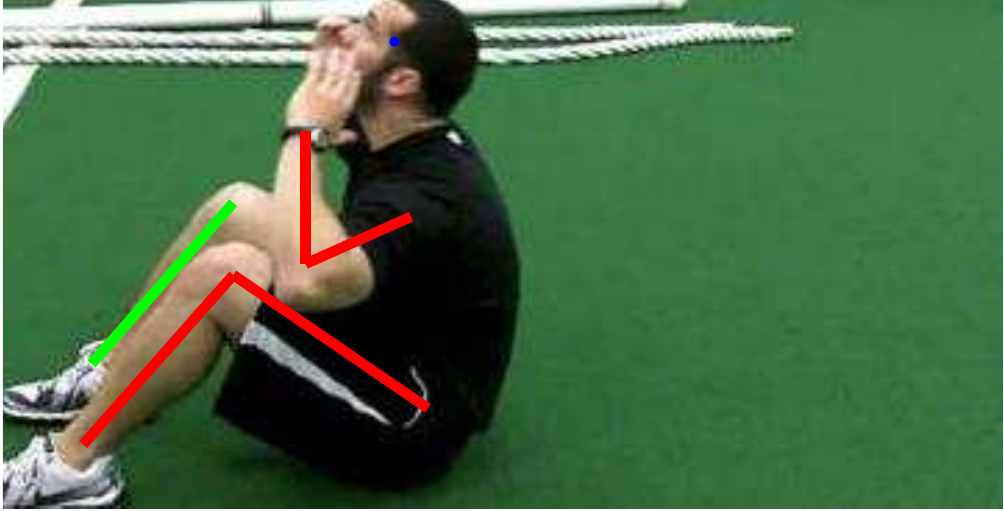}\hspace{0.15mm}
      \includegraphics[height=0.07\textwidth]{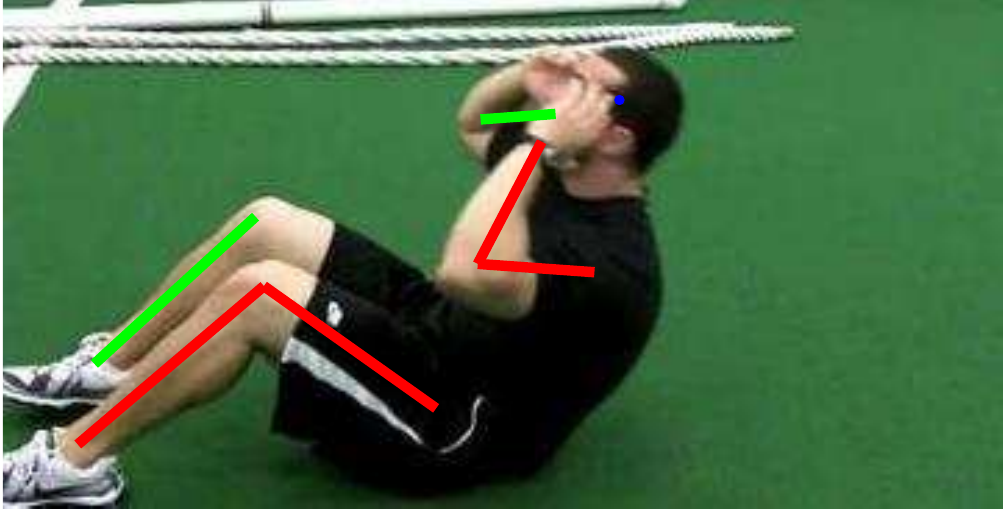}\hspace{0.15mm}
      \includegraphics[height=0.07\textwidth]{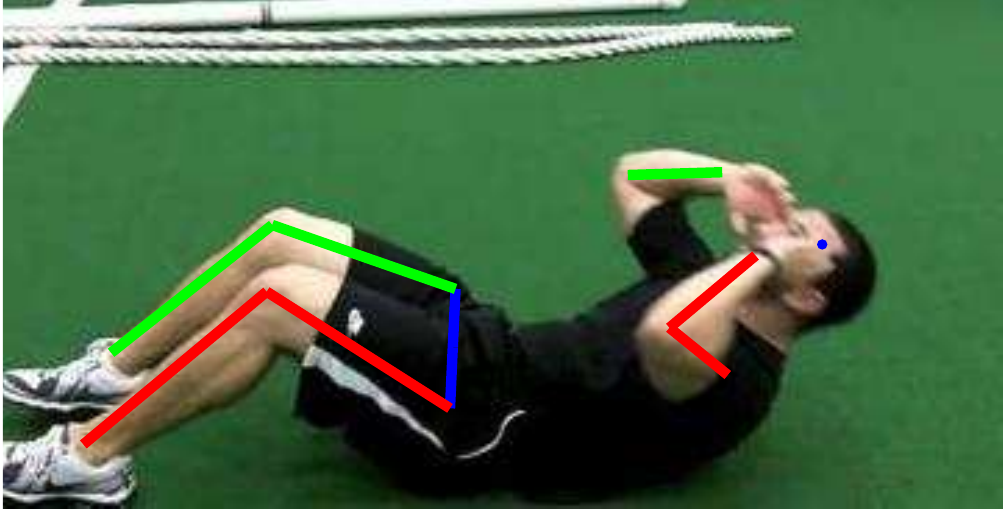}\hspace{0.15mm}
      \includegraphics[height=0.07\textwidth]{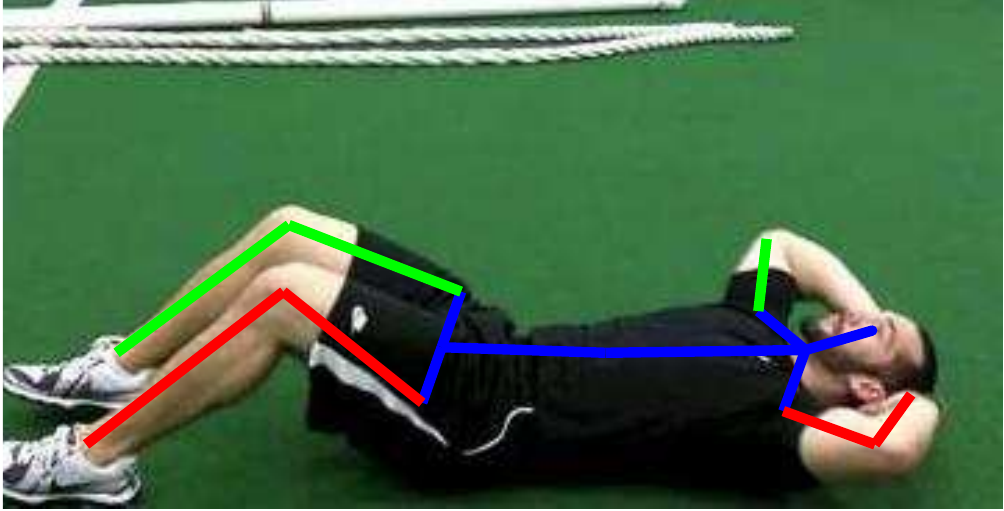}\hspace{0.15mm}
      \includegraphics[height=0.07\textwidth]{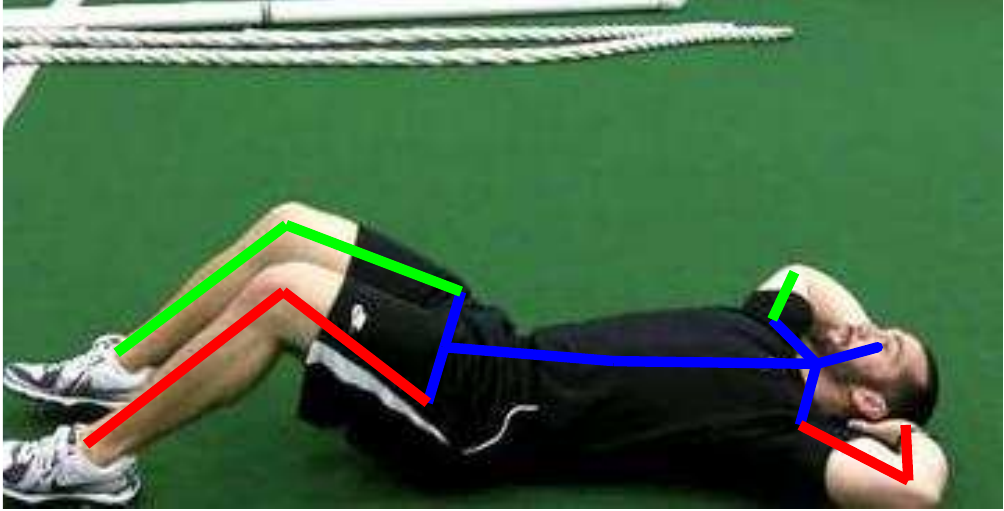}
      \\
      \hspace{-2.7mm}
      \includegraphics[height=0.07\textwidth]{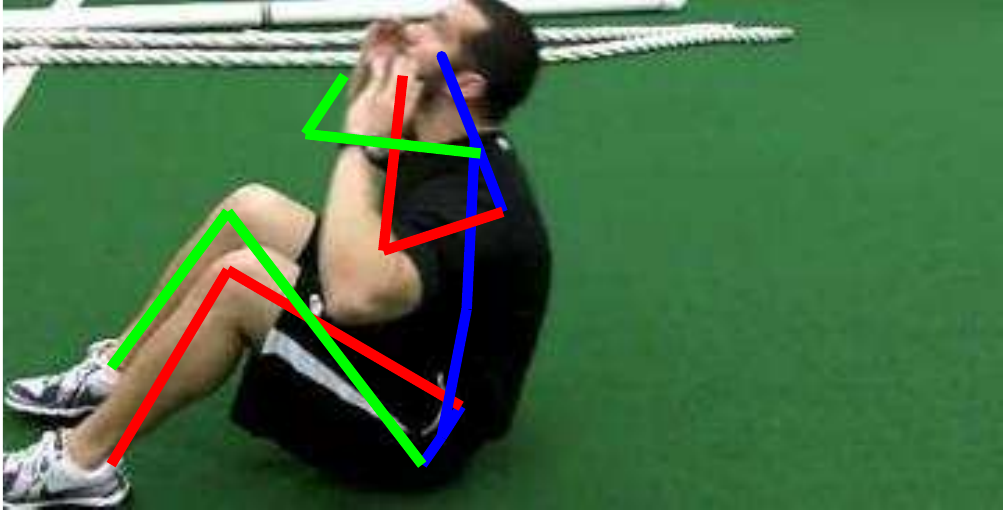}\hspace{0.15mm}
      \includegraphics[height=0.07\textwidth]{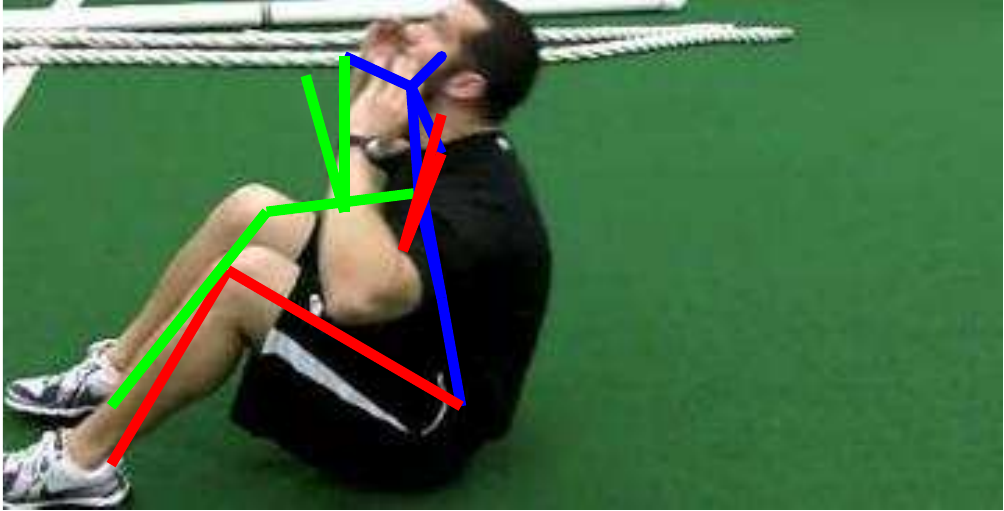}\hspace{0.15mm}
      \includegraphics[height=0.07\textwidth]{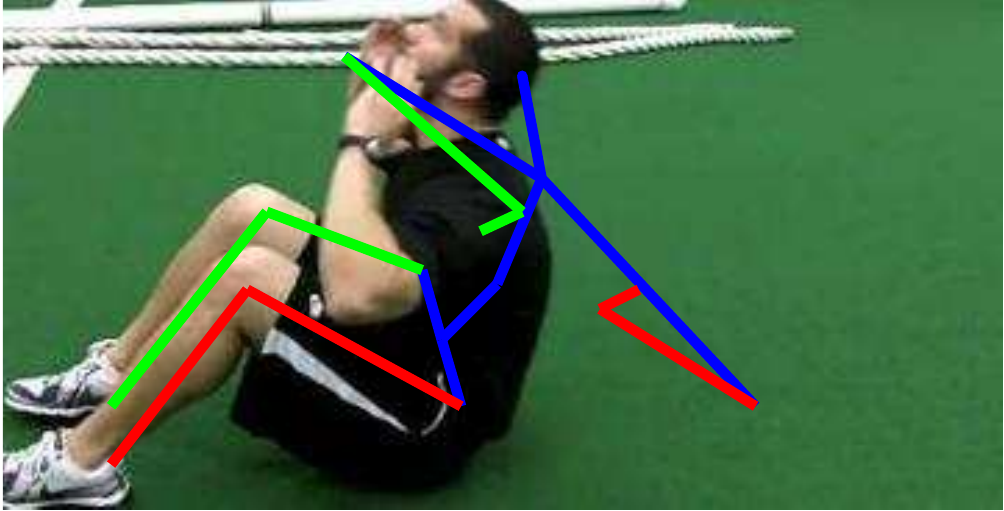}\hspace{0.15mm}
      \includegraphics[height=0.07\textwidth]{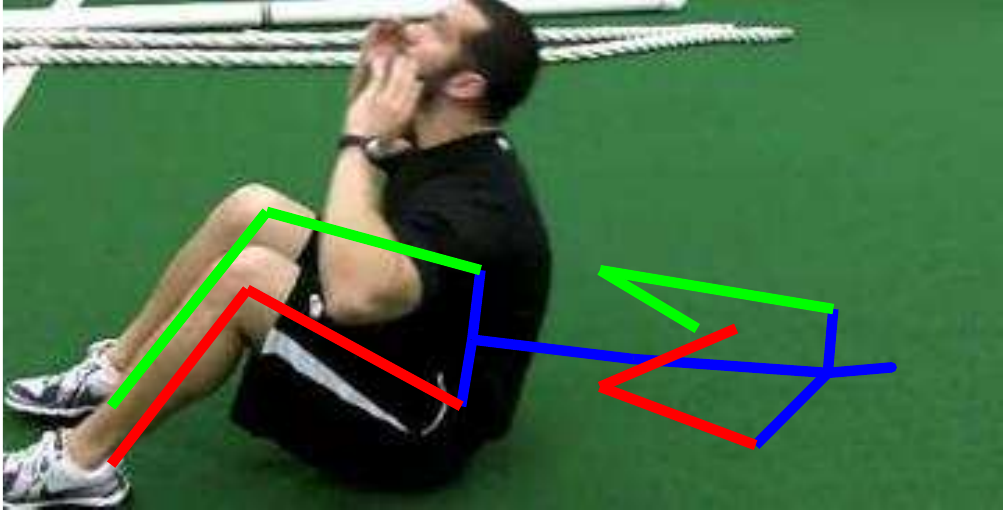}\hspace{0.15mm}
      \includegraphics[height=0.07\textwidth]{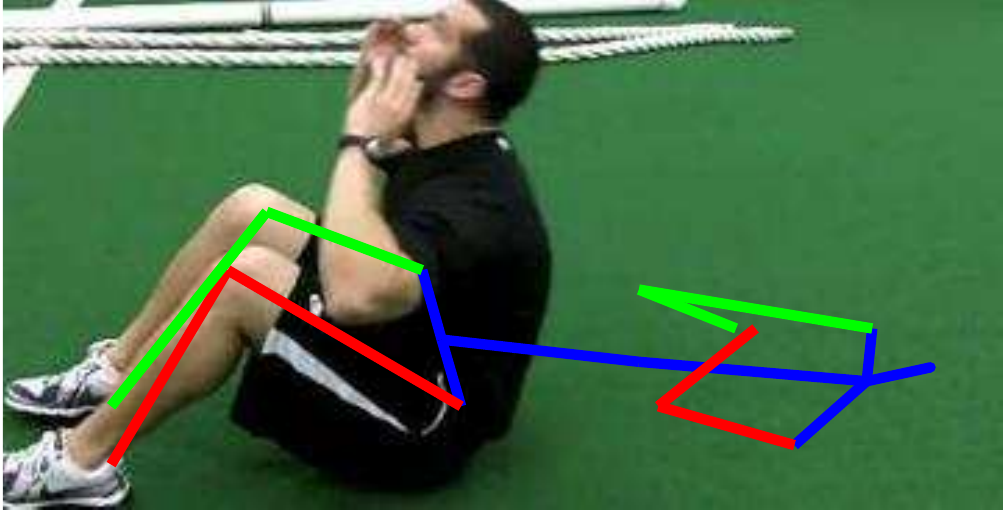}\hspace{0.15mm}
      \includegraphics[height=0.07\textwidth]{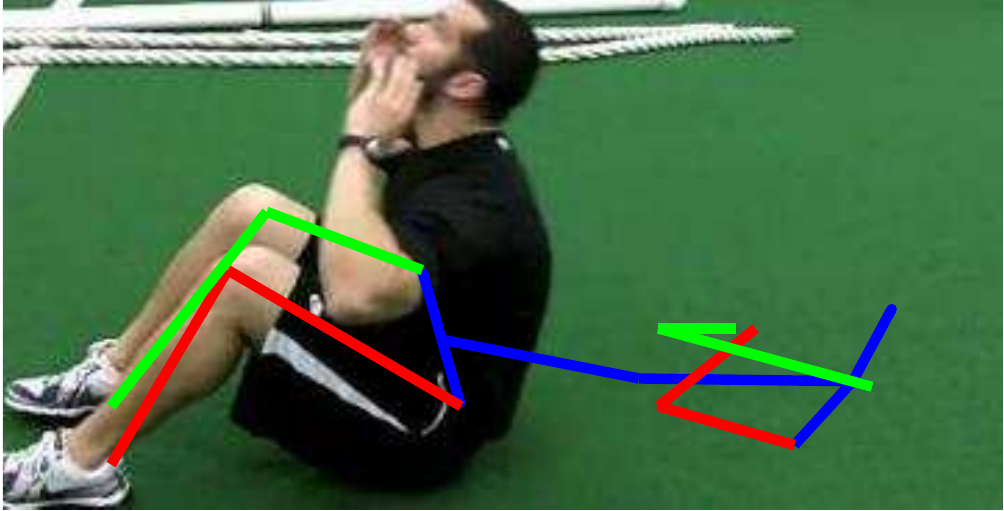}
      \\
      \includegraphics[height=0.10\textwidth]{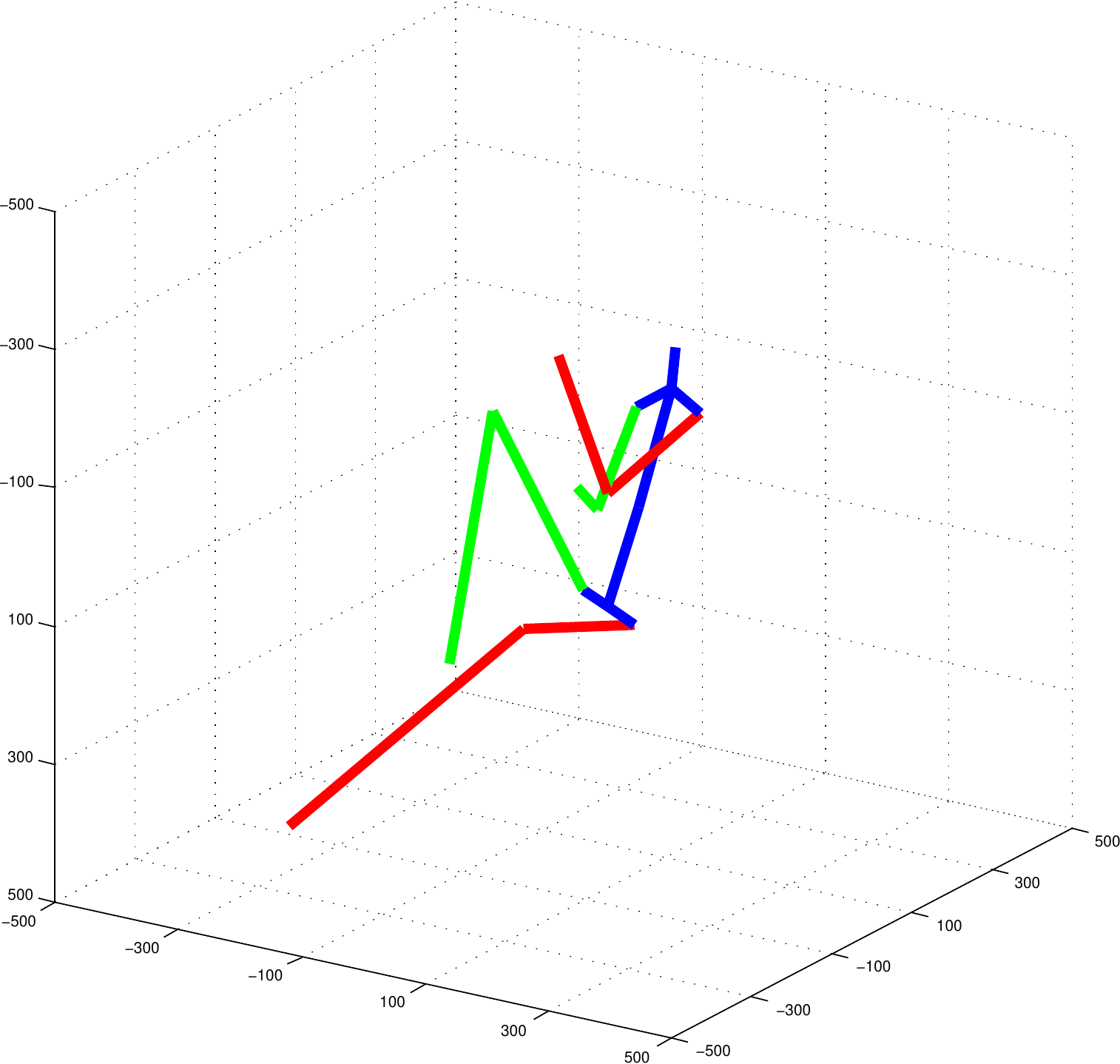}\hspace{1.9mm}
      \includegraphics[height=0.10\textwidth]{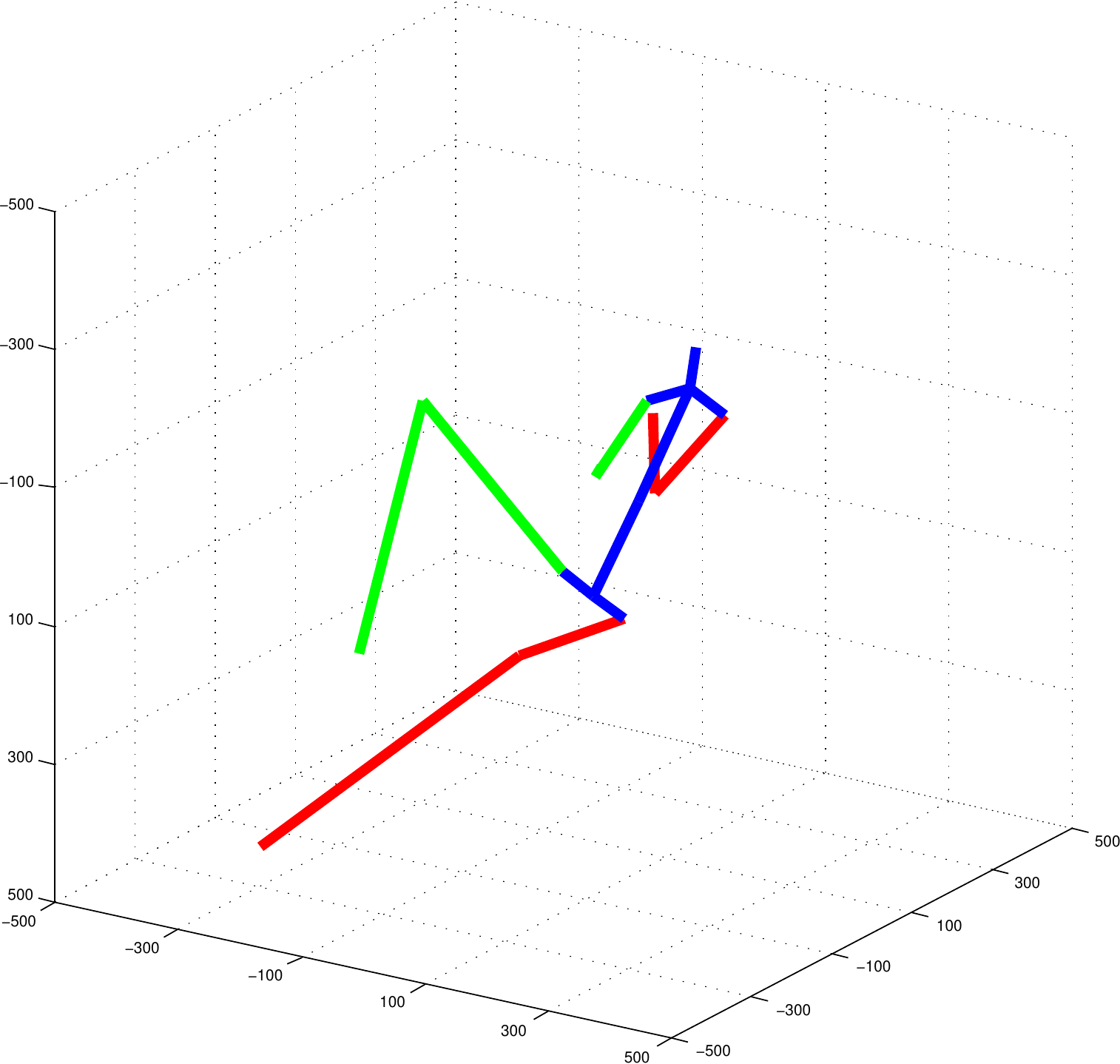}\hspace{1.9mm}
      \includegraphics[height=0.10\textwidth]{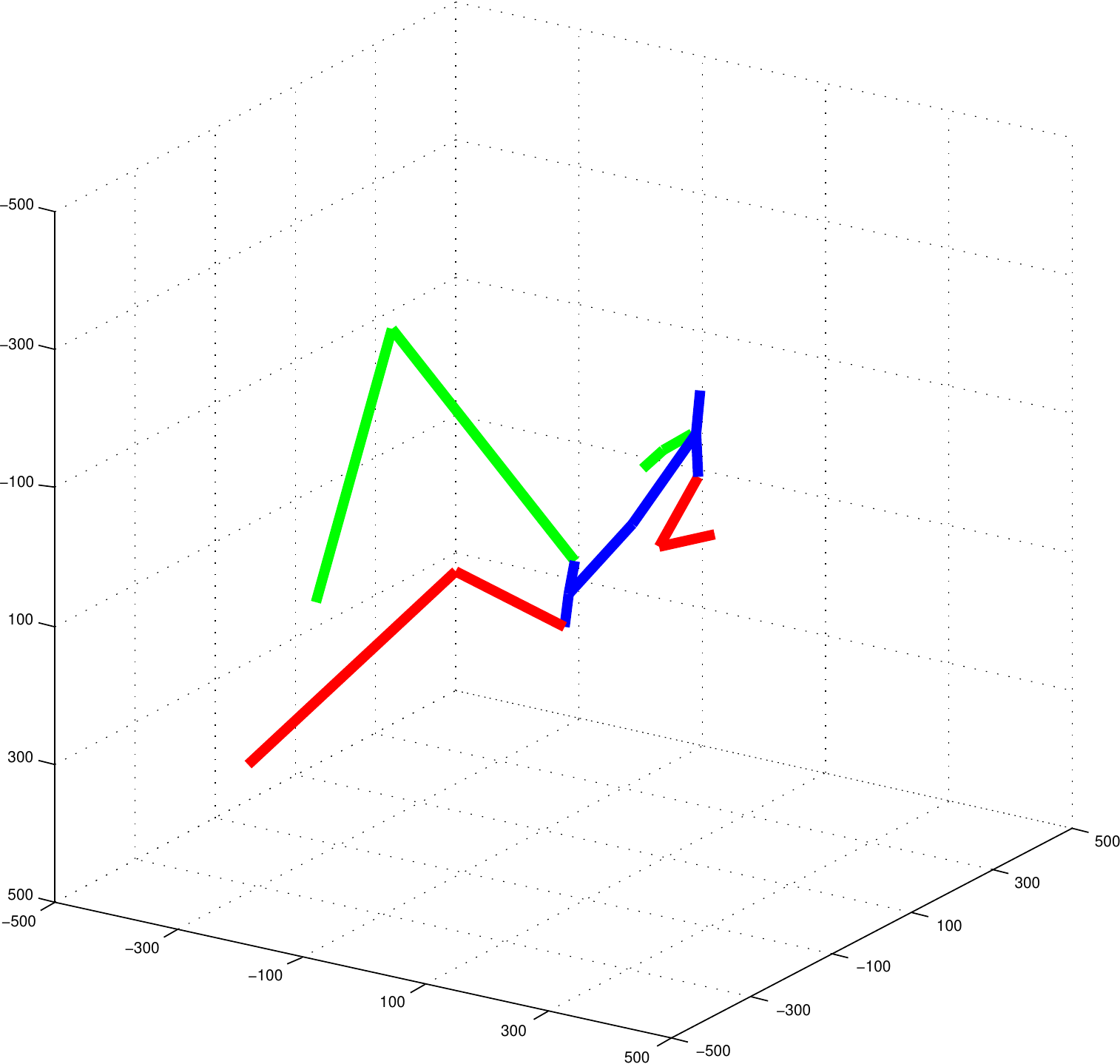}\hspace{1.9mm}
      \includegraphics[height=0.10\textwidth]{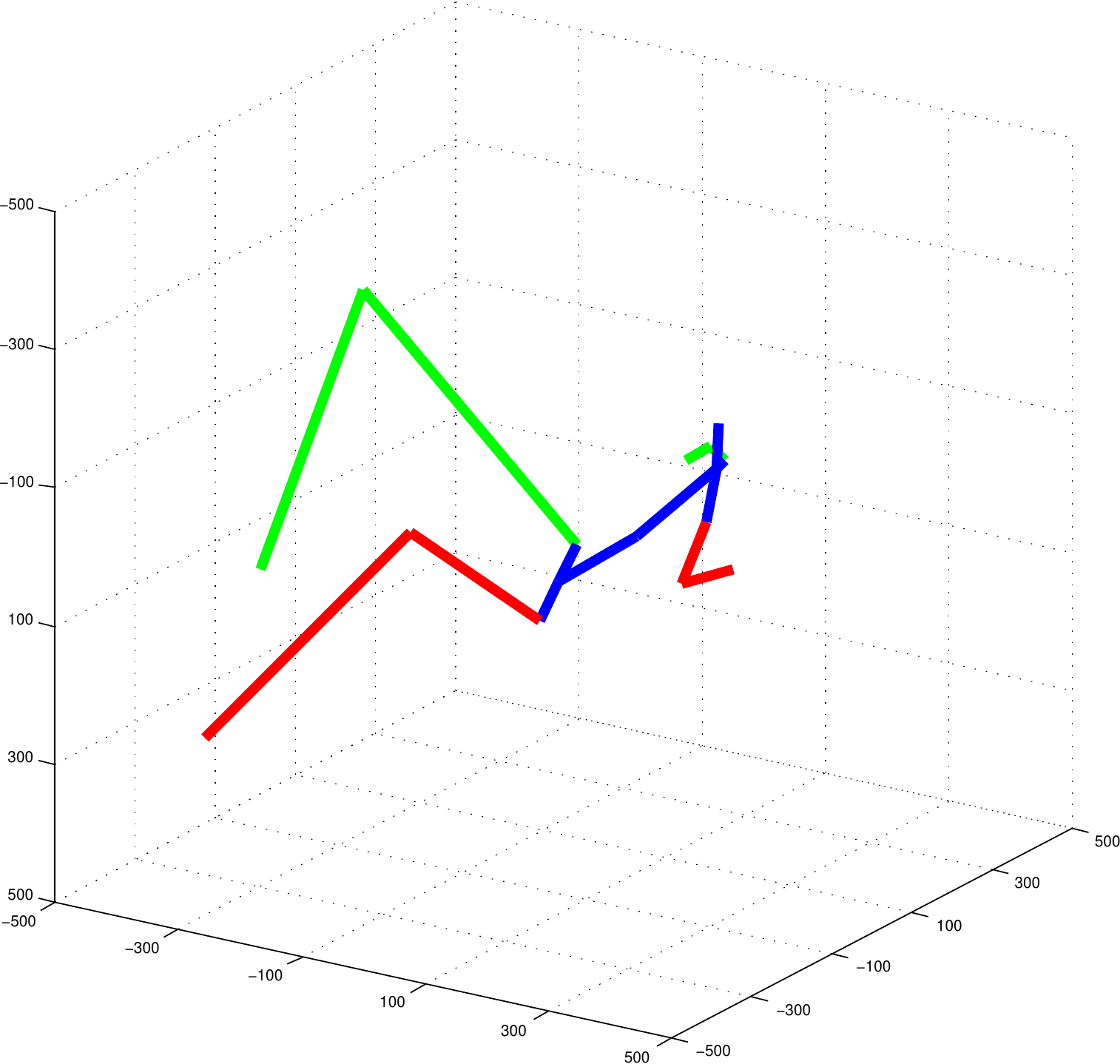}\hspace{1.9mm}
      \includegraphics[height=0.10\textwidth]{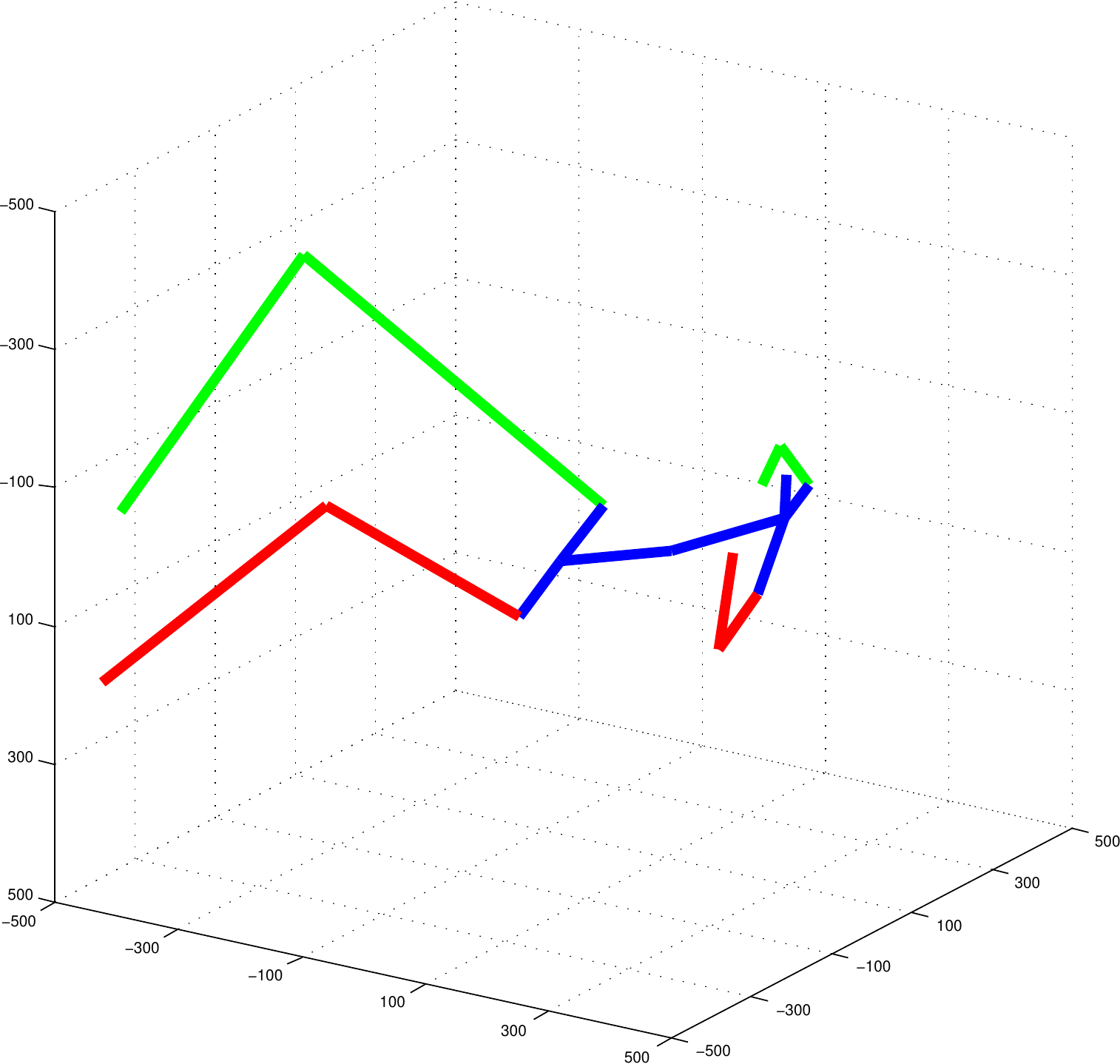}\hspace{1.9mm}
      \includegraphics[height=0.10\textwidth]{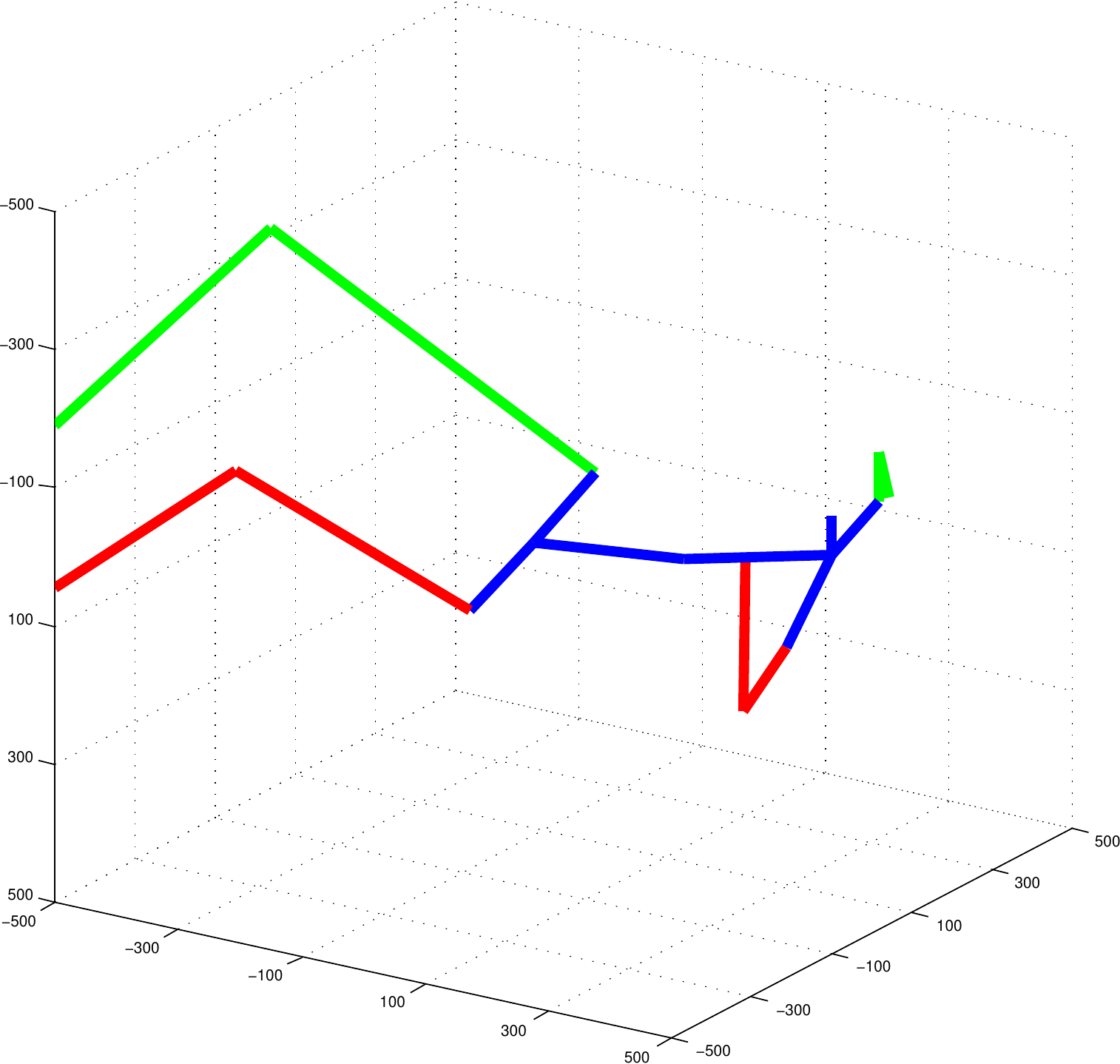}\hspace{1.9mm}
      \includegraphics[height=0.10\textwidth]{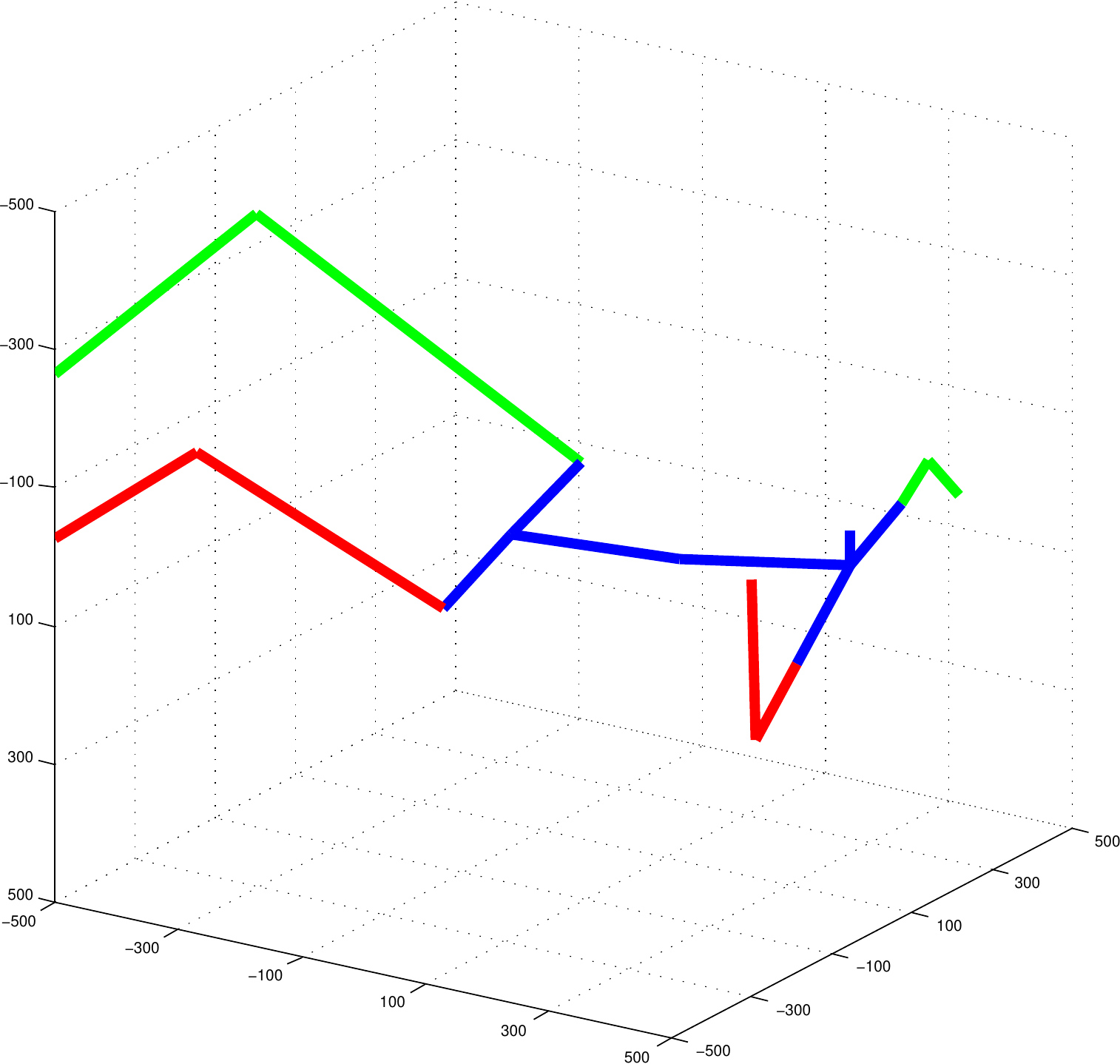}
    \end{tabular}
  \end{tabular}
  % \vspace{-2mm}
  \caption{\small Qualitative results of pose forecasting. The left column
shows the input images. For each input image, we show in the right column the
sequence of ground-truth frame and pose (top) and our forecasted pose sequence
in 2D (middle) and 3D (bottom). Note that some keypoints are not shown since
they are labeled as invisible in the ground-truth poses.}
  \vspace{-2mm}  % for arXiv compiler
  \label{fig:forecast}
\end{figure*}

\vspace{-3mm}

\begin{figure}[t]
  % \vspace{-2mm}
  % \centering
  \footnotesize
  \hspace{-0.8mm}
  \begin{subfigure}[c]{0.178\linewidth}
    \centering
    \includegraphics[height=1.61\textwidth]{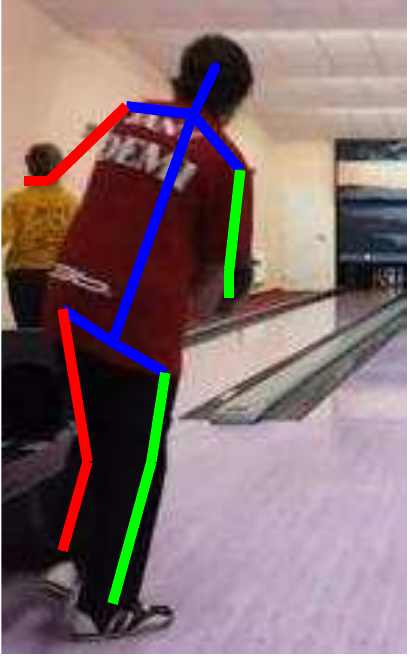}
  \end{subfigure}
  \begin{subfigure}[c]{0.05\linewidth}
    \centering
    \includegraphics[height=1\textwidth]{arrow.pdf}
  \end{subfigure}
  \hspace{-2.5mm}
  \begin{subfigure}[c]{0.178\linewidth}
    \centering
    \includegraphics[height=1.61\textwidth]{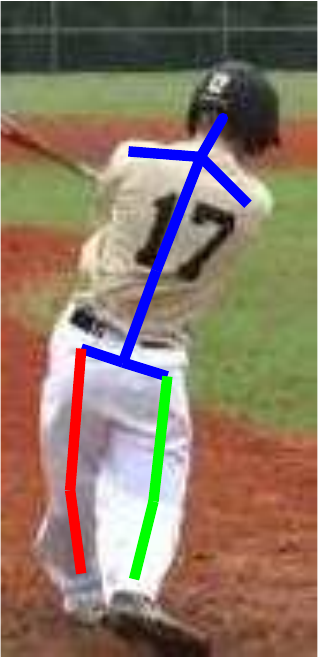}
  \end{subfigure}
  \hspace{-2.0mm}
  \begin{subfigure}[c]{0.05\linewidth}
    \centering
    \includegraphics[height=1\textwidth]{arrow.pdf}
  \end{subfigure}
  \hspace{-0.5mm}
  \begin{subfigure}[c]{0.178\linewidth}
    \centering
    \includegraphics[height=1.61\textwidth]{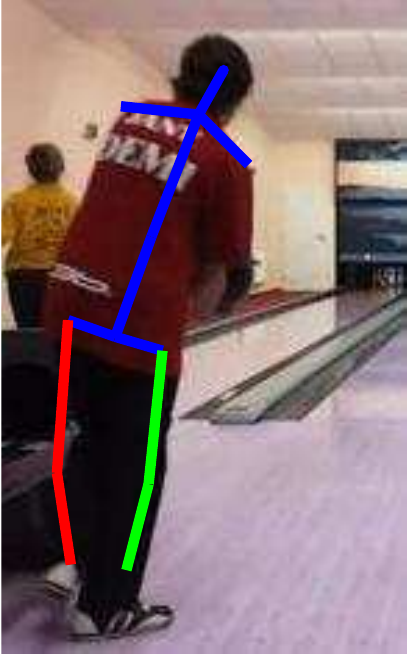}
  \end{subfigure}
  \begin{subfigure}[c]{0.178\linewidth}
    \centering
    \includegraphics[height=1.61\textwidth]{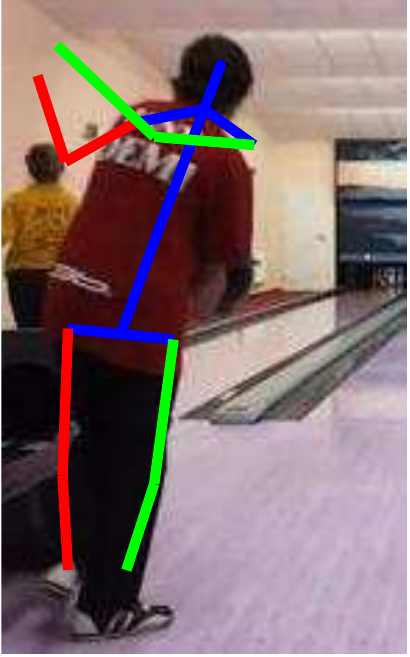}
  \end{subfigure}
  \begin{subfigure}[c]{0.178\linewidth}
    \centering
    \includegraphics[height=1.61\textwidth]{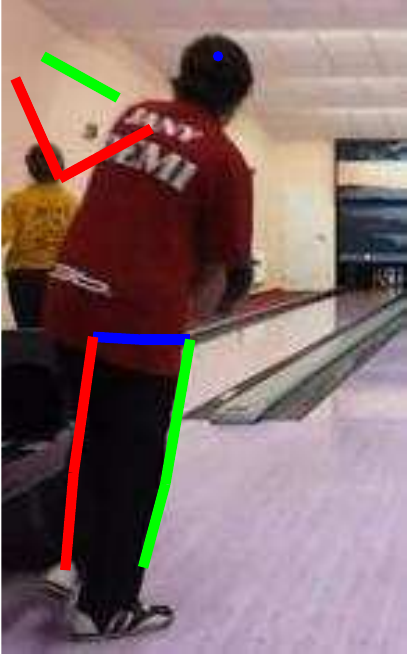}
  \end{subfigure}
  \\
  \hspace{-0.8mm}
  \begin{subfigure}[c]{0.178\linewidth}
    \centering
    \includegraphics[height=1.61\textwidth]{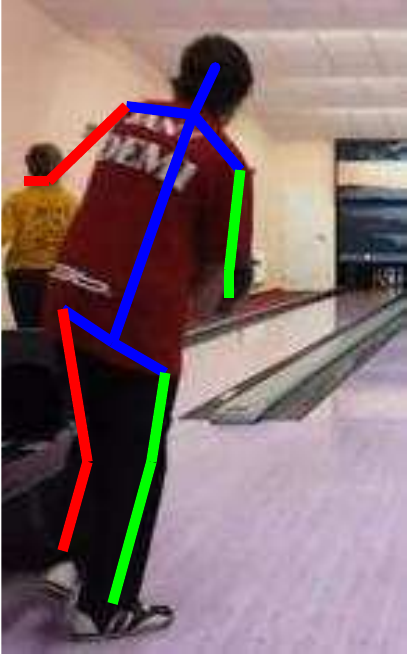}
  \end{subfigure}
  \begin{subfigure}[c]{0.05\linewidth}
    \centering
    \includegraphics[height=1\textwidth]{arrow.pdf}
  \end{subfigure}
  \hspace{-2.5mm}
  \begin{subfigure}[c]{0.178\linewidth}
    \centering
    \includegraphics[height=1.61\textwidth]{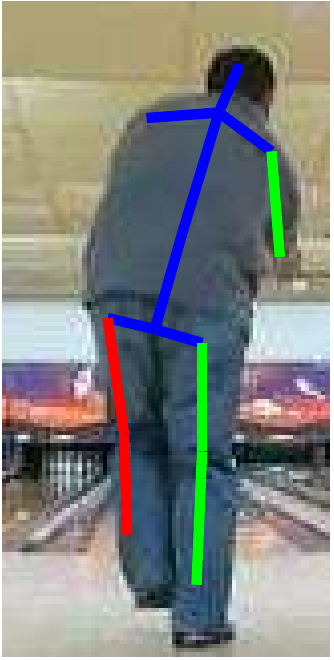}
  \end{subfigure}
  \hspace{-2.0mm}
  \begin{subfigure}[c]{0.05\linewidth}
    \centering
    \includegraphics[height=1\textwidth]{arrow.pdf}
  \end{subfigure}
  \hspace{-0.5mm}
  \begin{subfigure}[c]{0.178\linewidth}
    \centering
    \includegraphics[height=1.61\textwidth]{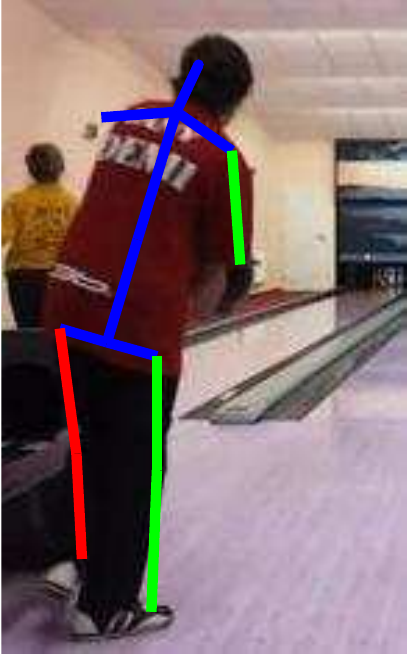}
  \end{subfigure}
  \begin{subfigure}[c]{0.178\linewidth}
    \centering
    \includegraphics[height=1.61\textwidth]{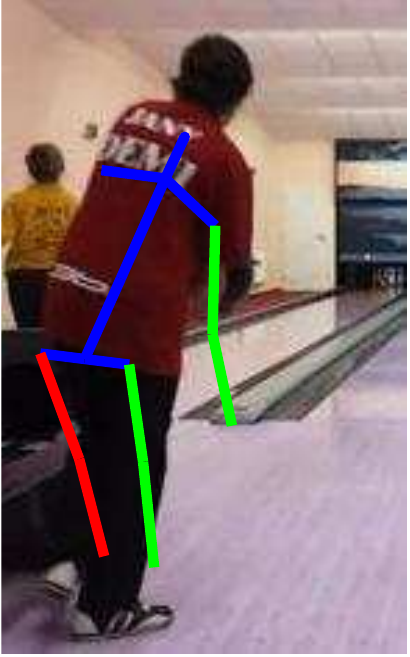}
  \end{subfigure}
  \begin{subfigure}[c]{0.178\linewidth}
    \centering
    \includegraphics[height=1.61\textwidth]{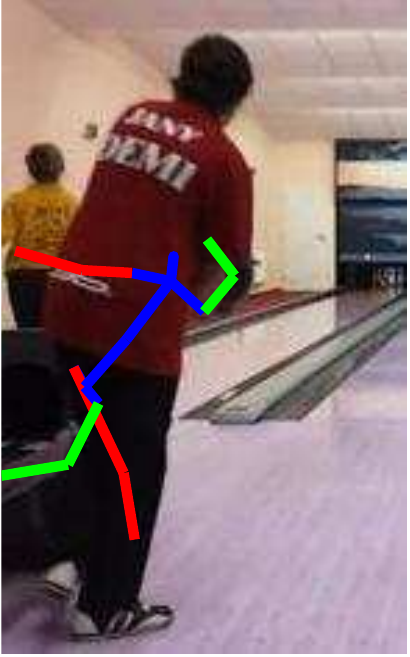}
  \end{subfigure}
  % \vspace{-2mm}
  \caption{\small Failure cases of the NN baselines. Top: NN-all. Bottom:
NN-oracle. Each row shows the input image with the estimated pose, the NN pose
in the training set, and the transformed pose sequence of the NN pose on the
input image.}
  \vspace{-2mm}  % for arXiv compiler
  \label{fig:nn}
\end{figure}

\paragraph{Results} Fig.~\ref{fig:pck} shows the PCK curves of our approach and
the baselines at different timesteps (timestep 1 corresponds to the current
frame). For all approaches, the PCK value decreases as timestep increases,
since prediction becomes more challenging due to increasing ambiguity as we
move further from the current observation. We also report the result of the
hourglass network used for our 3D-PFNet. Since the hourglass network can only
estimate the current pose, we only show its PCK curve in timestep 1. The three
NN baselines achieve similar performance at timestep 1. As the timestep
increases, NN-CaffNet gradually outperforms NN-all, verifying our hypothesis
that scene contexts can be used to reject irrelevant candidates and improve NN
results. Similarly, NN-oracle gradually outperforms NN-CaffeNet, since the
ground-truth action labels can improve the candidate set further. Finally, our
3D-PFNet outperforms all three baselines by significant margins.
Fig.~\ref{fig:forecast} shows qualitative examples of the poses forecasted by
our 3D-PFNet.~\footnote{Also see
\href{http://www.umich.edu/~ywchao/image-play/}{\texttt{http://www.umich.edu/$\sim$ywchao/image-play/}}.}
Our 3D-PFNet can predict reasonable pose sequences in both 2D and 3D space.
Fig.~\ref{fig:nn} shows failure cases of the NN baselines. The retrieved
sequence of NN-all (top) is inconsistent with the context (i.e. a bowling
alley) when the NN pose is from a different action class (i.e. baseball swing).
By exploiting ground-truth action labels, NN-oracle (bottom) is able to
retrieve a similar pose in the same context. However, the retrieved sequence
still fails due to a small error in pose alignment, i.e. the person should be
moving slightly toward the right rather than straight ahead. We believe the
internal feature representation learned by our 3D-PFNet can better align human
poses in the given context and thus can generate more accurate predictions.
Tab.~\ref{tab:pck} reports the PCK with threshold 0.05 (PCK@0.05) for all 16
timesteps. Note that all PCK values stop decreasing after timestep 8. This is
due to the subset of test sequences with repetitive ending frames, since
prediction is easier for those cases as we only need to learn to stop and
repeat the last predicted pose.

\subsection{3D Pose Recovery}

We separately evaluate the task of per-frame 3D skeleton recovery from 2D
heatmaps on Human3.6M \cite{ionescu:pami2014}.

\vspace{-3mm}

\paragraph{Setup} We use the same data split as in training 3D-PFNet. However,
we use video frames and generate heatmaps from hourglass rather than using
synthetic data. For evaluation, we construct a validation set of 16150 images
by sampling every 40 frames from all sequences and cameras of S5 and S6. Each
frame is cropped with the tightest window that encloses the person bounding box
while keeping the principal point at image center. We evaluate the predicted 3D
keypoint positions relative to their center (i.e. $\Delta$) with mean per joint
position error (MPJPE) proposed in \cite{ionescu:pami2014}. For training, we
first fine-tune the hourglass on Human3.6M. We initialize the 3D skeleton
converter with weights trained on synthetic data, and further train it with
heatmaps from the hourglass.

\begin{table*}[t]
  \centering
  \small
  \setlength{\tabcolsep}{3.9pt}
  \begin{tabular}{l C{0.84cm} C{0.84cm} C{0.84cm} C{0.84cm} C{0.84cm} C{0.84cm} C{0.84cm} C{0.84cm}
                    C{0.84cm} C{0.84cm} C{0.84cm} C{0.84cm} C{0.84cm} C{0.84cm} C{0.84cm} C{0.84cm}}
    \hline \TBstrut
                                 &  Head & R.Sho & L.Sho & R.Elb & L.Elb & R.Wri & L.Wri & R.Hip & L.Hip & R.Kne & L.Kne & R.Ank & L.Ank &   Avg  \\
    \hline \Tstrut
    Convex \cite{zhou:cvpr2015}  & 145.3 & 123.5 & 122.8 & 139.1 & 129.5 & 162.2 & 153.0 & 115.2 & 111.8 & 172.1 & 171.7 & 257.4 & 258.5 & 158.6  \\
    SMPLify \cite{bogo:eccv2016} & 132.3 & 117.4 & 119.3 & 149.6 & 149.5 & 204.3 & 192.8 & 140.9 & 124.0 & 131.9 & 135.3 & 202.3 & 213.6 & 154.9  \\ \Bstrut
    Ours                         &~~\textbf{72.3}&~~\textbf{64.7}&~~\textbf{63.5}&~~\textbf{93.9}&~~\textbf{88.8}& \textbf{135.1}& \textbf{124.2}
                                 &~~\textbf{59.1}&~~\textbf{57.5}&~~\textbf{75.7}&~~\textbf{76.5}& \textbf{113.6}& \textbf{113.4}&~~\textbf{87.6} \\
    \hline
  \end{tabular}
  \vspace{-2mm}
  \caption{\small Mean per joint position errors (mm) on Human3.6M. Our 3D
converter achieves a lower error than the baselines on all joints.}
  \vspace{-2mm}  % for arXiv compiler
  \label{tab:mpjpe}
\end{table*}

\begin{figure*}[t]
  % \vspace{-2mm}
  % \centering
  \footnotesize
  \begin{tabular}{L{0.49\linewidth}@{\hspace{1.6mm}} L{0.49\linewidth}}
    \hspace{-2.5mm}
    \includegraphics[height=0.116\textwidth]{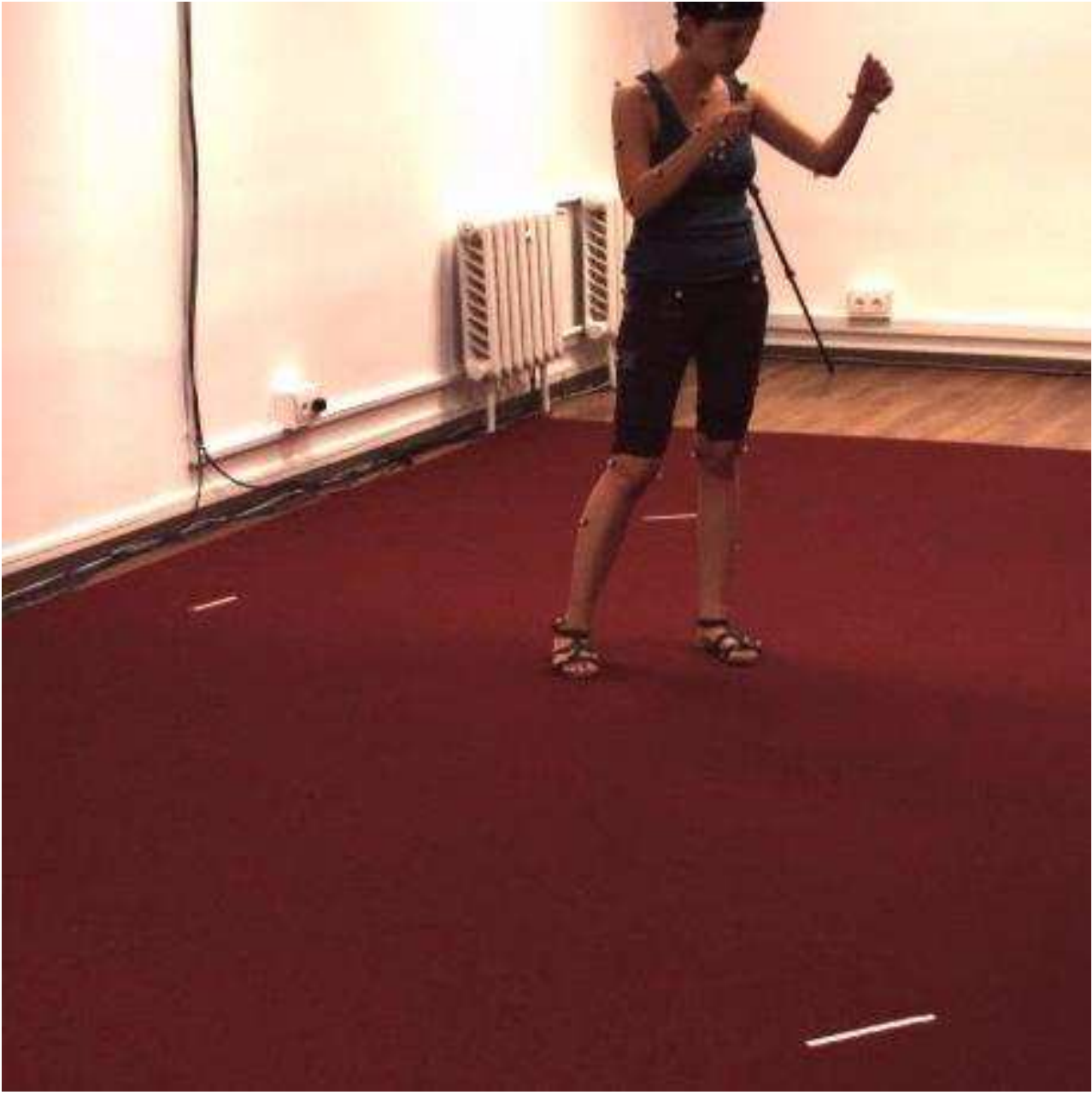}
    \includegraphics[height=0.116\textwidth]{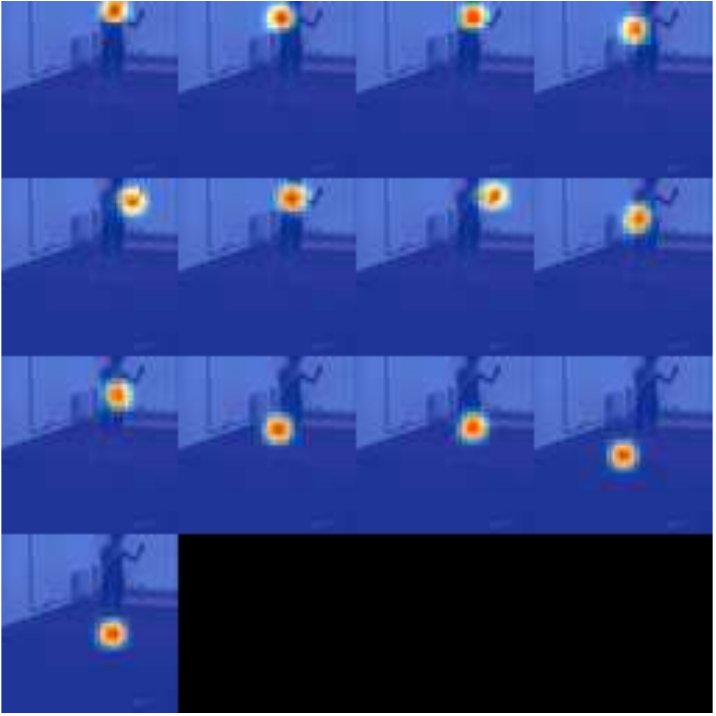}
    \includegraphics[height=0.116\textwidth]{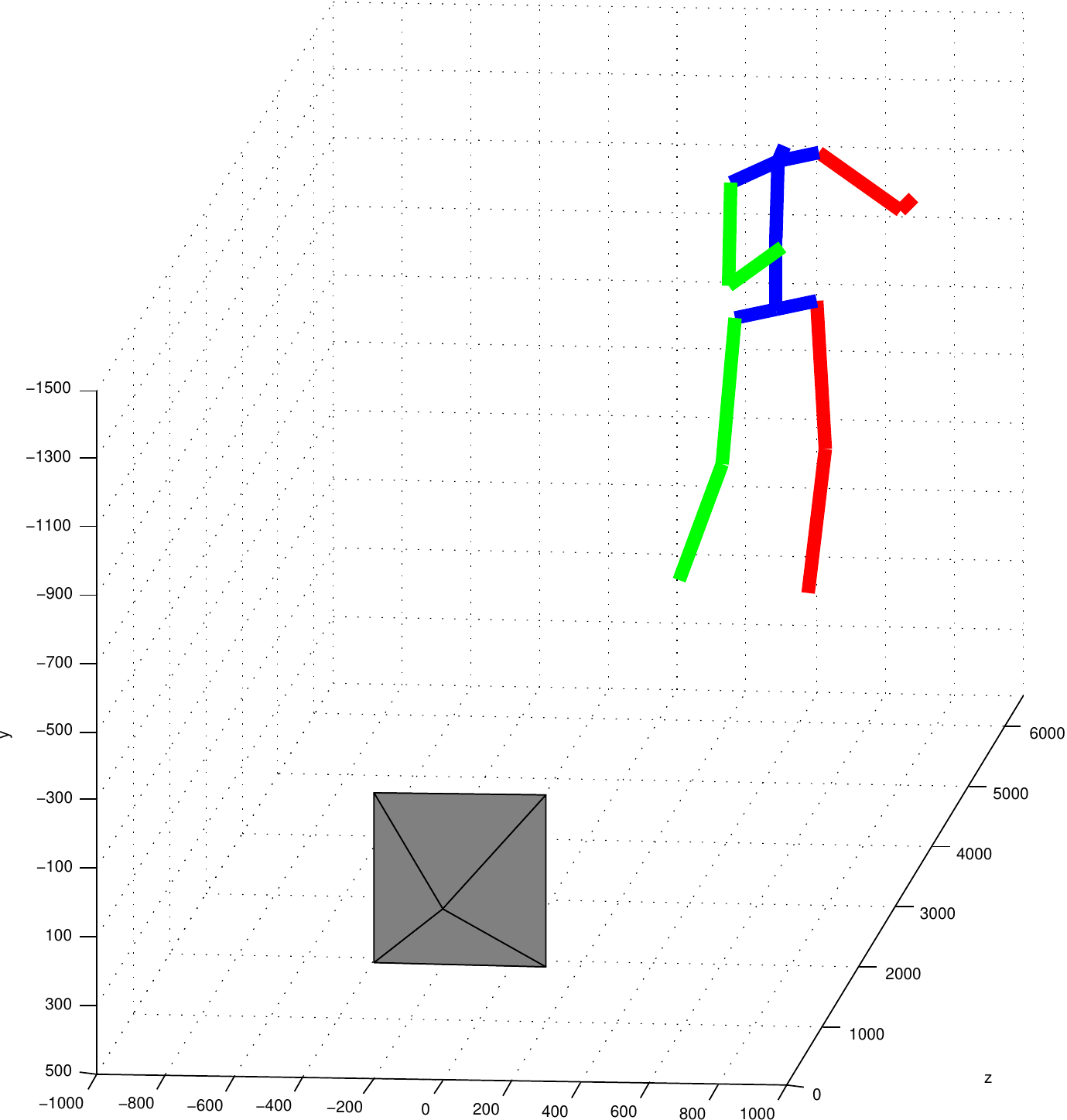}
    \includegraphics[height=0.116\textwidth]{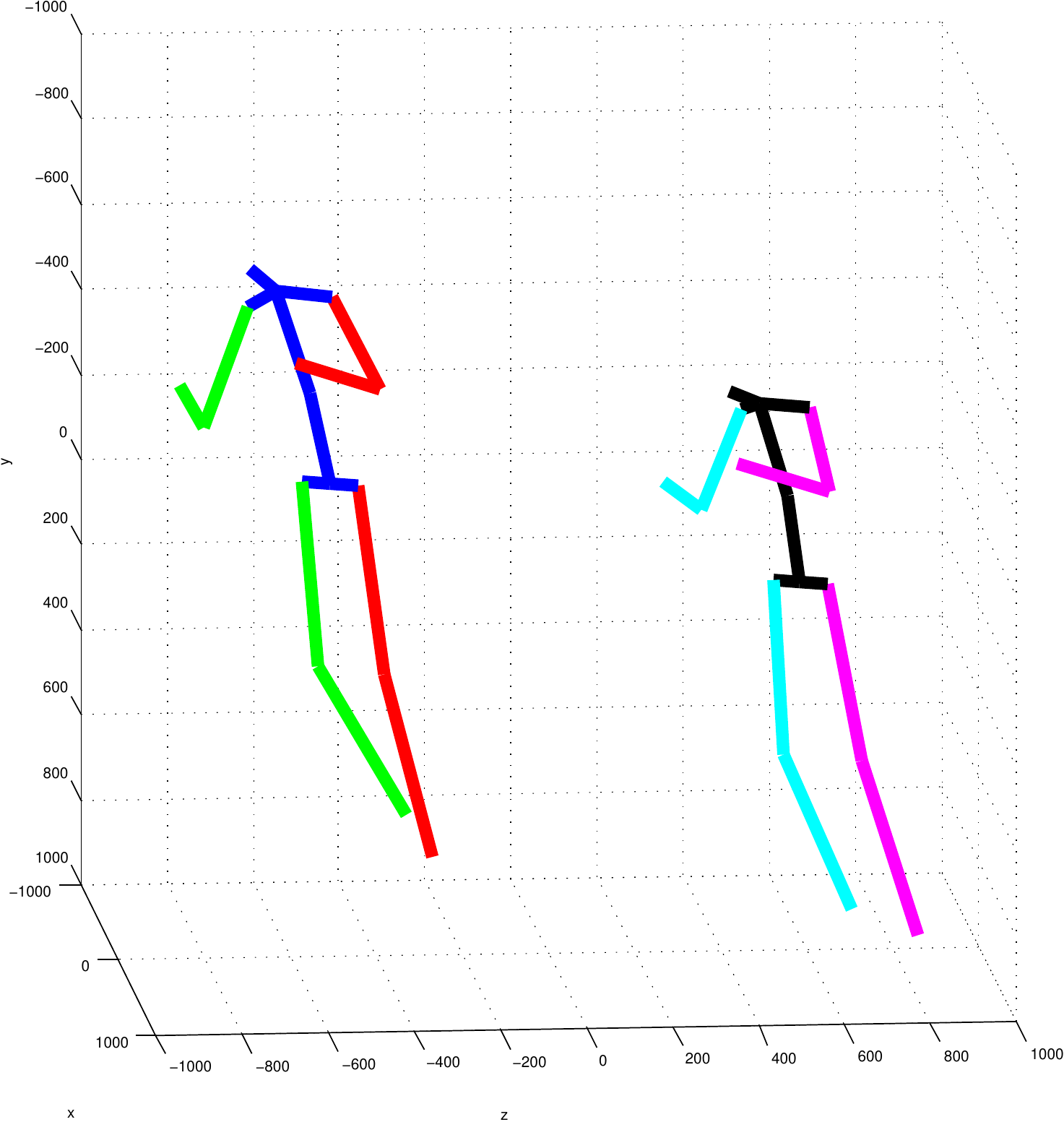}
    &
    \includegraphics[height=0.116\textwidth]{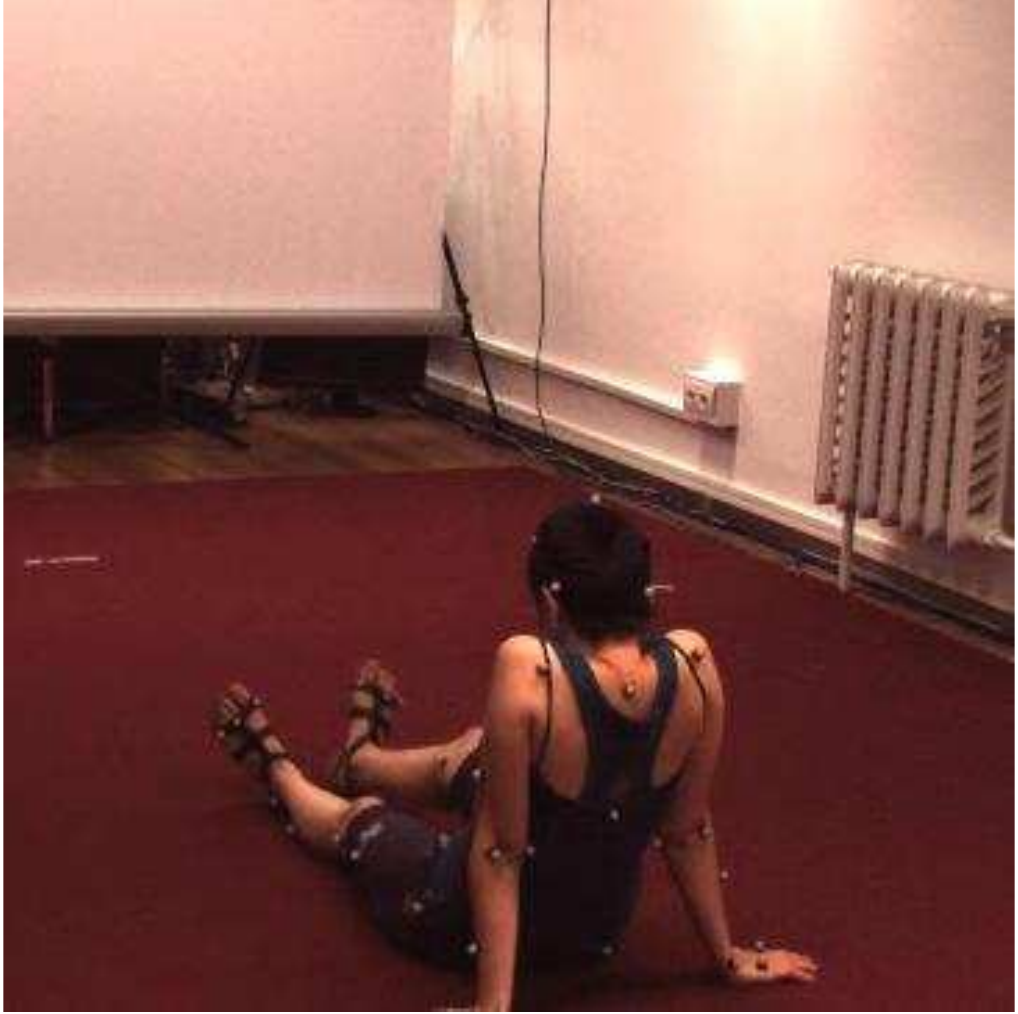}
    \includegraphics[height=0.116\textwidth]{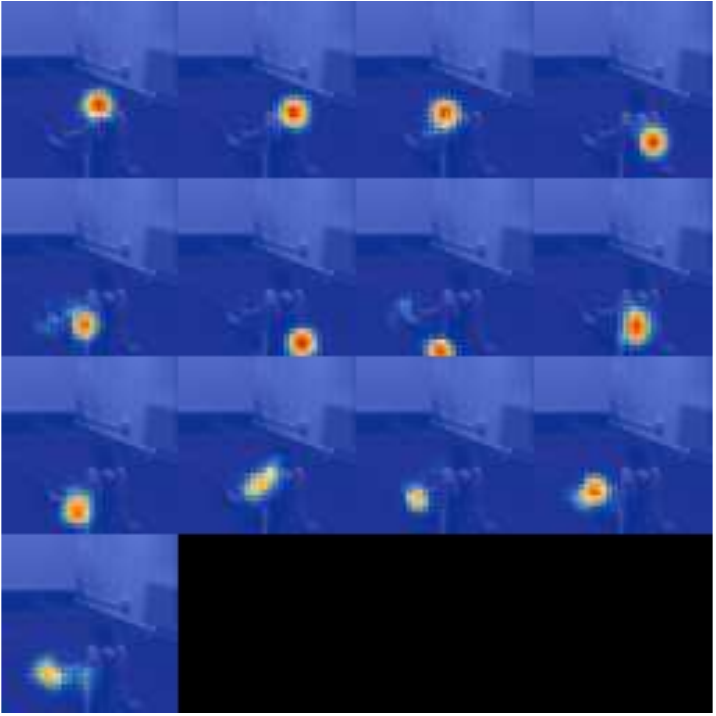}
    \includegraphics[height=0.116\textwidth]{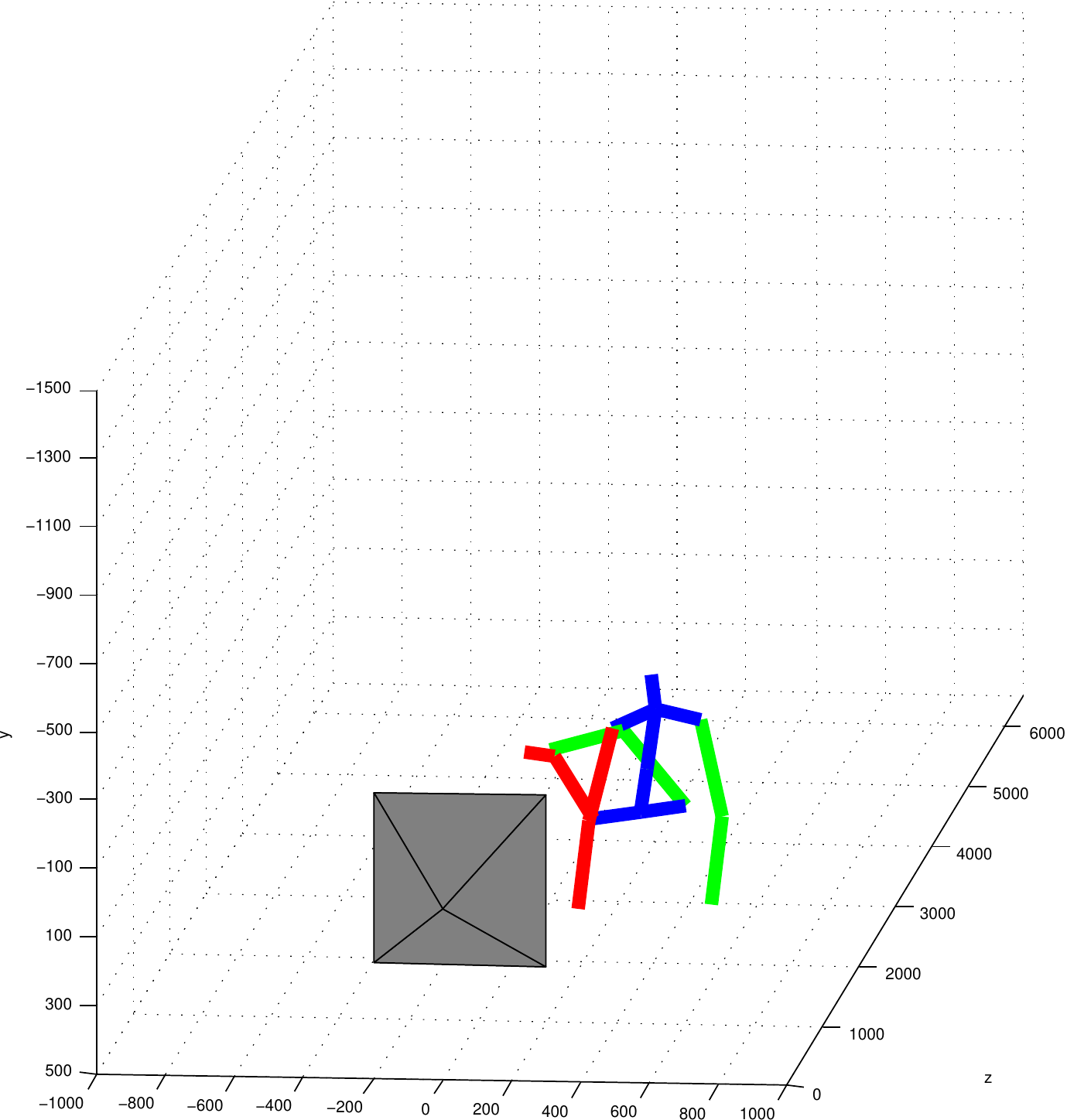}
    \includegraphics[height=0.116\textwidth]{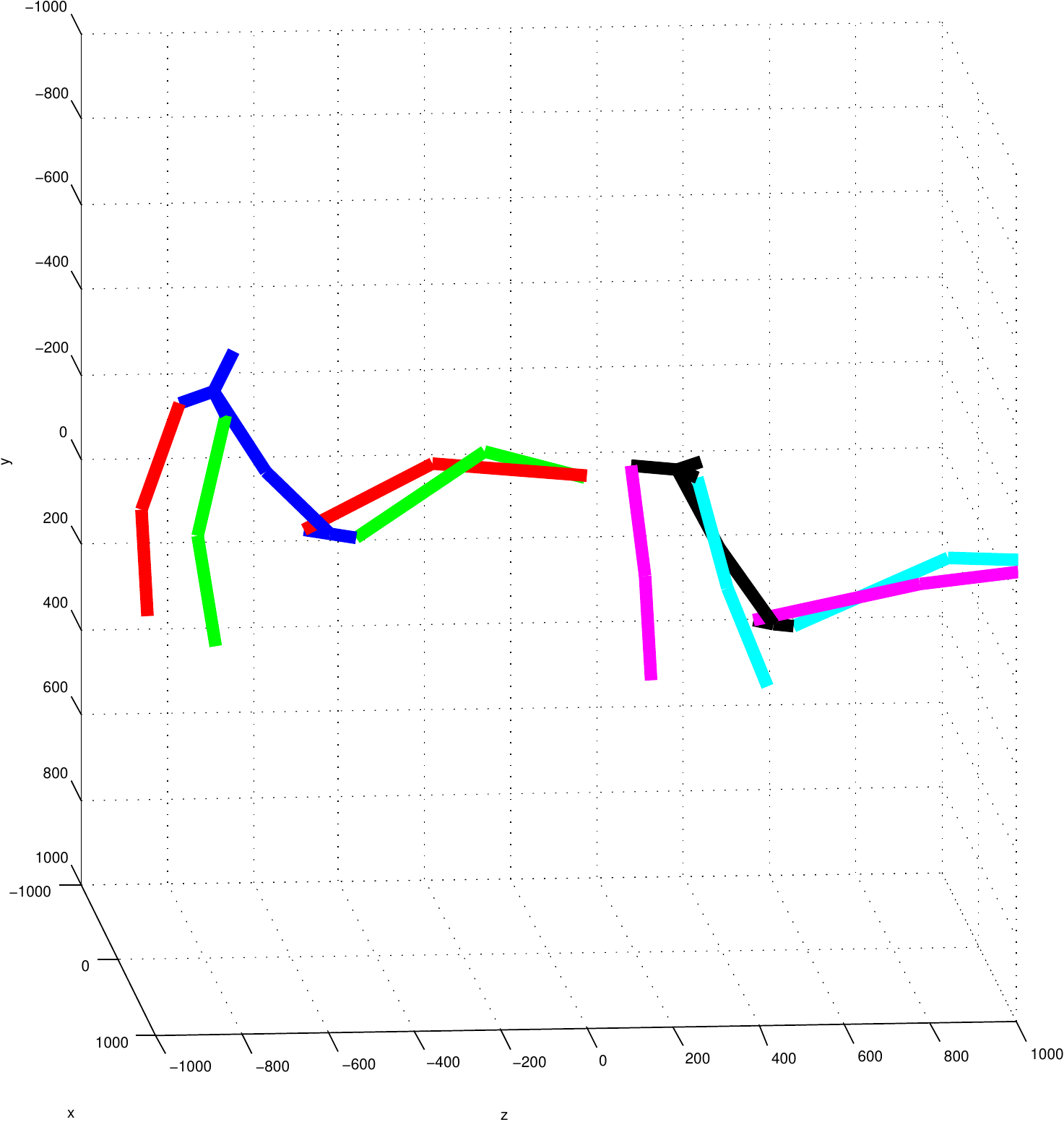}
    \\
    \hspace{-2.5mm}
    \includegraphics[height=0.116\textwidth]{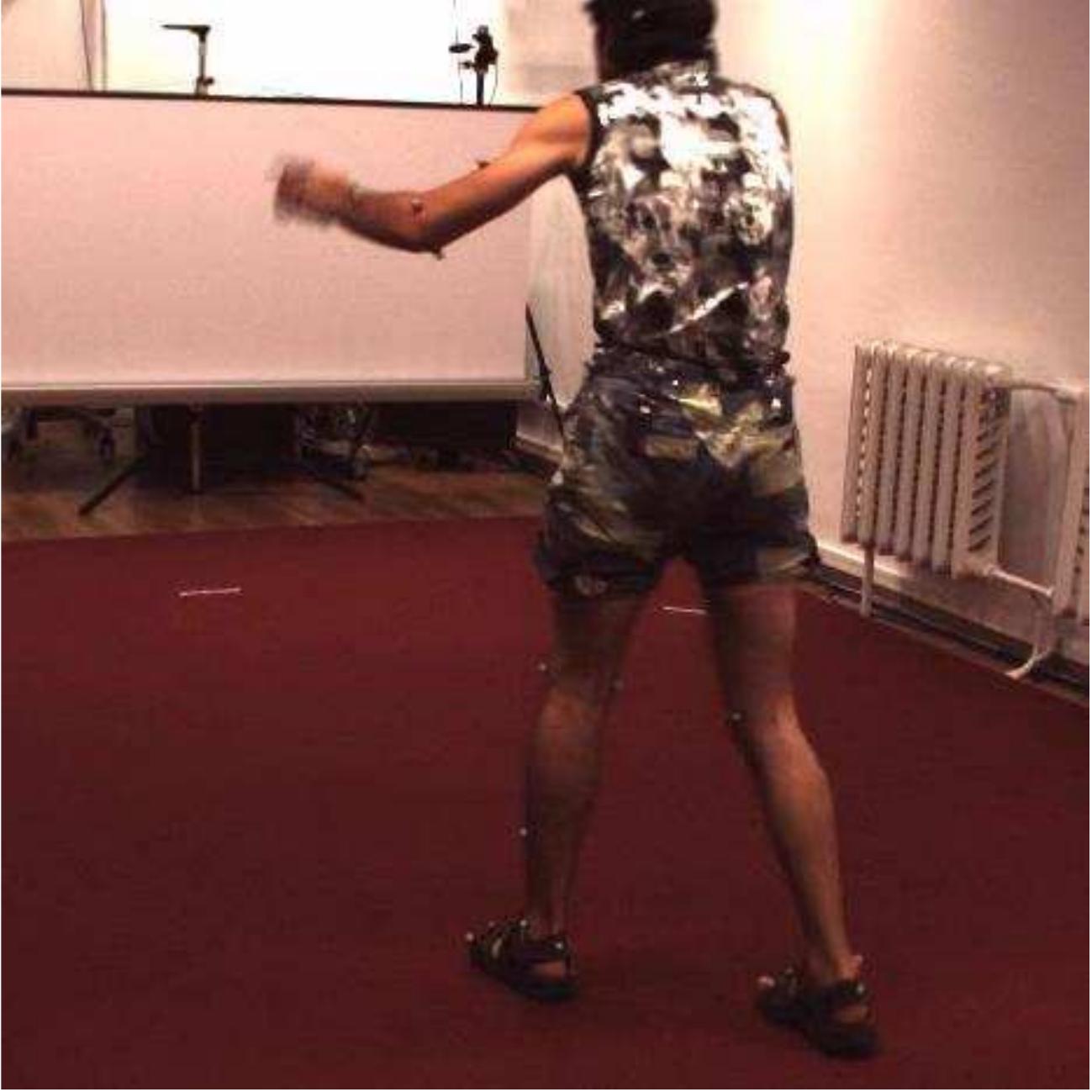}
    \includegraphics[height=0.116\textwidth]{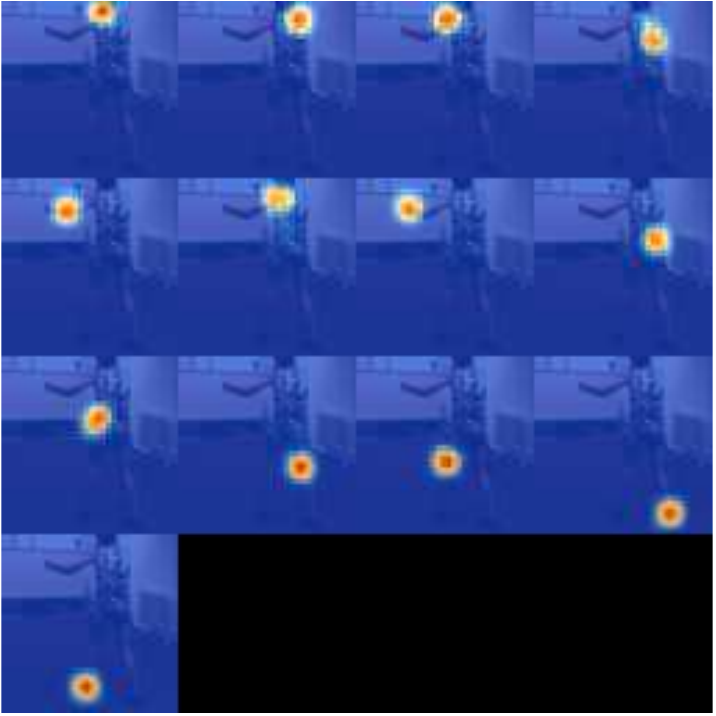}
    \includegraphics[height=0.116\textwidth]{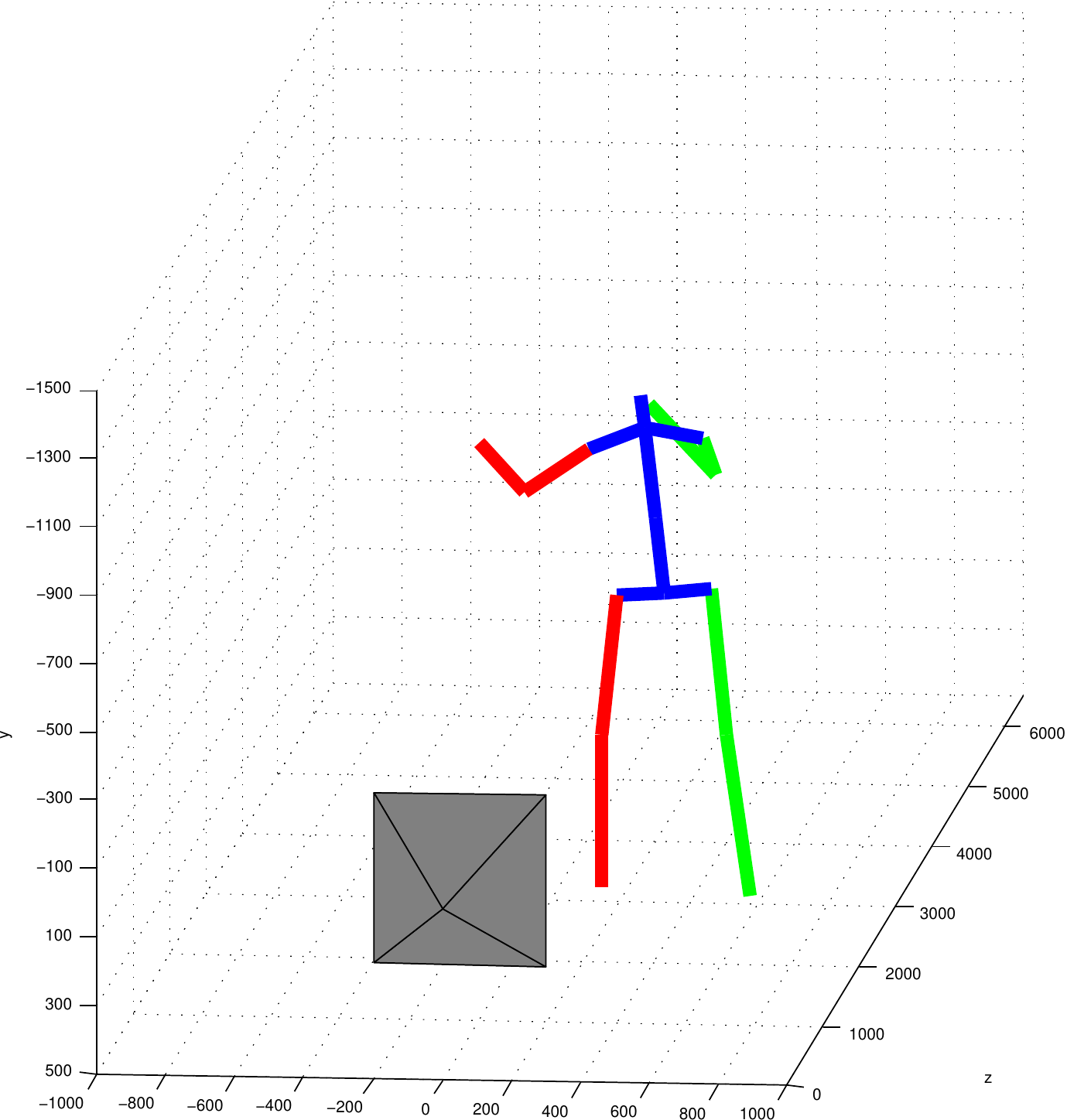}
    \includegraphics[height=0.116\textwidth]{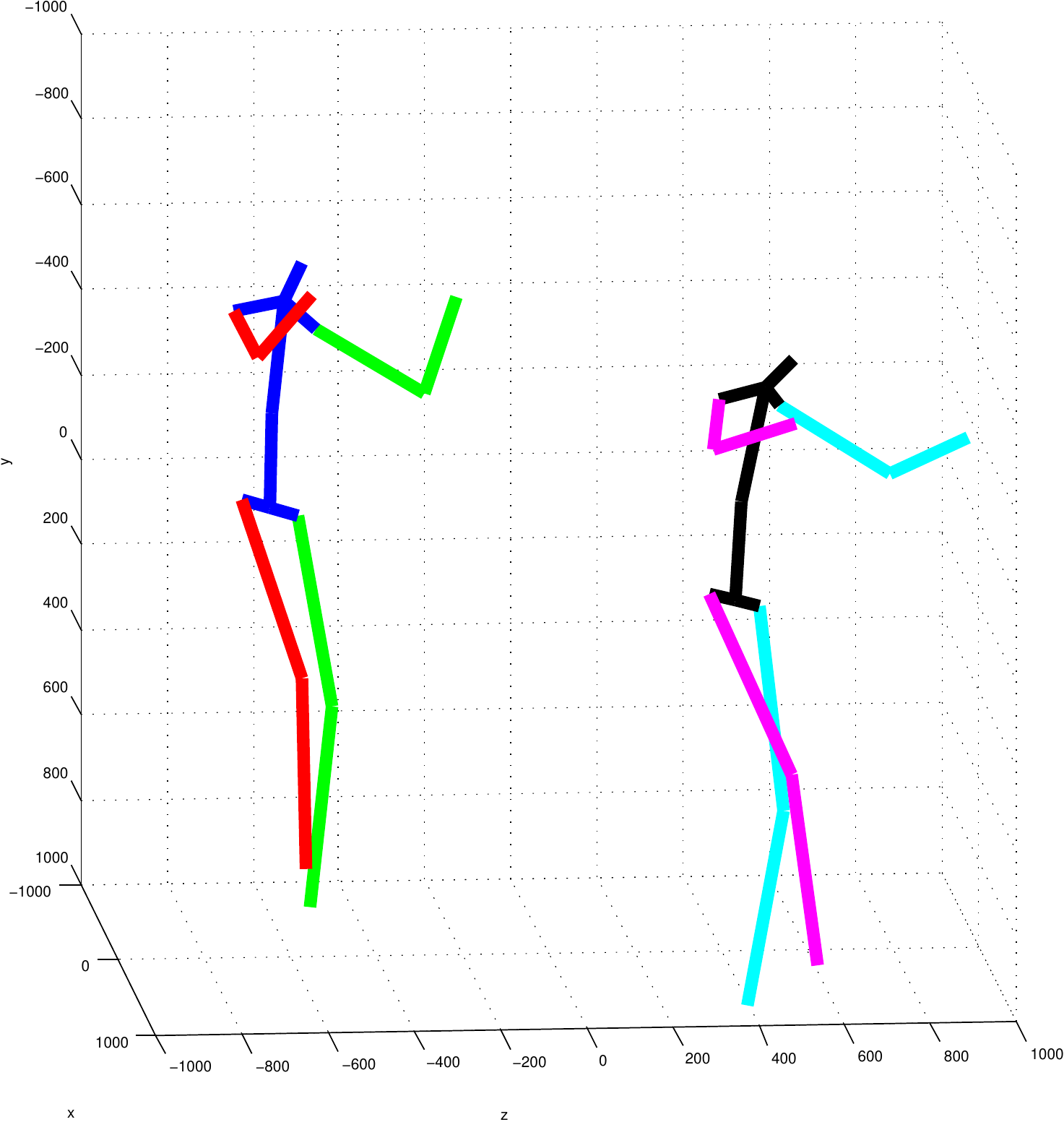}
    &
    \includegraphics[height=0.116\textwidth]{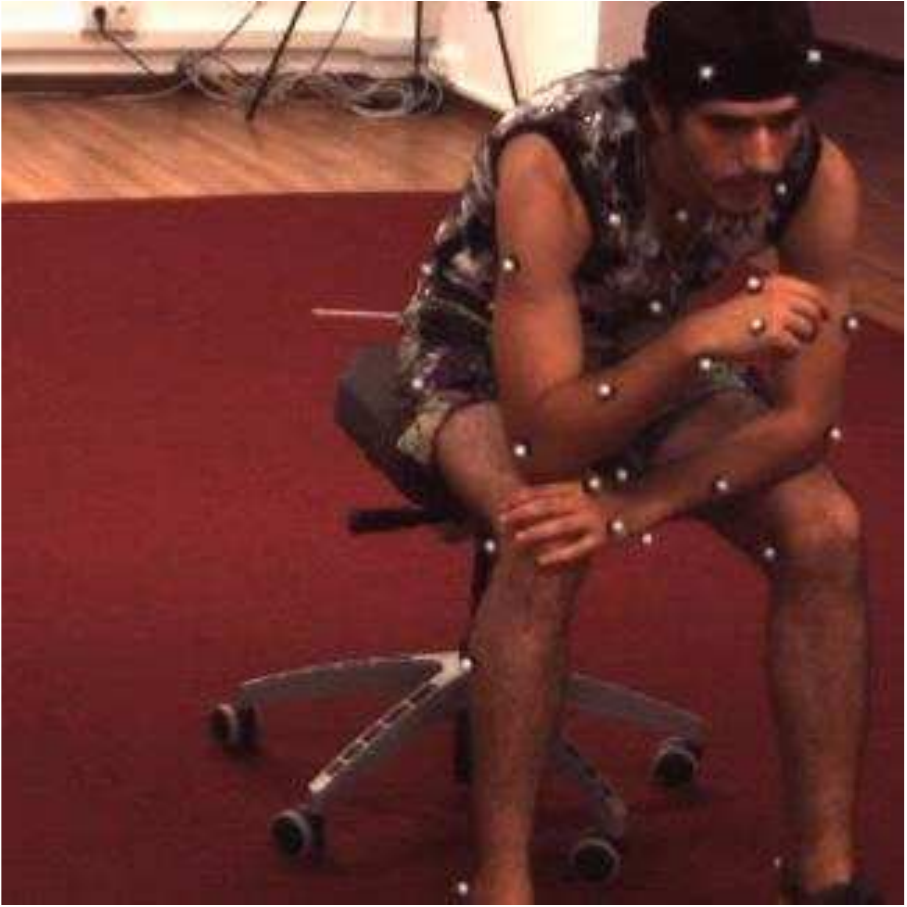}
    \includegraphics[height=0.116\textwidth]{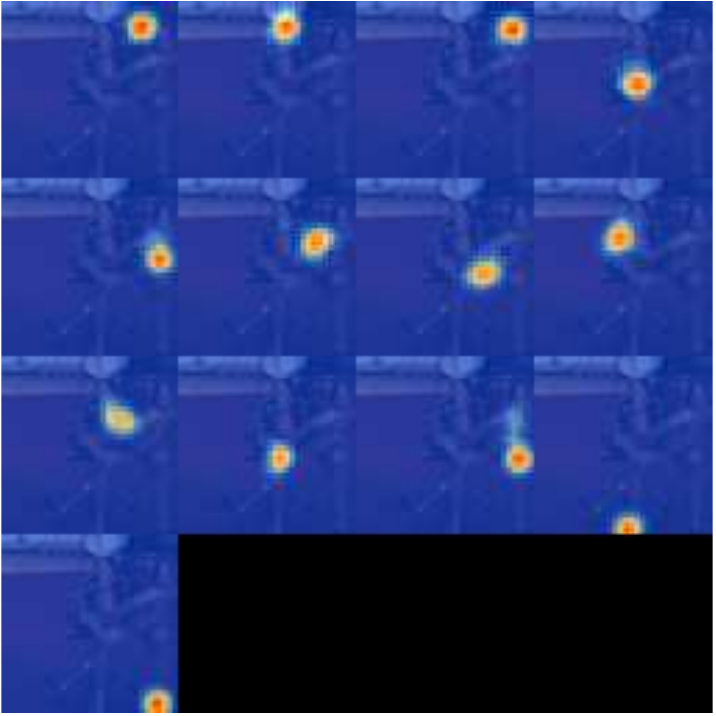}
    \includegraphics[height=0.116\textwidth]{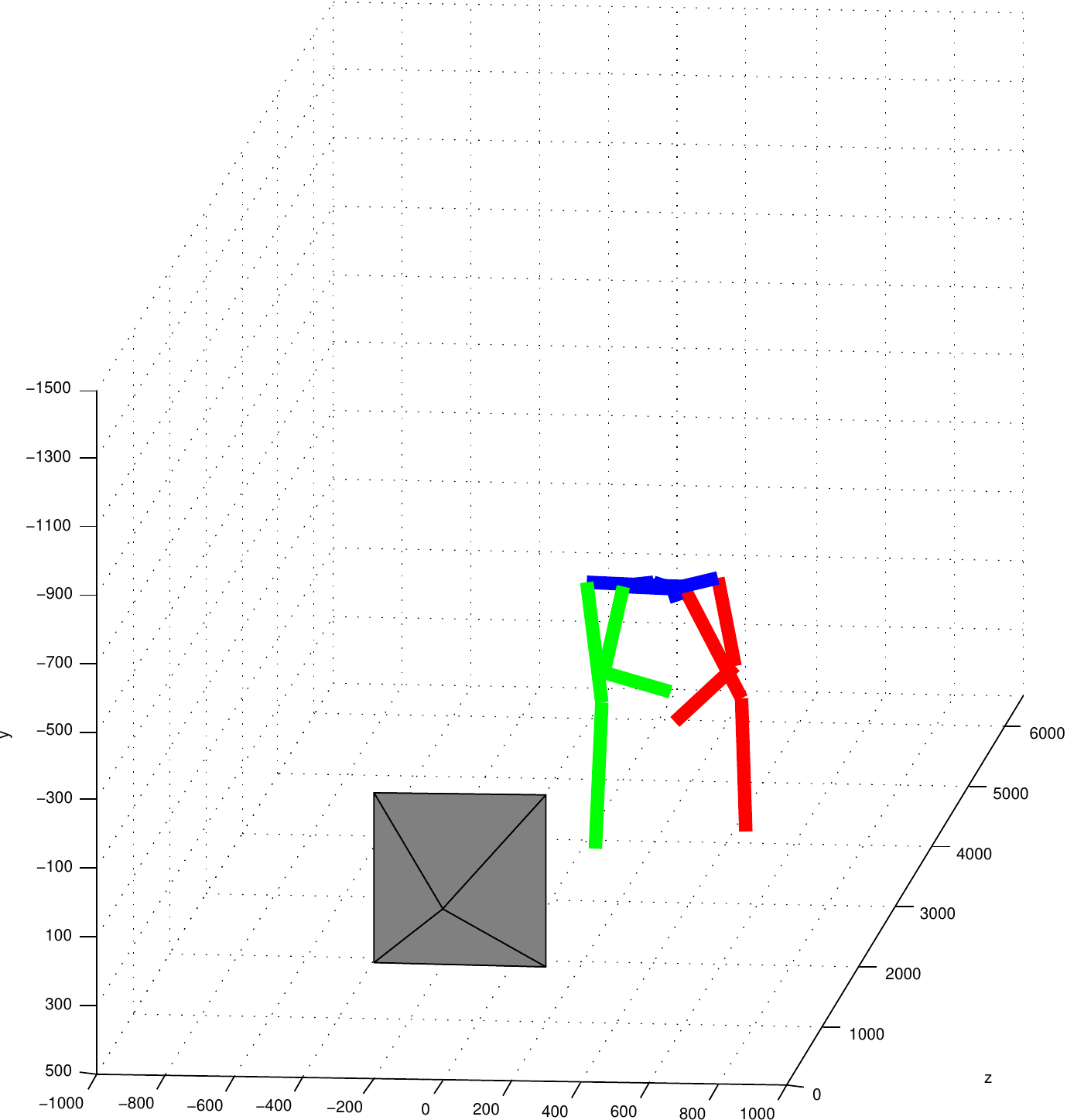}
    \includegraphics[height=0.116\textwidth]{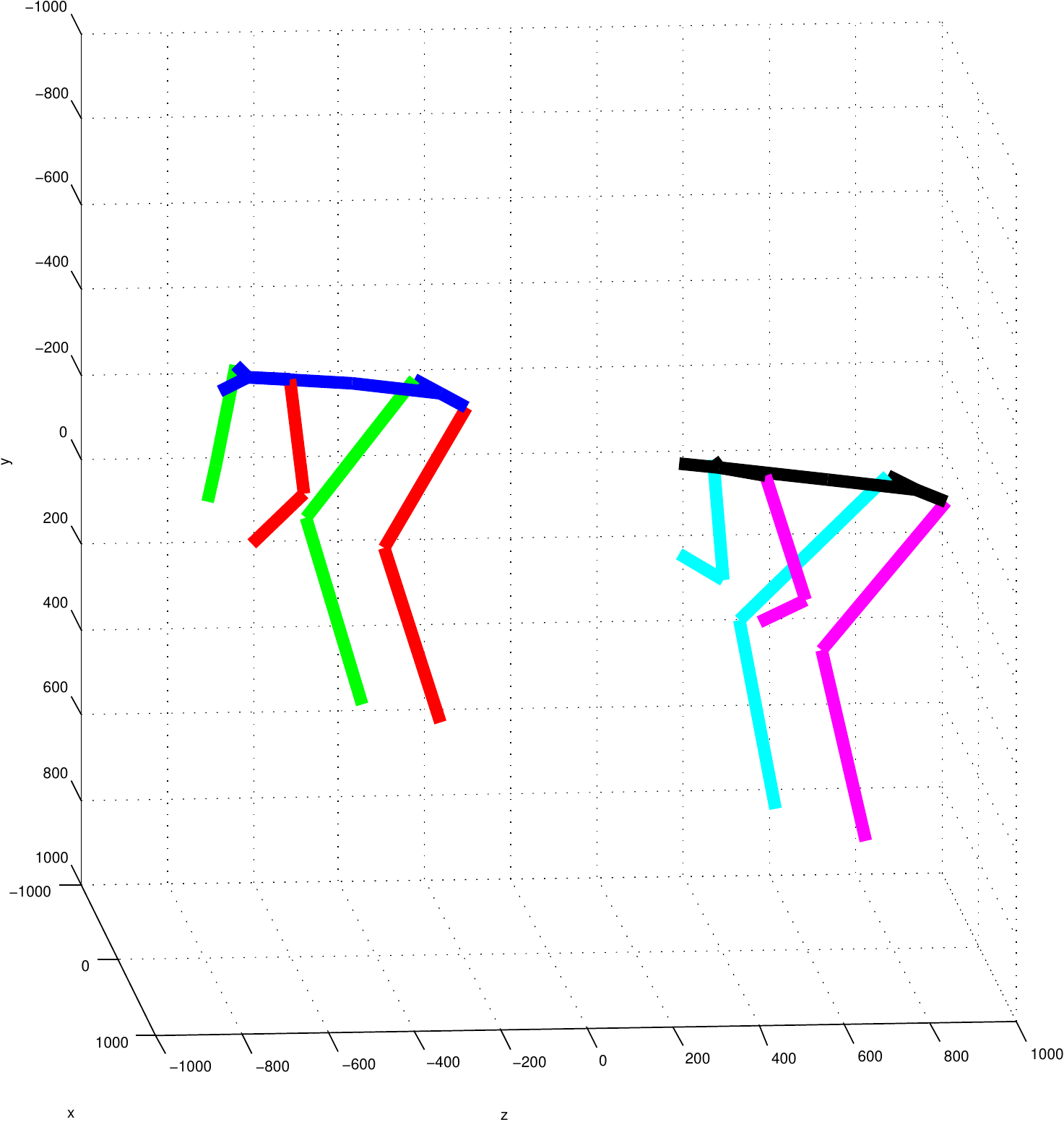}
  \end{tabular}
  % \vspace{-2mm}
  \caption{\small Qualitative results of 3D pose recovery. Each sample shows
the input image, the heatmaps output of the hourglass, the estimated 3D pose
and camera, and a side-by-side comparison with the ground-truth pose (colored
by black, cyan, and magenta).}
  \vspace{-2mm}
  \label{fig:skel3d}
\end{figure*}

\vspace{-3mm}

\paragraph{Baselines} We compare our 3D skeleton converter with two top-down
approaches: the convex optimization based approach (Convex) proposed by Zhou
\emph{et al.} \cite{zhou:cvpr2015} and SMPLify \cite{bogo:eccv2016}. Since
Convex assumes a weak perspective camera model, it can only estimate keypoint
positions relative to their center up to a scaling factor. To generate poses
with absolute scale, we first learn a prior on the length of human body limbs
using the training data in Human3.6M, and scale their output pose to minimize
the error between the predicted limb lengths and the prior. Besides, since
Convex takes input of 2D keypoint coordinates rather than heatmaps, we sample
2D coordinates for each keypoint by searching for the maximum response in the
heatmap. We also re-train the pose dictionary of Convex using the same training
set of Human3.6M.

\vspace{-3mm}

\paragraph{Results} Tab.~\ref{tab:mpjpe} shows the comparison of our approach
against the baselines on 13 body joints. Our 3D skeleton converter achieves a
lower error on all 13 body joints by a significant margin. The improvement over
Convex is especially significant on the keypoints of knees and ankles (e.g. for
left knee, from 171.7 to 76.5mm, and for left ankle, from 258.5 to 113.4mm). As
pointed out in \cite{wu:eccv2016}, Zhou \emph{et al.}'s method assumes the
input keypoint coordinates to be clean, which is not true for the hourglass
output. Our approach, by training on heatmaps, can be adjusted to noisy input.
Furthermore, our DNN-based, bottom-up approach, without using any pose priors,
enjoys advantages over two top-down baselines, by learning to directly regress
the 3D keypoint positions with a sufficiently complex model and a vast amount
of training data. We show qualitative examples of our reconstructed 3D poses as
well as the estimated camera poses in Fig.~\ref{fig:skel3d}.

\section{Conclusion}

This paper presents the first study on forecasting human dynamics from static
images. Our proposed 3D Pose Forecasting Network (3D-PFNet) integrates recent
advances on single-image human pose estimation and sequence prediction, and
further converts the 2D predictions into 3D space. We train the 3D-PFNet using
a three-step training strategy to leverage a diverse source of training data,
including image and video based human pose datasets and 3D MoCap data. We
demonstrate competitive performance of our 3D-PFNet on 2D pose forecasting and
3D pose recovery through quantitative and qualitative results.

{\small
\bibliographystyle{ieee}
\bibliography{egbib}
}

\appendix

% for arXiv compiler; was after \section{Supplementary Material}
\begin{table*}[t]
  \centering
  \small
  \setlength{\tabcolsep}{4.74pt}
  \begin{tabular}{l c c c c c c c c c c c c c c c c|r}
    \hline \TBstrut
    Timestep \#                      & 1    & 2    & 3    & 4    & 5    & 6    & 7    & 8    & 9    & 10   & 11   & 12   & 13   & 14   & 15   & 16   & \# Tr \\
    \hline\hline \Tstrut
    Baseball pitch                   & 79.7 & 51.2 & 37.4 & 30.3 & 26.3 & 23.6 & 22.2 & 21.5 & 20.8 & 20.6 & 20.5 & 20.7 & 20.8 & 20.7 & 20.6 & 20.5 &  94   \\
    Baseball swing                   & 81.2 & 69.0 & 54.9 & 46.7 & 42.3 & 40.2 & 39.1 & 38.7 & 38.8 & 38.9 & 38.7 & 38.9 & 39.0 & 38.8 & 38.8 & 38.7 & 104   \\
    Bench press                      & 69.1 & 60.6 & 52.6 & 50.1 & 48.8 & 48.7 & 48.9 & 49.3 & 49.9 & 50.5 & 51.3 & 52.1 & 52.9 & 53.6 & 54.1 & 54.3 &  63   \\
    Bowl                             & 68.8 & 53.1 & 41.1 & 34.9 & 31.7 & 30.0 & 28.9 & 28.4 & 27.7 & 27.3 & 27.0 & 26.9 & 26.9 & 27.0 & 26.9 & 27.0 & 123   \\
    Clean and jerk                   & 87.5 & 60.1 & 52.7 & 47.9 & 44.6 & 41.6 & 39.9 & 38.5 & 38.0 & 37.5 & 37.1 & 36.8 & 36.9 & 37.0 & 37.1 & 37.1 &  39   \\
    Golf swing                       & 82.1 & 68.7 & 59.4 & 54.2 & 51.6 & 50.3 & 49.8 & 49.3 & 48.6 & 47.5 & 47.3 & 47.6 & 48.0 & 48.0 & 47.8 & 47.6 &  81   \\
    Jump rope                        & 83.6 & 69.4 & 60.6 & 61.1 & 65.4 & 69.2 & 65.6 & 61.9 & 62.2 & 64.9 & 66.1 & 64.6 & 64.2 & 65.6 & 67.2 & 67.6 &  36   \\
    Jumping jacks                    & 85.0 & 63.9 & 47.1 & 41.3 & 40.7 & 42.9 & 46.7 & 50.0 & 52.6 & 53.9 & 55.4 & 57.9 & 60.5 & 62.9 & 64.9 & 65.5 &  51   \\
    Pullup                           & 81.4 & 65.7 & 50.9 & 44.3 & 42.1 & 42.3 & 43.4 & 44.8 & 46.7 & 48.8 & 50.8 & 52.5 & 54.4 & 55.7 & 56.4 & 56.5 &  89   \\
    Pushup                           & 73.3 & 65.5 & 57.5 & 53.1 & 51.4 & 51.3 & 51.9 & 53.2 & 54.9 & 56.6 & 58.4 & 60.1 & 61.6 & 62.7 & 63.2 & 63.2 &  94   \\
    Situp                            & 67.1 & 48.0 & 41.6 & 38.9 & 37.6 & 37.1 & 37.4 & 38.0 & 39.0 & 39.6 & 40.4 & 41.2 & 41.8 & 42.3 & 42.6 & 42.8 &  45   \\
    Squat                            & 81.3 & 58.4 & 46.1 & 42.3 & 40.8 & 41.1 & 42.3 & 43.7 & 45.5 & 47.4 & 49.3 & 51.2 & 53.0 & 54.8 & 56.0 & 56.0 & 104   \\
    Strum guitar                     & 62.4 & 61.6 & 61.5 & 61.2 & 61.1 & 61.6 & 61.1 & 60.7 & 60.3 & 60.2 & 59.7 & 59.2 & 58.6 & 58.5 & 58.4 & 58.3 &  42   \\
    Tennis forehand                  & 80.9 & 59.3 & 40.8 & 31.7 & 27.4 & 24.7 & 22.9 & 22.0 & 21.0 & 20.5 & 20.1 & 19.9 & 19.8 & 19.7 & 19.7 & 19.6 &  73   \\
    Tennis serve                     & 78.8 & 56.4 & 41.3 & 34.1 & 29.5 & 26.4 & 24.3 & 22.8 & 21.6 & 20.7 & 20.3 & 20.0 & 20.0 & 20.3 & 20.3 & 20.2 & 104   \\
    \hline
  \end{tabular}
  \vspace{-2mm}
  \caption{\small PCK@0.05 of 3D-PFNet on individual action classes.}
  \vspace{-2mm}
  \label{tab:pck}
\end{table*}

\section{Supplementary Material}

\subsection{Human Character Rendering}

\begin{figure*}[t]
  % \vspace{-2mm}
  \centering
  \footnotesize
  \begin{tabular}{L{0.12\linewidth}@{\hspace{1.0mm}}|L{0.9\linewidth}@{\hspace{-0.0mm}}}
    \includegraphics[height=0.182\textwidth]{forecast_qual-0518-input.jpg}
    &
    \vspace{-2.7mm}
    \begin{tabular}{L{1.0\linewidth}}
      \hspace{-2.7mm}
      \includegraphics[height=0.10\textwidth]{forecast_qual-0518-skel2d_007-01.pdf}\hspace{0.9mm}
      \includegraphics[height=0.10\textwidth]{forecast_qual-0518-skel2d_007-02.pdf}\hspace{0.9mm}
      \includegraphics[height=0.10\textwidth]{forecast_qual-0518-skel2d_007-03.pdf}\hspace{0.9mm}
      \includegraphics[height=0.10\textwidth]{forecast_qual-0518-skel2d_007-05.pdf}\hspace{0.9mm}
      \includegraphics[height=0.10\textwidth]{forecast_qual-0518-skel2d_007-06.pdf}\hspace{0.9mm}
      \includegraphics[height=0.10\textwidth]{forecast_qual-0518-skel2d_007-07.pdf}\hspace{0.9mm}
      \includegraphics[height=0.10\textwidth]{forecast_qual-0518-skel2d_007-09.pdf}\hspace{0.9mm}
      \includegraphics[height=0.10\textwidth]{forecast_qual-0518-skel2d_007-10.pdf}\hspace{0.9mm}
      \includegraphics[height=0.10\textwidth]{forecast_qual-0518-skel2d_007-11.pdf}\hspace{0.9mm}
      \includegraphics[height=0.10\textwidth]{forecast_qual-0518-skel2d_007-13.pdf}\hspace{0.9mm}
      \includegraphics[height=0.10\textwidth]{forecast_qual-0518-skel2d_007-14.pdf}\hspace{0.9mm}
      \includegraphics[height=0.10\textwidth]{forecast_qual-0518-skel2d_007-15.pdf}
      \\
      \includegraphics[height=0.10\textwidth]{forecast_qual-0518-skel3d_007-01.pdf}\hspace{1.9mm}
      \includegraphics[height=0.10\textwidth]{forecast_qual-0518-skel3d_007-03.pdf}\hspace{1.9mm}
      \includegraphics[height=0.10\textwidth]{forecast_qual-0518-skel3d_007-06.pdf}\hspace{1.9mm}
      \includegraphics[height=0.10\textwidth]{forecast_qual-0518-skel3d_007-08.pdf}\hspace{1.9mm}
      \includegraphics[height=0.10\textwidth]{forecast_qual-0518-skel3d_007-11.pdf}\hspace{1.9mm}
      \includegraphics[height=0.10\textwidth]{forecast_qual-0518-skel3d_007-13.pdf}\hspace{1.9mm}
      \includegraphics[height=0.10\textwidth]{forecast_qual-0518-skel3d_007-15.pdf}
      \\
      \hspace{-2.7mm}
      \includegraphics[height=0.10\textwidth]{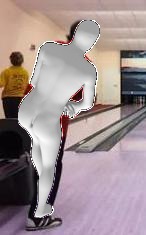}\hspace{0.96mm}
      \includegraphics[height=0.10\textwidth]{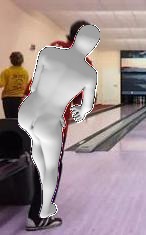}\hspace{0.96mm}
      \includegraphics[height=0.10\textwidth]{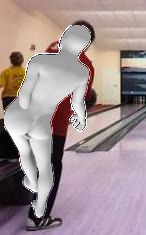}\hspace{0.96mm}
      \includegraphics[height=0.10\textwidth]{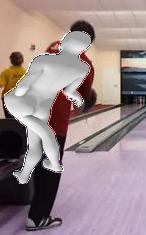}\hspace{0.96mm}
      \includegraphics[height=0.10\textwidth]{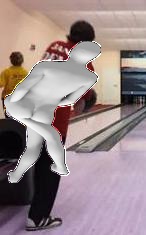}\hspace{0.96mm}
      \includegraphics[height=0.10\textwidth]{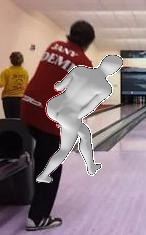}\hspace{0.96mm}
      \includegraphics[height=0.10\textwidth]{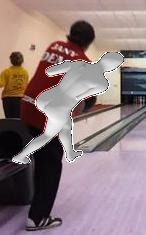}\hspace{0.96mm}
      \includegraphics[height=0.10\textwidth]{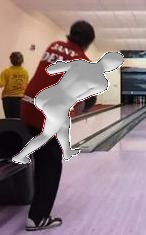}\hspace{0.96mm}
      \includegraphics[height=0.10\textwidth]{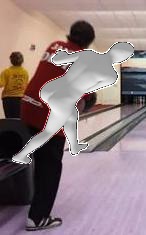}\hspace{0.96mm}
      \includegraphics[height=0.10\textwidth]{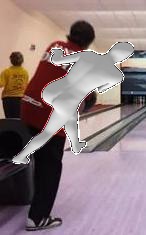}\hspace{0.96mm}
      \includegraphics[height=0.10\textwidth]{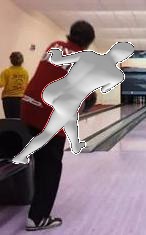}\hspace{0.96mm}
      \includegraphics[height=0.10\textwidth]{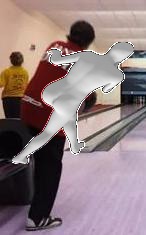}
      \\
      \hspace{-2.7mm}
      \includegraphics[height=0.10\textwidth]{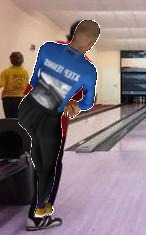}\hspace{0.96mm}
      \includegraphics[height=0.10\textwidth]{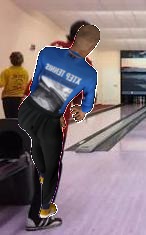}\hspace{0.96mm}
      \includegraphics[height=0.10\textwidth]{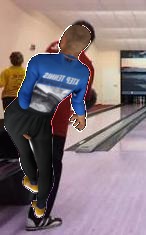}\hspace{0.96mm}
      \includegraphics[height=0.10\textwidth]{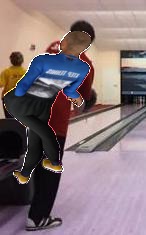}\hspace{0.96mm}
      \includegraphics[height=0.10\textwidth]{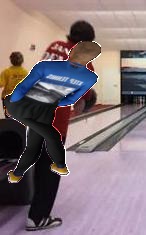}\hspace{0.96mm}
      \includegraphics[height=0.10\textwidth]{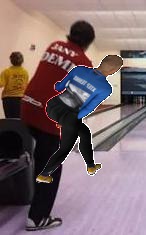}\hspace{0.96mm}
      \includegraphics[height=0.10\textwidth]{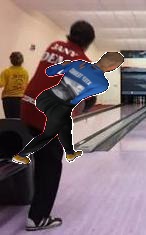}\hspace{0.96mm}
      \includegraphics[height=0.10\textwidth]{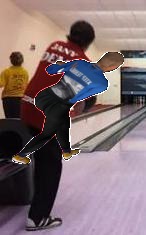}\hspace{0.96mm}
      \includegraphics[height=0.10\textwidth]{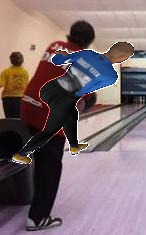}\hspace{0.96mm}
      \includegraphics[height=0.10\textwidth]{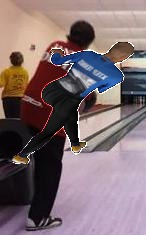}\hspace{0.96mm}
      \includegraphics[height=0.10\textwidth]{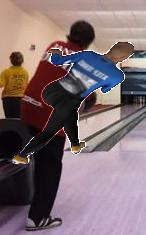}\hspace{0.96mm}
      \includegraphics[height=0.10\textwidth]{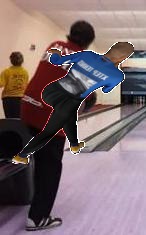}
    \end{tabular}
    \\ [-0.5em] & \\
    \includegraphics[height=0.232\textwidth]{forecast_qual-0787-input.jpg}
    &
    \vspace{-2.7mm}
    \begin{tabular}{L{1.0\linewidth}}
      \hspace{-2.7mm}
      \includegraphics[height=0.10\textwidth]{forecast_qual-0787-skel2d_024-01.pdf}\hspace{0.82mm}
      \includegraphics[height=0.10\textwidth]{forecast_qual-0787-skel2d_024-02.pdf}\hspace{0.82mm}
      \includegraphics[height=0.10\textwidth]{forecast_qual-0787-skel2d_024-03.pdf}\hspace{0.82mm}
      \includegraphics[height=0.10\textwidth]{forecast_qual-0787-skel2d_024-04.pdf}\hspace{0.82mm}
      \includegraphics[height=0.10\textwidth]{forecast_qual-0787-skel2d_024-05.pdf}\hspace{0.82mm}
      \includegraphics[height=0.10\textwidth]{forecast_qual-0787-skel2d_024-06.pdf}\hspace{0.82mm}
      \includegraphics[height=0.10\textwidth]{forecast_qual-0787-skel2d_024-07.pdf}\hspace{0.82mm}
      \includegraphics[height=0.10\textwidth]{forecast_qual-0787-skel2d_024-08.pdf}\hspace{0.82mm}
      \includegraphics[height=0.10\textwidth]{forecast_qual-0787-skel2d_024-09.pdf}\hspace{0.82mm}
      \includegraphics[height=0.10\textwidth]{forecast_qual-0787-skel2d_024-10.pdf}\hspace{0.82mm}
      \includegraphics[height=0.10\textwidth]{forecast_qual-0787-skel2d_024-11.pdf}\hspace{0.82mm}
      \includegraphics[height=0.10\textwidth]{forecast_qual-0787-skel2d_024-12.pdf}\hspace{0.82mm}
      \includegraphics[height=0.10\textwidth]{forecast_qual-0787-skel2d_024-13.pdf}\hspace{0.82mm}
      \includegraphics[height=0.10\textwidth]{forecast_qual-0787-skel2d_024-14.pdf}\hspace{0.82mm}
      \includegraphics[height=0.10\textwidth]{forecast_qual-0787-skel2d_024-15.pdf}
      \\
      \includegraphics[height=0.10\textwidth]{forecast_qual-0787-skel3d_024-01.pdf}\hspace{1.9mm}
      \includegraphics[height=0.10\textwidth]{forecast_qual-0787-skel3d_024-03.pdf}\hspace{1.9mm}
      \includegraphics[height=0.10\textwidth]{forecast_qual-0787-skel3d_024-06.pdf}\hspace{1.9mm}
      \includegraphics[height=0.10\textwidth]{forecast_qual-0787-skel3d_024-08.pdf}\hspace{1.9mm}
      \includegraphics[height=0.10\textwidth]{forecast_qual-0787-skel3d_024-11.pdf}\hspace{1.9mm}
      \includegraphics[height=0.10\textwidth]{forecast_qual-0787-skel3d_024-13.pdf}\hspace{1.9mm}
      \includegraphics[height=0.10\textwidth]{forecast_qual-0787-skel3d_024-15.pdf}
      \\
      \hspace{-2.7mm}
      \includegraphics[height=0.10\textwidth]{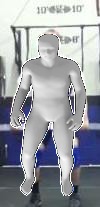}\hspace{0.86mm}
      \includegraphics[height=0.10\textwidth]{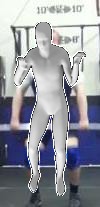}\hspace{0.86mm}
      \includegraphics[height=0.10\textwidth]{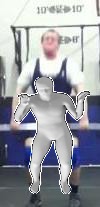}\hspace{0.86mm}
      \includegraphics[height=0.10\textwidth]{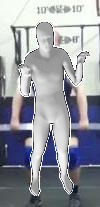}\hspace{0.86mm}
      \includegraphics[height=0.10\textwidth]{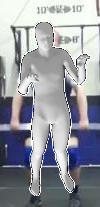}\hspace{0.86mm}
      \includegraphics[height=0.10\textwidth]{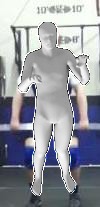}\hspace{0.86mm}
      \includegraphics[height=0.10\textwidth]{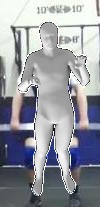}\hspace{0.86mm}
      \includegraphics[height=0.10\textwidth]{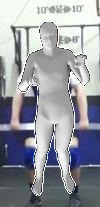}\hspace{0.86mm}
      \includegraphics[height=0.10\textwidth]{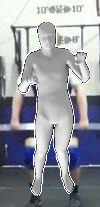}\hspace{0.86mm}
      \includegraphics[height=0.10\textwidth]{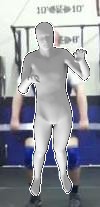}\hspace{0.86mm}
      \includegraphics[height=0.10\textwidth]{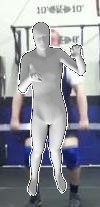}\hspace{0.86mm}
      \includegraphics[height=0.10\textwidth]{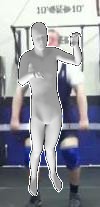}\hspace{0.86mm}
      \includegraphics[height=0.10\textwidth]{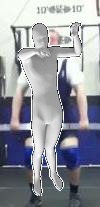}\hspace{0.86mm}
      \includegraphics[height=0.10\textwidth]{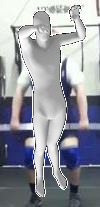}\hspace{0.86mm}
      \includegraphics[height=0.10\textwidth]{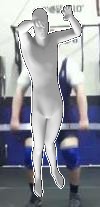}
      \\
      \hspace{-2.7mm}
      \includegraphics[height=0.10\textwidth]{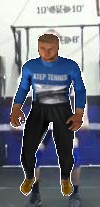}\hspace{0.86mm}
      \includegraphics[height=0.10\textwidth]{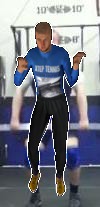}\hspace{0.86mm}
      \includegraphics[height=0.10\textwidth]{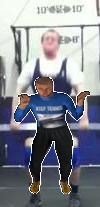}\hspace{0.86mm}
      \includegraphics[height=0.10\textwidth]{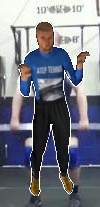}\hspace{0.86mm}
      \includegraphics[height=0.10\textwidth]{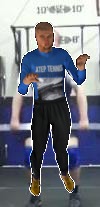}\hspace{0.86mm}
      \includegraphics[height=0.10\textwidth]{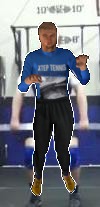}\hspace{0.86mm}
      \includegraphics[height=0.10\textwidth]{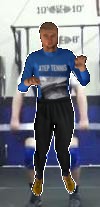}\hspace{0.86mm}
      \includegraphics[height=0.10\textwidth]{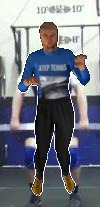}\hspace{0.86mm}
      \includegraphics[height=0.10\textwidth]{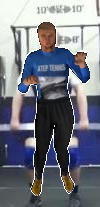}\hspace{0.86mm}
      \includegraphics[height=0.10\textwidth]{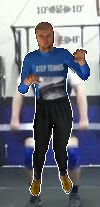}\hspace{0.86mm}
      \includegraphics[height=0.10\textwidth]{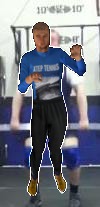}\hspace{0.86mm}
      \includegraphics[height=0.10\textwidth]{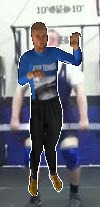}\hspace{0.86mm}
      \includegraphics[height=0.10\textwidth]{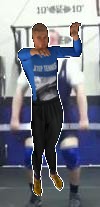}\hspace{0.86mm}
      \includegraphics[height=0.10\textwidth]{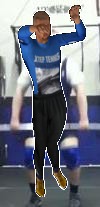}\hspace{0.86mm}
      \includegraphics[height=0.10\textwidth]{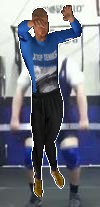}
    \end{tabular}
    \\ [-0.5em] & \\
    \includegraphics[height=0.057\textwidth]{forecast_qual-1582-input.jpg}
    &
    \vspace{-2.7mm}
    \begin{tabular}{L{1.0\linewidth}}
      \hspace{-2.7mm}
      \includegraphics[height=0.07\textwidth]{forecast_qual-1582-skel2d_032-01.pdf}\hspace{0.15mm}
      \includegraphics[height=0.07\textwidth]{forecast_qual-1582-skel2d_032-03.pdf}\hspace{0.15mm}
      \includegraphics[height=0.07\textwidth]{forecast_qual-1582-skel2d_032-05.pdf}\hspace{0.15mm}
      \includegraphics[height=0.07\textwidth]{forecast_qual-1582-skel2d_032-07.pdf}\hspace{0.15mm}
      \includegraphics[height=0.07\textwidth]{forecast_qual-1582-skel2d_032-09.pdf}\hspace{0.15mm}
      \includegraphics[height=0.07\textwidth]{forecast_qual-1582-skel2d_032-11.pdf}
      \\
      \includegraphics[height=0.10\textwidth]{forecast_qual-1582-skel3d_032-01.pdf}\hspace{1.9mm}
      \includegraphics[height=0.10\textwidth]{forecast_qual-1582-skel3d_032-03.pdf}\hspace{1.9mm}
      \includegraphics[height=0.10\textwidth]{forecast_qual-1582-skel3d_032-05.pdf}\hspace{1.9mm}
      \includegraphics[height=0.10\textwidth]{forecast_qual-1582-skel3d_032-06.pdf}\hspace{1.9mm}
      \includegraphics[height=0.10\textwidth]{forecast_qual-1582-skel3d_032-08.pdf}\hspace{1.9mm}
      \includegraphics[height=0.10\textwidth]{forecast_qual-1582-skel3d_032-10.pdf}\hspace{1.9mm}
      \includegraphics[height=0.10\textwidth]{forecast_qual-1582-skel3d_032-12.pdf}
      \\
      \hspace{-2.7mm}
      \includegraphics[height=0.07\textwidth]{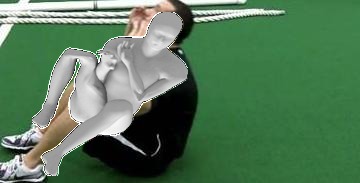}\hspace{0.12mm}
      \includegraphics[height=0.07\textwidth]{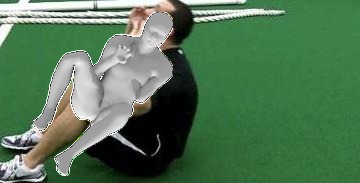}\hspace{0.12mm}
      \includegraphics[height=0.07\textwidth]{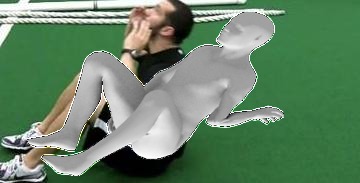}\hspace{0.12mm}
      \includegraphics[height=0.07\textwidth]{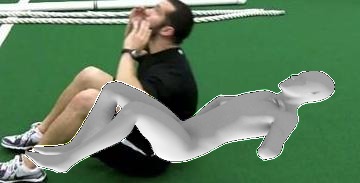}\hspace{0.12mm}
      \includegraphics[height=0.07\textwidth]{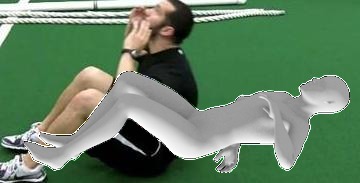}\hspace{0.12mm}
      \includegraphics[height=0.07\textwidth]{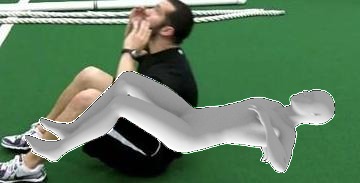}
      \\
      \hspace{-2.7mm}
      \includegraphics[height=0.07\textwidth]{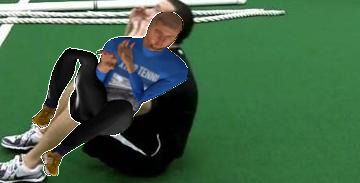}\hspace{0.12mm}
      \includegraphics[height=0.07\textwidth]{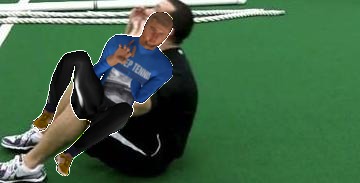}\hspace{0.12mm}
      \includegraphics[height=0.07\textwidth]{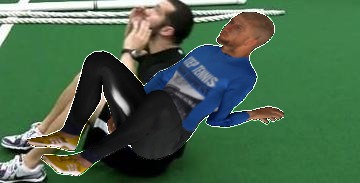}\hspace{0.12mm}
      \includegraphics[height=0.07\textwidth]{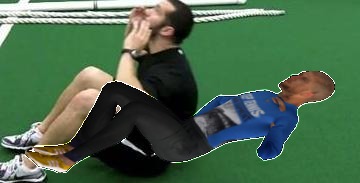}\hspace{0.12mm}
      \includegraphics[height=0.07\textwidth]{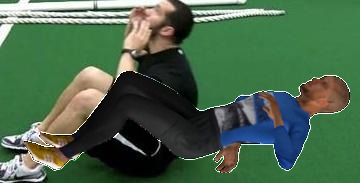}\hspace{0.12mm}
      \includegraphics[height=0.07\textwidth]{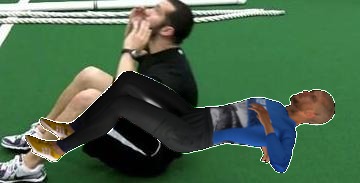}
    \end{tabular}
  \end{tabular}
  % \vspace{-2mm}
  \caption{\small Rendering human characters from the forecasted 3D skeletons.
The left column shows the input images. For each input image, we show in the
right column our forecasted pose sequence in 2D (row 1) and 3D (row 2), and the
rendered human body without texture (row 3) and with skin and cloth textures
(row 4). We use the rendering code provided by \cite{chen:3dv2016}.}
  % \vspace{-2mm}
  \label{fig:render}
\end{figure*}

We demonstrate one potential application of 3D pose forecasting by rendering
human characters from 3D skeletal poses. We use the public code provided by
Chen \emph{et al.} \cite{chen:3dv2016}: We first produce a 3D human shape model
from each 3D skeletal pose using SCAPE. We then transfer skin and clothing
textures to the 3D human model. Finally, the 3D model is rendered and overlaid
on the person's projected bounding box in the input image.
Fig.~\ref{fig:render} shows the rendered human characters, both textureless and
textured, for the qualitative results shown in the Fig.~{\textcolor{red} 8} of
the paper. We believe the capability of pose forecasting with 3D human
rendering may trigger further applications in augmented reality.

\subsection{Performance on Individual Action Classes}

Tab.~\ref{tab:pck} shows the PCK and the number of training videos of each
action class. We see that actions with holistic joint motions (e.g. baseball
pitch) are more challenging for pose forecasting and thus have lower PCK values
even with more training samples, while actions with only partial joint motions
(e.g. jump rope) are the opposite.

\subsection{Additional Qualitative Results}

\begin{figure*}[t]
  % \vspace{-2mm}
  \centering
  \footnotesize
  \begin{tabular}{L{0.12\linewidth}@{\hspace{1.0mm}}|L{0.9\linewidth}@{\hspace{-0.0mm}}}
    \includegraphics[height=0.278\textwidth]{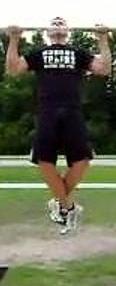}
    &
    \vspace{-2.7mm}
    \begin{tabular}{L{1.0\linewidth}}
      \hspace{-2.7mm}
      \includegraphics[height=0.133\textwidth]{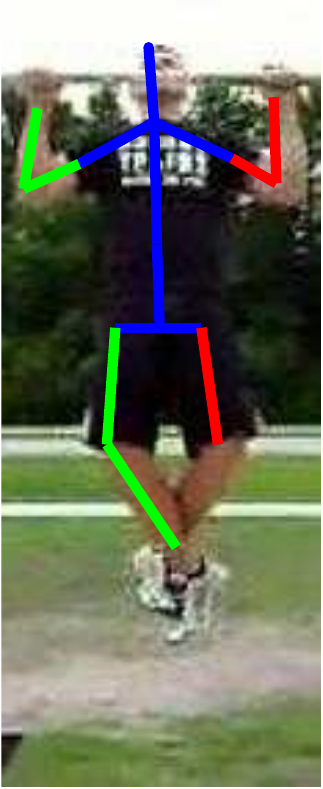}\hspace{2.41mm}
      \includegraphics[height=0.133\textwidth]{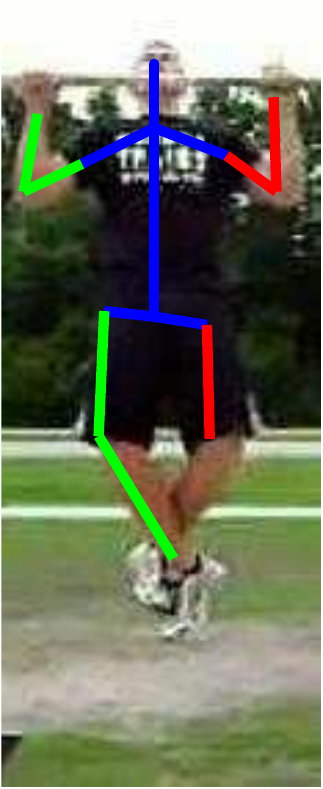}\hspace{2.41mm}
      \includegraphics[height=0.133\textwidth]{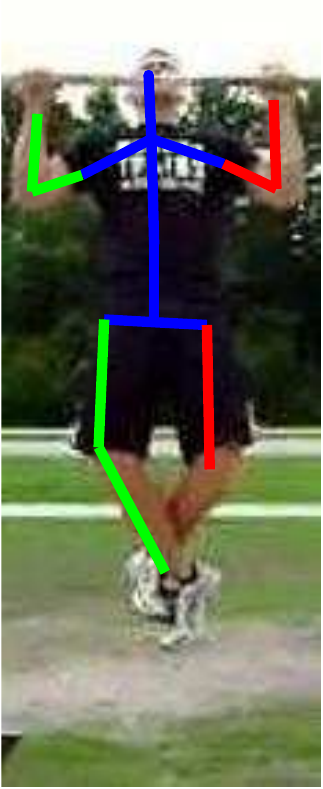}\hspace{2.41mm}
      \includegraphics[height=0.133\textwidth]{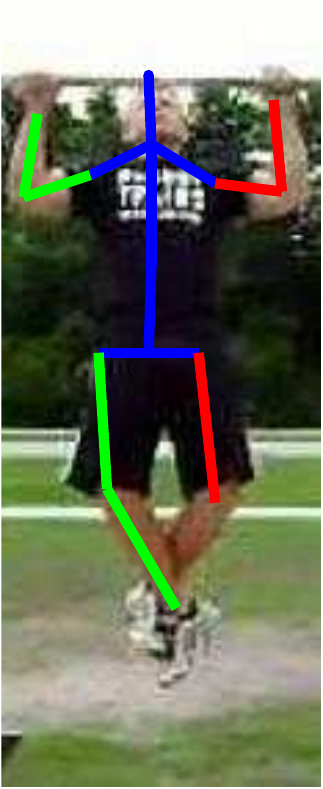}\hspace{2.41mm}
      \includegraphics[height=0.133\textwidth]{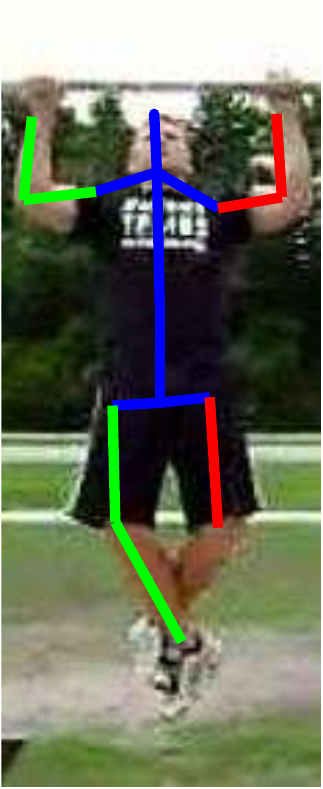}\hspace{2.41mm}
      \includegraphics[height=0.133\textwidth]{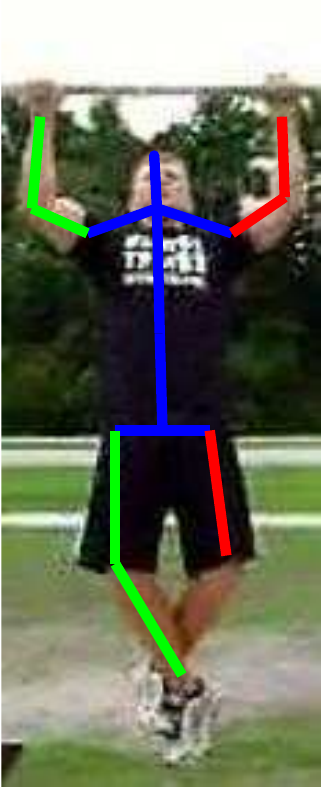}\hspace{2.41mm}
      \includegraphics[height=0.133\textwidth]{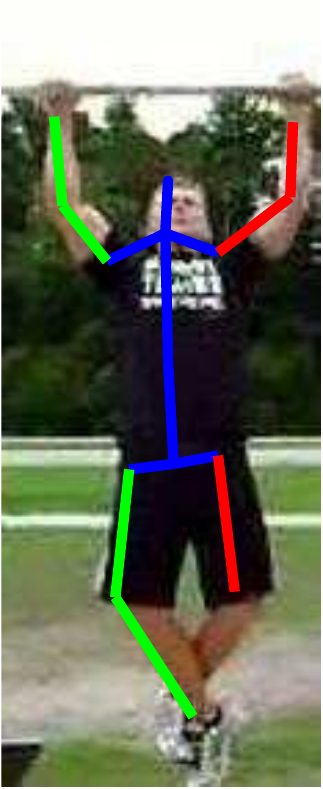}\hspace{2.41mm}
      \includegraphics[height=0.133\textwidth]{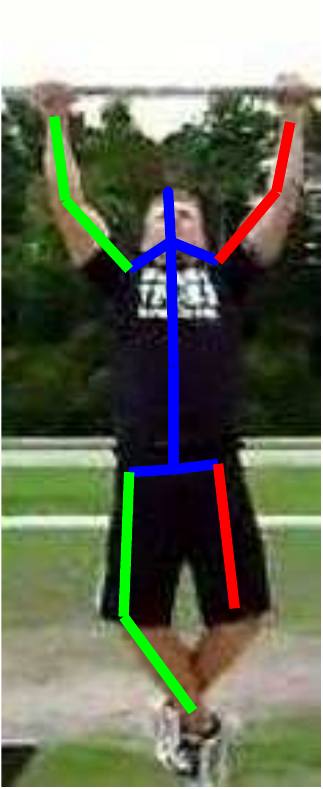}\hspace{2.41mm}
      \includegraphics[height=0.133\textwidth]{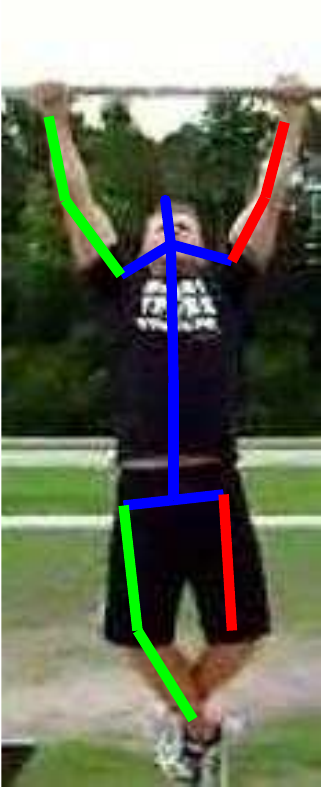}\hspace{2.41mm}
      \includegraphics[height=0.133\textwidth]{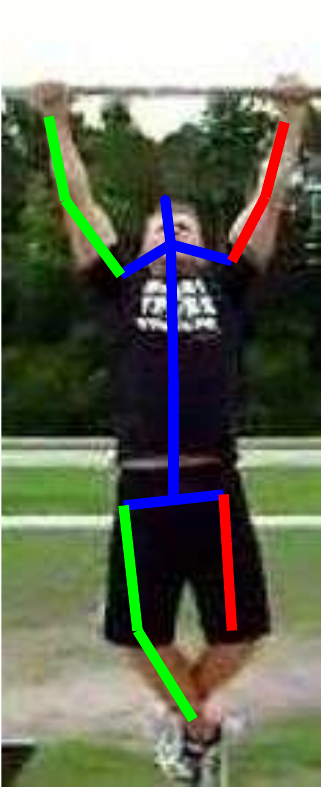}\hspace{2.41mm}
      \includegraphics[height=0.133\textwidth]{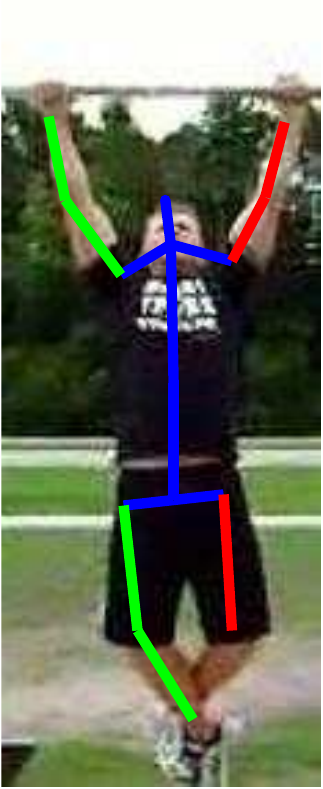}\hspace{2.41mm}
      \includegraphics[height=0.133\textwidth]{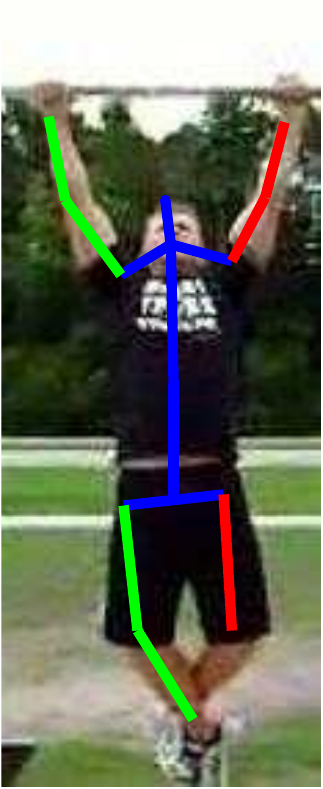}
      \\
      \hspace{-2.7mm}
      \includegraphics[height=0.133\textwidth]{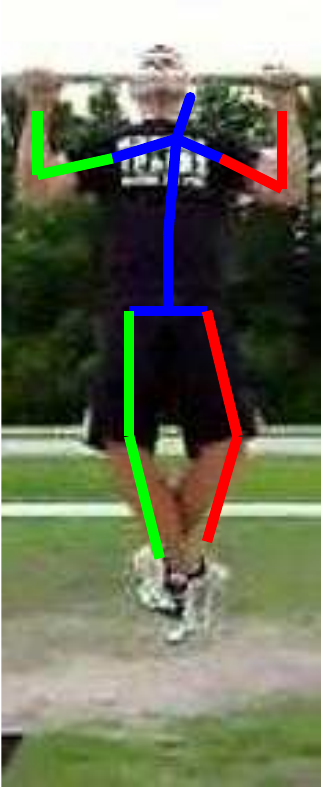}\hspace{2.41mm}
      \includegraphics[height=0.133\textwidth]{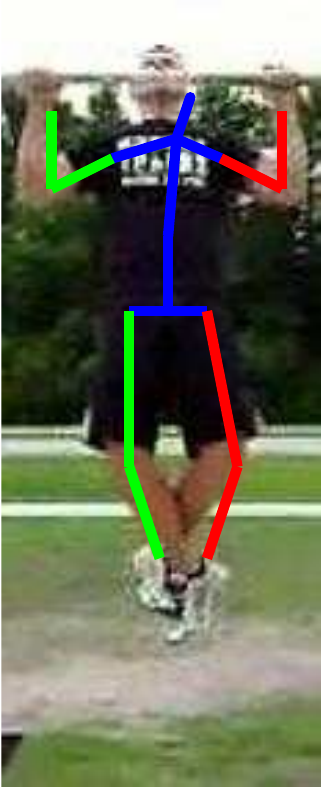}\hspace{2.41mm}
      \includegraphics[height=0.133\textwidth]{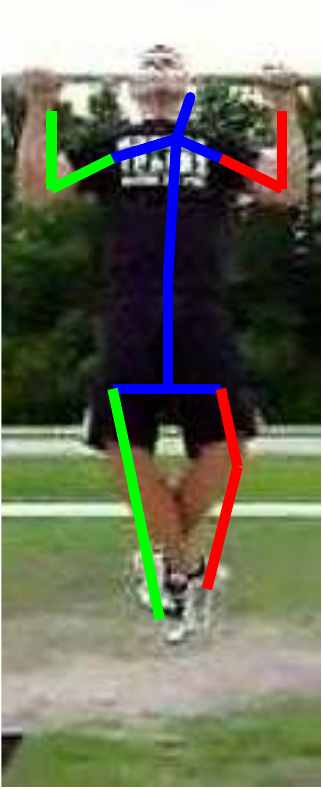}\hspace{2.41mm}
      \includegraphics[height=0.133\textwidth]{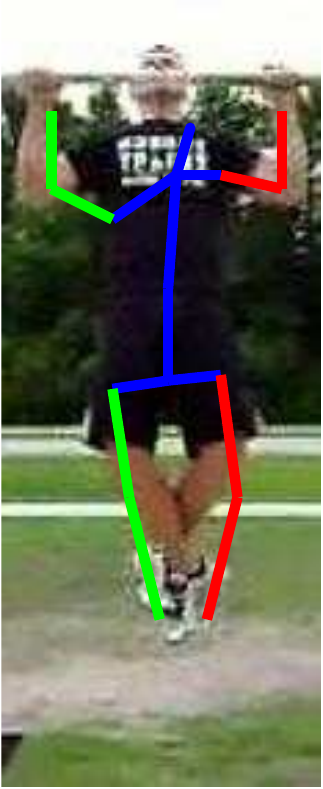}\hspace{2.41mm}
      \includegraphics[height=0.133\textwidth]{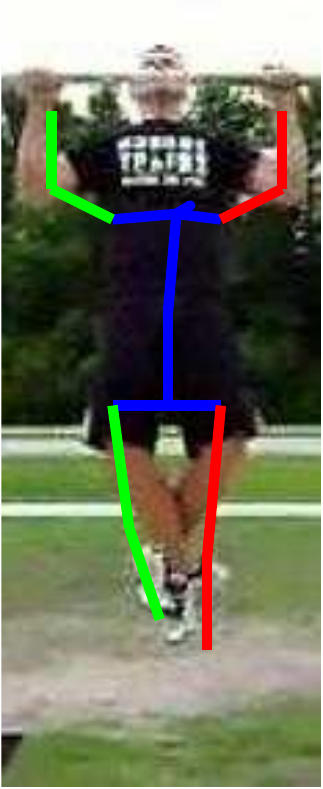}\hspace{2.41mm}
      \includegraphics[height=0.133\textwidth]{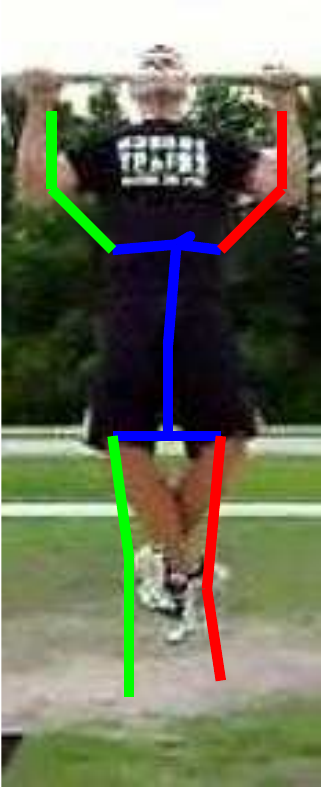}\hspace{2.41mm}
      \includegraphics[height=0.133\textwidth]{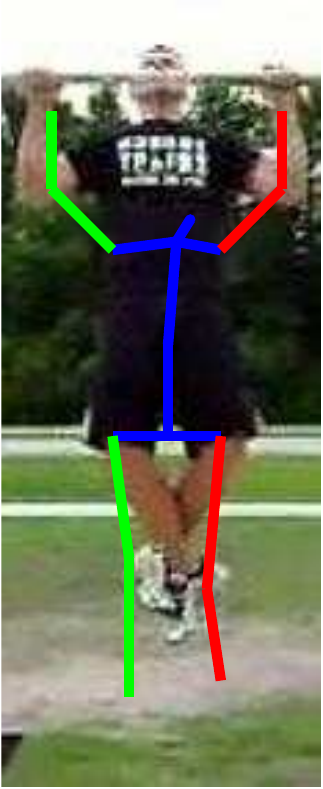}\hspace{2.41mm}
      \includegraphics[height=0.133\textwidth]{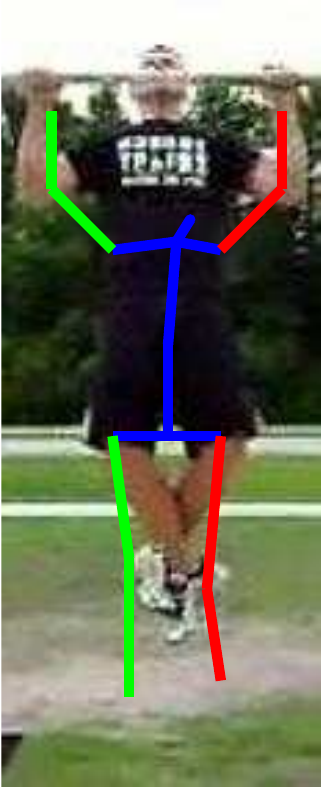}\hspace{2.41mm}
      \includegraphics[height=0.133\textwidth]{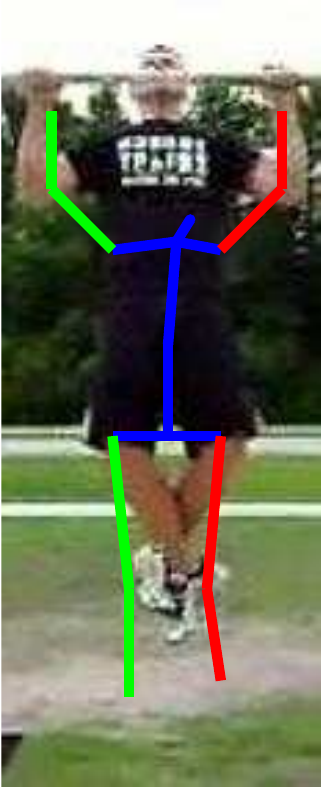}\hspace{2.41mm}
      \includegraphics[height=0.133\textwidth]{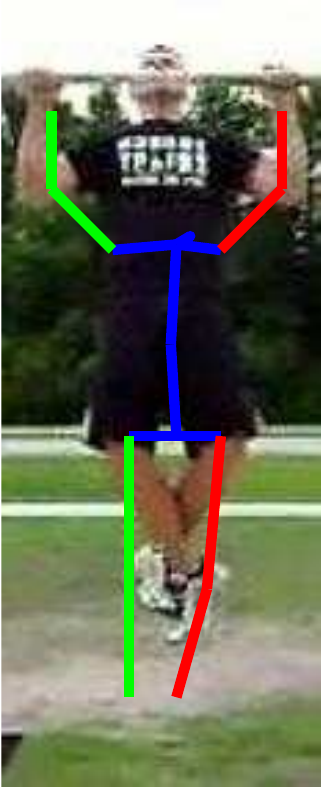}\hspace{2.41mm}
      \includegraphics[height=0.133\textwidth]{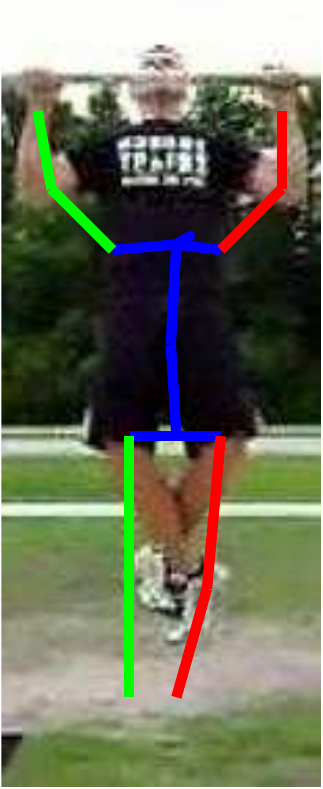}\hspace{2.41mm}
      \includegraphics[height=0.133\textwidth]{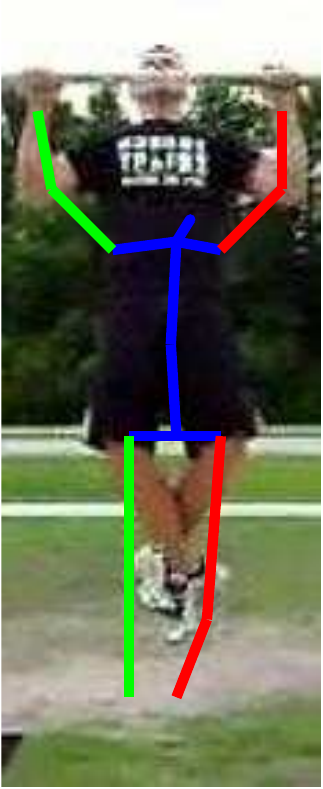}
      \\
      \includegraphics[height=0.10\textwidth]{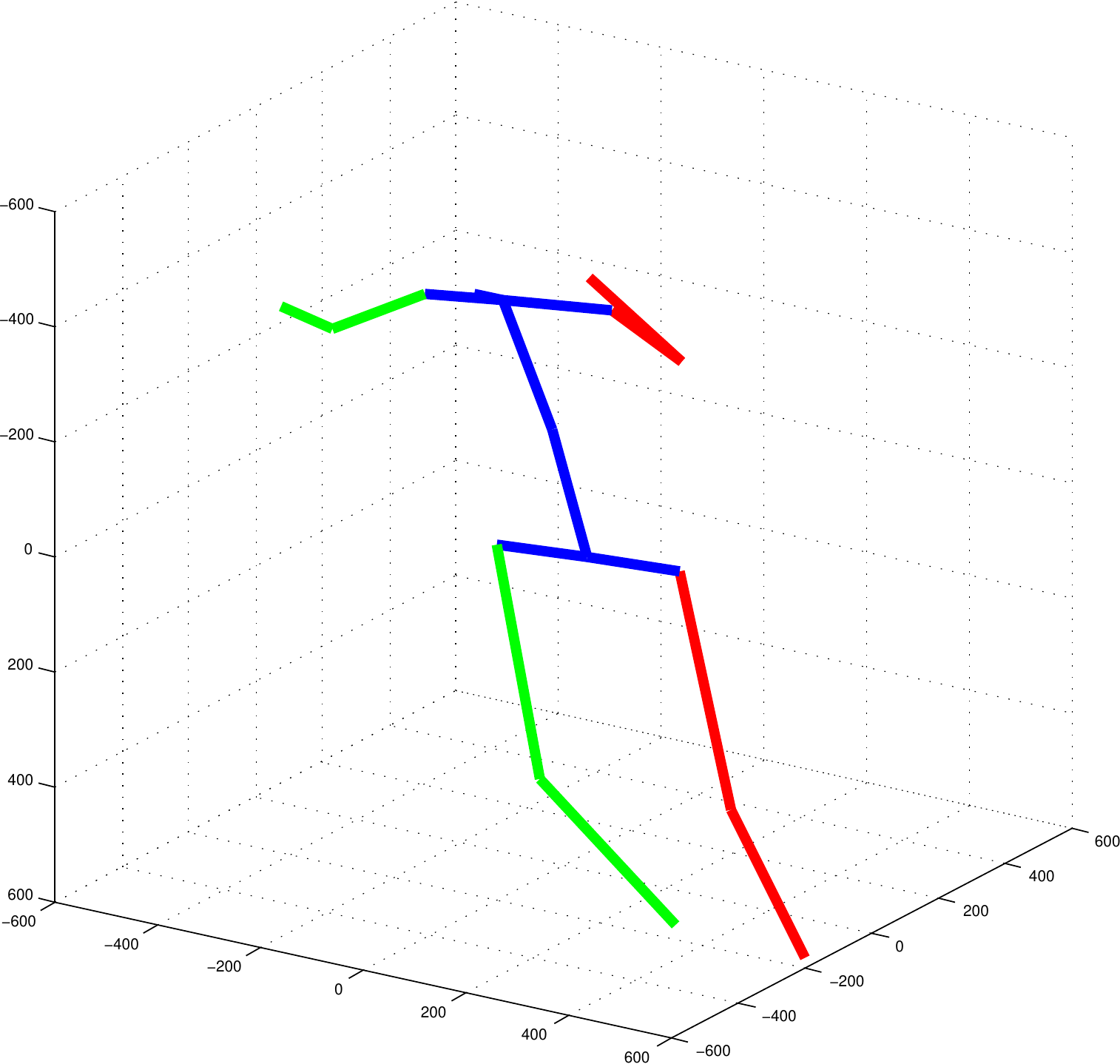}\hspace{1.9mm}
      \includegraphics[height=0.10\textwidth]{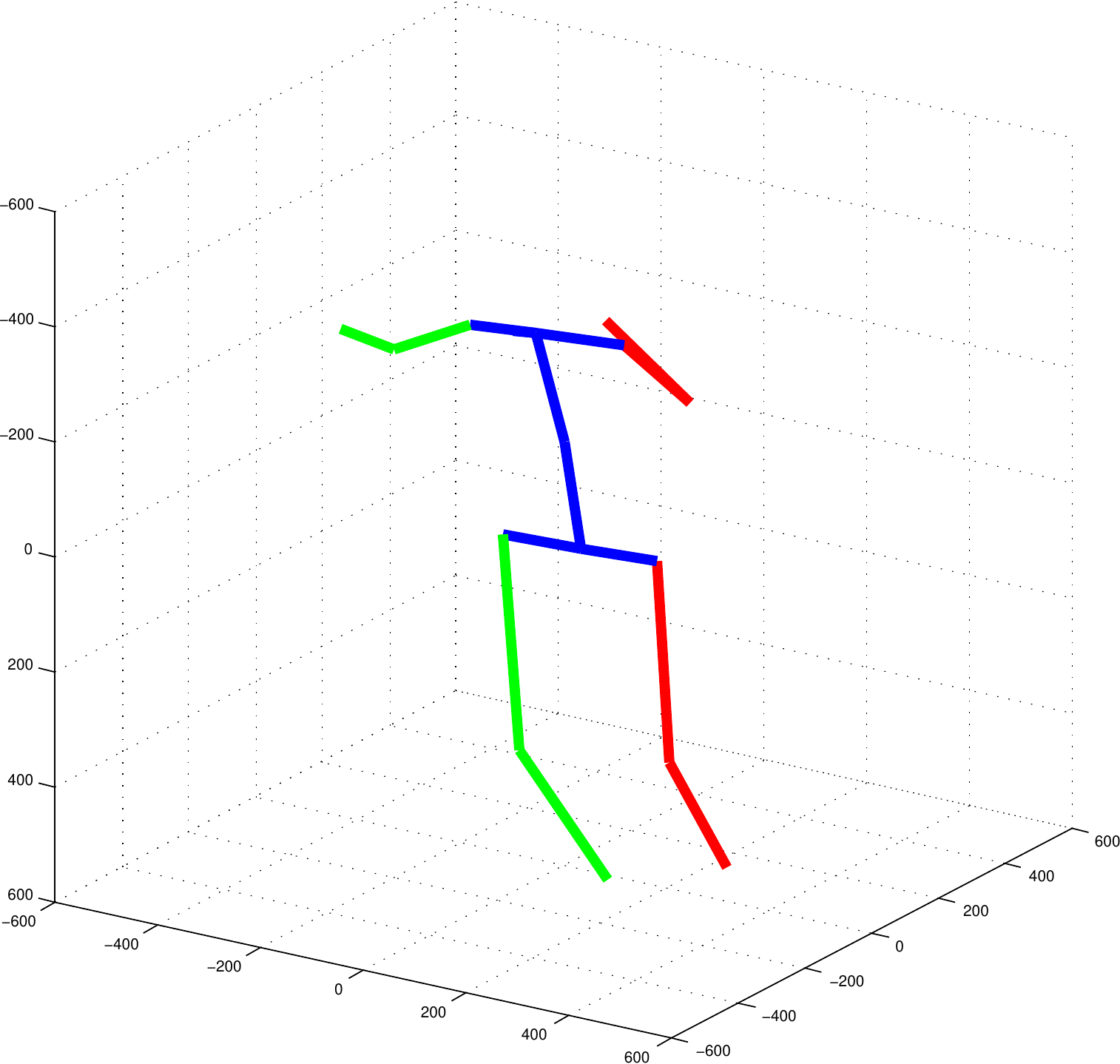}\hspace{1.9mm}
      \includegraphics[height=0.10\textwidth]{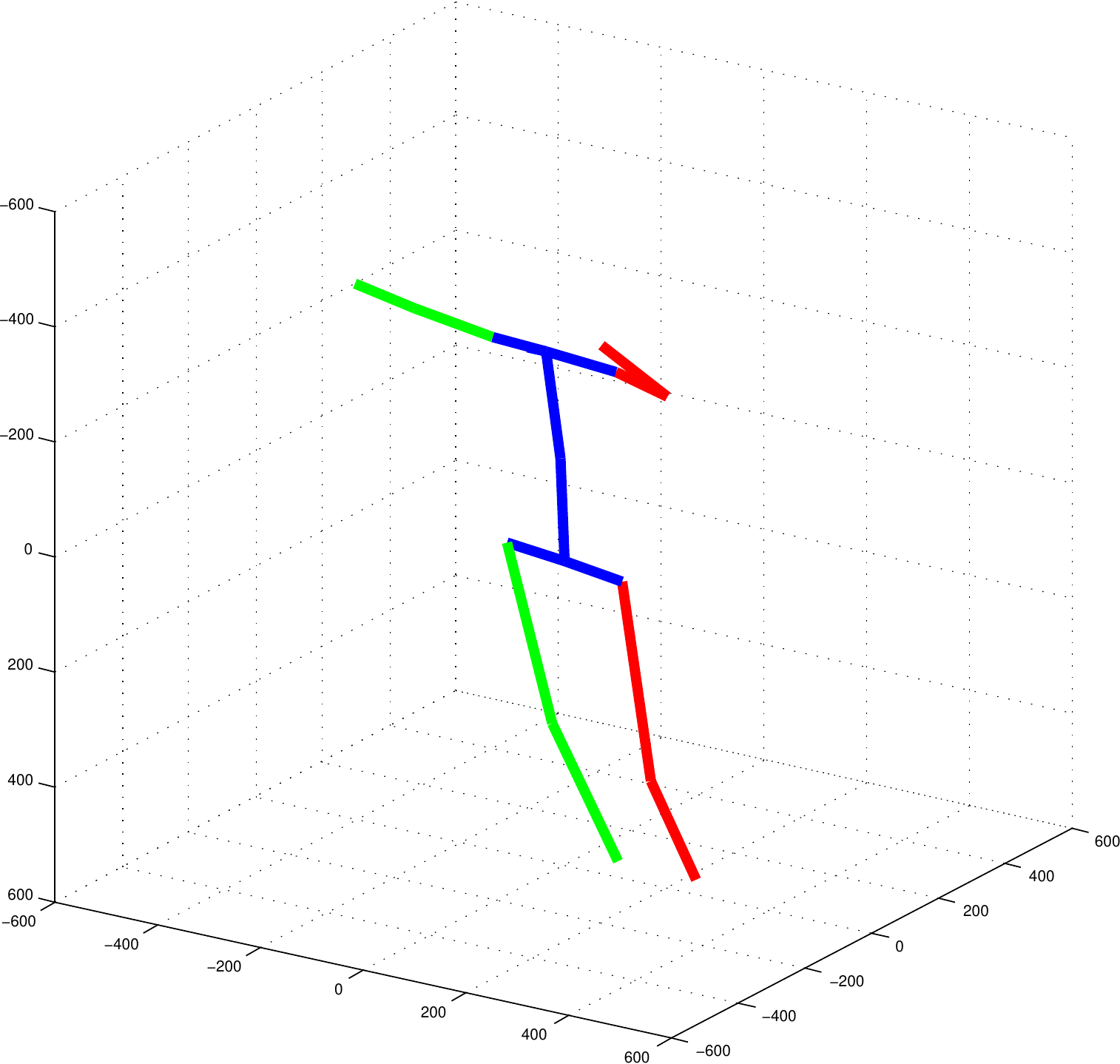}\hspace{1.9mm}
      \includegraphics[height=0.10\textwidth]{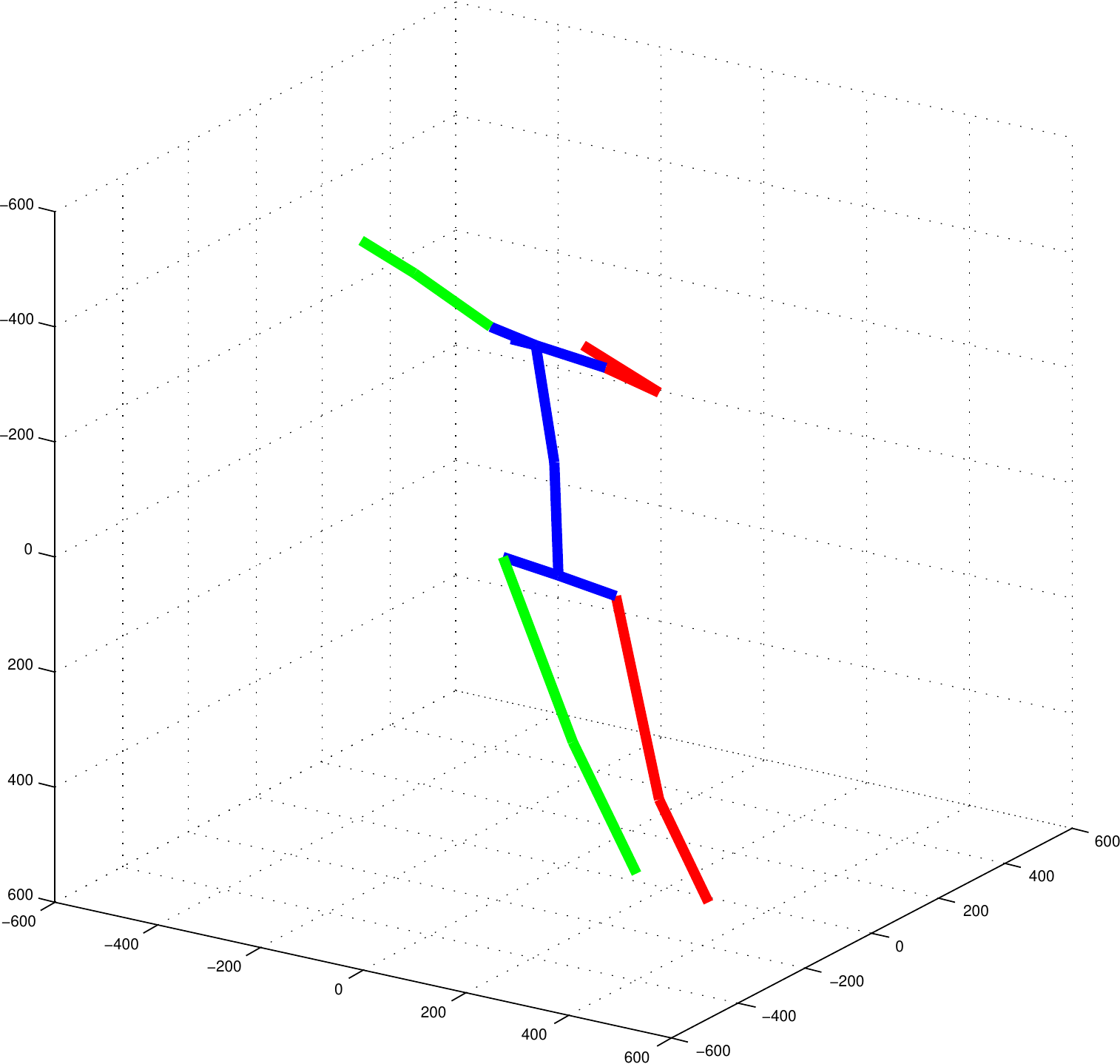}\hspace{1.9mm}
      \includegraphics[height=0.10\textwidth]{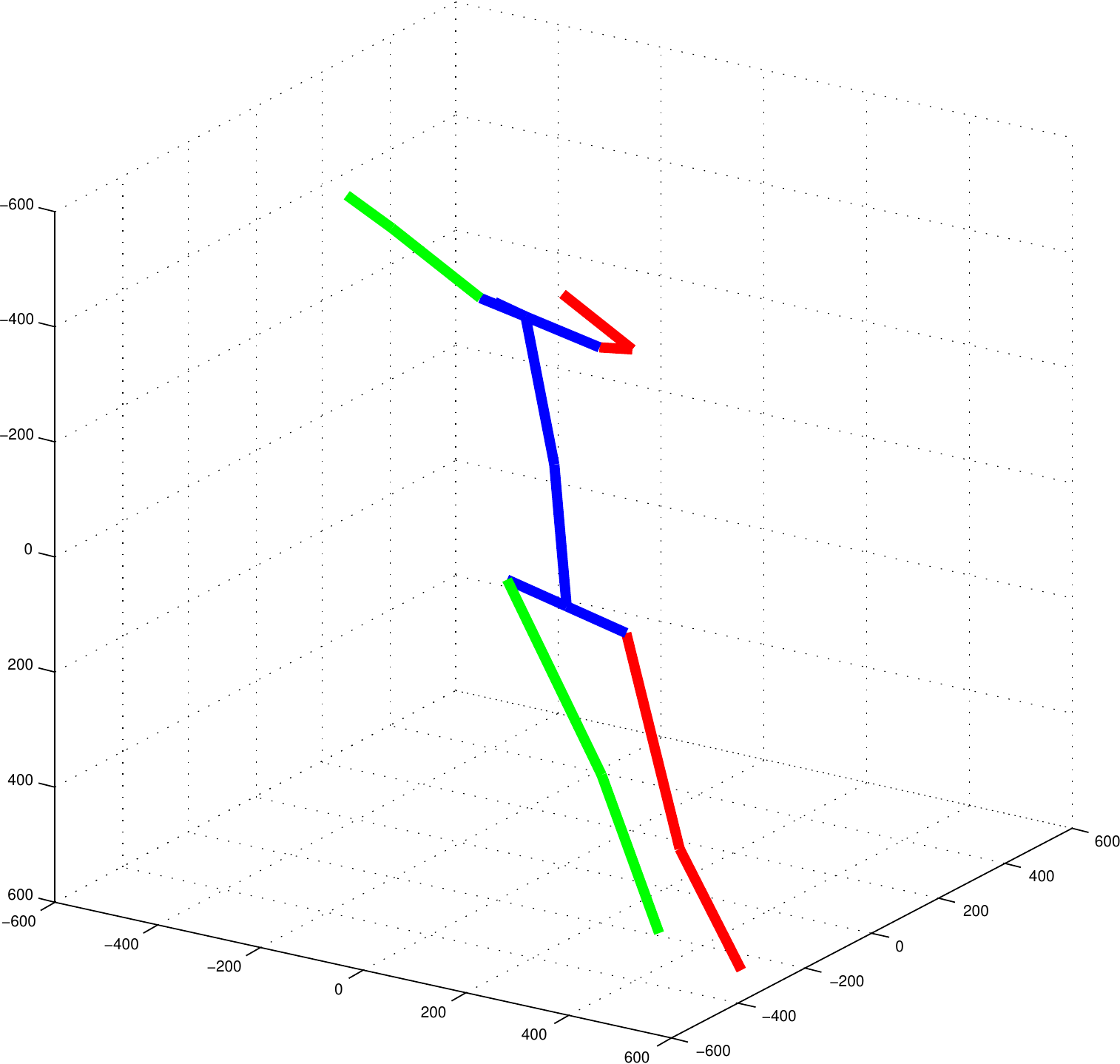}\hspace{1.9mm}
      \includegraphics[height=0.10\textwidth]{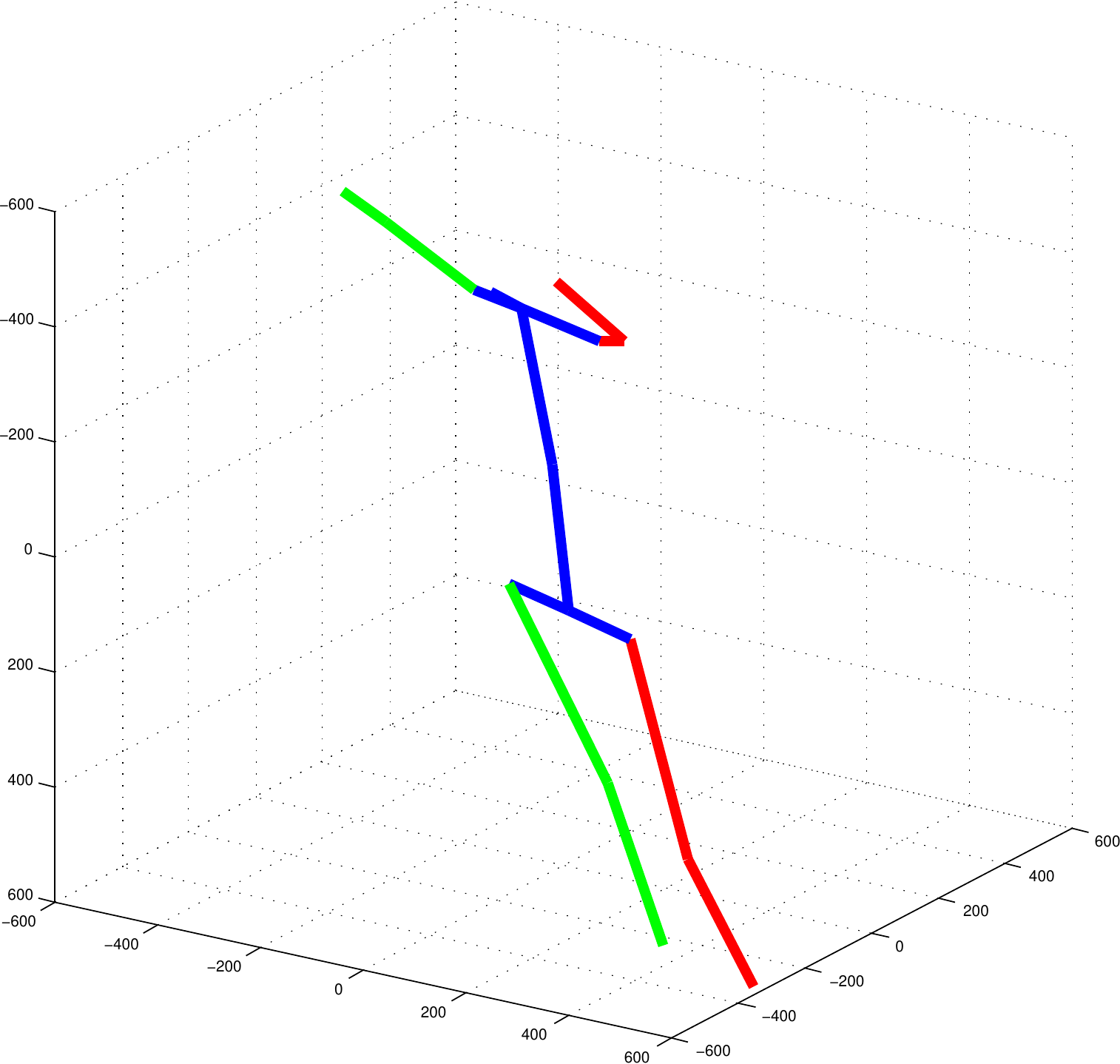}\hspace{1.9mm}
      \includegraphics[height=0.10\textwidth]{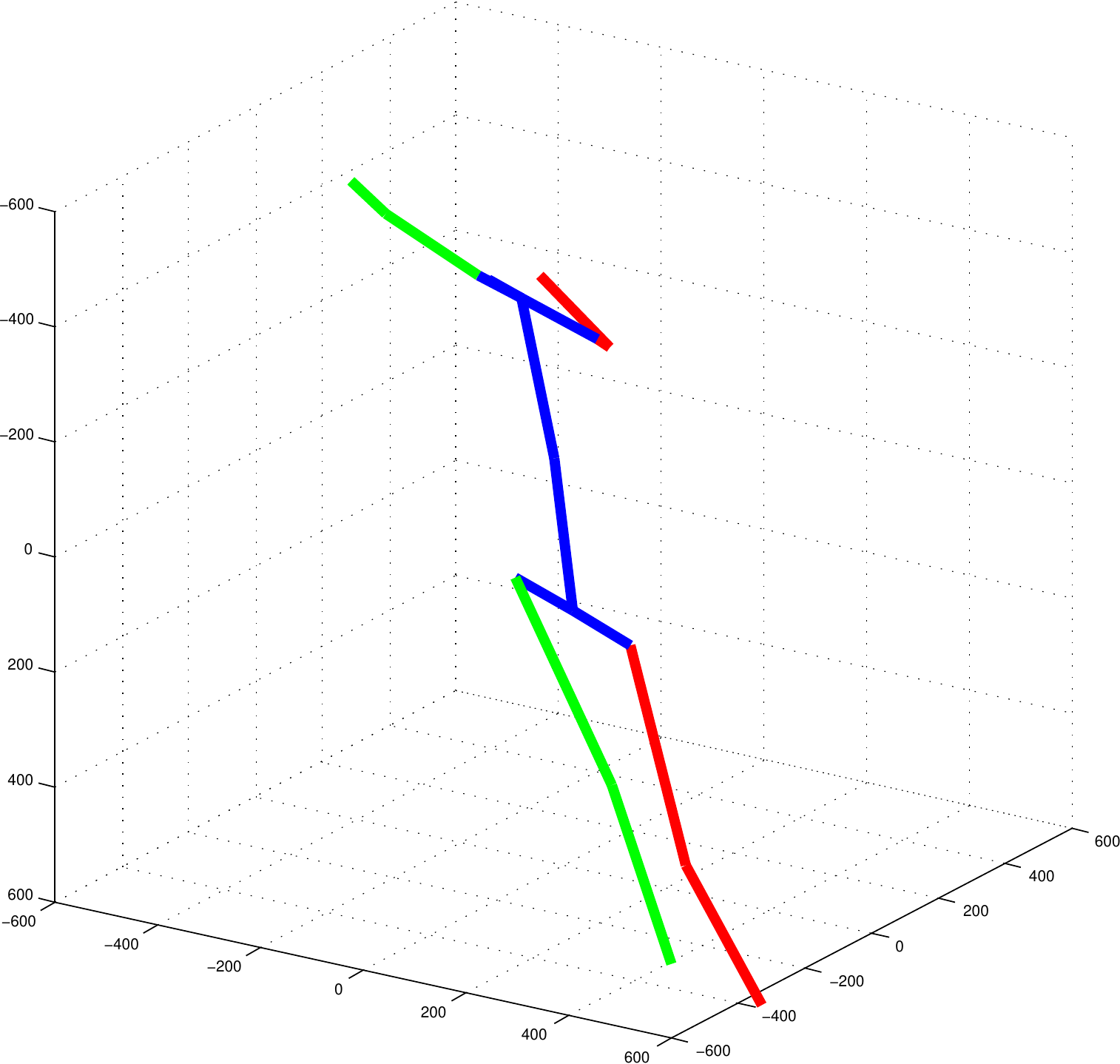}
      \\
      \hspace{-2.7mm}
      \includegraphics[height=0.133\textwidth]{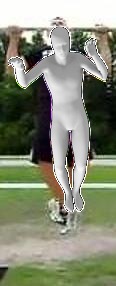}\hspace{2.52mm}
      \includegraphics[height=0.133\textwidth]{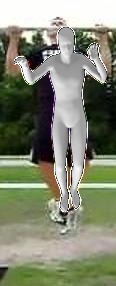}\hspace{2.52mm}
      \includegraphics[height=0.133\textwidth]{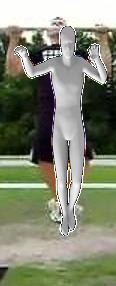}\hspace{2.52mm}
      \includegraphics[height=0.133\textwidth]{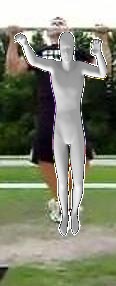}\hspace{2.52mm}
      \includegraphics[height=0.133\textwidth]{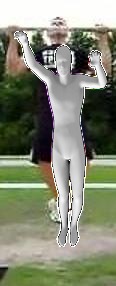}\hspace{2.52mm}
      \includegraphics[height=0.133\textwidth]{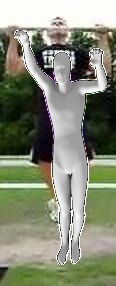}\hspace{2.52mm}
      \includegraphics[height=0.133\textwidth]{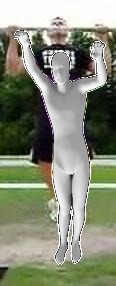}\hspace{2.52mm}
      \includegraphics[height=0.133\textwidth]{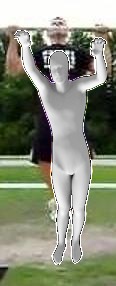}\hspace{2.52mm}
      \includegraphics[height=0.133\textwidth]{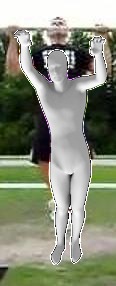}\hspace{2.52mm}
      \includegraphics[height=0.133\textwidth]{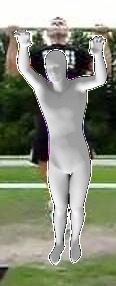}\hspace{2.52mm}
      \includegraphics[height=0.133\textwidth]{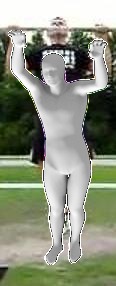}\hspace{2.52mm}
      \includegraphics[height=0.133\textwidth]{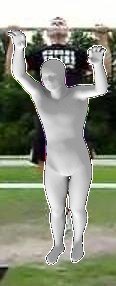}
      \\
      \hspace{-2.7mm}
      \includegraphics[height=0.133\textwidth]{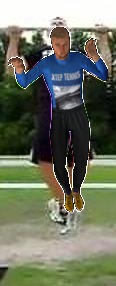}\hspace{2.52mm}
      \includegraphics[height=0.133\textwidth]{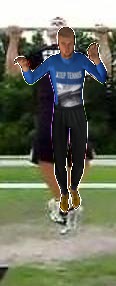}\hspace{2.52mm}
      \includegraphics[height=0.133\textwidth]{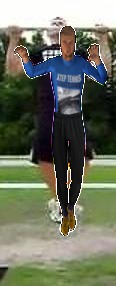}\hspace{2.52mm}
      \includegraphics[height=0.133\textwidth]{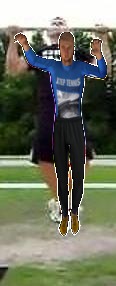}\hspace{2.52mm}
      \includegraphics[height=0.133\textwidth]{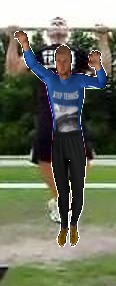}\hspace{2.52mm}
      \includegraphics[height=0.133\textwidth]{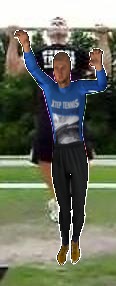}\hspace{2.52mm}
      \includegraphics[height=0.133\textwidth]{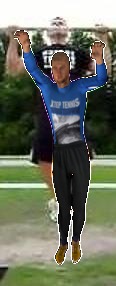}\hspace{2.52mm}
      \includegraphics[height=0.133\textwidth]{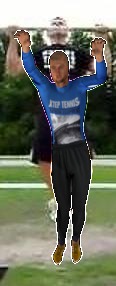}\hspace{2.52mm}
      \includegraphics[height=0.133\textwidth]{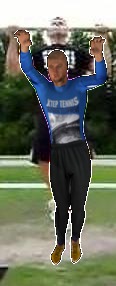}\hspace{2.52mm}
      \includegraphics[height=0.133\textwidth]{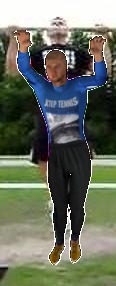}\hspace{2.52mm}
      \includegraphics[height=0.133\textwidth]{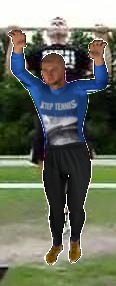}\hspace{2.52mm}
      \includegraphics[height=0.133\textwidth]{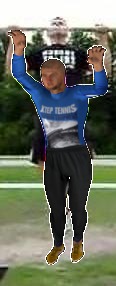}
    \end{tabular}
    \\ [-0.0em] & \\
    \includegraphics[height=0.157\textwidth]{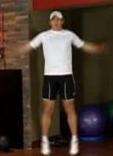}
    &
    \vspace{-2.7mm}
    \begin{tabular}{L{1.0\linewidth}}
      \hspace{-2.7mm}
      \includegraphics[height=0.10\textwidth]{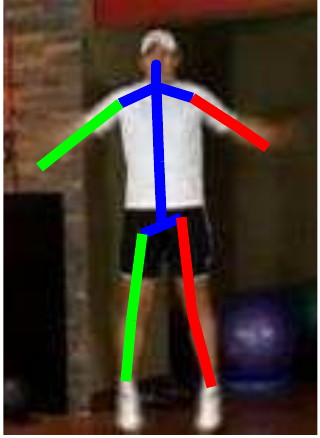}\hspace{0.12mm}
      \includegraphics[height=0.10\textwidth]{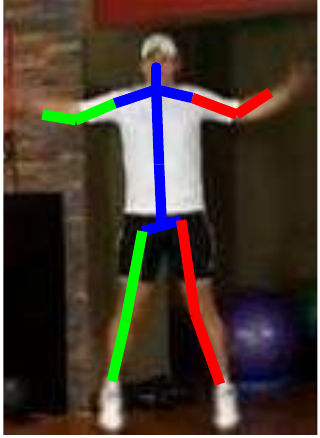}\hspace{0.12mm}
      \includegraphics[height=0.10\textwidth]{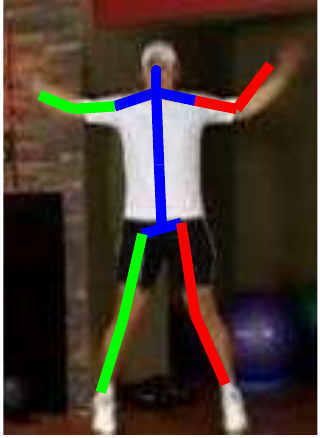}\hspace{0.12mm}
      \includegraphics[height=0.10\textwidth]{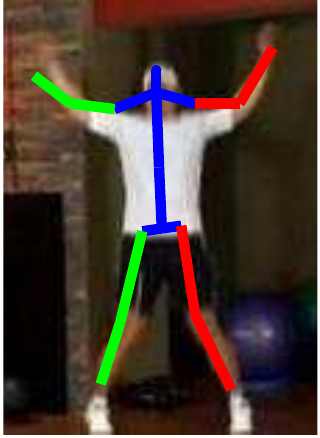}\hspace{0.12mm}
      \includegraphics[height=0.10\textwidth]{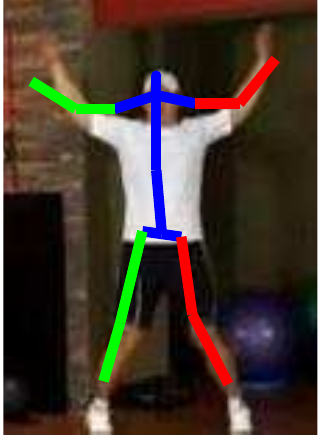}\hspace{0.12mm}
      \includegraphics[height=0.10\textwidth]{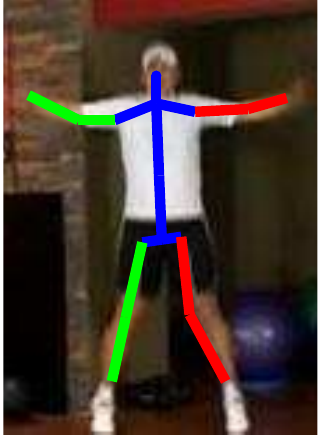}\hspace{0.12mm}
      \includegraphics[height=0.10\textwidth]{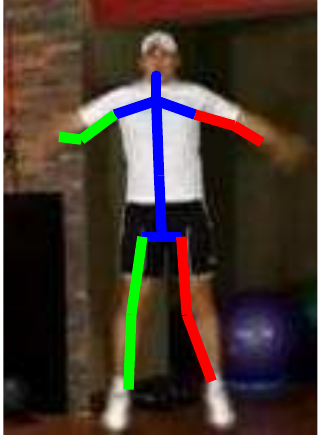}\hspace{0.12mm}
      \includegraphics[height=0.10\textwidth]{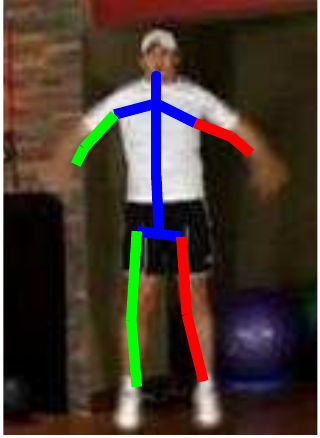}\hspace{0.12mm}
      \includegraphics[height=0.10\textwidth]{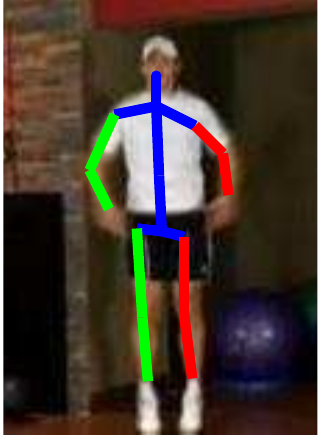}\hspace{0.12mm}
      \includegraphics[height=0.10\textwidth]{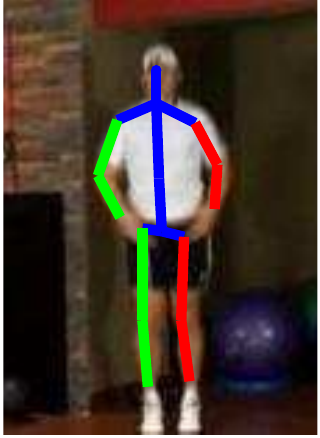}\hspace{0.12mm}
      \includegraphics[height=0.10\textwidth]{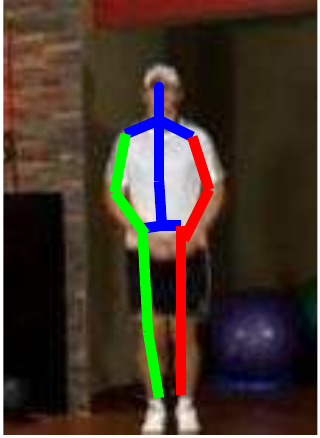}
      \\
      \hspace{-2.7mm}
      \includegraphics[height=0.10\textwidth]{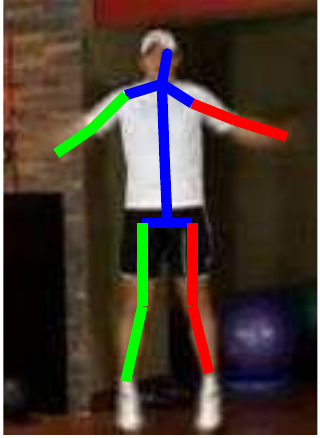}\hspace{0.12mm}
      \includegraphics[height=0.10\textwidth]{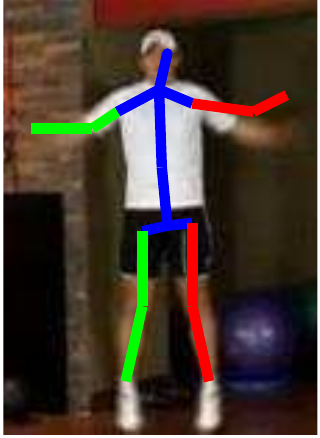}\hspace{0.12mm}
      \includegraphics[height=0.10\textwidth]{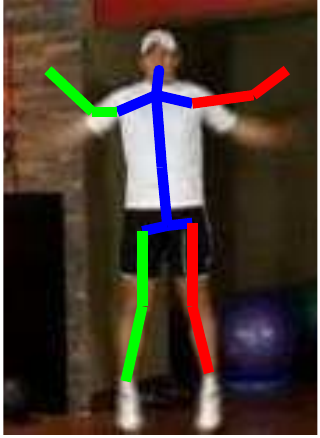}\hspace{0.12mm}
      \includegraphics[height=0.10\textwidth]{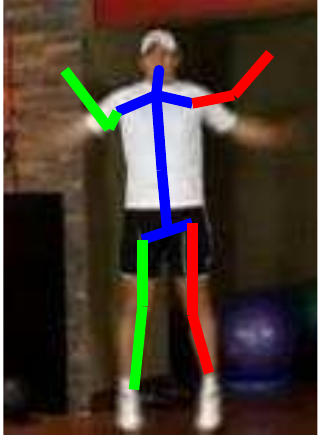}\hspace{0.12mm}
      \includegraphics[height=0.10\textwidth]{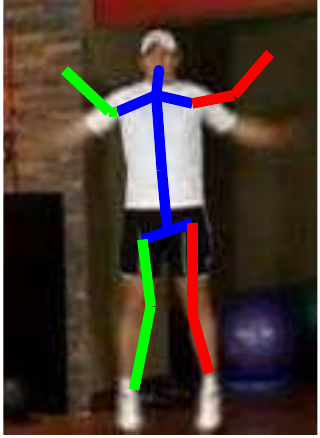}\hspace{0.12mm}
      \includegraphics[height=0.10\textwidth]{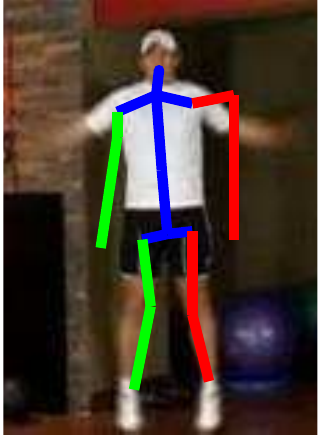}\hspace{0.12mm}
      \includegraphics[height=0.10\textwidth]{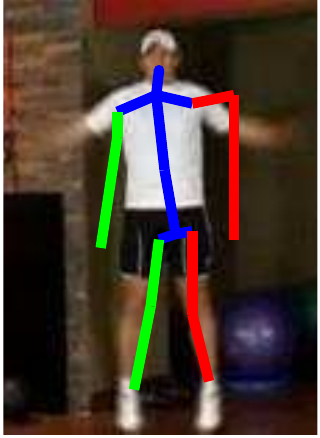}\hspace{0.12mm}
      \includegraphics[height=0.10\textwidth]{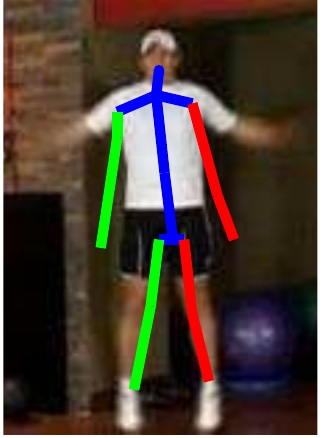}\hspace{0.12mm}
      \includegraphics[height=0.10\textwidth]{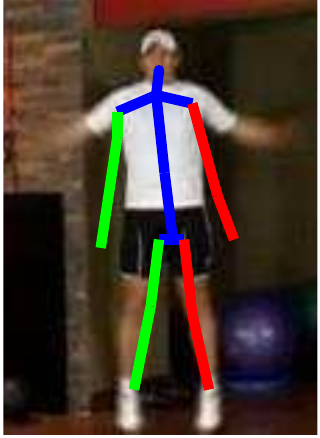}\hspace{0.12mm}
      \includegraphics[height=0.10\textwidth]{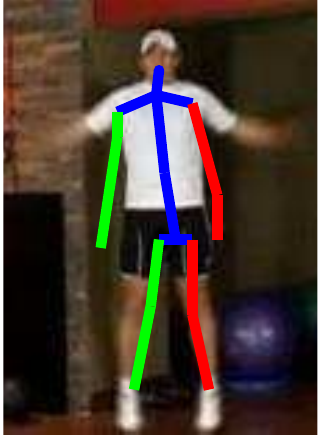}\hspace{0.12mm}
      \includegraphics[height=0.10\textwidth]{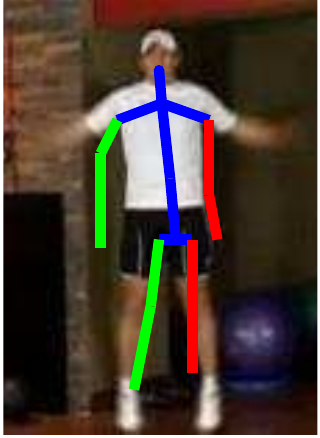}
      \\
      \includegraphics[height=0.10\textwidth]{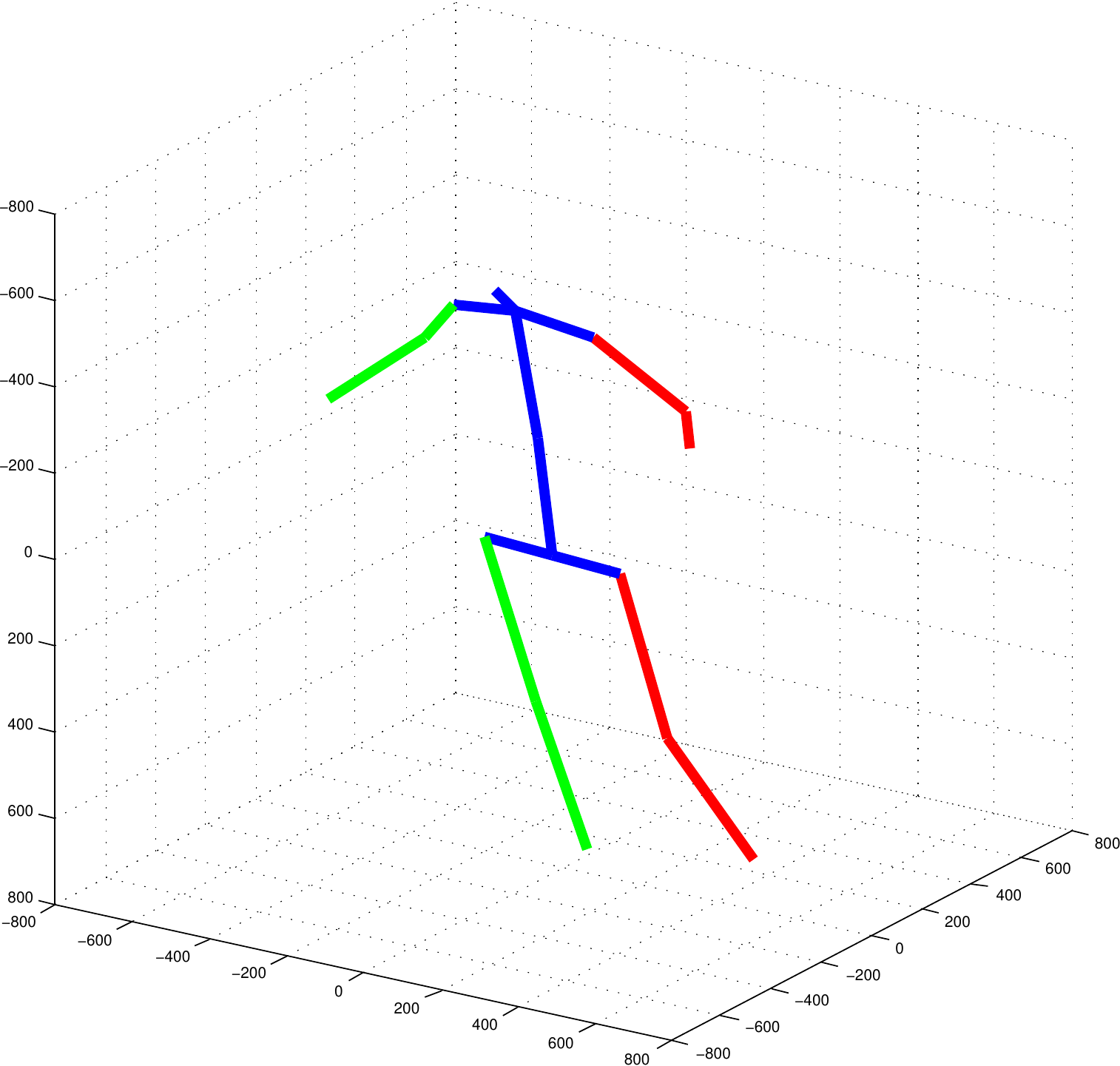}\hspace{1.9mm}
      \includegraphics[height=0.10\textwidth]{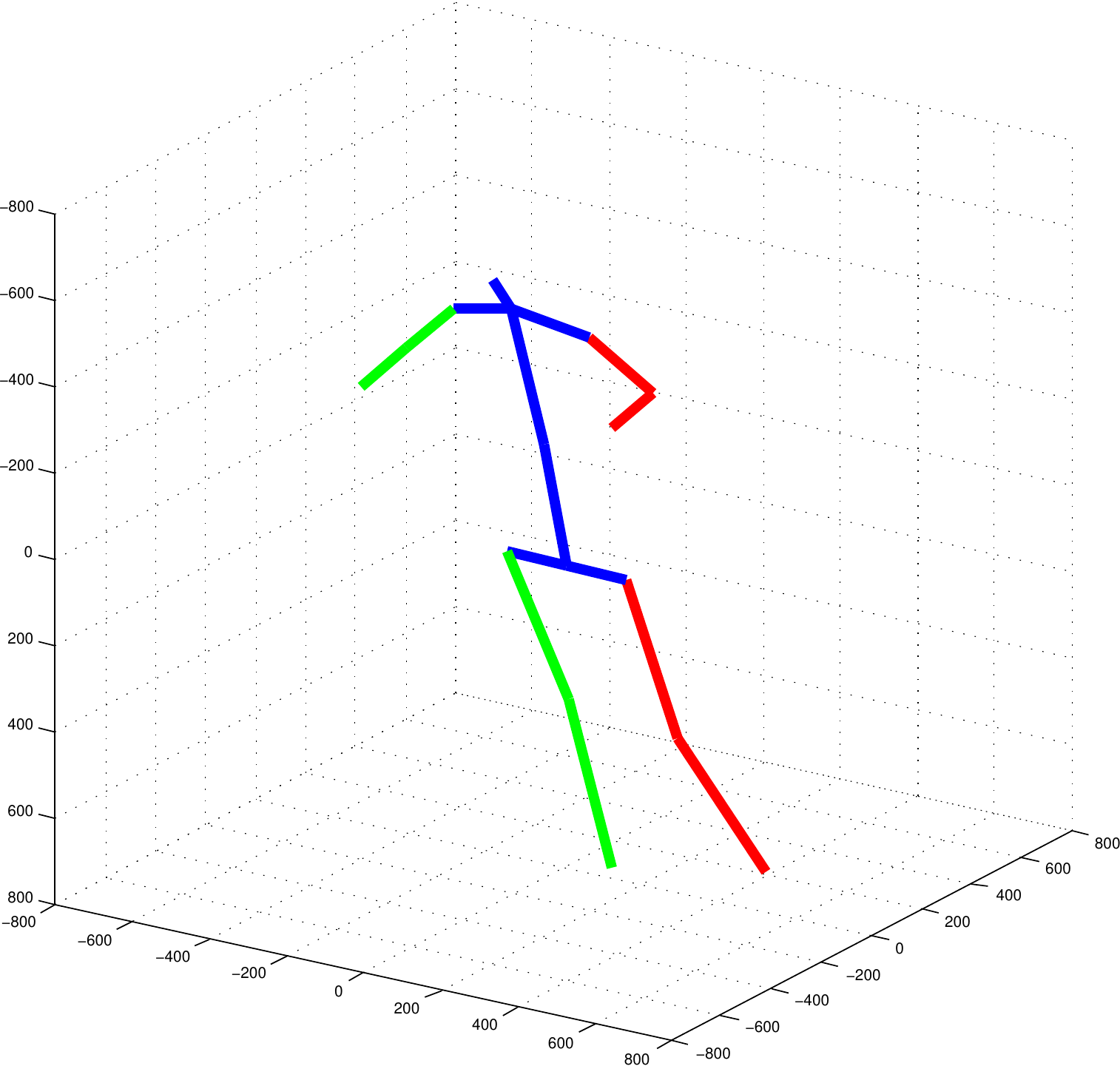}\hspace{1.9mm}
      \includegraphics[height=0.10\textwidth]{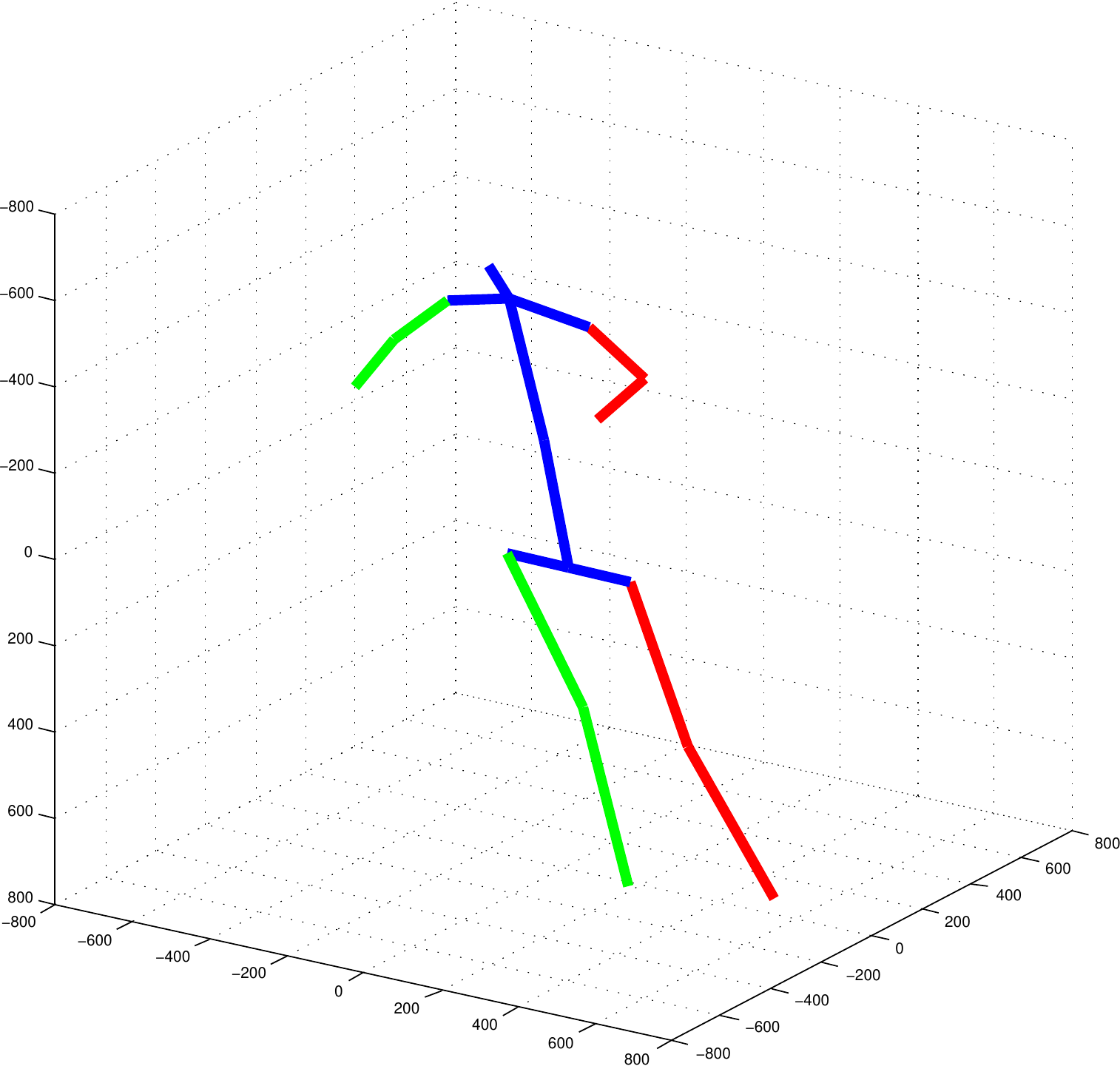}\hspace{1.9mm}
      \includegraphics[height=0.10\textwidth]{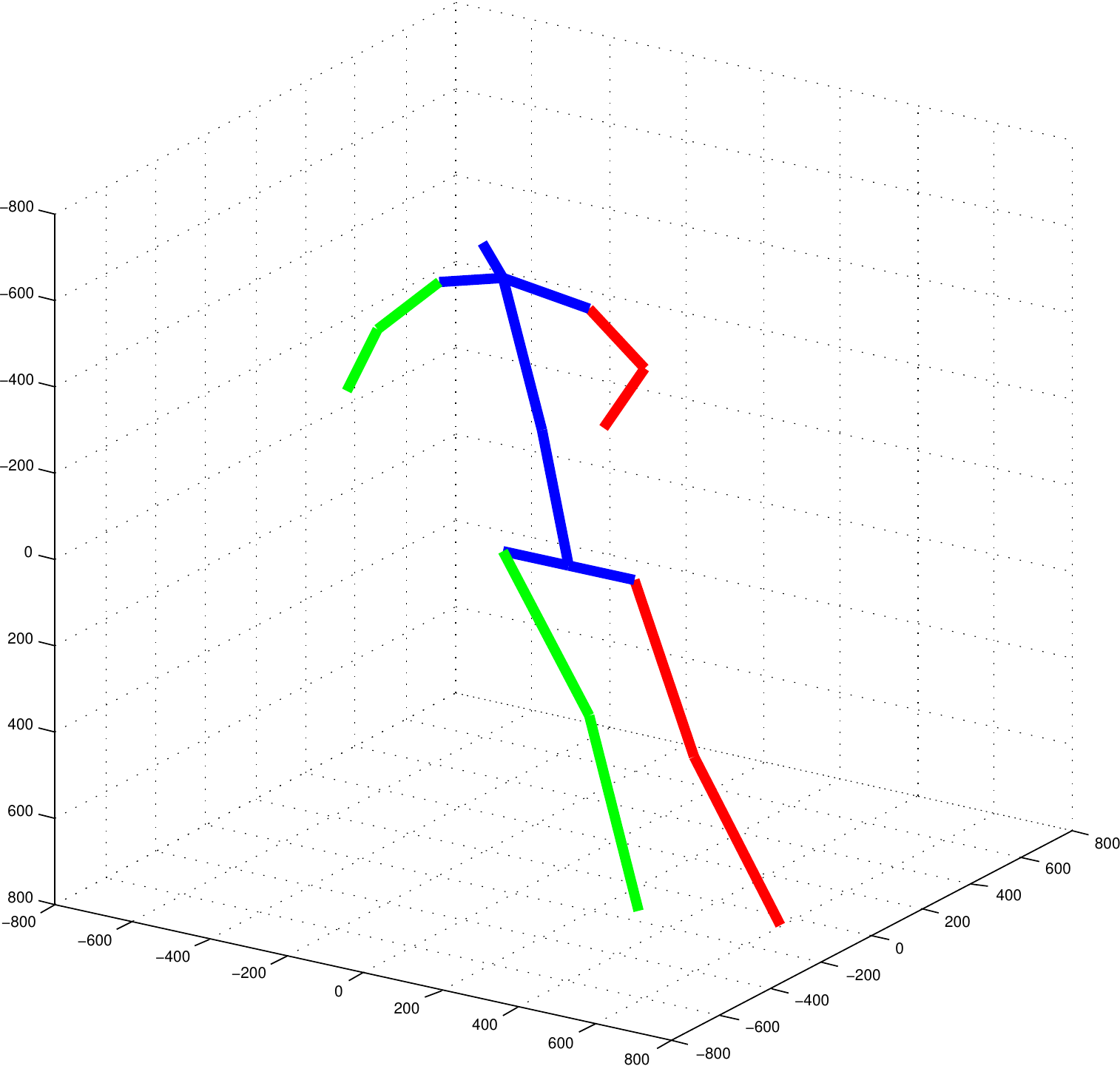}\hspace{1.9mm}
      \includegraphics[height=0.10\textwidth]{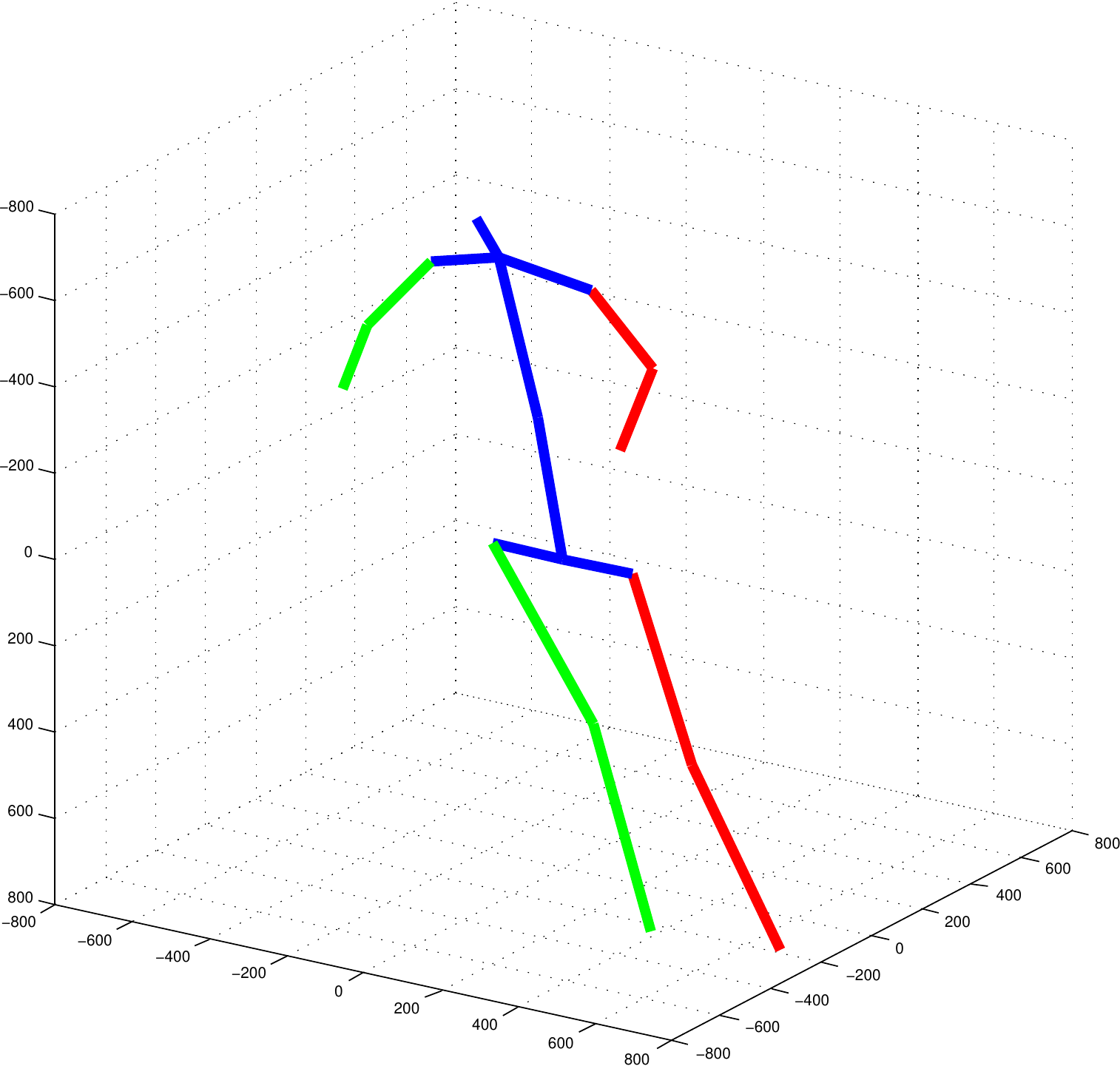}\hspace{1.9mm}
      \includegraphics[height=0.10\textwidth]{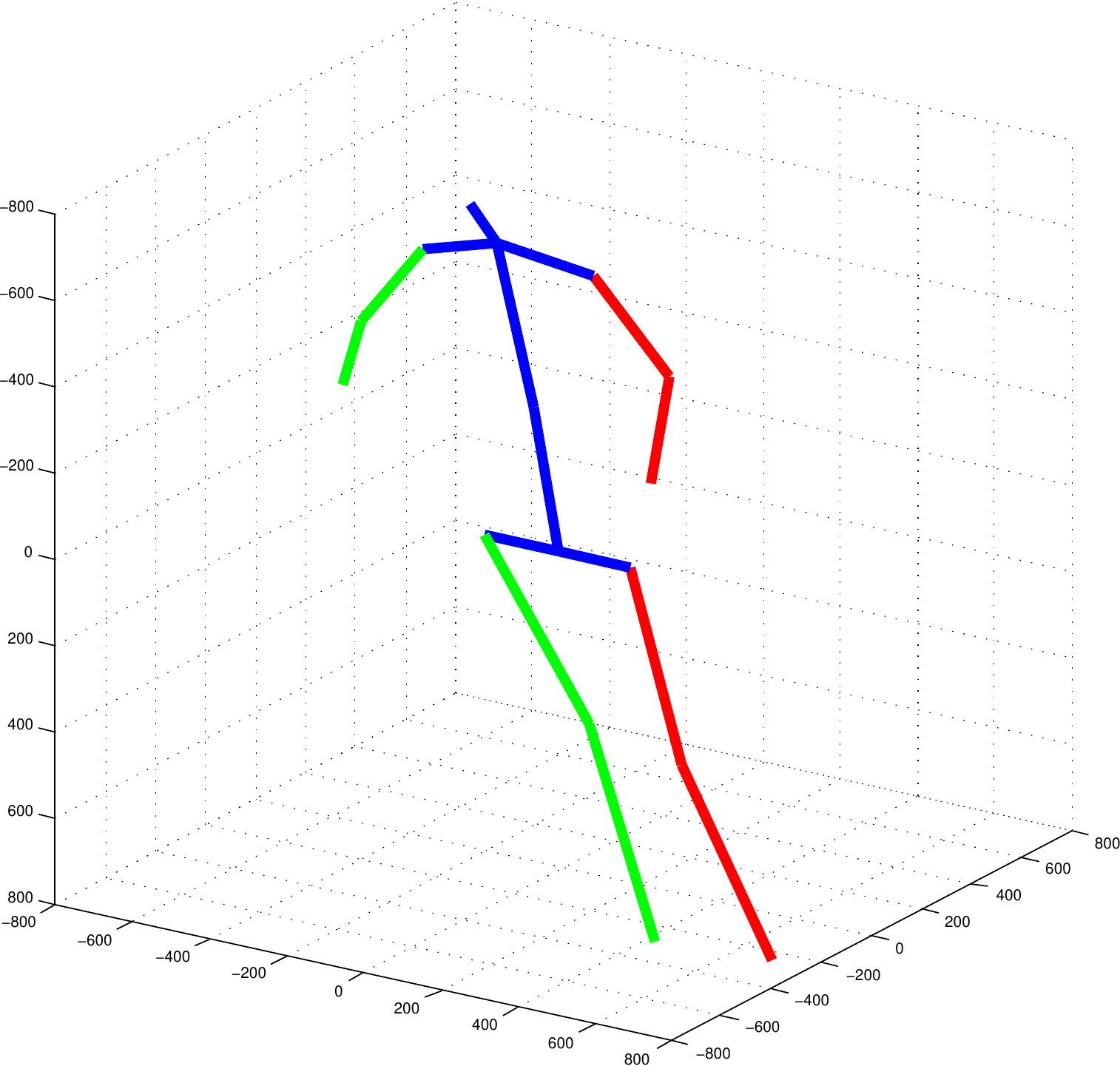}\hspace{1.9mm}
      \includegraphics[height=0.10\textwidth]{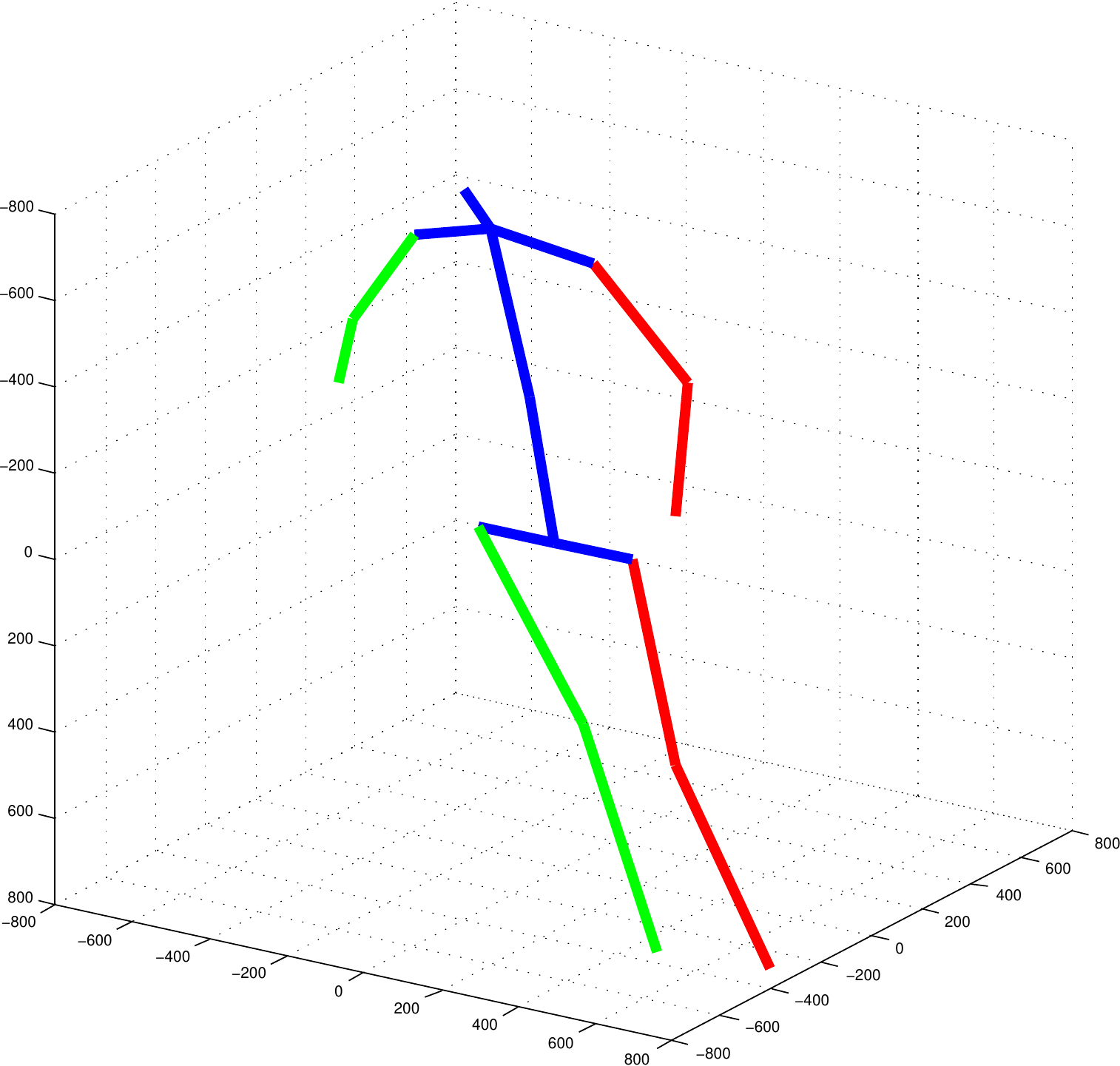}
      \\
      \hspace{-2.7mm}
      \includegraphics[height=0.10\textwidth]{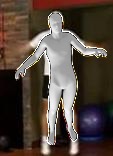}\hspace{0.22mm}
      \includegraphics[height=0.10\textwidth]{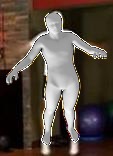}\hspace{0.22mm}
      \includegraphics[height=0.10\textwidth]{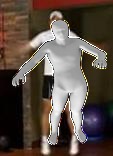}\hspace{0.22mm}
      \includegraphics[height=0.10\textwidth]{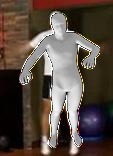}\hspace{0.22mm}
      \includegraphics[height=0.10\textwidth]{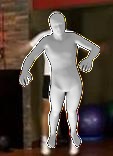}\hspace{0.22mm}
      \includegraphics[height=0.10\textwidth]{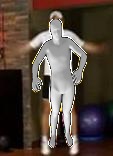}\hspace{0.22mm}
      \includegraphics[height=0.10\textwidth]{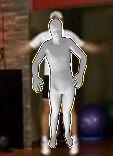}\hspace{0.22mm}
      \includegraphics[height=0.10\textwidth]{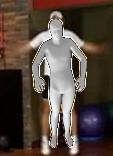}\hspace{0.22mm}
      \includegraphics[height=0.10\textwidth]{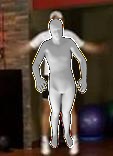}\hspace{0.22mm}
      \includegraphics[height=0.10\textwidth]{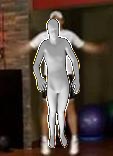}\hspace{0.22mm}
      \includegraphics[height=0.10\textwidth]{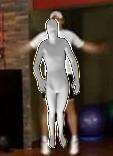}
      \\
      \hspace{-2.7mm}
      \includegraphics[height=0.10\textwidth]{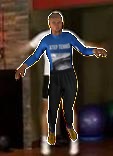}\hspace{0.22mm}
      \includegraphics[height=0.10\textwidth]{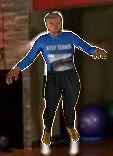}\hspace{0.22mm}
      \includegraphics[height=0.10\textwidth]{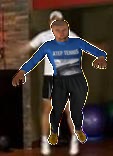}\hspace{0.22mm}
      \includegraphics[height=0.10\textwidth]{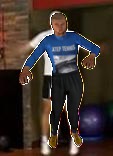}\hspace{0.22mm}
      \includegraphics[height=0.10\textwidth]{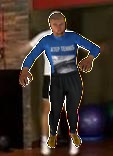}\hspace{0.22mm}
      \includegraphics[height=0.10\textwidth]{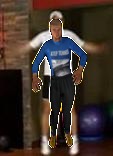}\hspace{0.22mm}
      \includegraphics[height=0.10\textwidth]{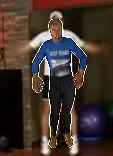}\hspace{0.22mm}
      \includegraphics[height=0.10\textwidth]{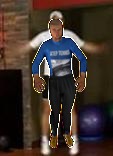}\hspace{0.22mm}
      \includegraphics[height=0.10\textwidth]{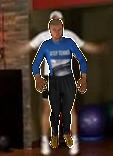}\hspace{0.22mm}
      \includegraphics[height=0.10\textwidth]{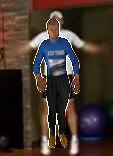}\hspace{0.22mm}
      \includegraphics[height=0.10\textwidth]{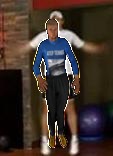}
    \end{tabular}
  \end{tabular}
  % \vspace{-2mm}
  \caption{\small Additional qualitative results of pose forecasting. The left
column shows the input images. For each input image, we show in the right
column the sequence of ground-truth frame and pose (row 1), our forecasted pose
sequence in 2D (row 2) and 3D (row 3), and the rendered human body without
texture (row 4) and with skin and cloth textures (row 5).}
  % \phantomcaption
  % \vspace{-2mm}
  \label{fig:additional1}
\end{figure*}

\begin{figure*}[t]%\ContinuedFloat
  % \vspace{-2mm}
  \centering
  \footnotesize
  \begin{tabular}{L{0.12\linewidth}@{\hspace{1.0mm}}|L{0.9\linewidth}@{\hspace{-0.0mm}}}
    \includegraphics[height=0.084\textwidth]{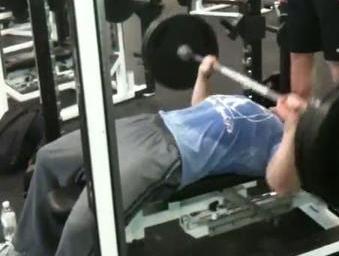}
    &
    \vspace{-2.7mm}
    \begin{tabular}{L{1.0\linewidth}}
      \hspace{-2.7mm}
      \includegraphics[height=0.10\textwidth]{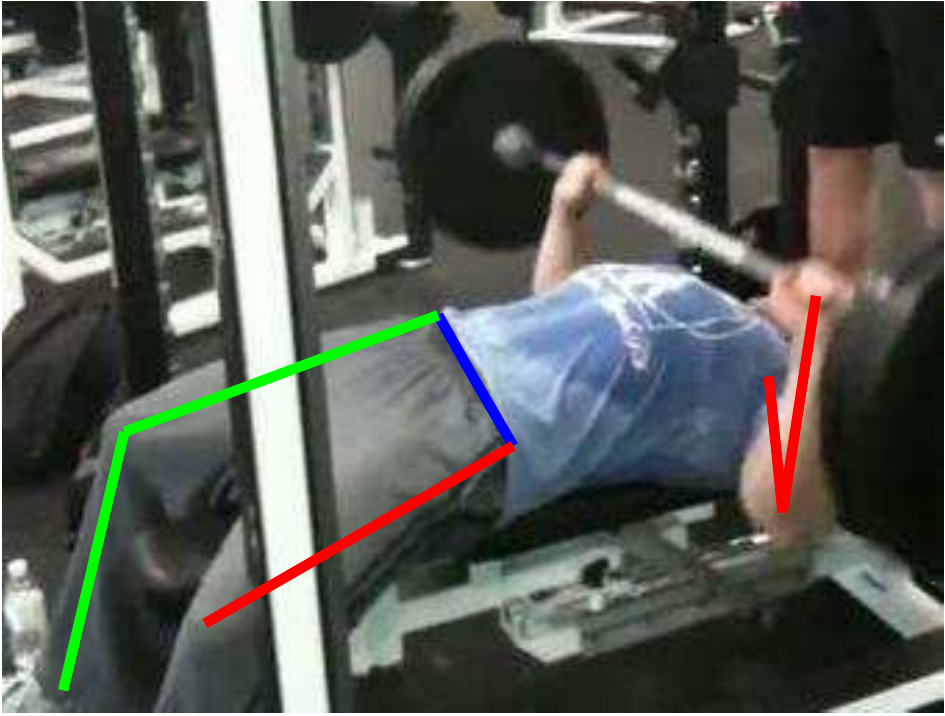}\hspace{1.33mm}
      \includegraphics[height=0.10\textwidth]{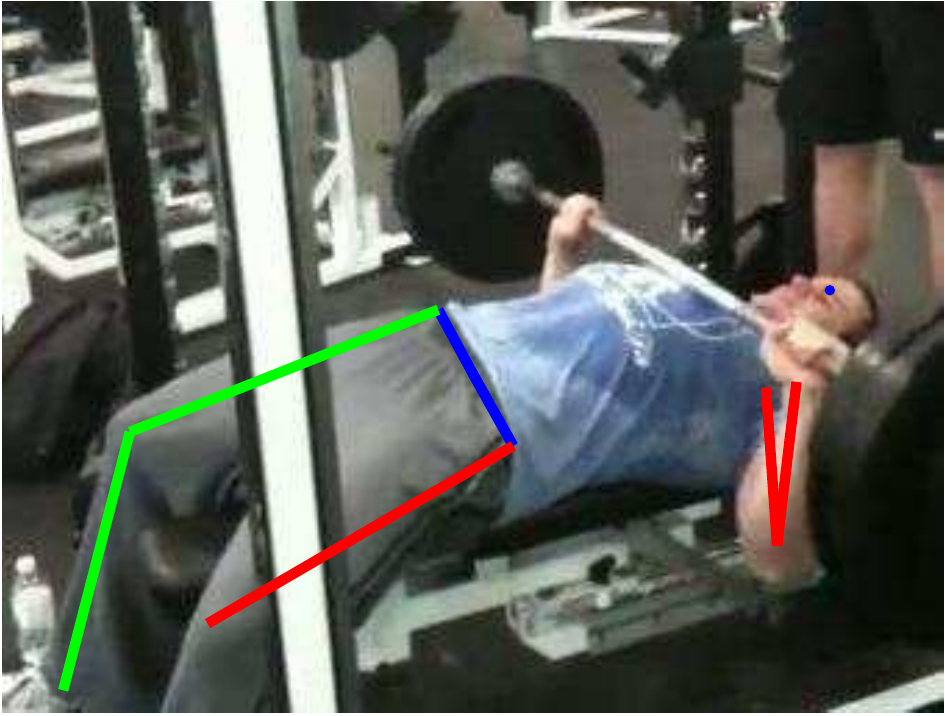}\hspace{1.33mm}
      \includegraphics[height=0.10\textwidth]{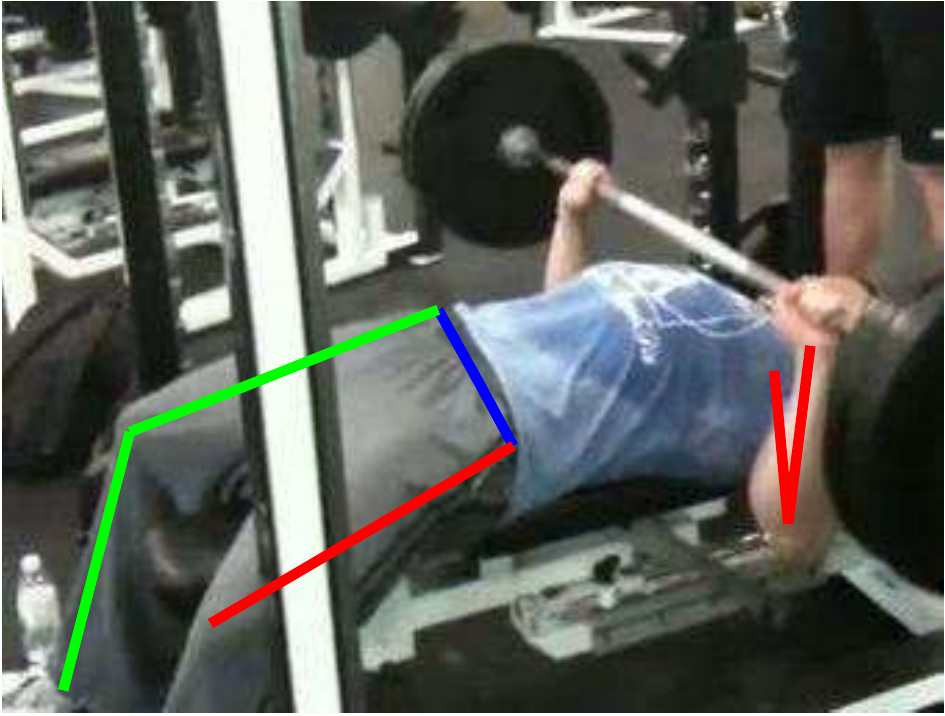}\hspace{1.33mm}
      \includegraphics[height=0.10\textwidth]{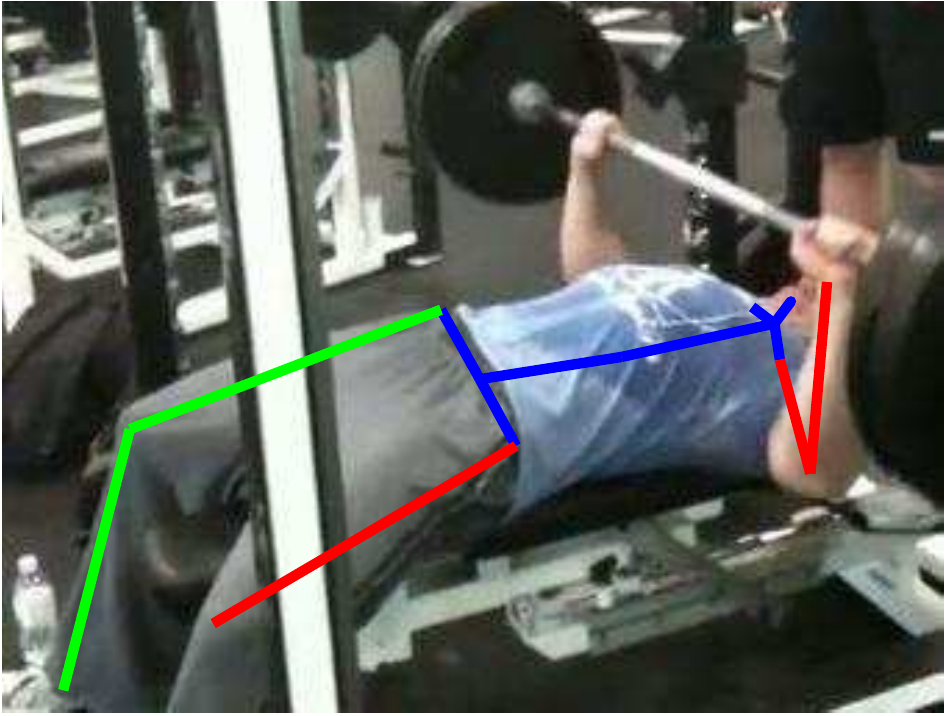}\hspace{1.33mm}
      \includegraphics[height=0.10\textwidth]{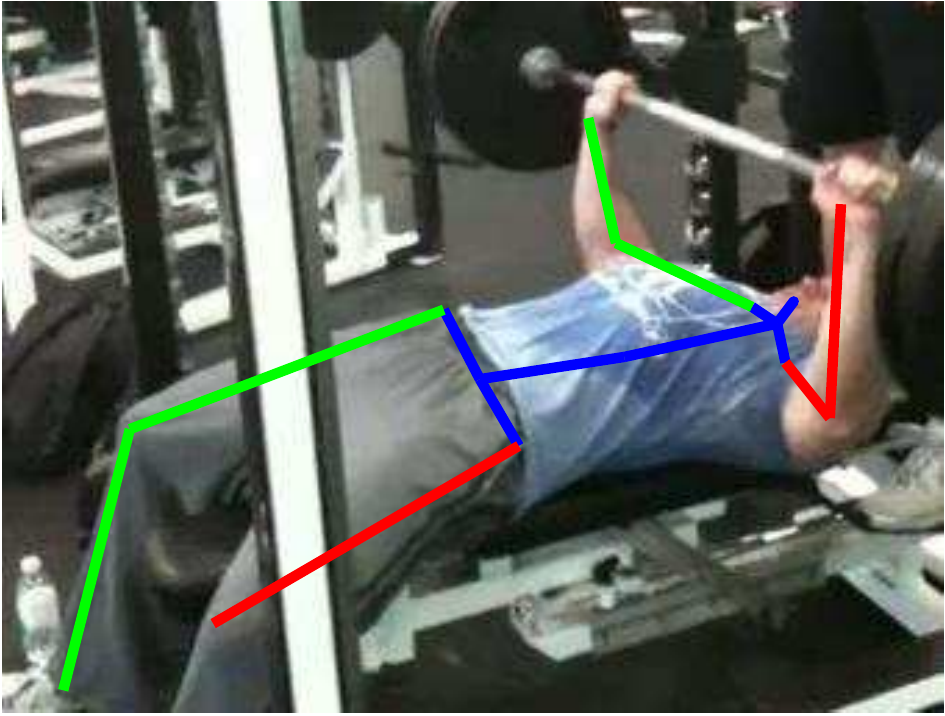}\hspace{1.33mm}
      \includegraphics[height=0.10\textwidth]{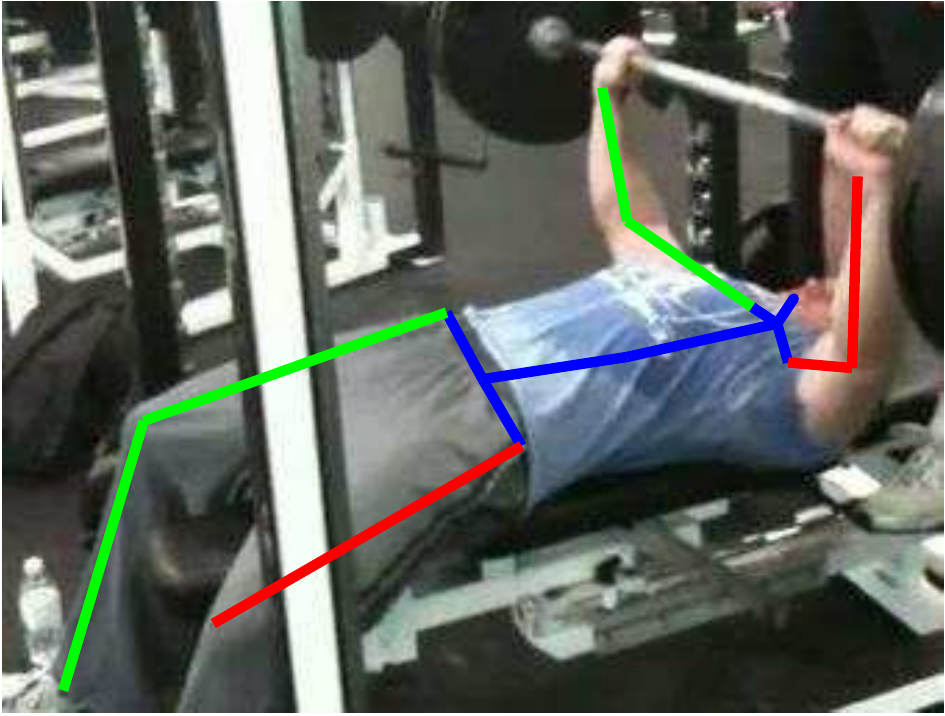}
      \\
      \hspace{-2.7mm}
      \includegraphics[height=0.10\textwidth]{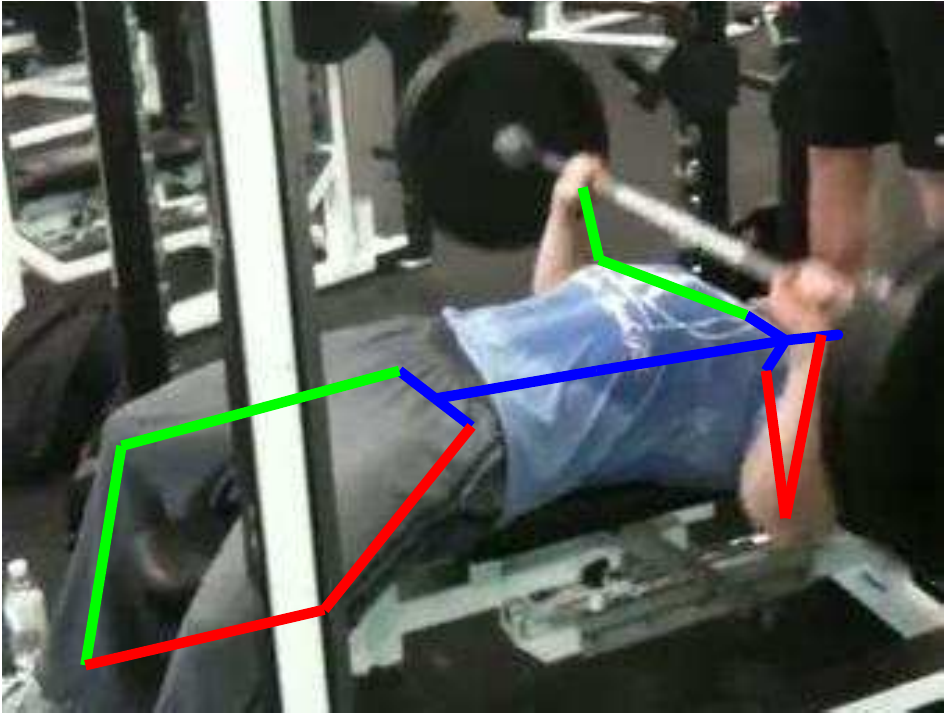}\hspace{1.33mm}
      \includegraphics[height=0.10\textwidth]{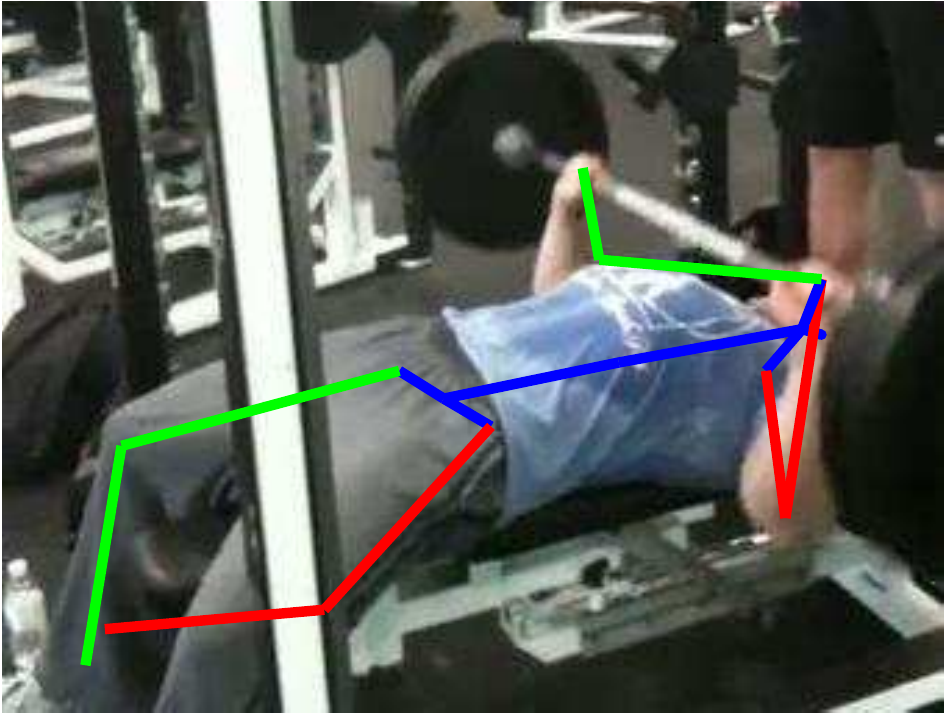}\hspace{1.33mm}
      \includegraphics[height=0.10\textwidth]{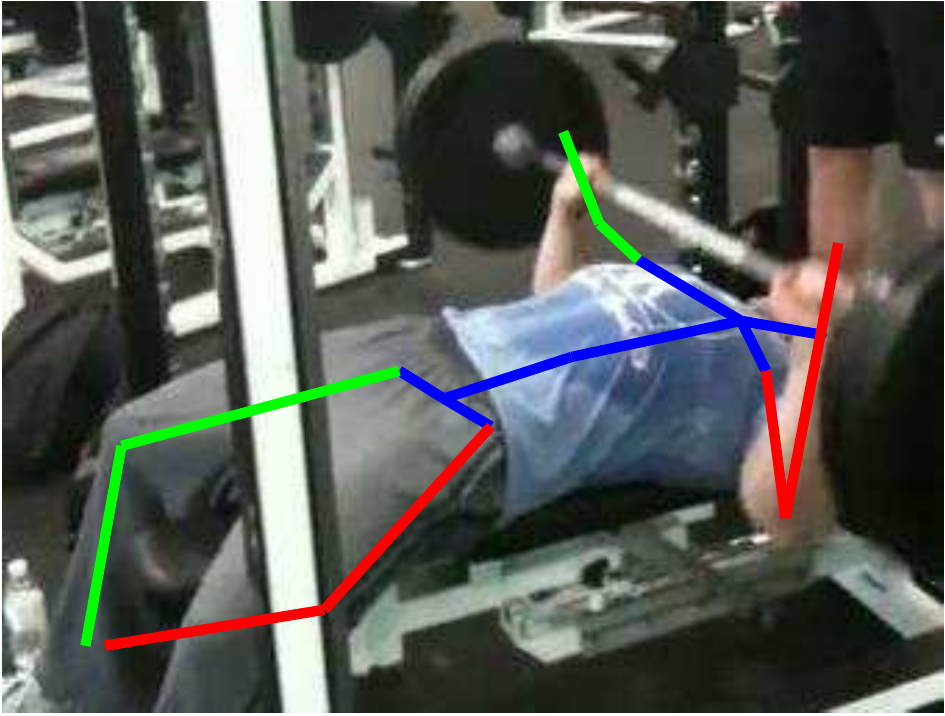}\hspace{1.33mm}
      \includegraphics[height=0.10\textwidth]{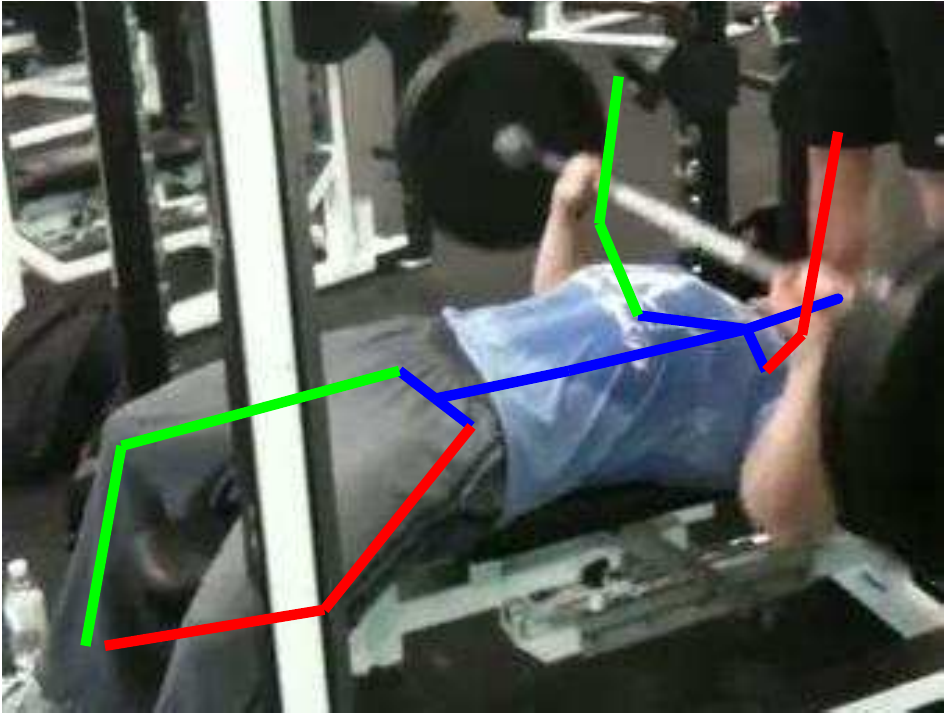}\hspace{1.33mm}
      \includegraphics[height=0.10\textwidth]{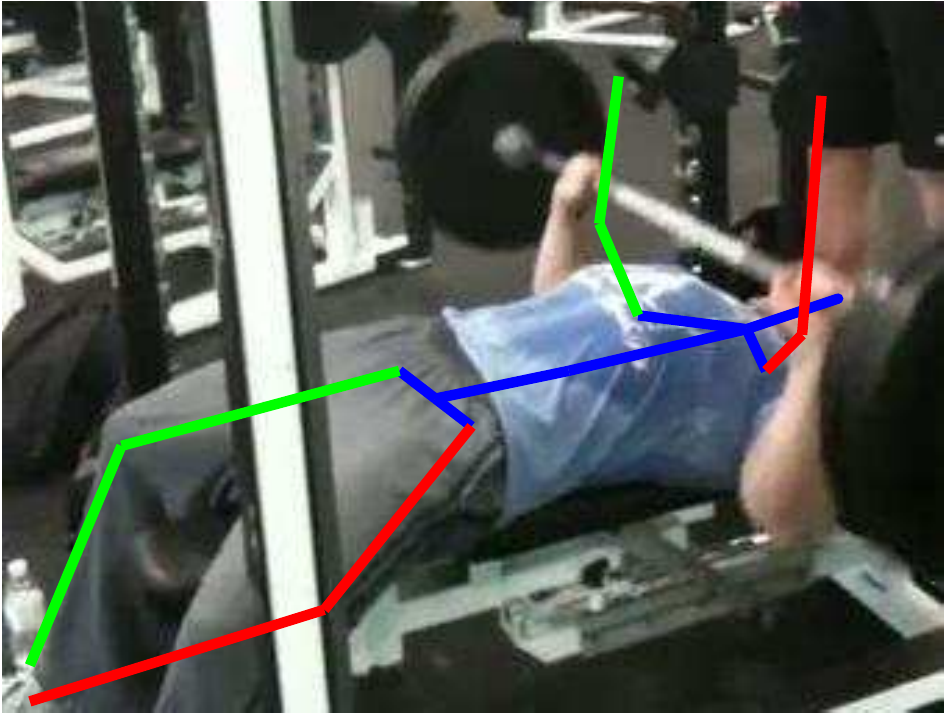}\hspace{1.33mm}
      \includegraphics[height=0.10\textwidth]{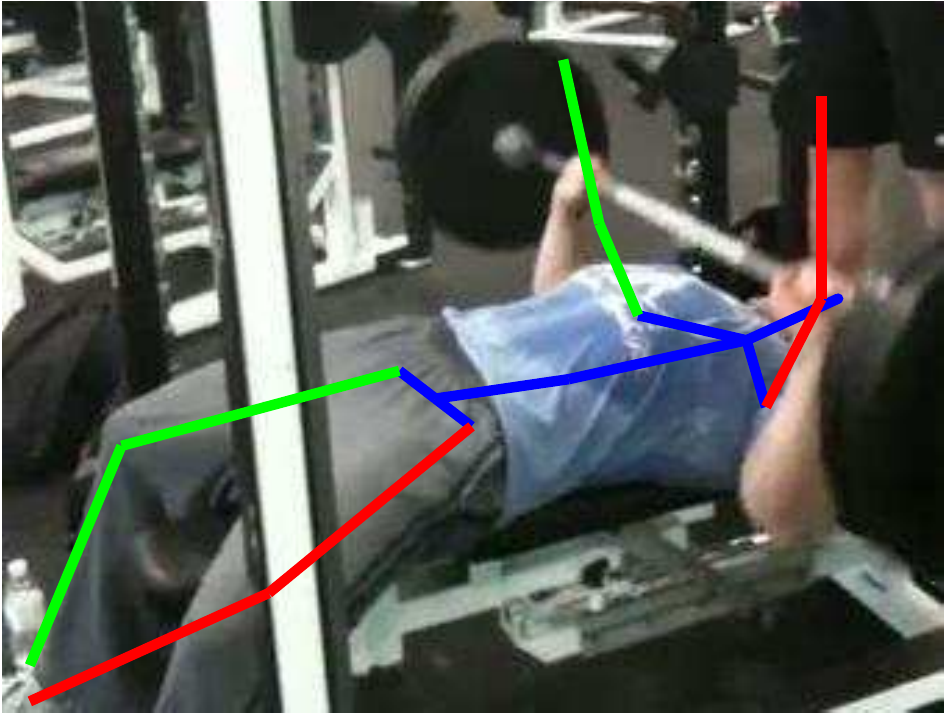}
      \\
      \includegraphics[height=0.10\textwidth]{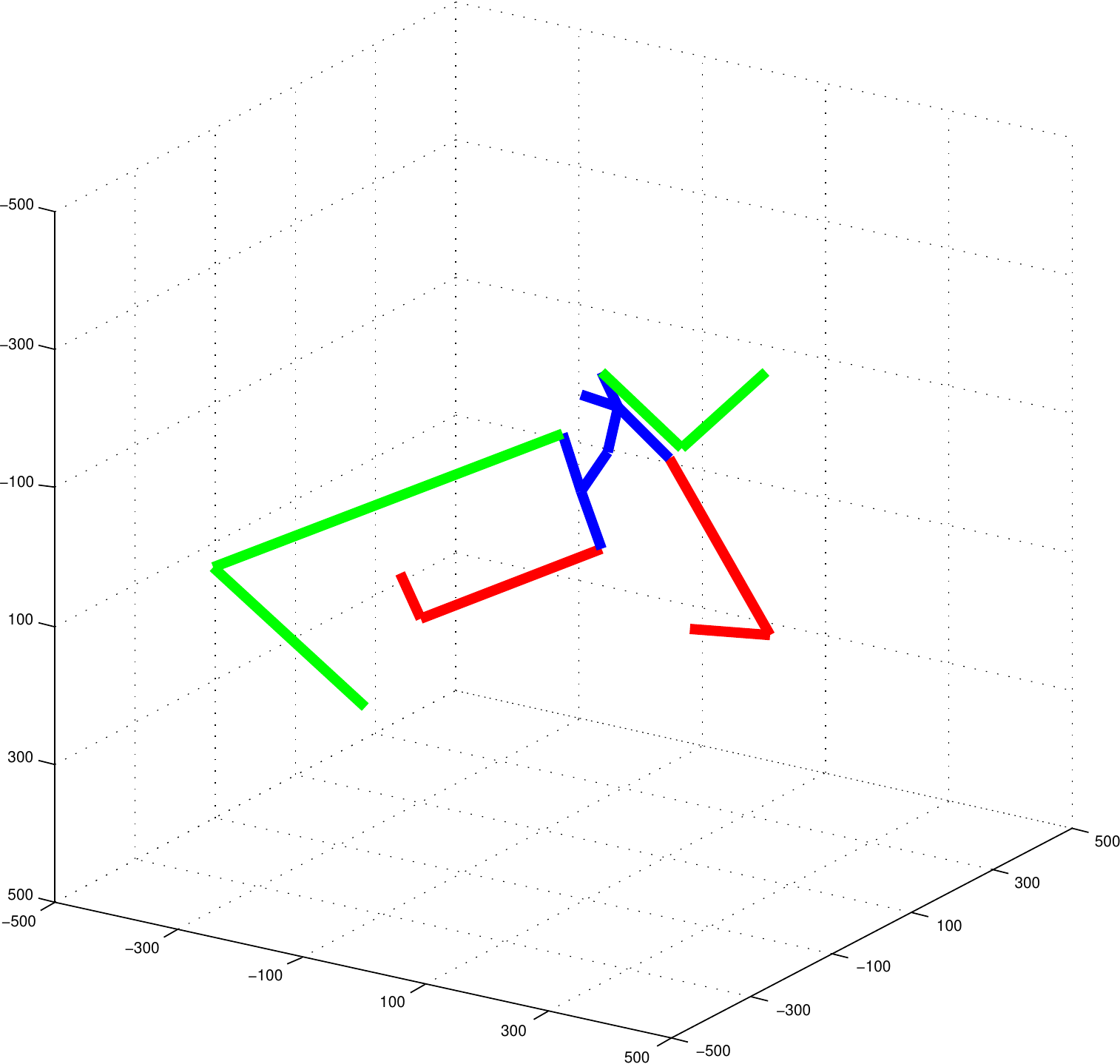}\hspace{1.9mm}
      \includegraphics[height=0.10\textwidth]{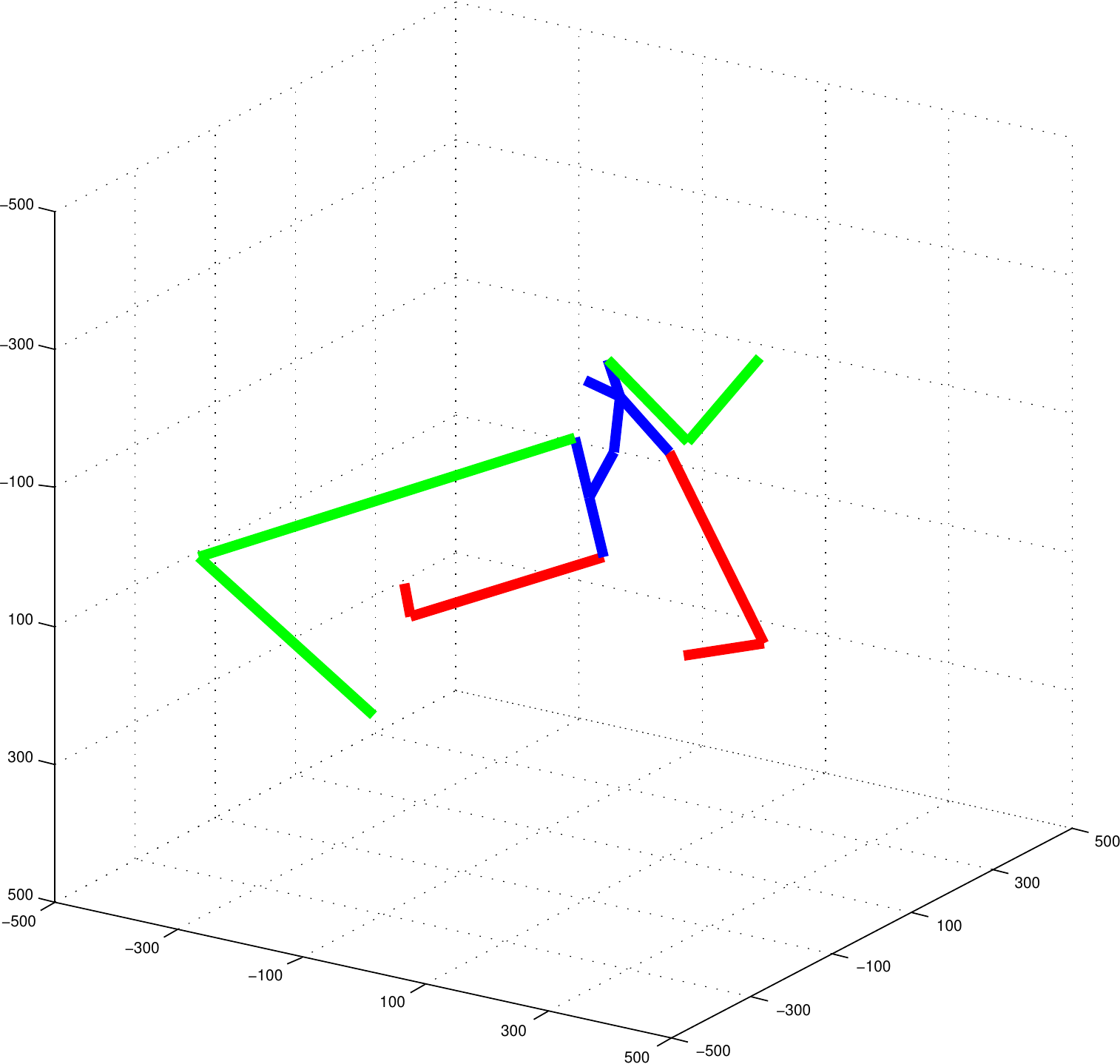}\hspace{1.9mm}
      \includegraphics[height=0.10\textwidth]{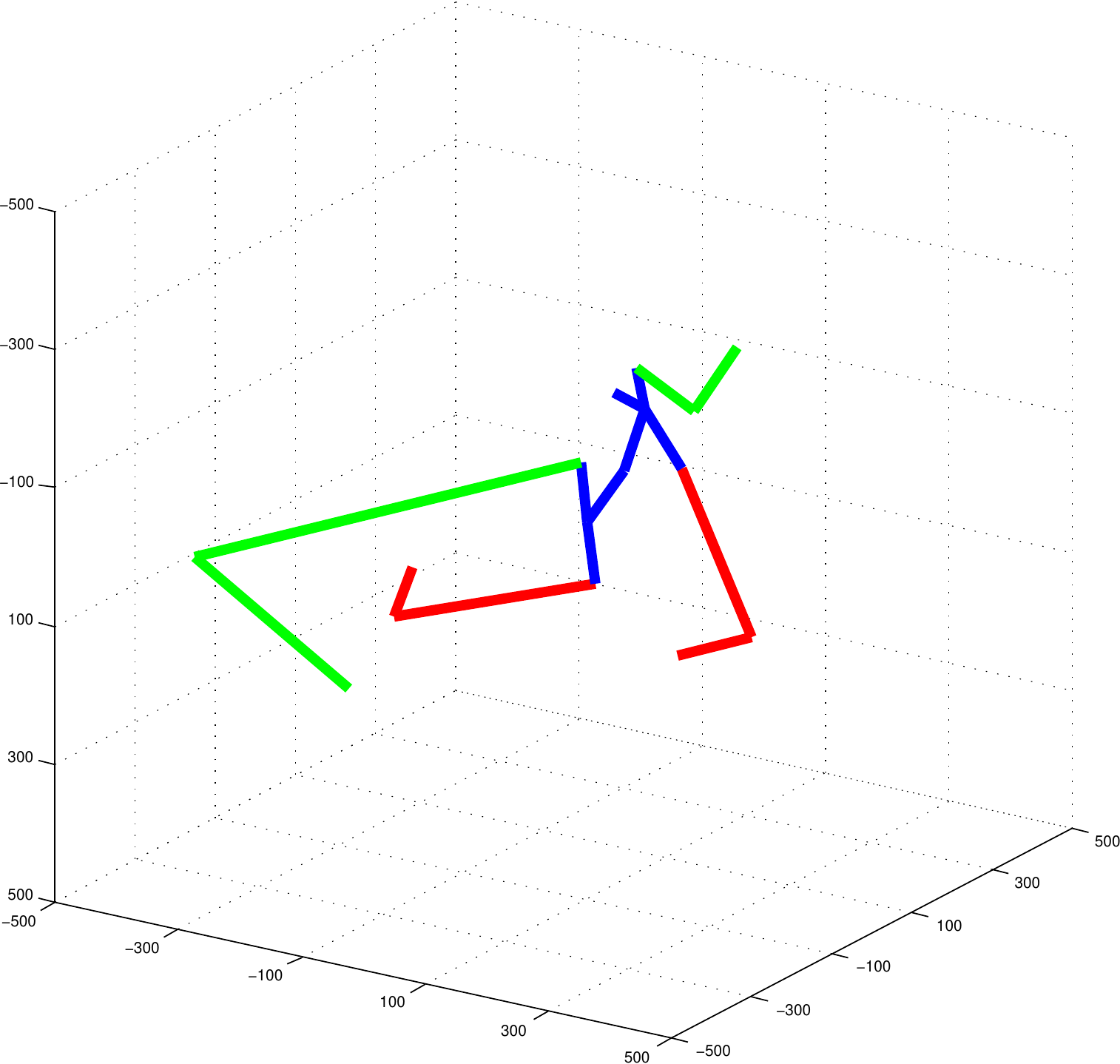}\hspace{1.9mm}
      \includegraphics[height=0.10\textwidth]{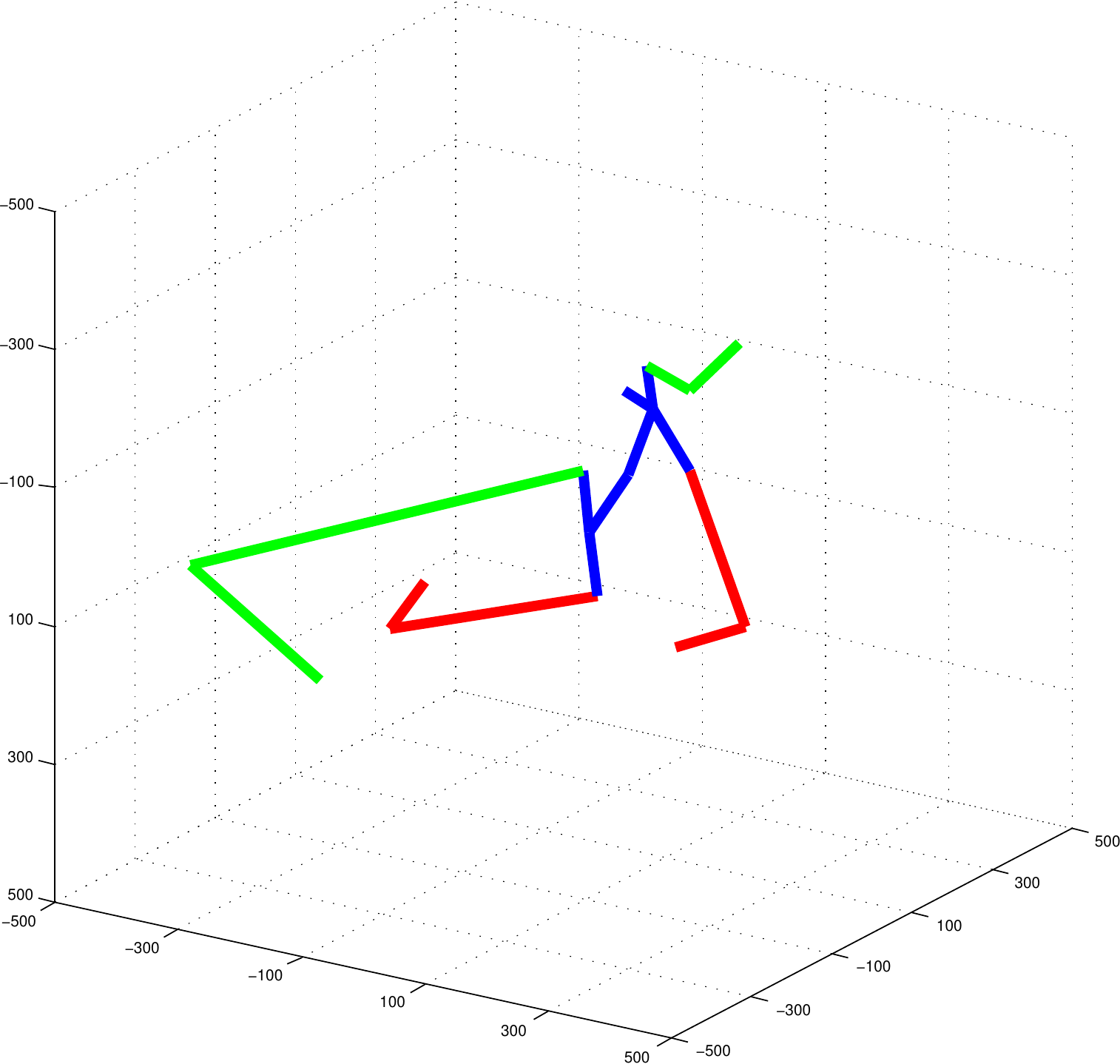}\hspace{1.9mm}
      \includegraphics[height=0.10\textwidth]{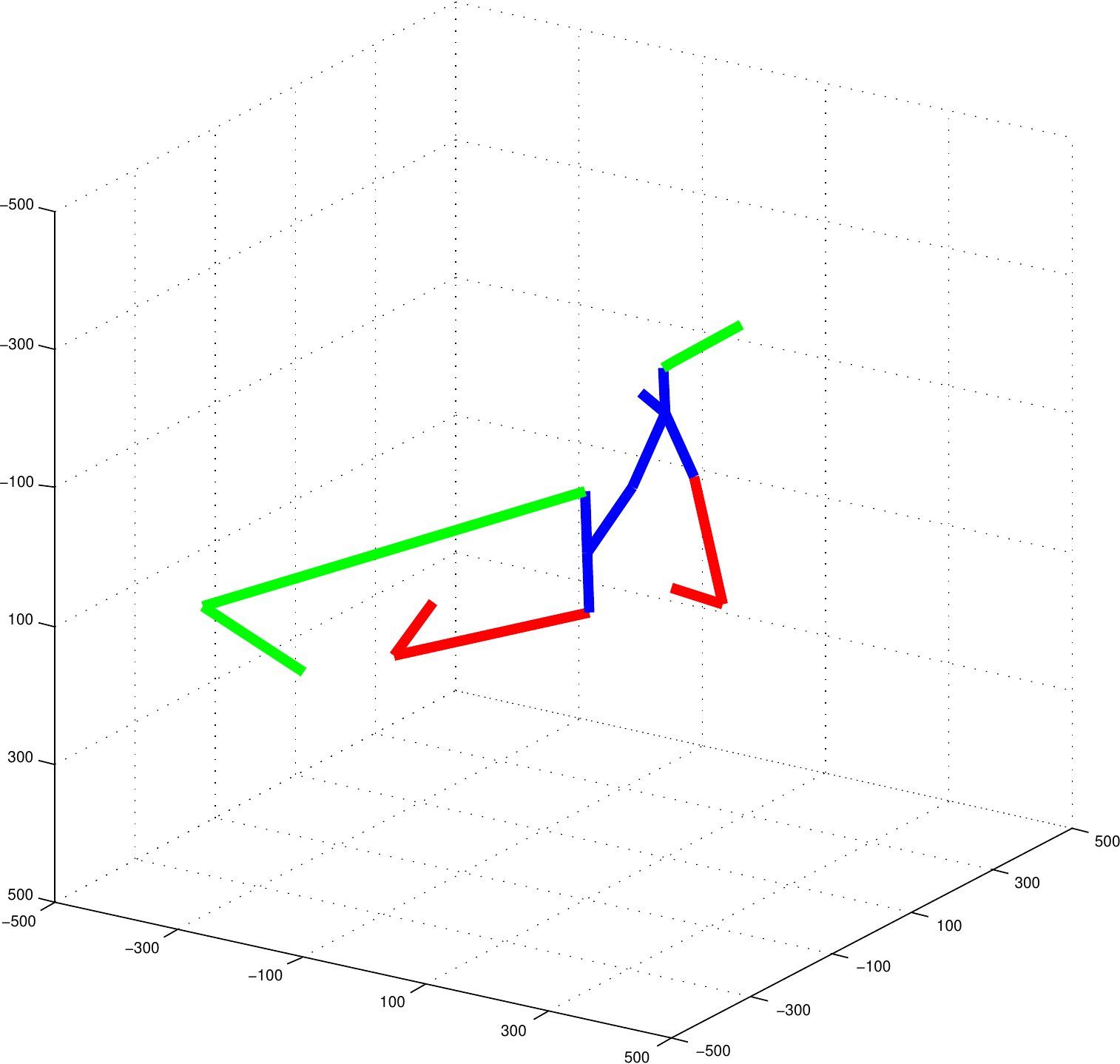}\hspace{1.9mm}
      \includegraphics[height=0.10\textwidth]{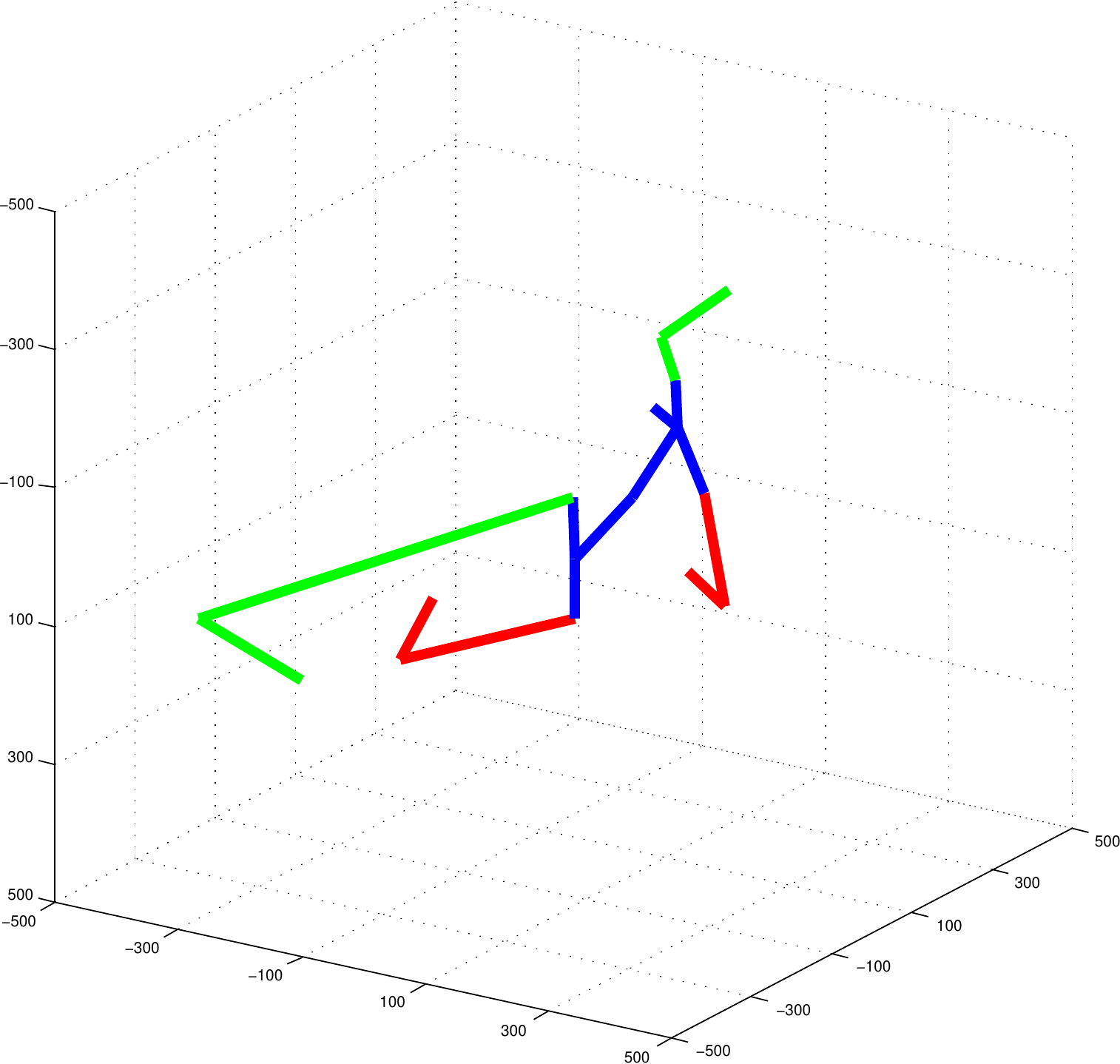}\hspace{1.9mm}
      \includegraphics[height=0.10\textwidth]{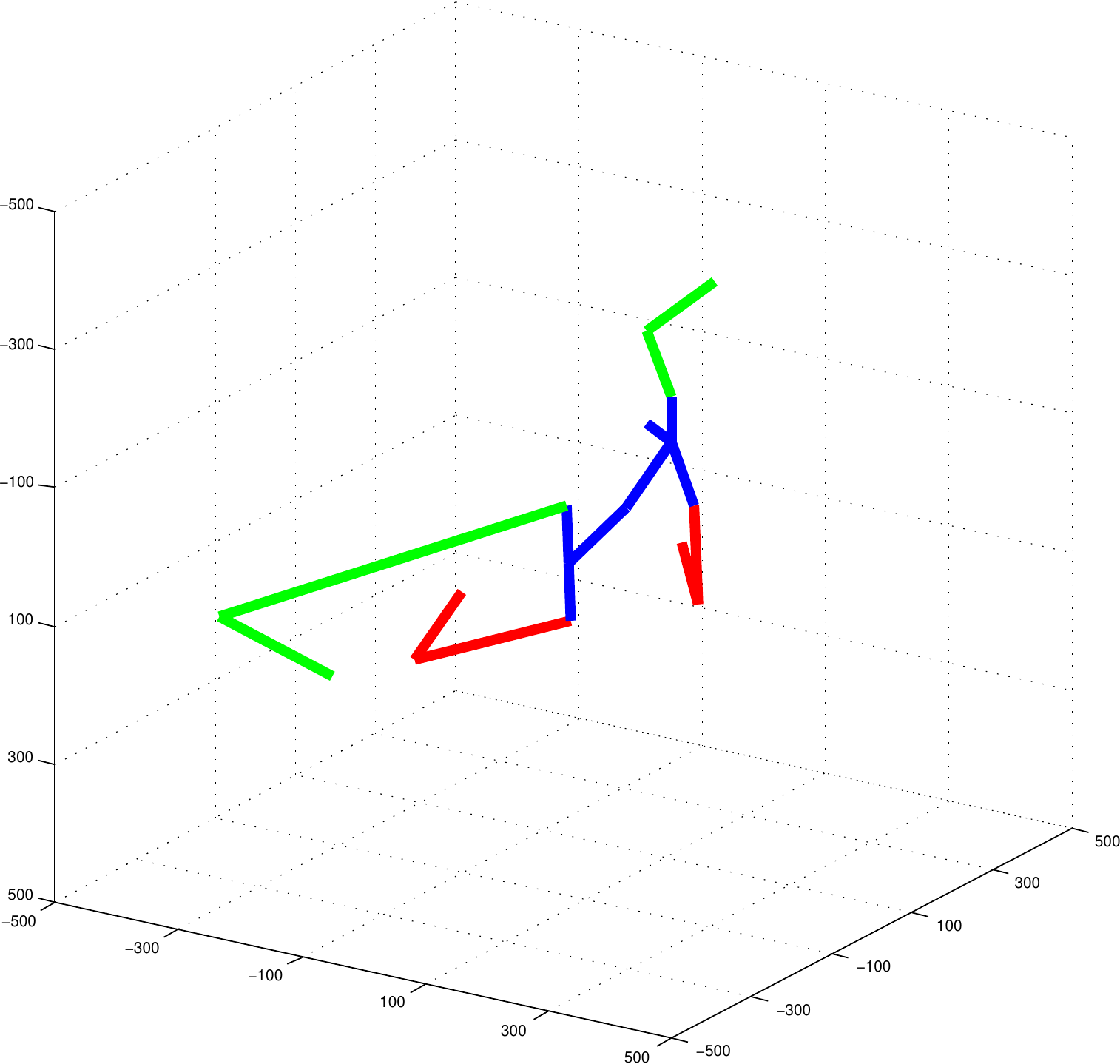}
      \\
      \hspace{-2.7mm}
      \includegraphics[height=0.10\textwidth]{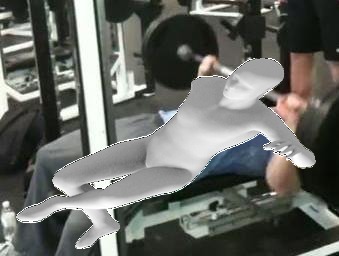}\hspace{1.24mm}
      \includegraphics[height=0.10\textwidth]{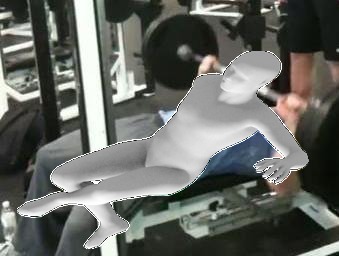}\hspace{1.24mm}
      \includegraphics[height=0.10\textwidth]{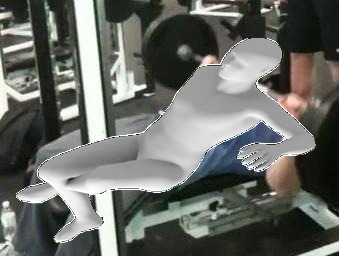}\hspace{1.24mm}
      \includegraphics[height=0.10\textwidth]{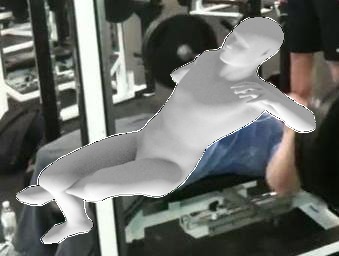}\hspace{1.24mm}
      \includegraphics[height=0.10\textwidth]{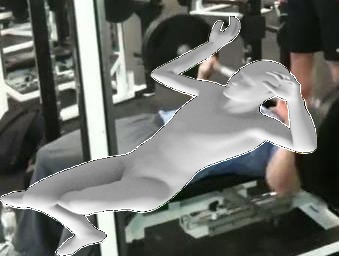}\hspace{1.24mm}
      \includegraphics[height=0.10\textwidth]{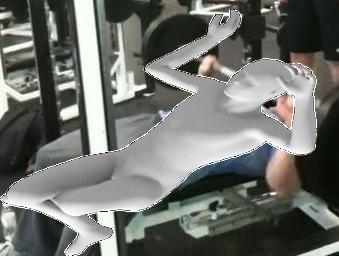}
      \\
      \hspace{-2.7mm}
      \includegraphics[height=0.10\textwidth]{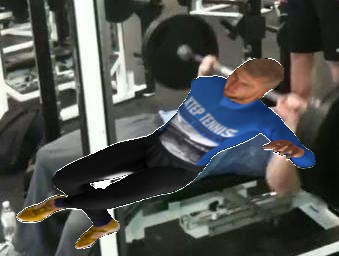}\hspace{1.24mm}
      \includegraphics[height=0.10\textwidth]{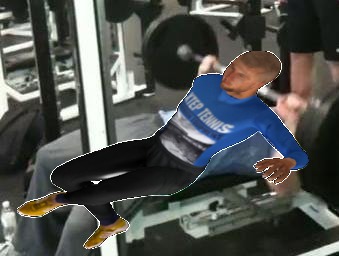}\hspace{1.24mm}
      \includegraphics[height=0.10\textwidth]{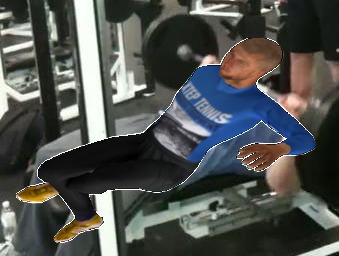}\hspace{1.24mm}
      \includegraphics[height=0.10\textwidth]{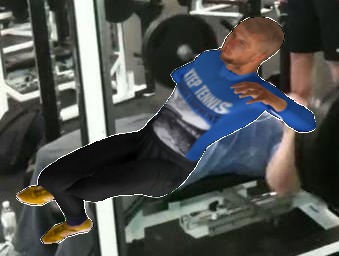}\hspace{1.24mm}
      \includegraphics[height=0.10\textwidth]{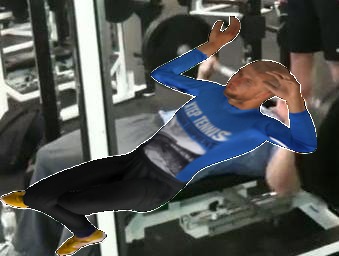}\hspace{1.24mm}
      \includegraphics[height=0.10\textwidth]{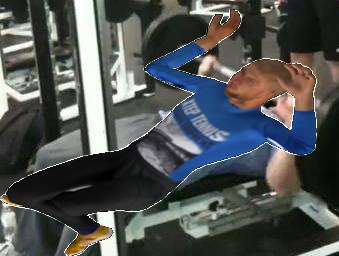}
    \end{tabular}
    \\ [-0.0em] & \\
    \includegraphics[height=0.127\textwidth]{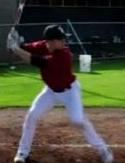}
    &
    \vspace{-2.7mm}
    \begin{tabular}{L{1.0\linewidth}}
      \hspace{-2.7mm}
      \includegraphics[height=0.10\textwidth]{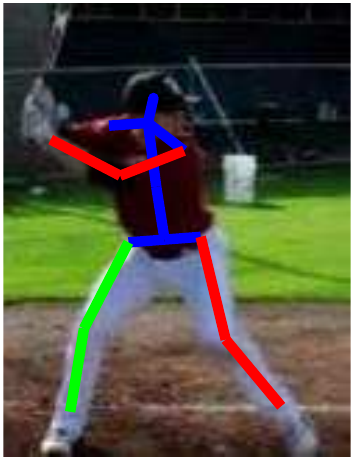}\hspace{0.82mm}
      \includegraphics[height=0.10\textwidth]{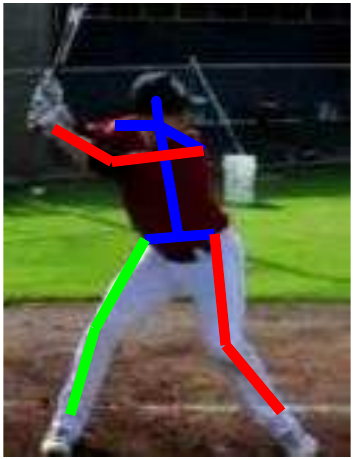}\hspace{0.82mm}
      \includegraphics[height=0.10\textwidth]{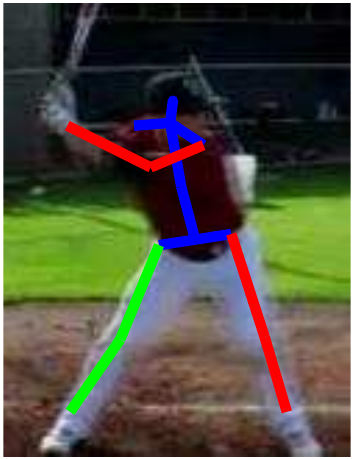}\hspace{0.82mm}
      \includegraphics[height=0.10\textwidth]{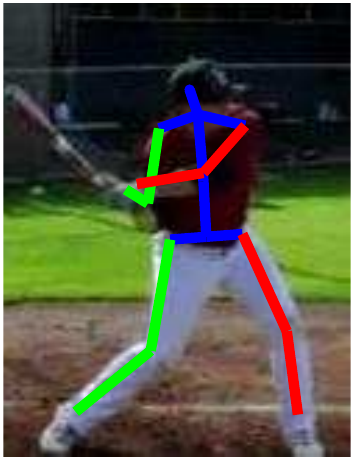}\hspace{0.82mm}
      \includegraphics[height=0.10\textwidth]{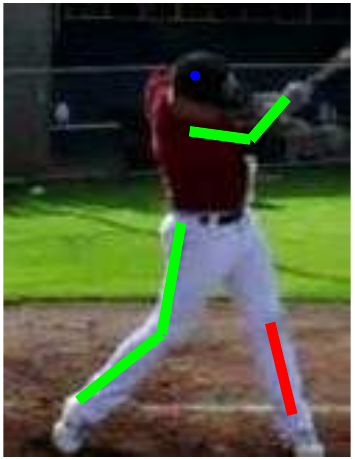}\hspace{0.82mm}
      \includegraphics[height=0.10\textwidth]{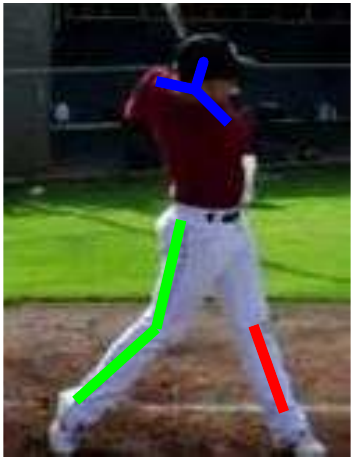}\hspace{0.82mm}
      \includegraphics[height=0.10\textwidth]{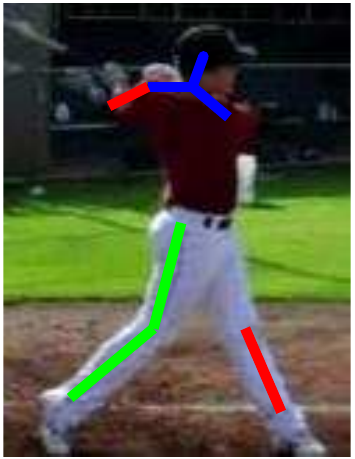}\hspace{0.82mm}
      \includegraphics[height=0.10\textwidth]{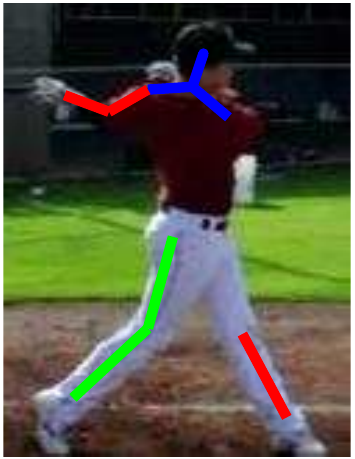}\hspace{0.82mm}
      \includegraphics[height=0.10\textwidth]{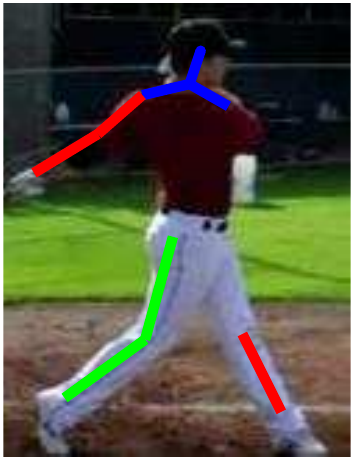}\hspace{0.82mm}
      \includegraphics[height=0.10\textwidth]{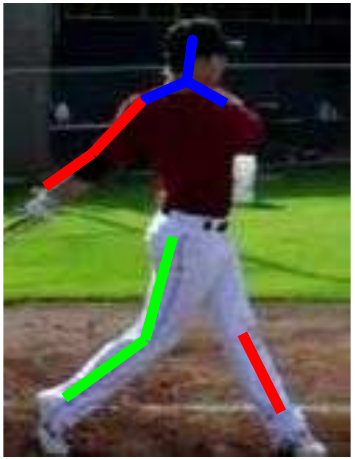}
      \\
      \hspace{-2.7mm}
      \includegraphics[height=0.10\textwidth]{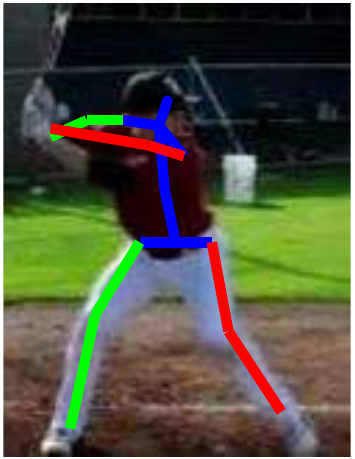}\hspace{0.82mm}
      \includegraphics[height=0.10\textwidth]{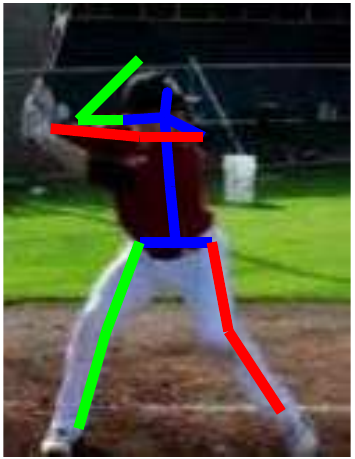}\hspace{0.82mm}
      \includegraphics[height=0.10\textwidth]{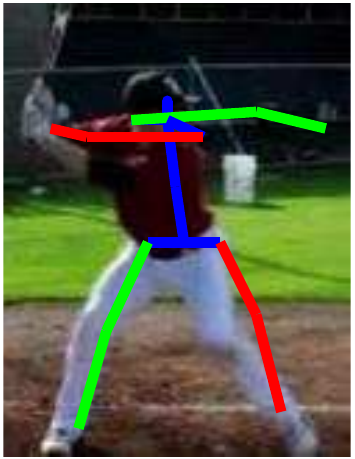}\hspace{0.82mm}
      \includegraphics[height=0.10\textwidth]{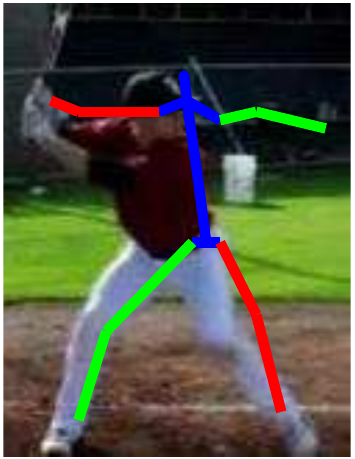}\hspace{0.82mm}
      \includegraphics[height=0.10\textwidth]{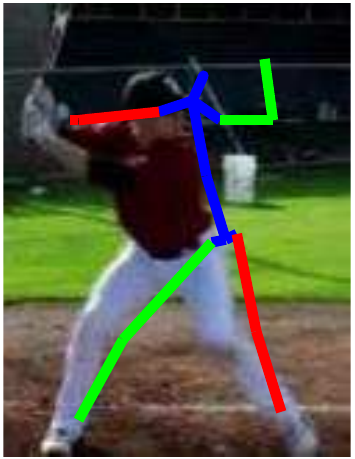}\hspace{0.82mm}
      \includegraphics[height=0.10\textwidth]{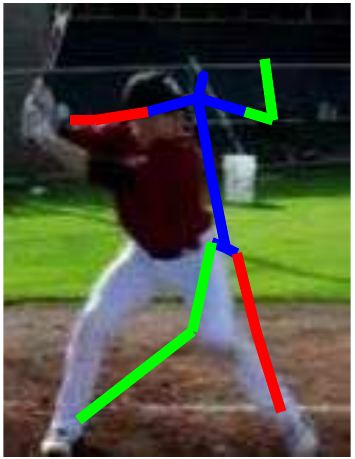}\hspace{0.82mm}
      \includegraphics[height=0.10\textwidth]{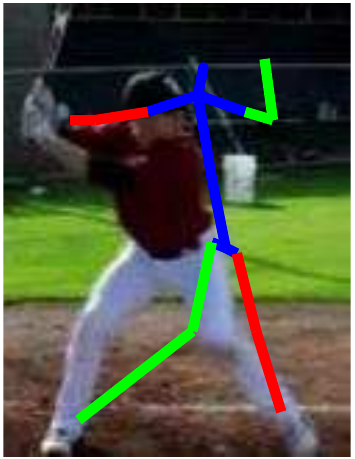}\hspace{0.82mm}
      \includegraphics[height=0.10\textwidth]{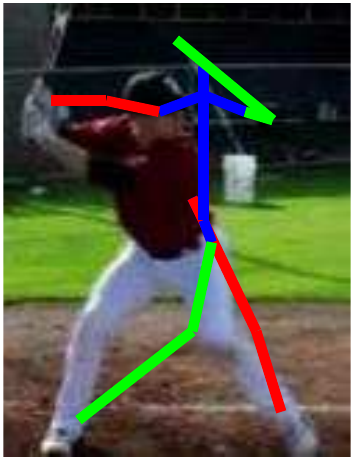}\hspace{0.82mm}
      \includegraphics[height=0.10\textwidth]{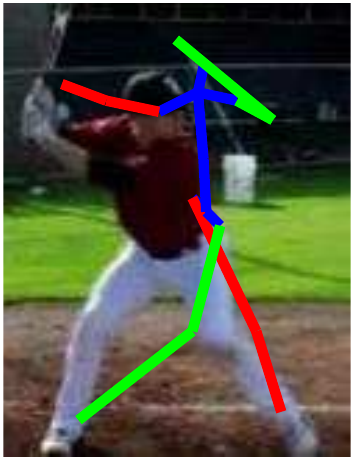}\hspace{0.82mm}
      \includegraphics[height=0.10\textwidth]{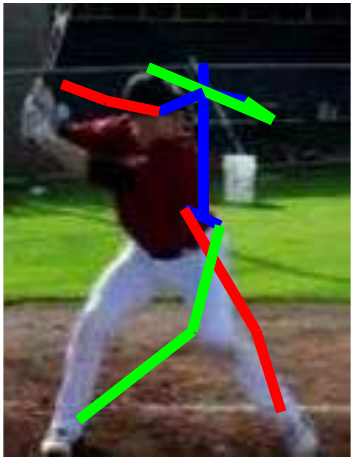}
      \\
      \includegraphics[height=0.10\textwidth]{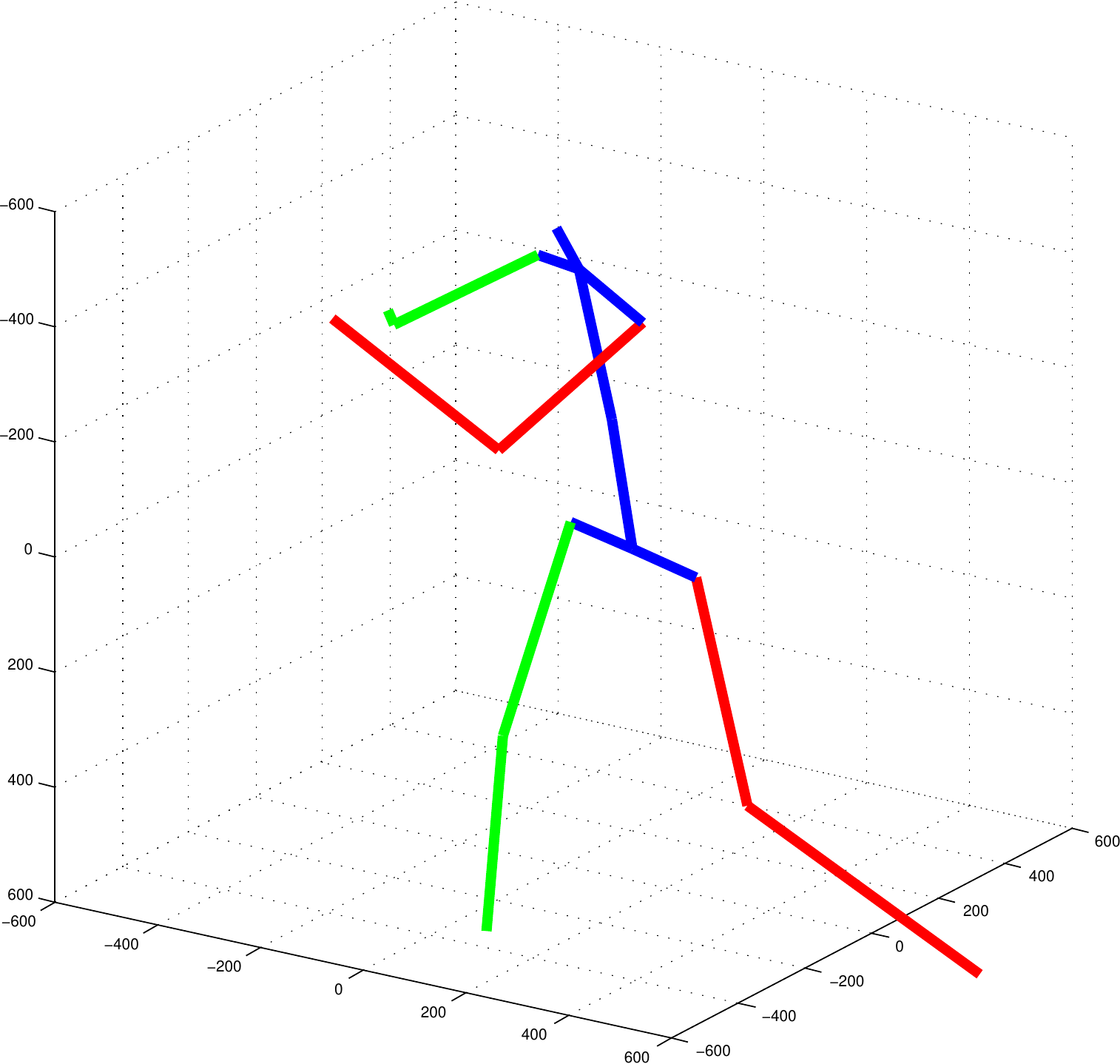}\hspace{1.9mm}
      \includegraphics[height=0.10\textwidth]{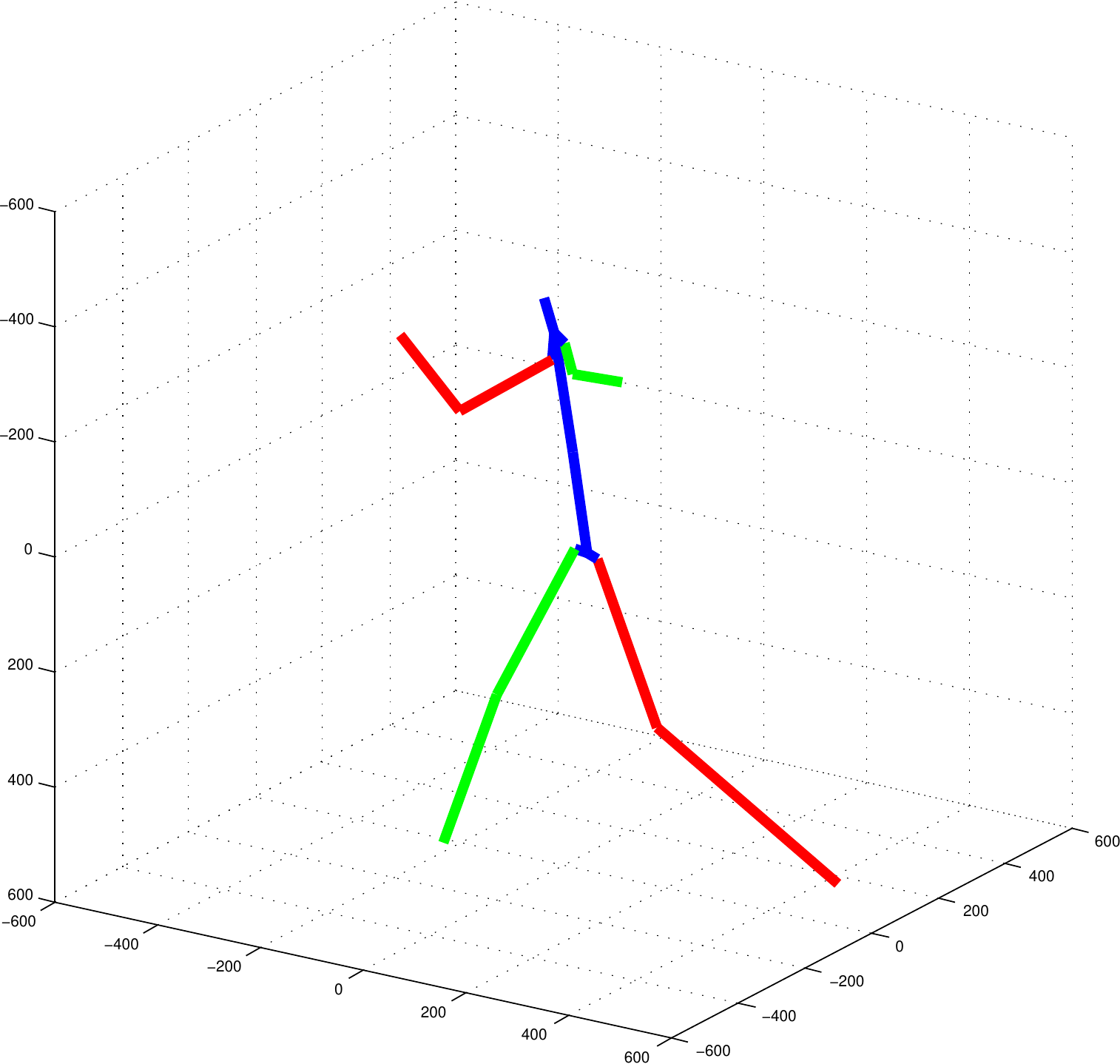}\hspace{1.9mm}
      \includegraphics[height=0.10\textwidth]{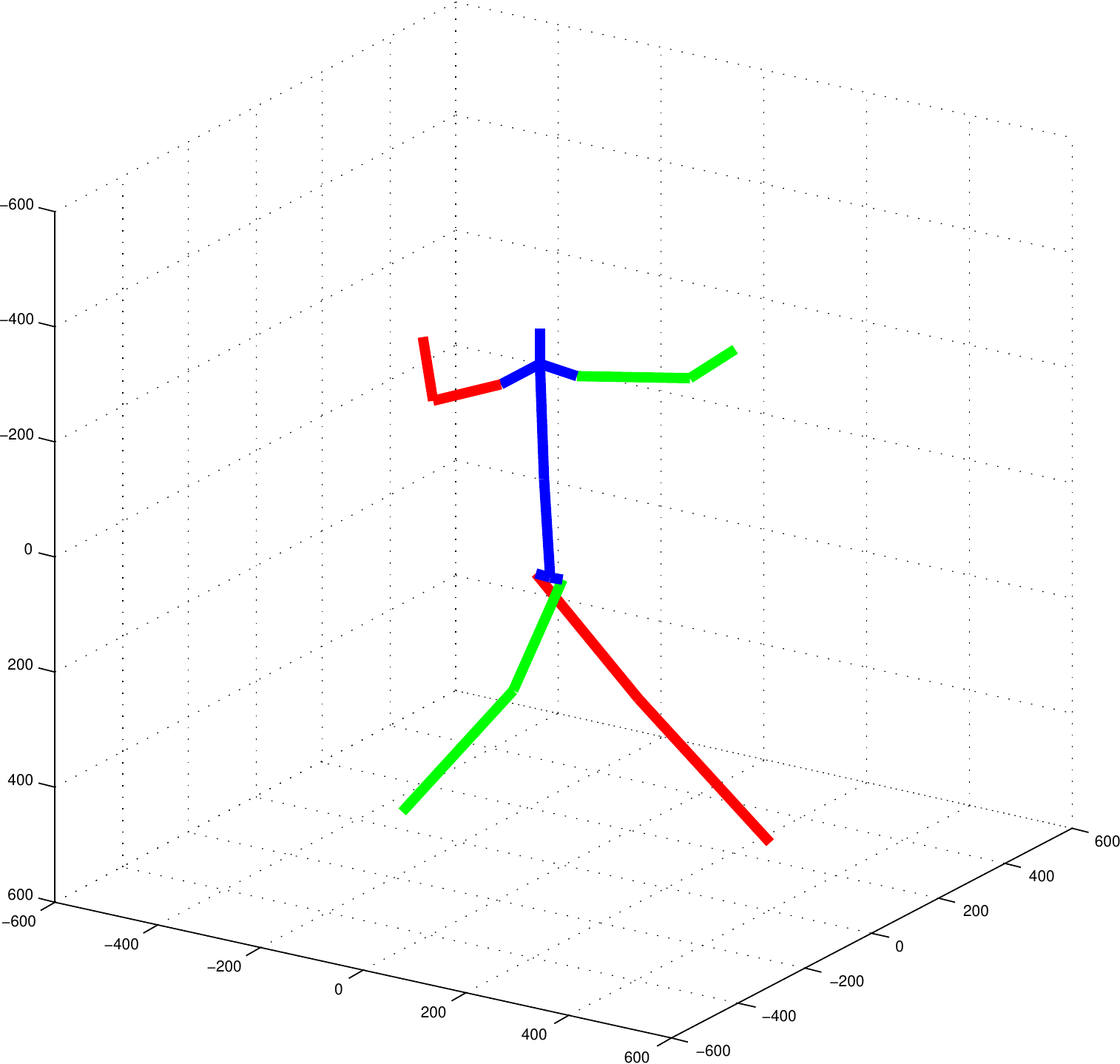}\hspace{1.9mm}
      \includegraphics[height=0.10\textwidth]{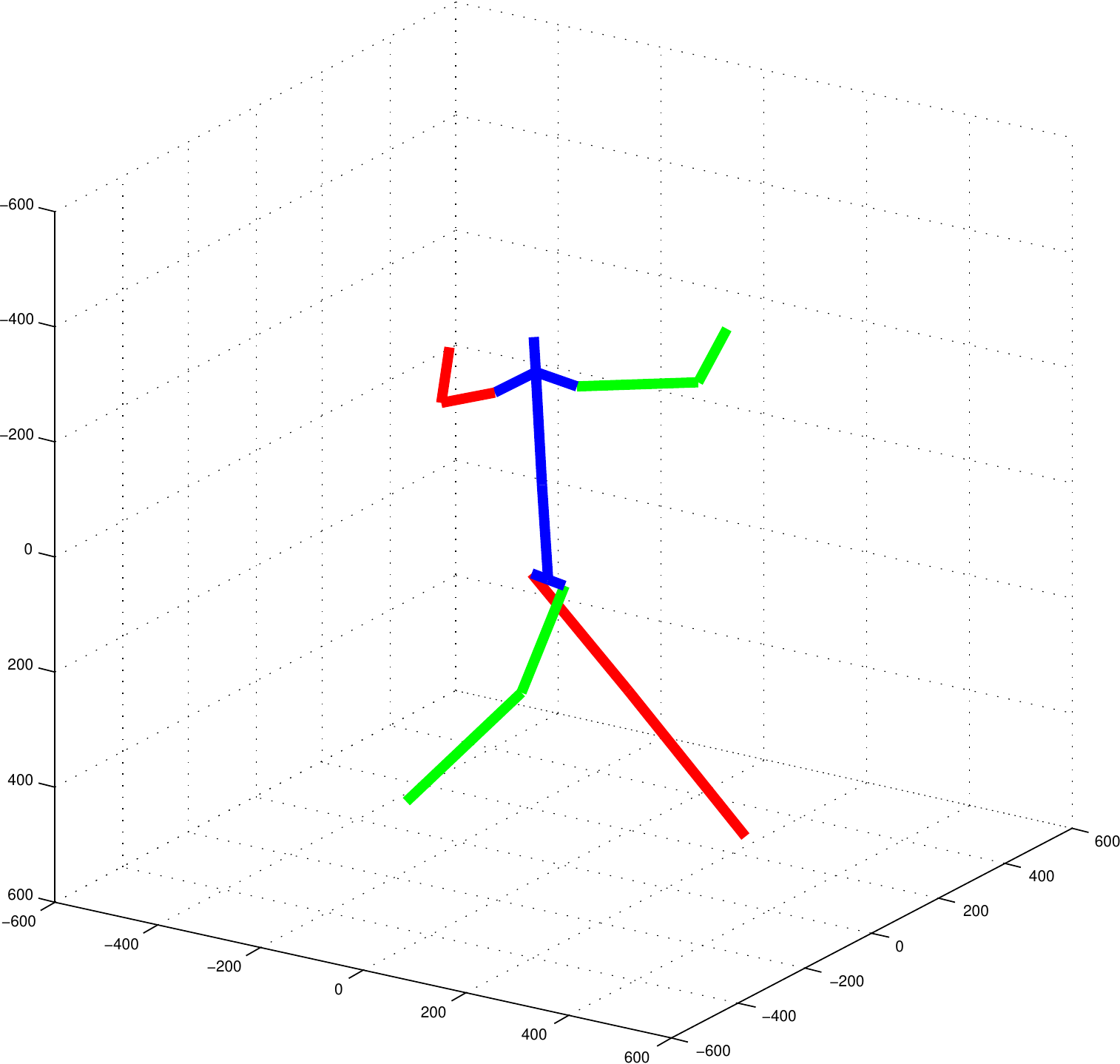}\hspace{1.9mm}
      \includegraphics[height=0.10\textwidth]{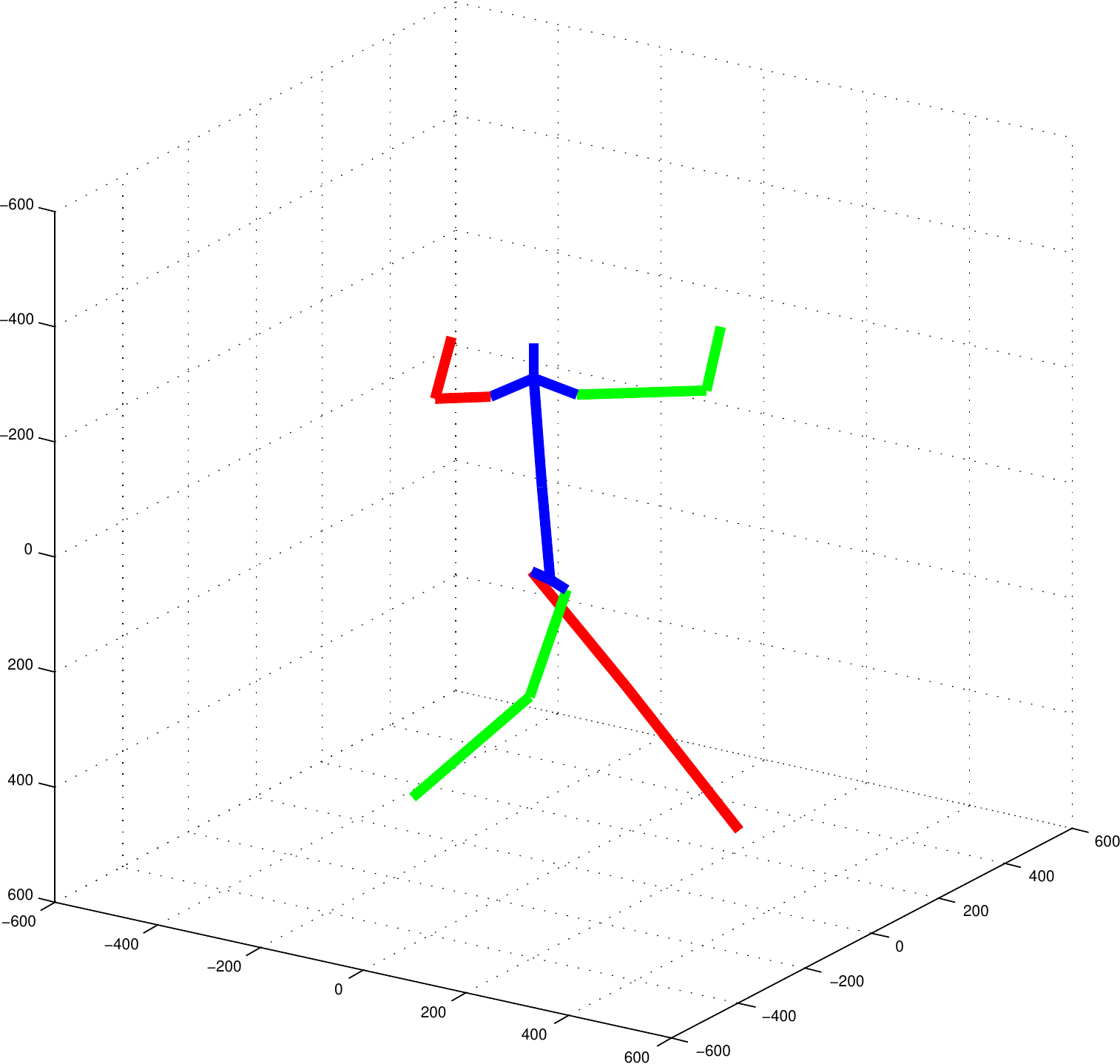}\hspace{1.9mm}
      \includegraphics[height=0.10\textwidth]{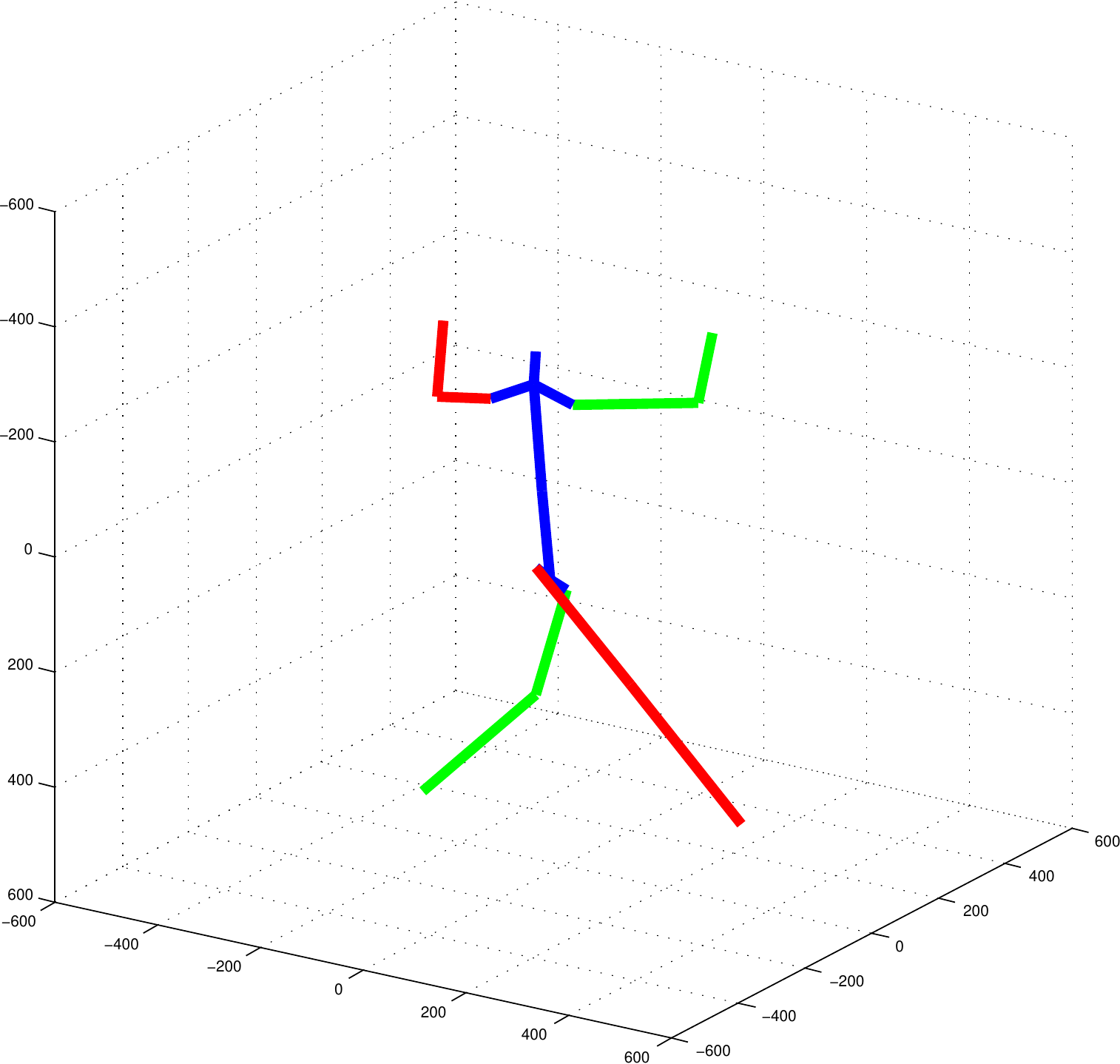}\hspace{1.9mm}
      \includegraphics[height=0.10\textwidth]{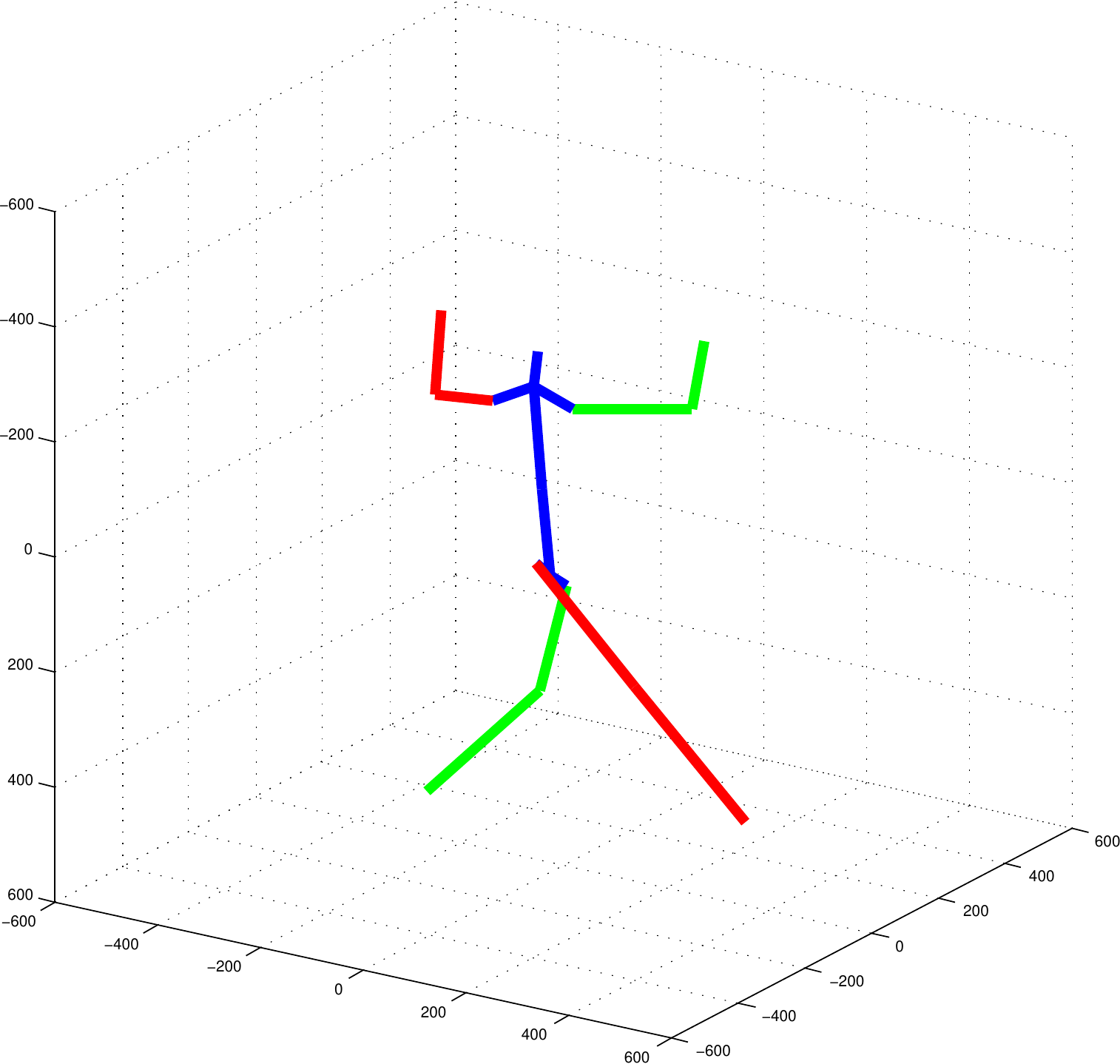}
      \\
      \hspace{-2.7mm}
      \includegraphics[height=0.10\textwidth]{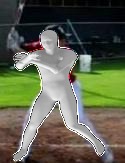}\hspace{0.92mm}
      \includegraphics[height=0.10\textwidth]{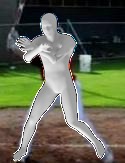}\hspace{0.92mm}
      \includegraphics[height=0.10\textwidth]{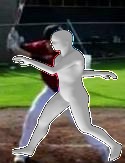}\hspace{0.92mm}
      \includegraphics[height=0.10\textwidth]{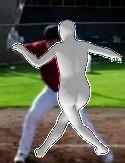}\hspace{0.92mm}
      \includegraphics[height=0.10\textwidth]{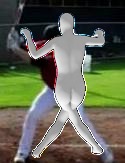}\hspace{0.92mm}
      \includegraphics[height=0.10\textwidth]{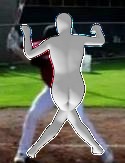}\hspace{0.92mm}
      \includegraphics[height=0.10\textwidth]{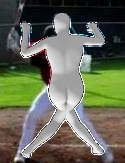}\hspace{0.92mm}
      \includegraphics[height=0.10\textwidth]{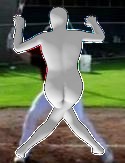}\hspace{0.92mm}
      \includegraphics[height=0.10\textwidth]{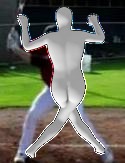}\hspace{0.92mm}
      \includegraphics[height=0.10\textwidth]{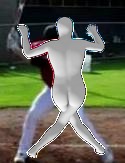}
      \\
      \hspace{-2.7mm}
      \includegraphics[height=0.10\textwidth]{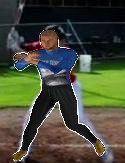}\hspace{0.92mm}
      \includegraphics[height=0.10\textwidth]{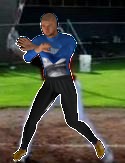}\hspace{0.92mm}
      \includegraphics[height=0.10\textwidth]{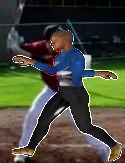}\hspace{0.92mm}
      \includegraphics[height=0.10\textwidth]{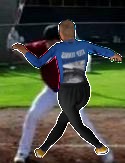}\hspace{0.92mm}
      \includegraphics[height=0.10\textwidth]{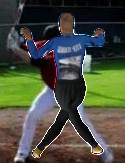}\hspace{0.92mm}
      \includegraphics[height=0.10\textwidth]{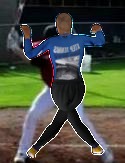}\hspace{0.92mm}
      \includegraphics[height=0.10\textwidth]{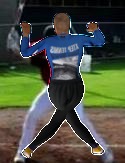}\hspace{0.92mm}
      \includegraphics[height=0.10\textwidth]{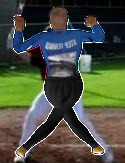}\hspace{0.92mm}
      \includegraphics[height=0.10\textwidth]{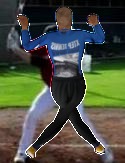}\hspace{0.92mm}
      \includegraphics[height=0.10\textwidth]{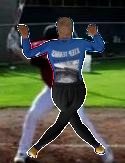}
    \end{tabular}
  \end{tabular}
  % \vspace{-2mm}
  \caption{\small Additional qualitative results of pose forecasting. The left
column shows the input images. For each input image, we show in the right
column the sequence of ground-truth frame and pose (row 1), our forecasted pose
sequence in 2D (row 2) and 3D (row 3), and the rendered human body without
texture (row 4) and with skin and cloth textures (row 5).}
  % \phantomcaption
  % \vspace{-2mm}
  \label{fig:additional2}
\end{figure*}

\begin{figure*}[t]%\ContinuedFloat
  % \vspace{-2mm}
  \centering
  \footnotesize
  \begin{tabular}{L{0.12\linewidth}@{\hspace{1.0mm}}|L{0.9\linewidth}@{\hspace{-0.0mm}}}
    \includegraphics[height=0.152\textwidth]{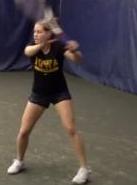}
    &
    \vspace{-2.7mm}
    \begin{tabular}{L{1.0\linewidth}}
      \hspace{-2.7mm}
      \includegraphics[height=0.10\textwidth]{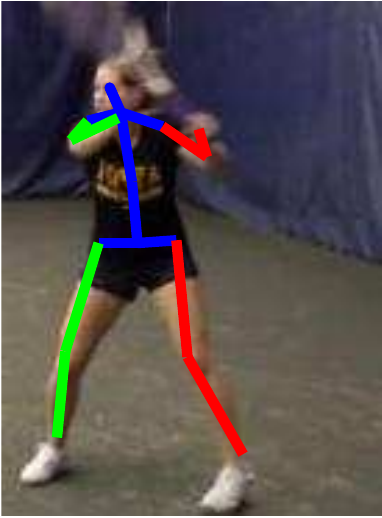}\hspace{1.47mm}
      \includegraphics[height=0.10\textwidth]{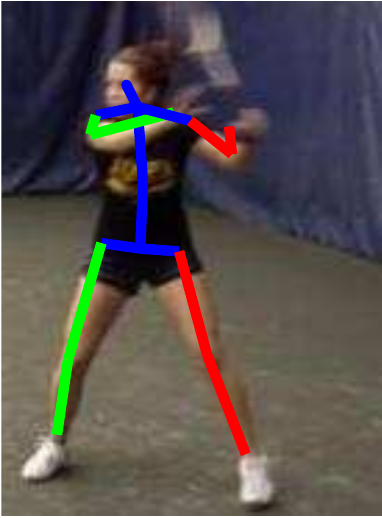}\hspace{1.47mm}
      \includegraphics[height=0.10\textwidth]{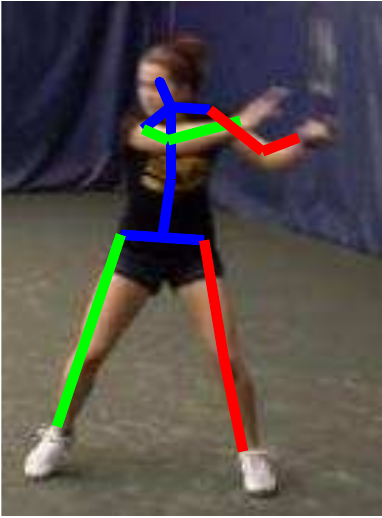}\hspace{1.47mm}
      \includegraphics[height=0.10\textwidth]{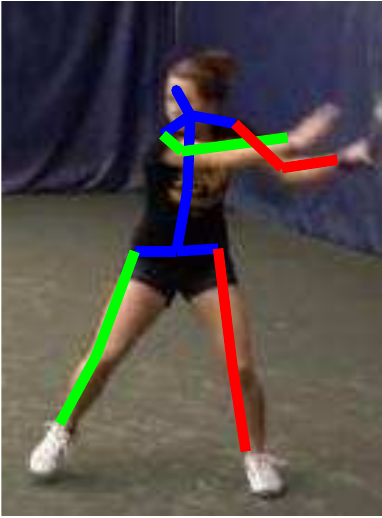}\hspace{1.47mm}
      \includegraphics[height=0.10\textwidth]{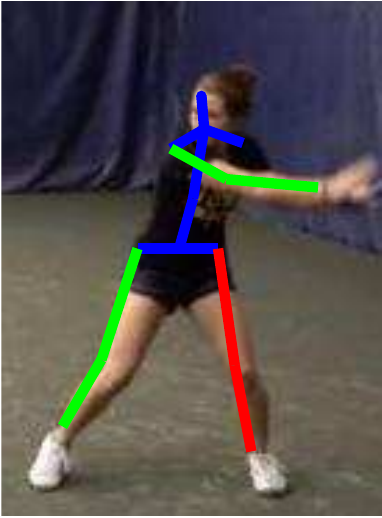}\hspace{1.47mm}
      \includegraphics[height=0.10\textwidth]{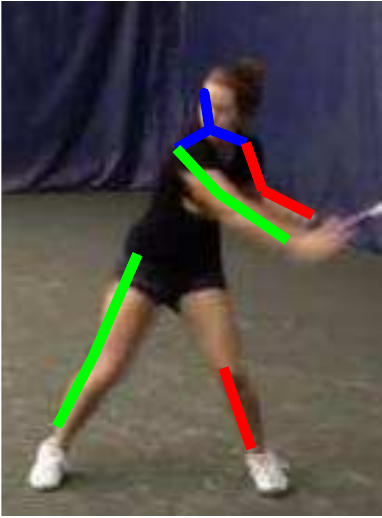}\hspace{1.47mm}
      \includegraphics[height=0.10\textwidth]{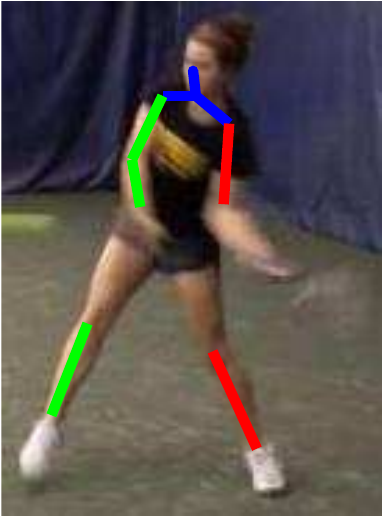}\hspace{1.47mm}
      \includegraphics[height=0.10\textwidth]{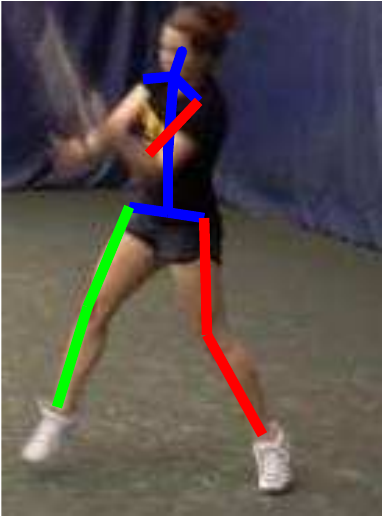}\hspace{1.47mm}
      \includegraphics[height=0.10\textwidth]{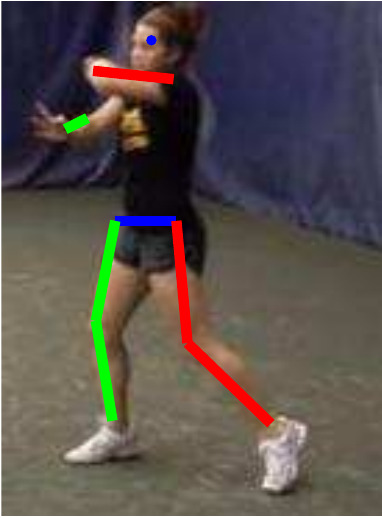}\hspace{1.47mm}
      \includegraphics[height=0.10\textwidth]{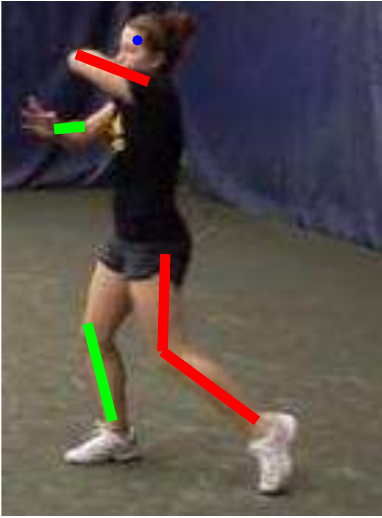}
      \\
      \hspace{-2.7mm}
      \includegraphics[height=0.10\textwidth]{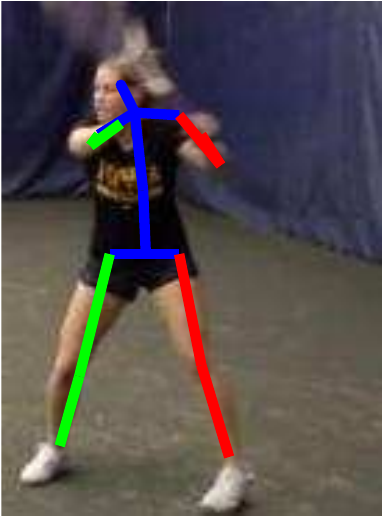}\hspace{1.47mm}
      \includegraphics[height=0.10\textwidth]{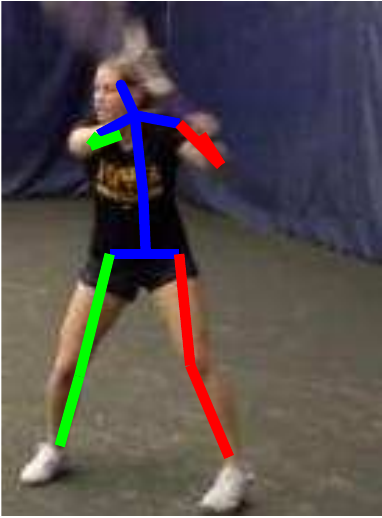}\hspace{1.47mm}
      \includegraphics[height=0.10\textwidth]{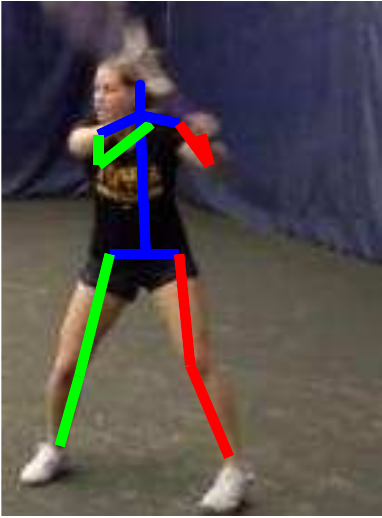}\hspace{1.47mm}
      \includegraphics[height=0.10\textwidth]{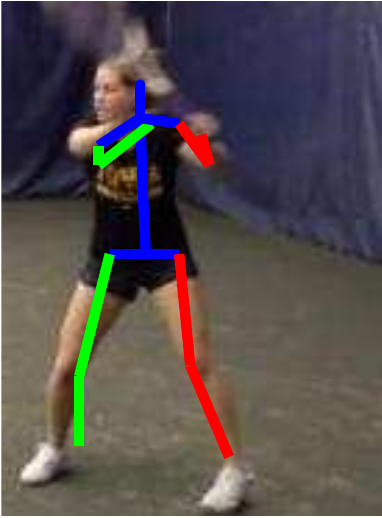}\hspace{1.47mm}
      \includegraphics[height=0.10\textwidth]{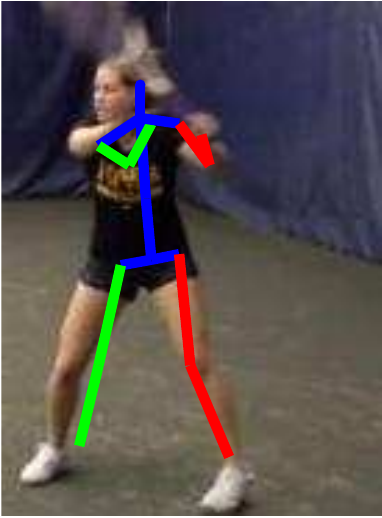}\hspace{1.47mm}
      \includegraphics[height=0.10\textwidth]{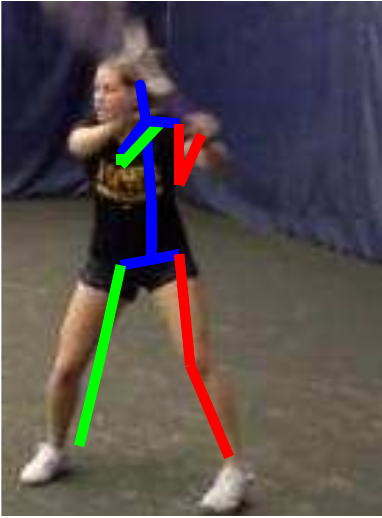}\hspace{1.47mm}
      \includegraphics[height=0.10\textwidth]{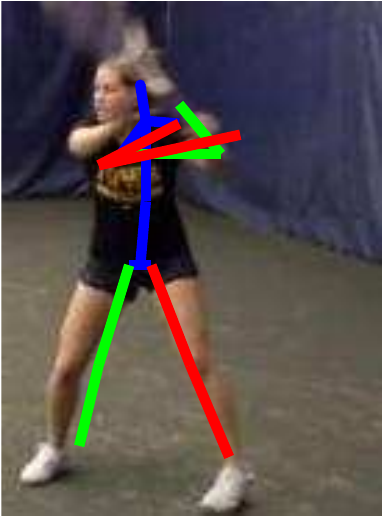}\hspace{1.47mm}
      \includegraphics[height=0.10\textwidth]{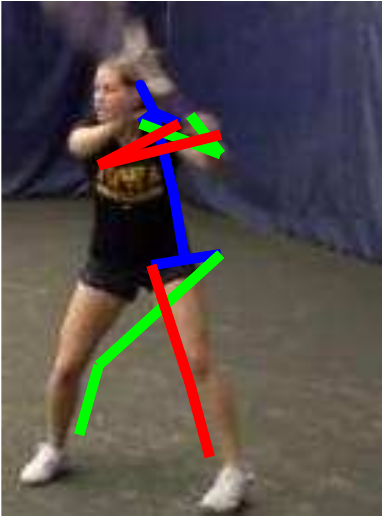}\hspace{1.47mm}
      \includegraphics[height=0.10\textwidth]{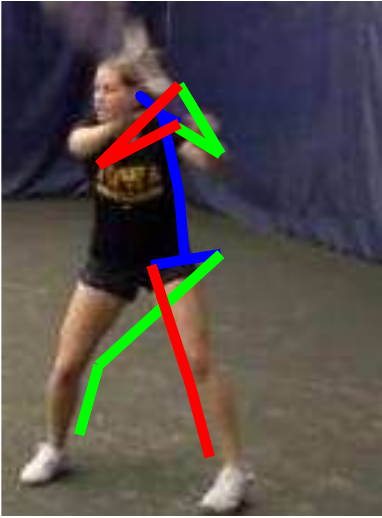}\hspace{1.47mm}
      \includegraphics[height=0.10\textwidth]{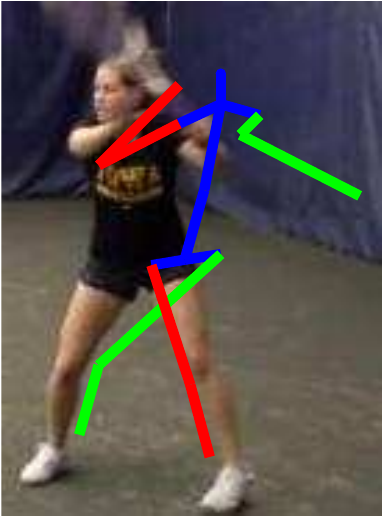}
      \\
      \includegraphics[height=0.10\textwidth]{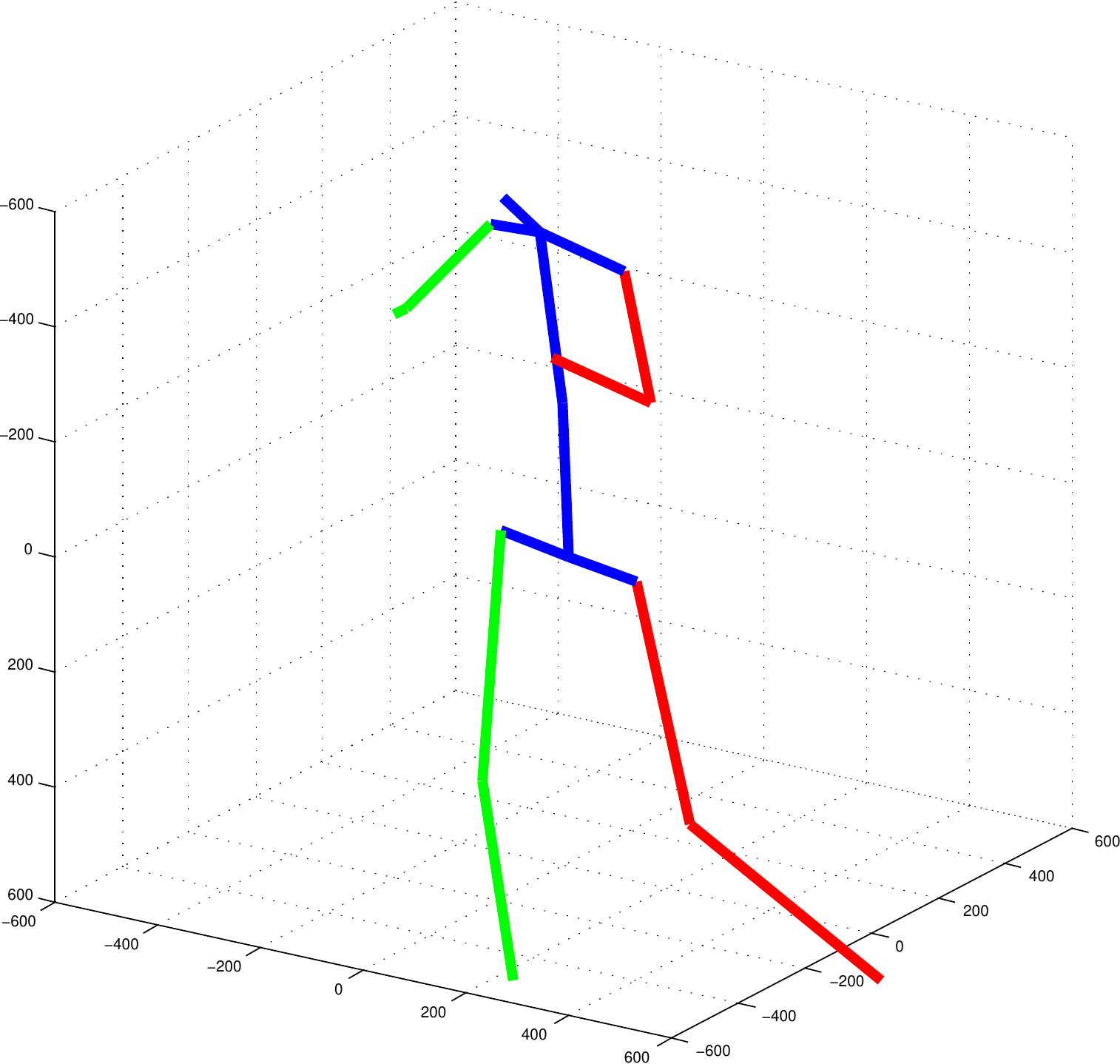}\hspace{1.9mm}
      \includegraphics[height=0.10\textwidth]{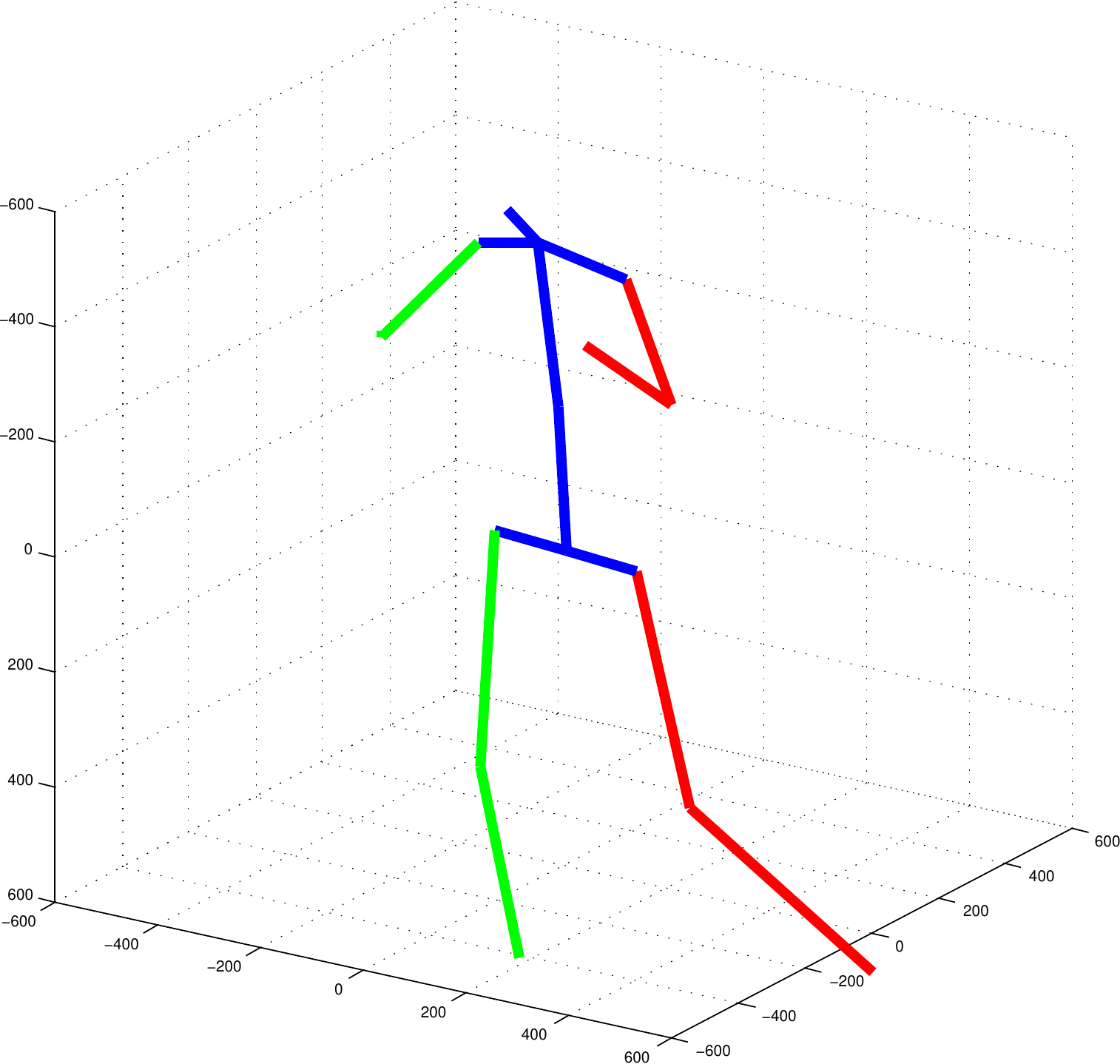}\hspace{1.9mm}
      \includegraphics[height=0.10\textwidth]{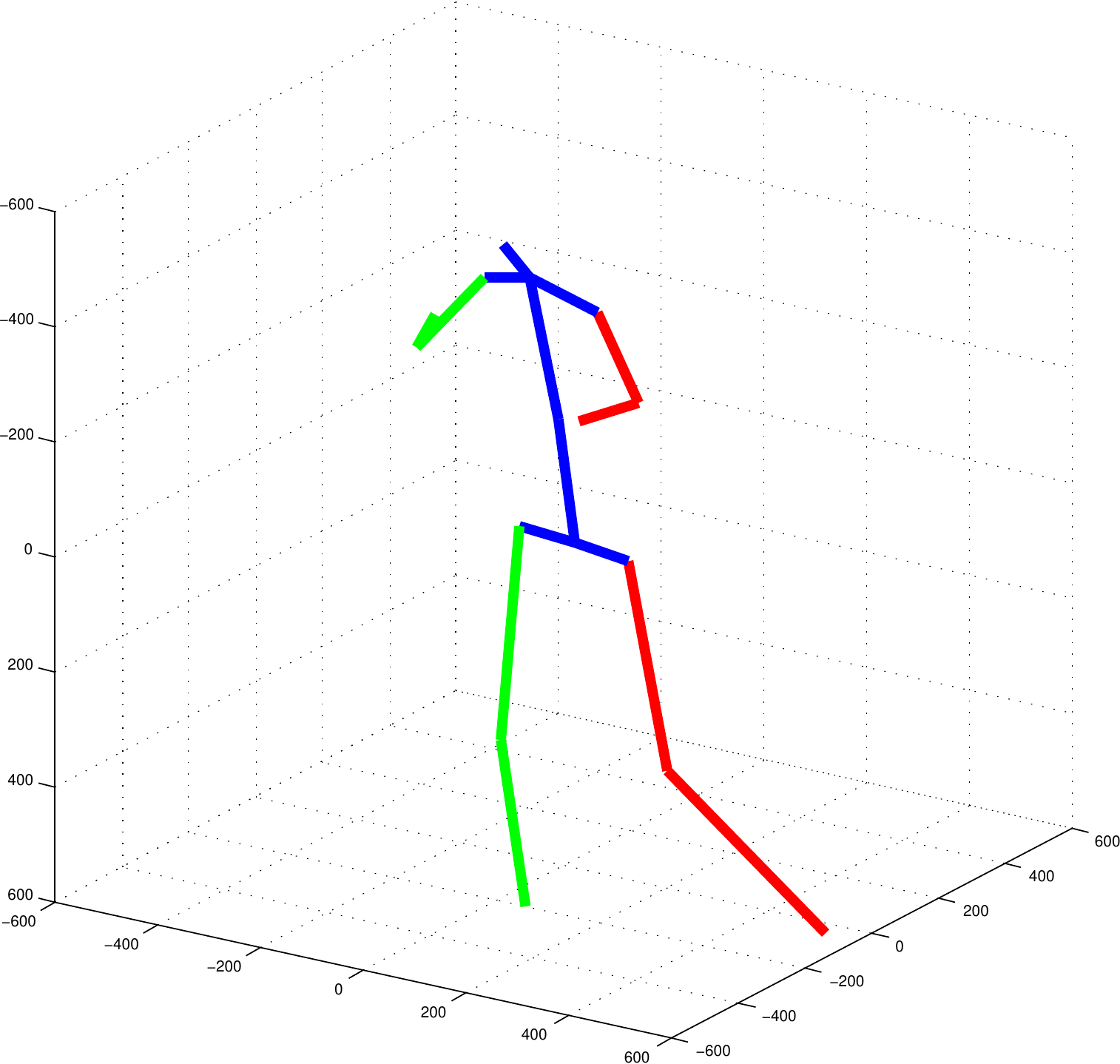}\hspace{1.9mm}
      \includegraphics[height=0.10\textwidth]{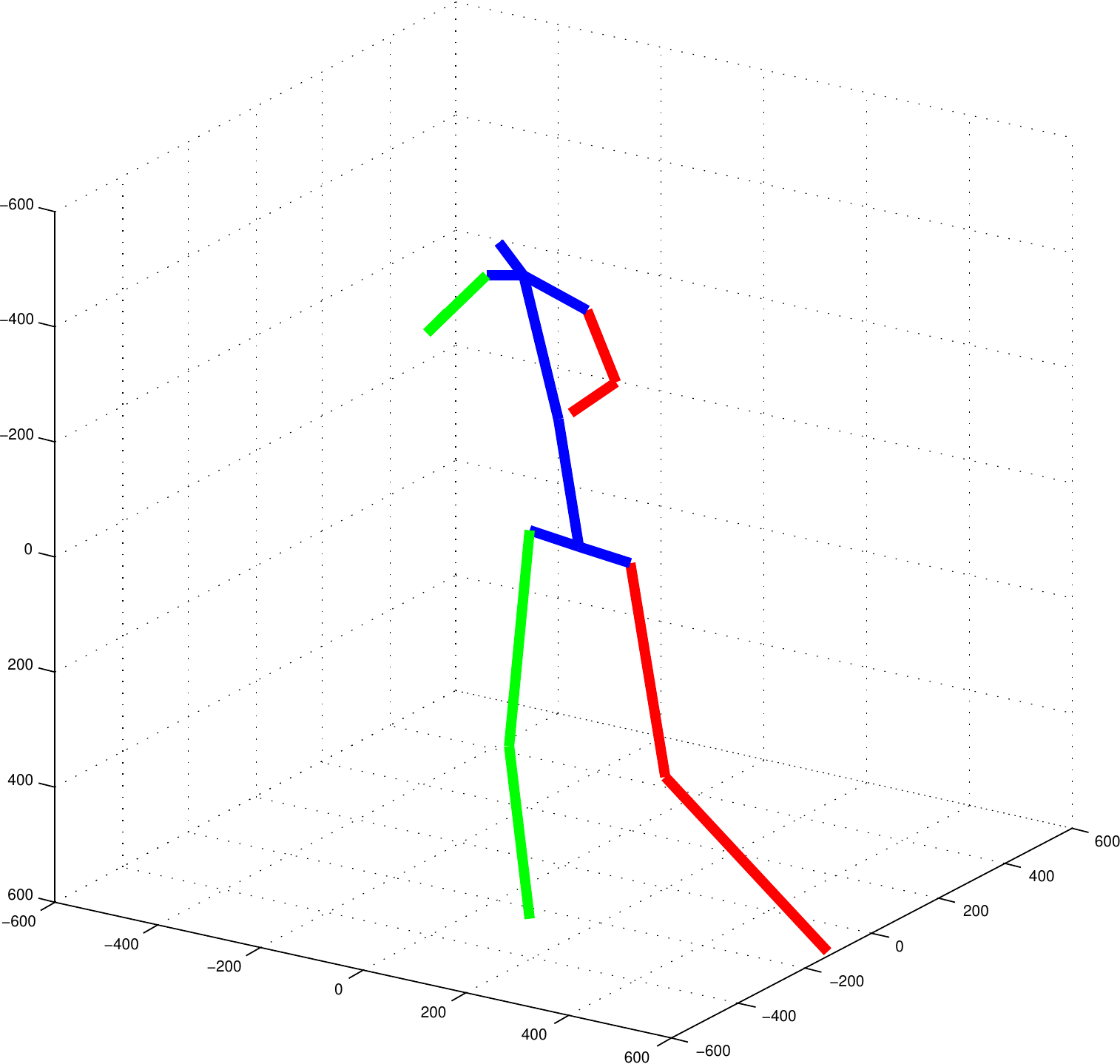}\hspace{1.9mm}
      \includegraphics[height=0.10\textwidth]{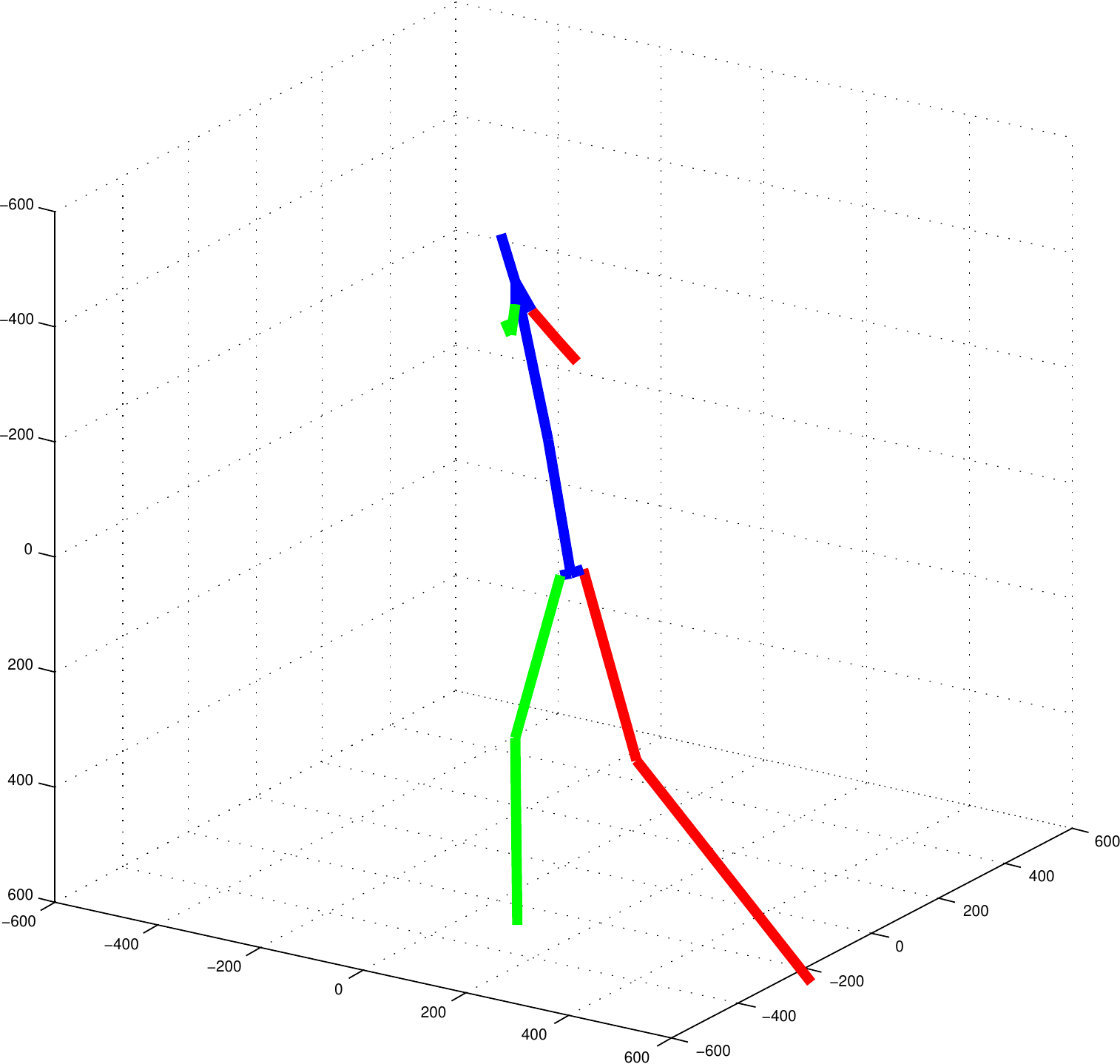}\hspace{1.9mm}
      \includegraphics[height=0.10\textwidth]{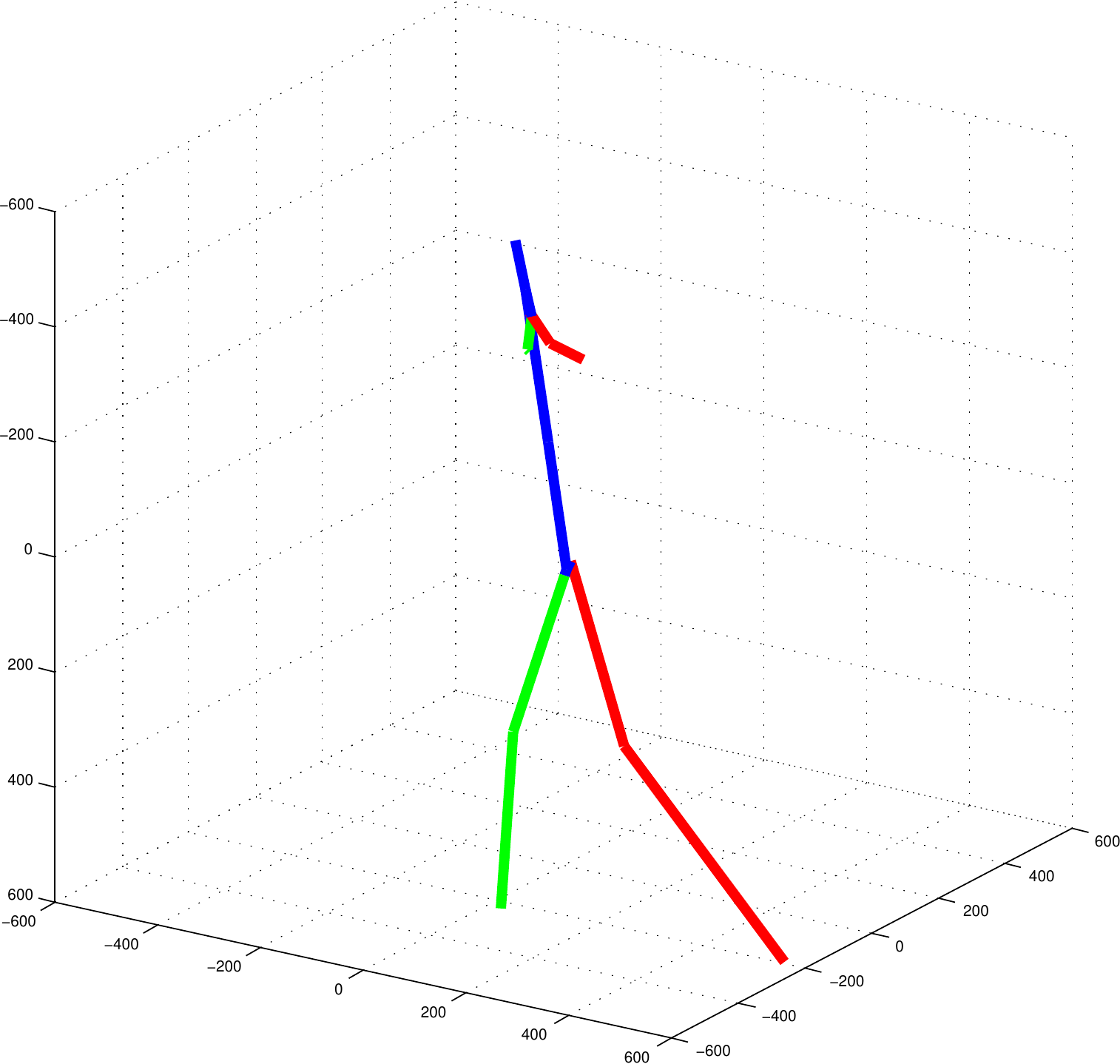}\hspace{1.9mm}
      \includegraphics[height=0.10\textwidth]{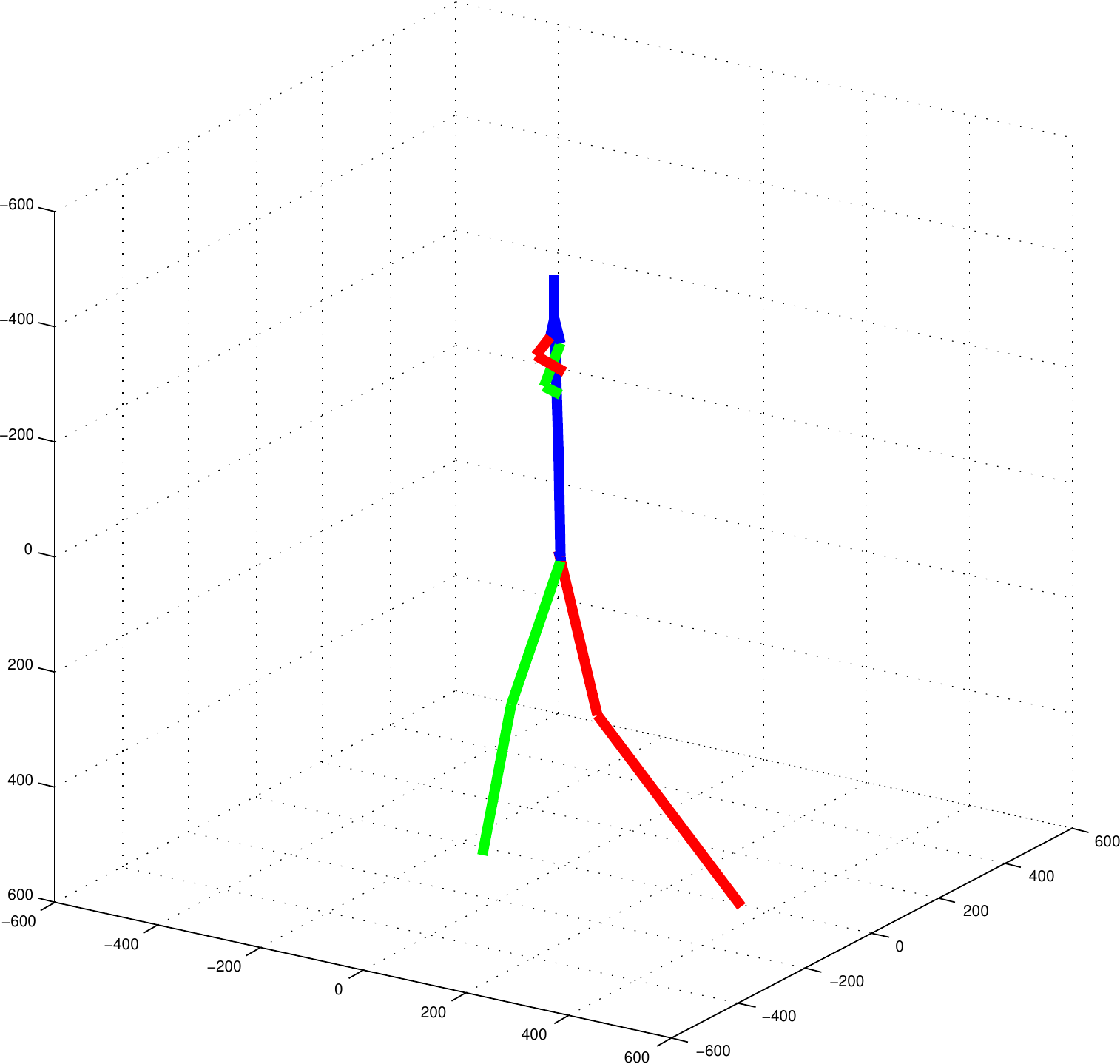}
      \\
      \hspace{-2.7mm}
      \includegraphics[height=0.10\textwidth]{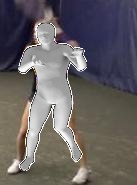}\hspace{1.43mm}
      \includegraphics[height=0.10\textwidth]{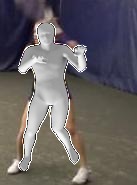}\hspace{1.43mm}
      \includegraphics[height=0.10\textwidth]{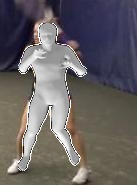}\hspace{1.43mm}
      \includegraphics[height=0.10\textwidth]{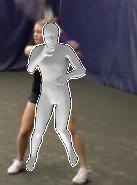}\hspace{1.43mm}
      \includegraphics[height=0.10\textwidth]{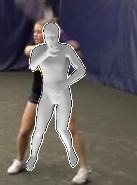}\hspace{1.43mm}
      \includegraphics[height=0.10\textwidth]{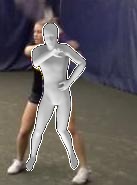}\hspace{1.43mm}
      \includegraphics[height=0.10\textwidth]{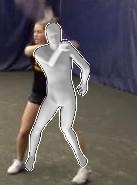}\hspace{1.43mm}
      \includegraphics[height=0.10\textwidth]{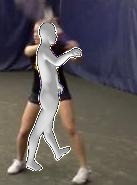}\hspace{1.43mm}
      \includegraphics[height=0.10\textwidth]{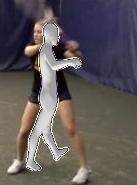}\hspace{1.43mm}
      \includegraphics[height=0.10\textwidth]{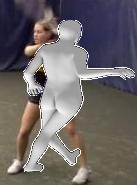}
      \\
      \hspace{-2.7mm}
      \includegraphics[height=0.10\textwidth]{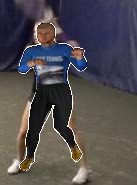}\hspace{1.43mm}
      \includegraphics[height=0.10\textwidth]{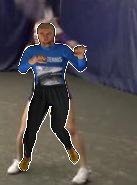}\hspace{1.43mm}
      \includegraphics[height=0.10\textwidth]{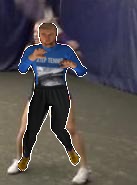}\hspace{1.43mm}
      \includegraphics[height=0.10\textwidth]{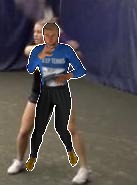}\hspace{1.43mm}
      \includegraphics[height=0.10\textwidth]{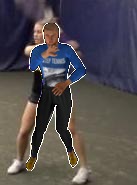}\hspace{1.43mm}
      \includegraphics[height=0.10\textwidth]{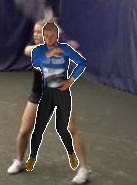}\hspace{1.43mm}
      \includegraphics[height=0.10\textwidth]{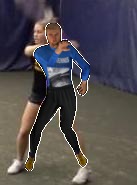}\hspace{1.43mm}
      \includegraphics[height=0.10\textwidth]{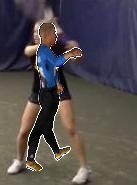}\hspace{1.43mm}
      \includegraphics[height=0.10\textwidth]{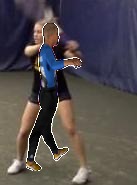}\hspace{1.43mm}
      \includegraphics[height=0.10\textwidth]{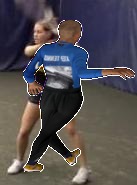}
    \end{tabular}
    \\ [-0.0em] & \\
    \includegraphics[height=0.19\textwidth]{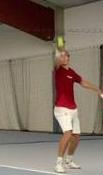}
    &
    \vspace{-2.7mm}
    \begin{tabular}{L{1.0\linewidth}}
      \hspace{-2.7mm}
      \includegraphics[height=0.10\textwidth]{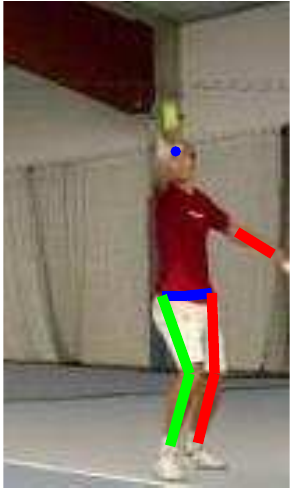}\hspace{0.41mm}
      \includegraphics[height=0.10\textwidth]{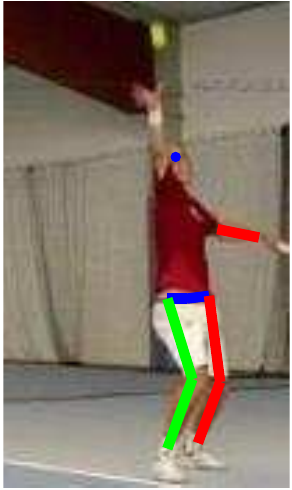}\hspace{0.41mm}
      \includegraphics[height=0.10\textwidth]{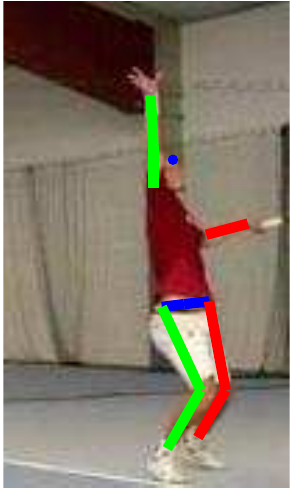}\hspace{0.41mm}
      \includegraphics[height=0.10\textwidth]{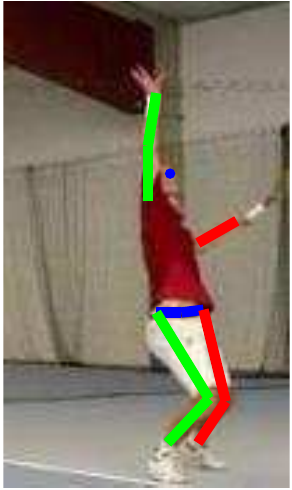}\hspace{0.41mm}
      \includegraphics[height=0.10\textwidth]{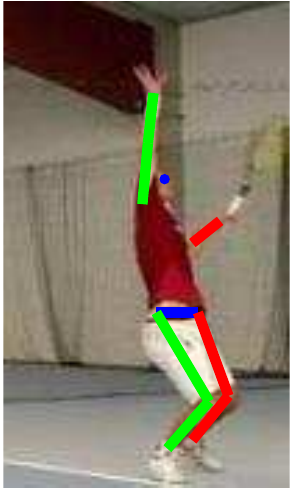}\hspace{0.41mm}
      \includegraphics[height=0.10\textwidth]{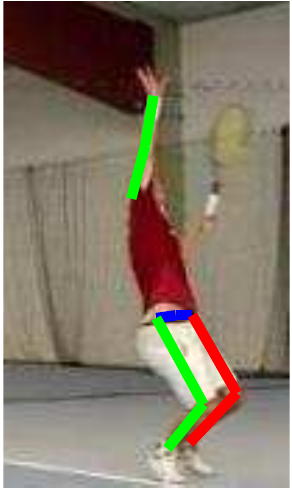}\hspace{0.41mm}
      \includegraphics[height=0.10\textwidth]{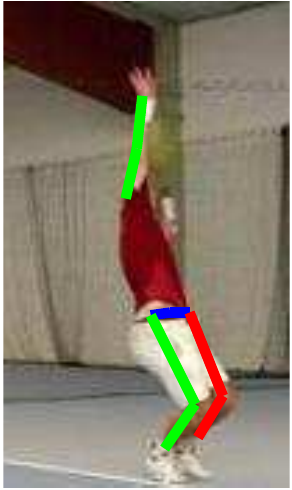}\hspace{0.41mm}
      \includegraphics[height=0.10\textwidth]{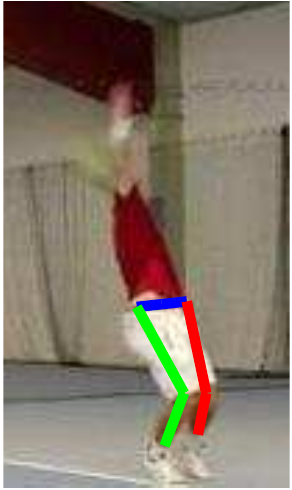}\hspace{0.41mm}
      \includegraphics[height=0.10\textwidth]{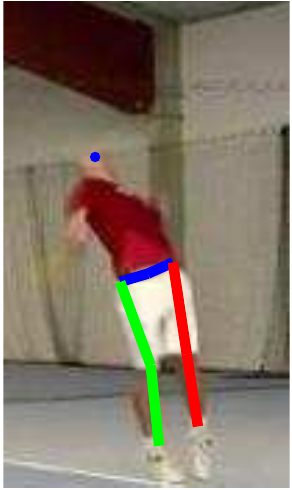}\hspace{0.41mm}
      \includegraphics[height=0.10\textwidth]{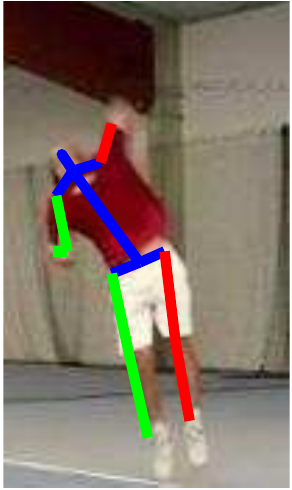}\hspace{0.41mm}
      \includegraphics[height=0.10\textwidth]{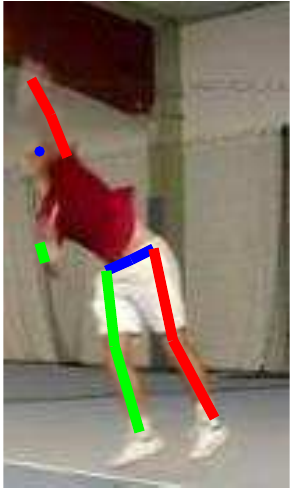}\hspace{0.41mm}
      \includegraphics[height=0.10\textwidth]{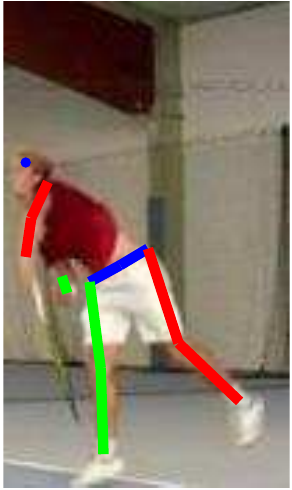}\hspace{0.41mm}
      \includegraphics[height=0.10\textwidth]{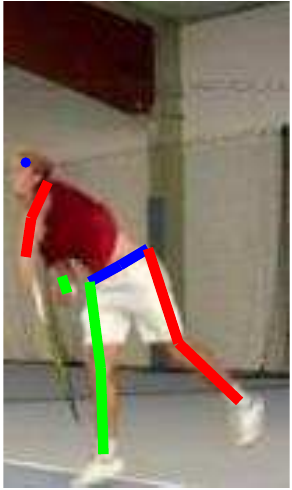}
      \\
      \hspace{-2.7mm}
      \includegraphics[height=0.10\textwidth]{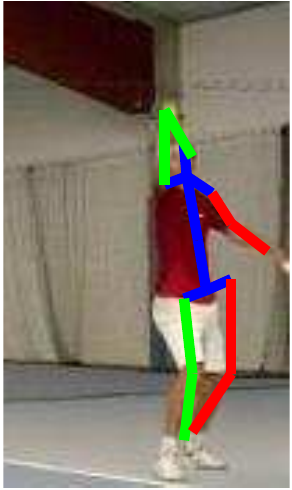}\hspace{0.41mm}
      \includegraphics[height=0.10\textwidth]{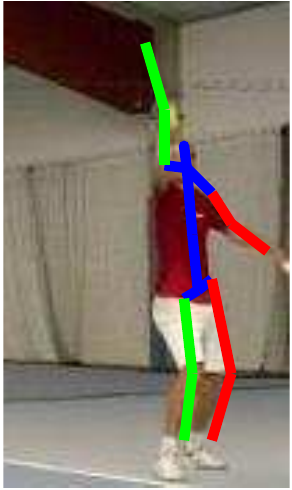}\hspace{0.41mm}
      \includegraphics[height=0.10\textwidth]{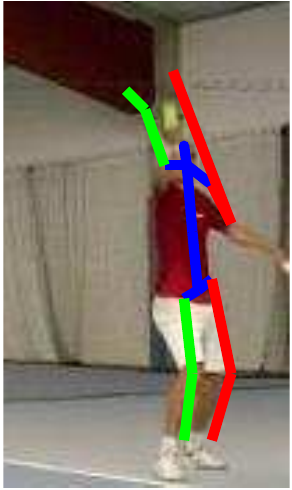}\hspace{0.41mm}
      \includegraphics[height=0.10\textwidth]{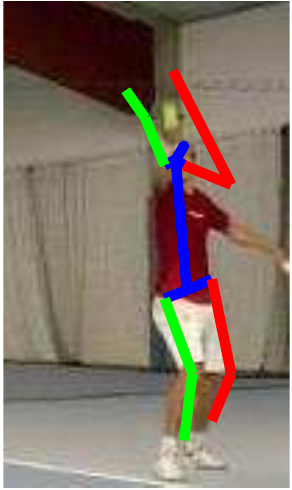}\hspace{0.41mm}
      \includegraphics[height=0.10\textwidth]{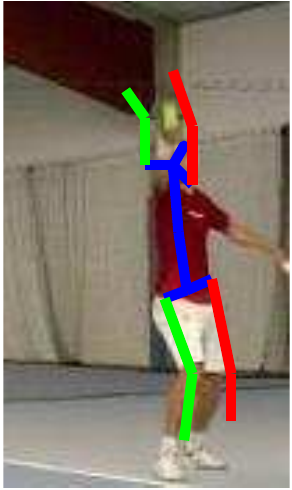}\hspace{0.41mm}
      \includegraphics[height=0.10\textwidth]{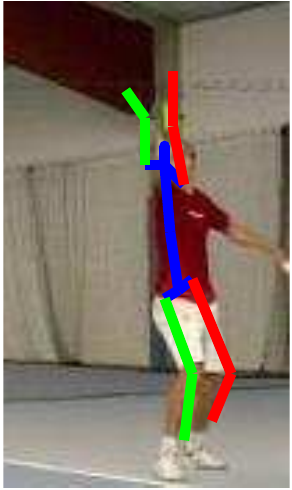}\hspace{0.41mm}
      \includegraphics[height=0.10\textwidth]{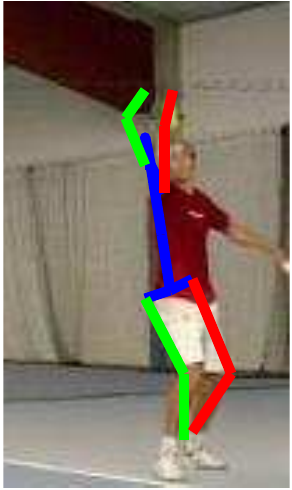}\hspace{0.41mm}
      \includegraphics[height=0.10\textwidth]{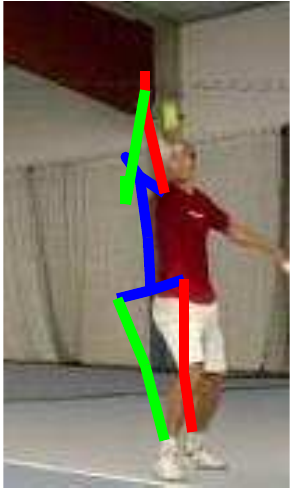}\hspace{0.41mm}
      \includegraphics[height=0.10\textwidth]{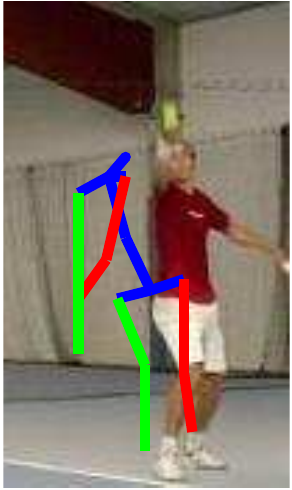}\hspace{0.41mm}
      \includegraphics[height=0.10\textwidth]{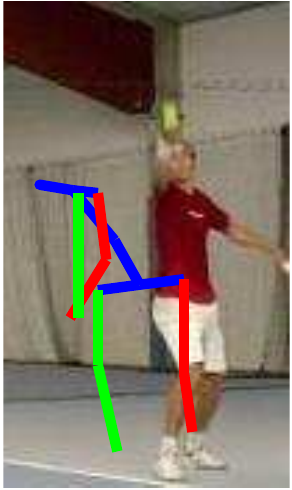}\hspace{0.41mm}
      \includegraphics[height=0.10\textwidth]{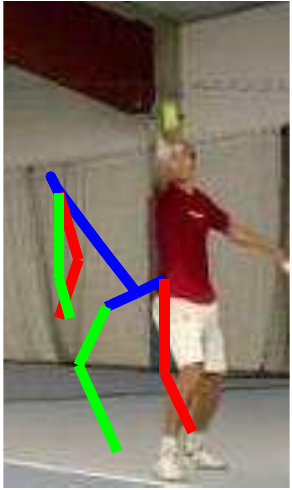}\hspace{0.41mm}
      \includegraphics[height=0.10\textwidth]{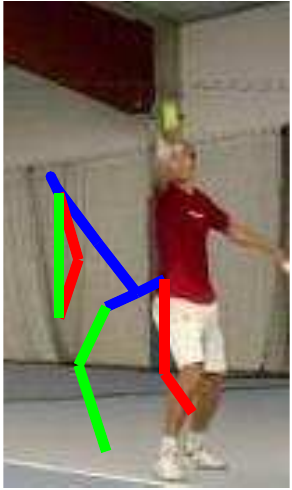}\hspace{0.41mm}
      \includegraphics[height=0.10\textwidth]{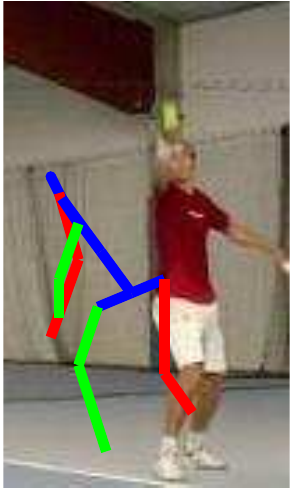}
      \\
      \includegraphics[height=0.10\textwidth]{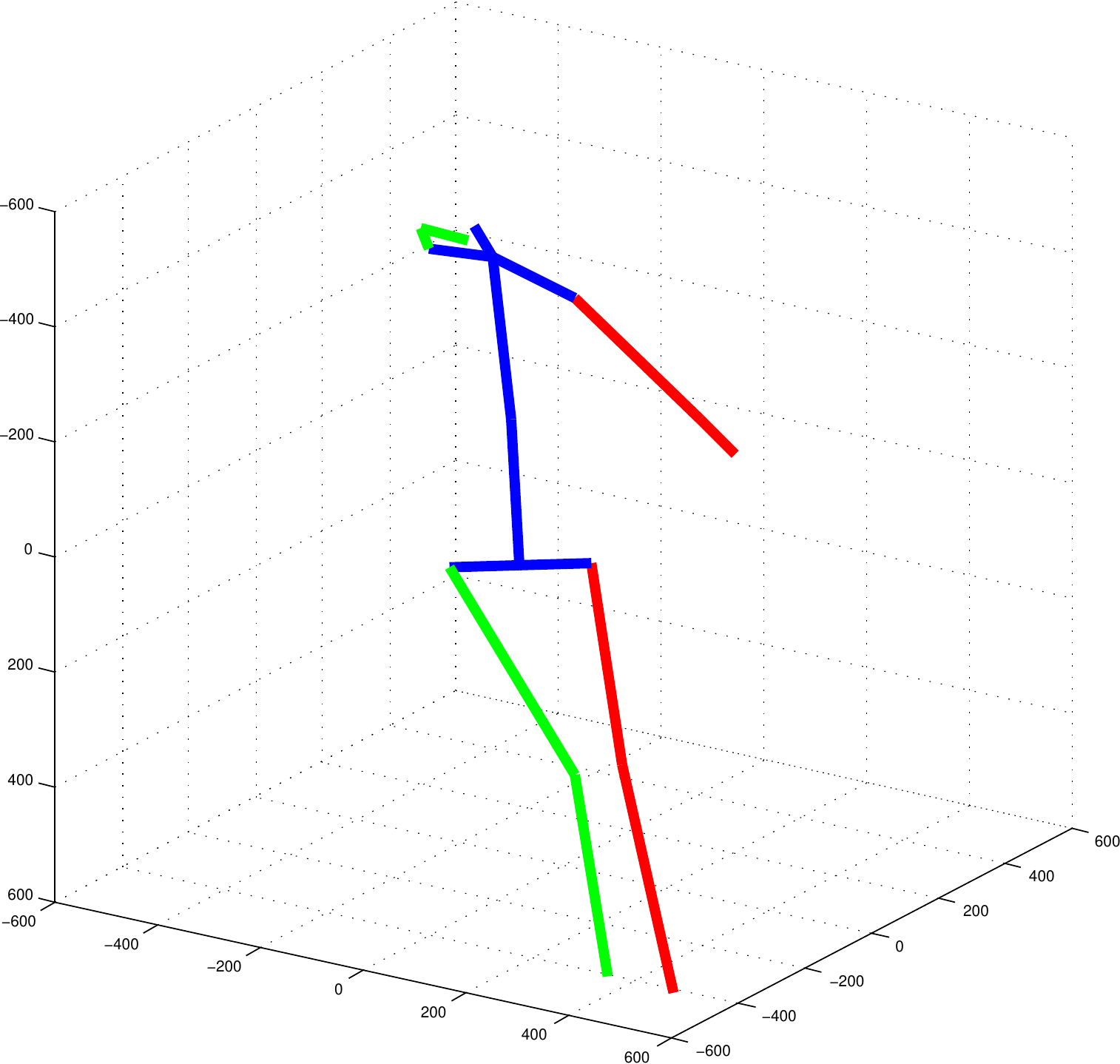}\hspace{1.9mm}
      \includegraphics[height=0.10\textwidth]{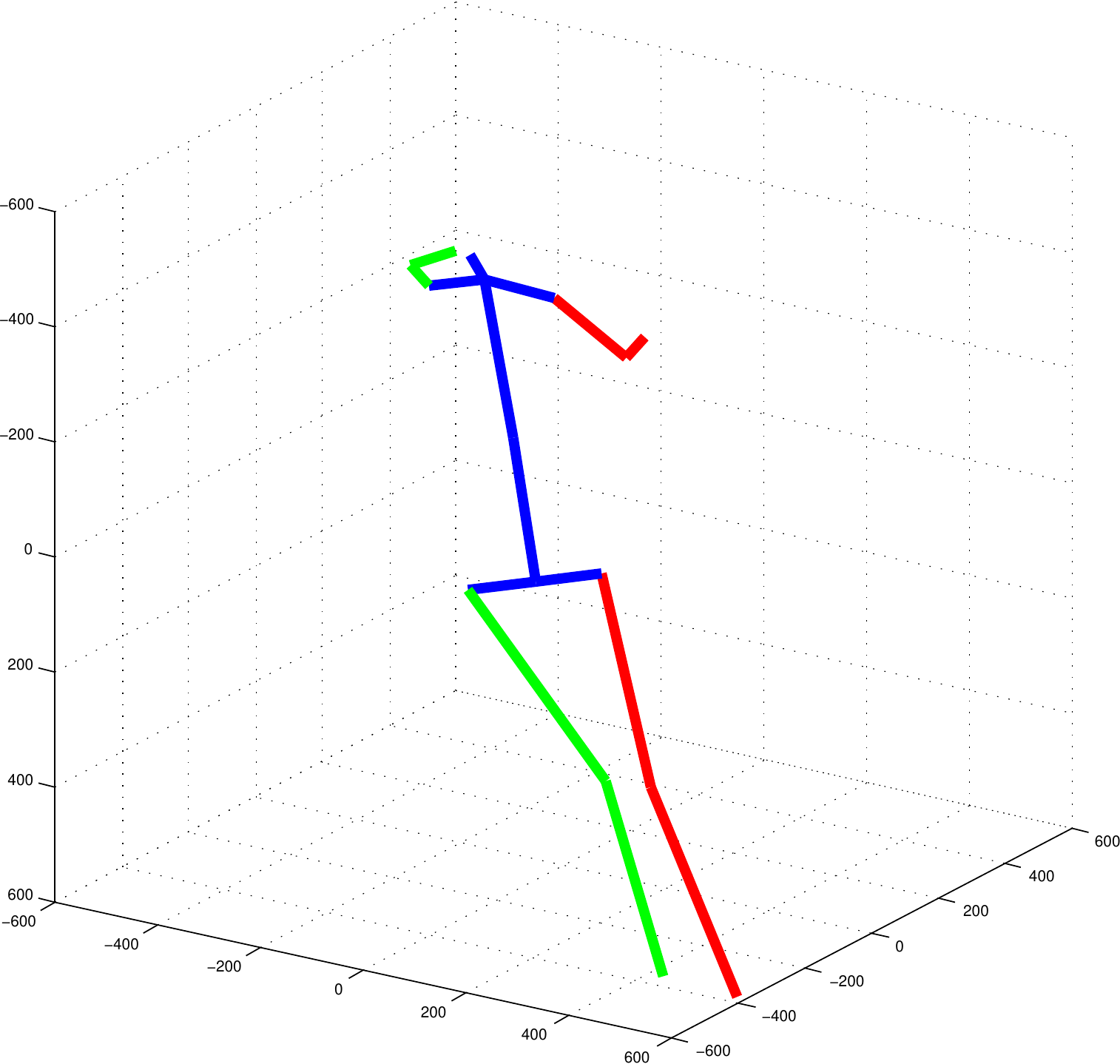}\hspace{1.9mm}
      \includegraphics[height=0.10\textwidth]{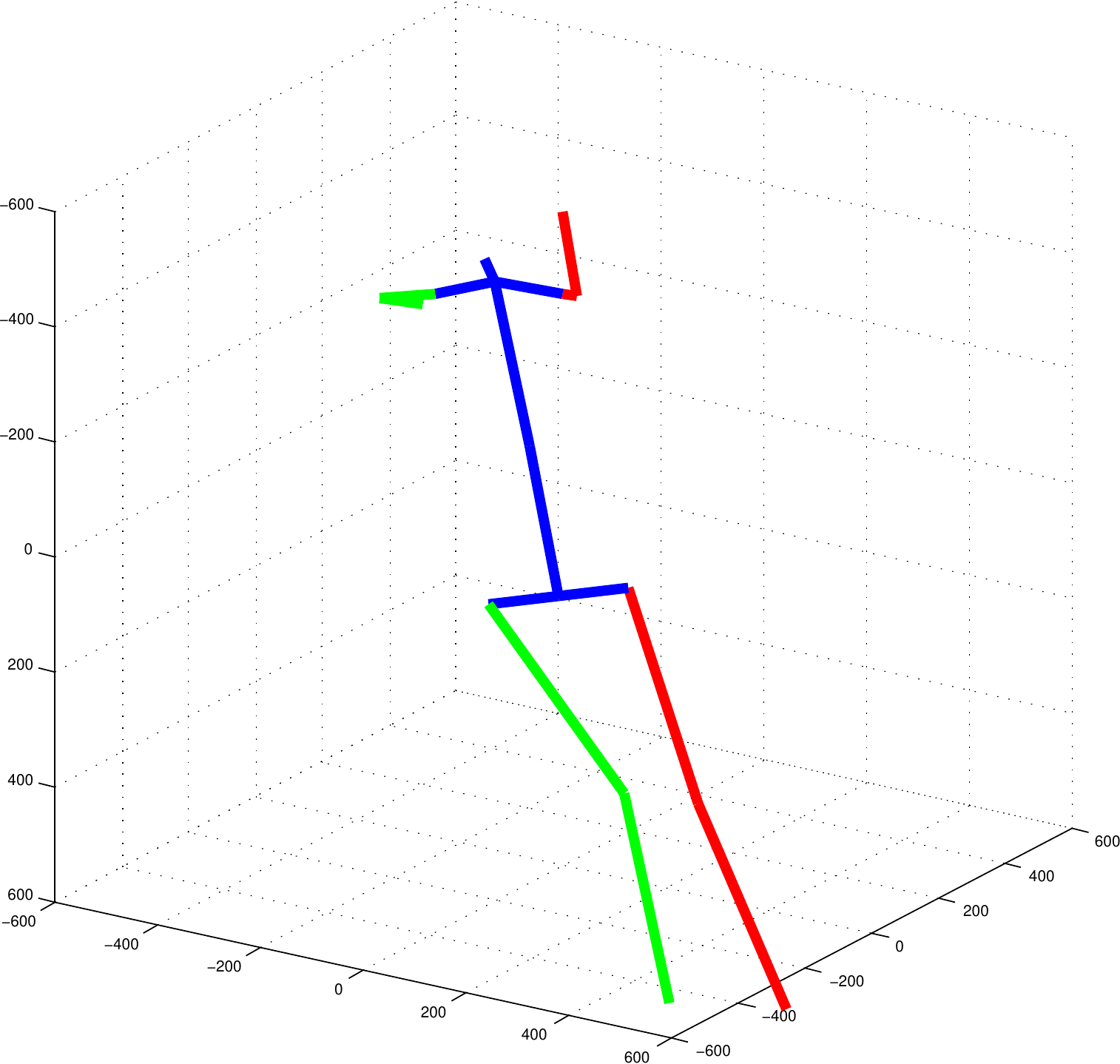}\hspace{1.9mm}
      \includegraphics[height=0.10\textwidth]{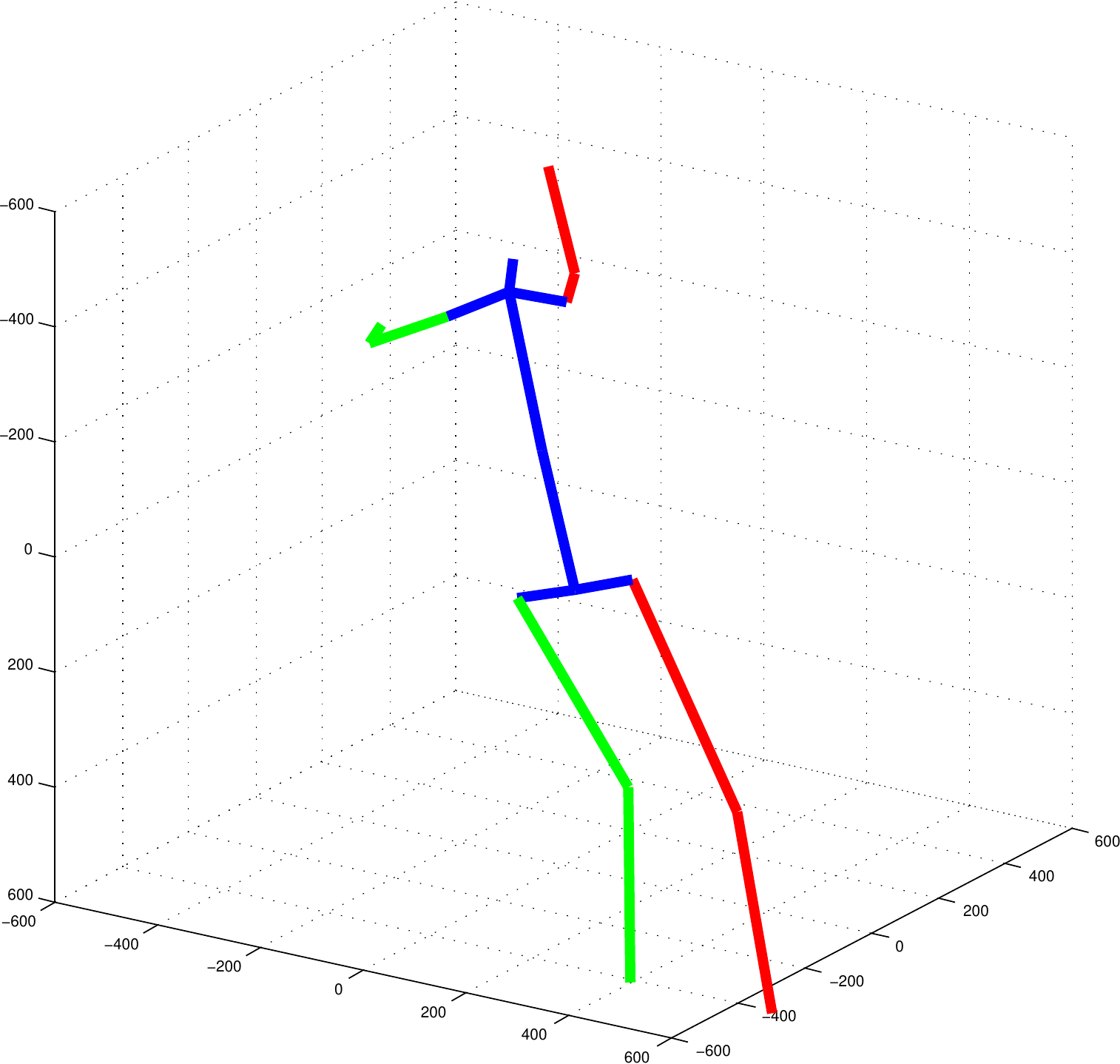}\hspace{1.9mm}
      \includegraphics[height=0.10\textwidth]{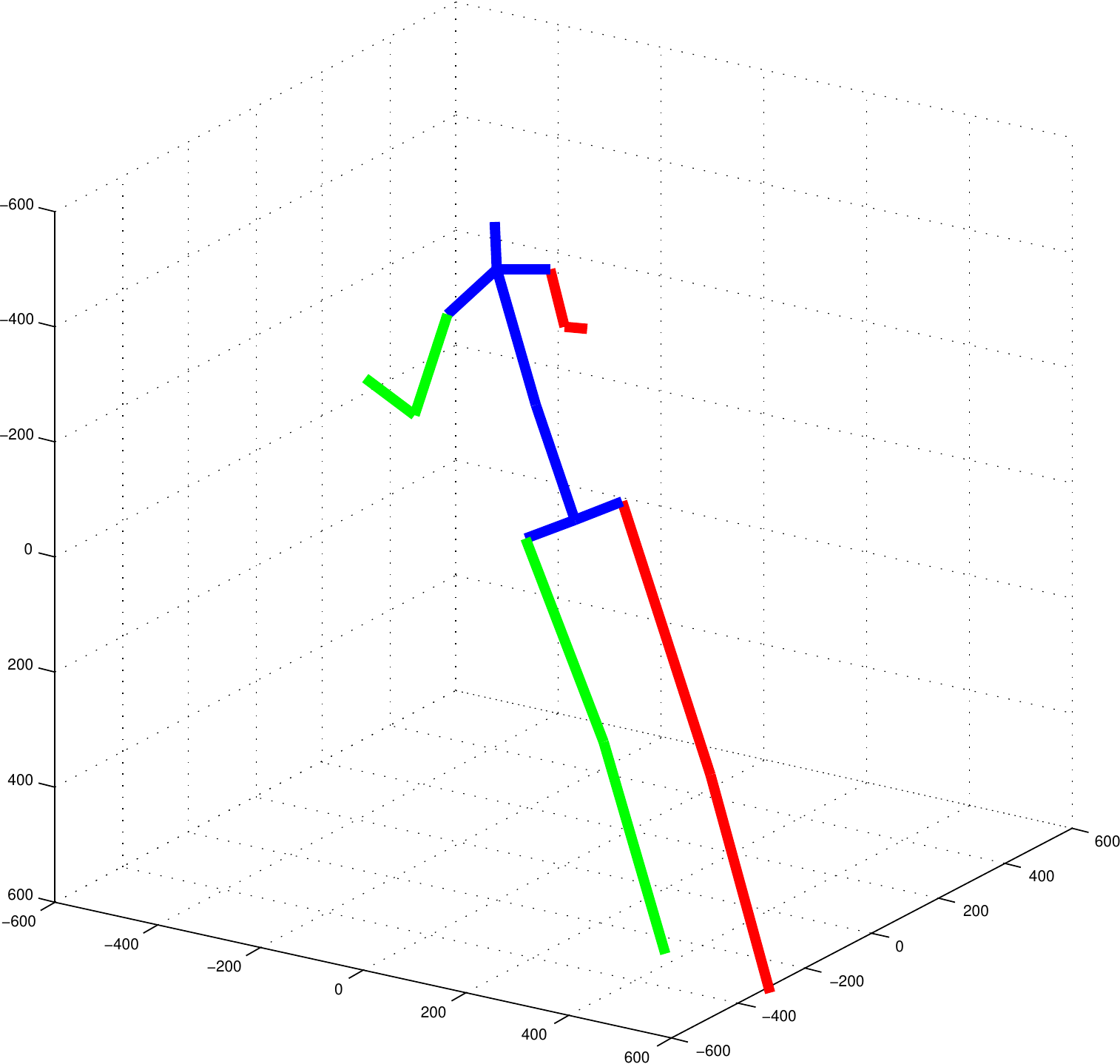}\hspace{1.9mm}
      \includegraphics[height=0.10\textwidth]{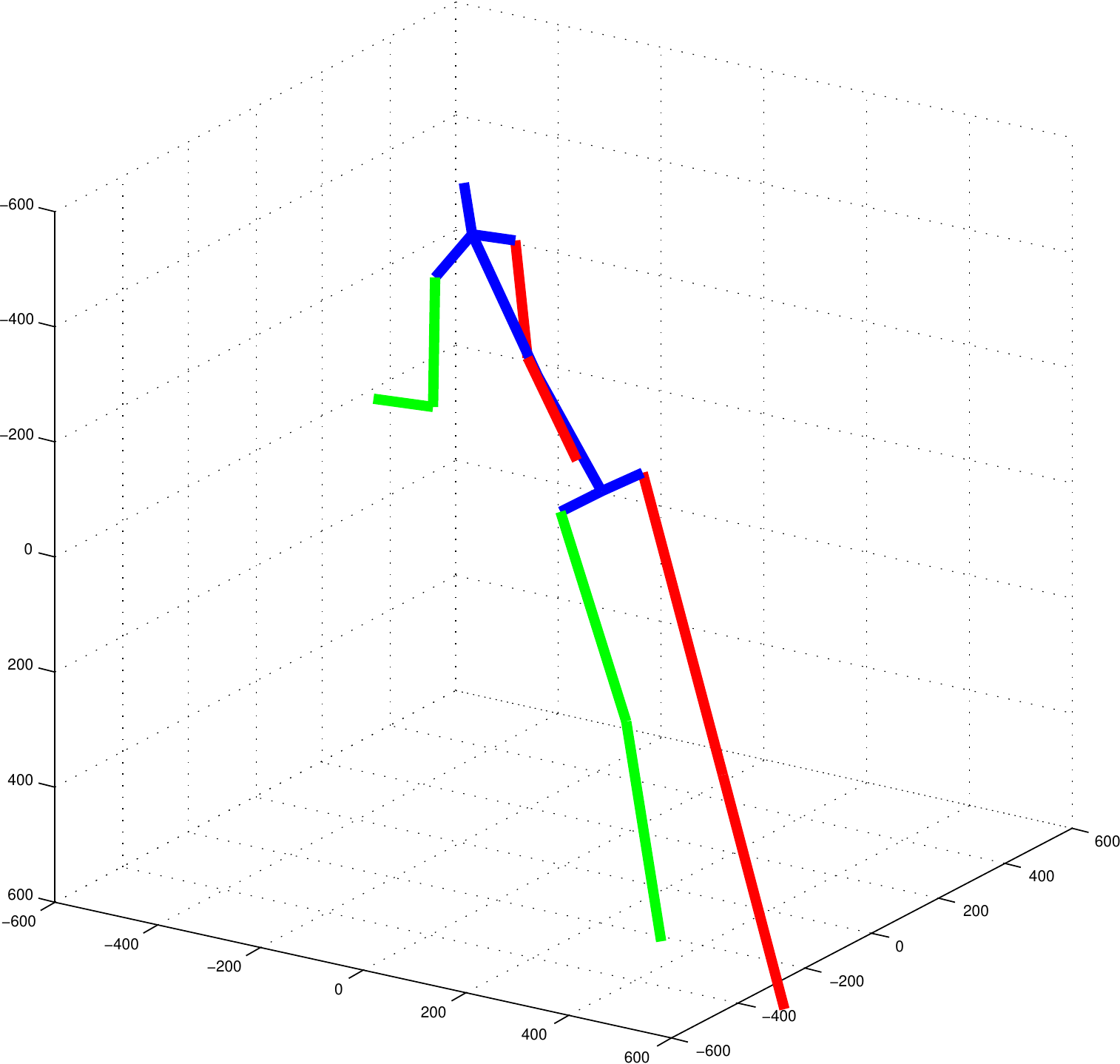}\hspace{1.9mm}
      \includegraphics[height=0.10\textwidth]{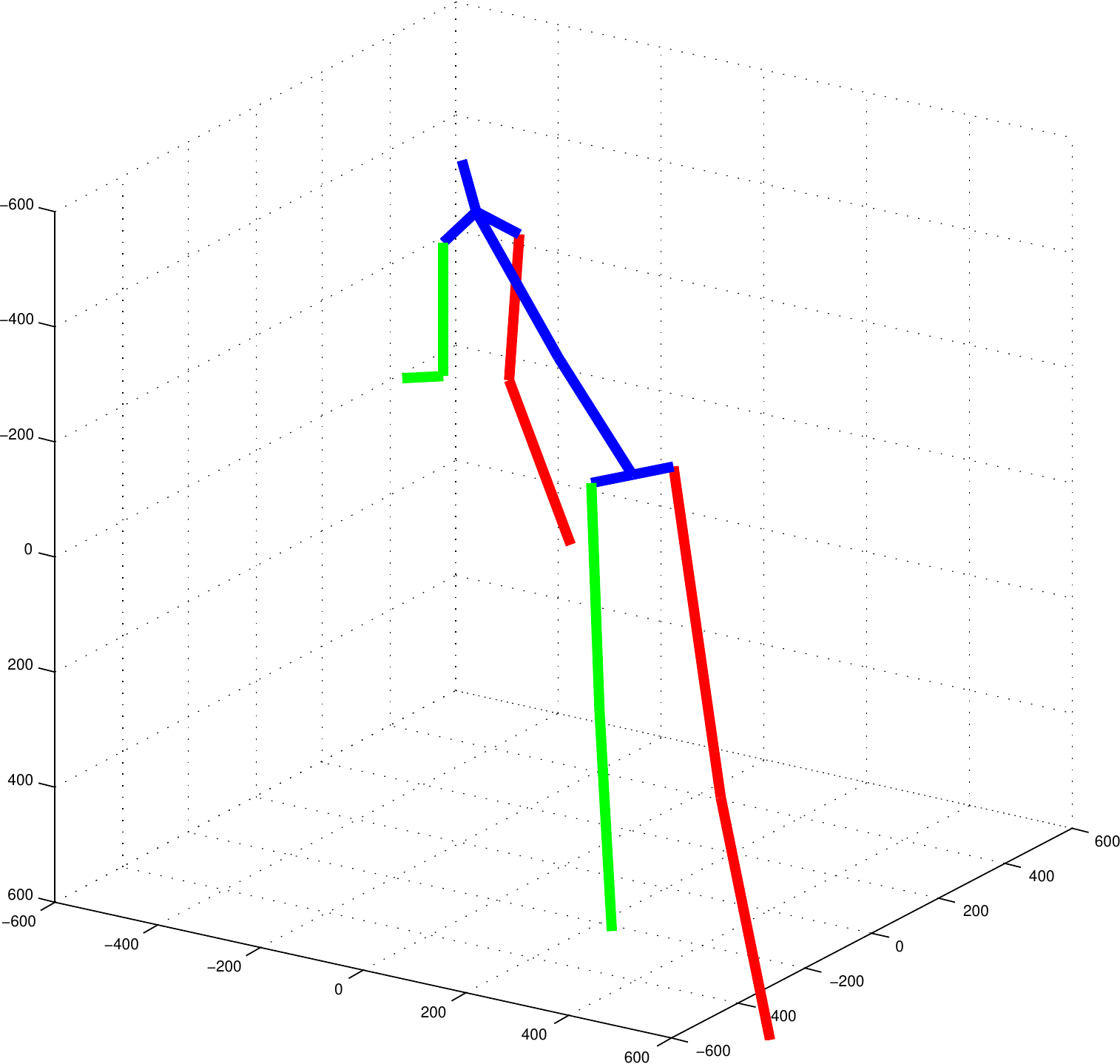}
      \\
      \hspace{-2.7mm}
      \includegraphics[height=0.10\textwidth]{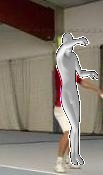}\hspace{0.54mm}
      \includegraphics[height=0.10\textwidth]{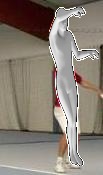}\hspace{0.54mm}
      \includegraphics[height=0.10\textwidth]{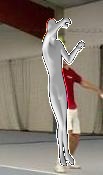}\hspace{0.54mm}
      \includegraphics[height=0.10\textwidth]{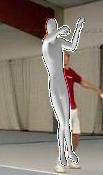}\hspace{0.54mm}
      \includegraphics[height=0.10\textwidth]{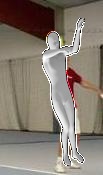}\hspace{0.54mm}
      \includegraphics[height=0.10\textwidth]{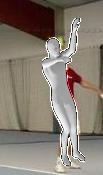}\hspace{0.54mm}
      \includegraphics[height=0.10\textwidth]{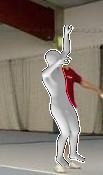}\hspace{0.54mm}
      \includegraphics[height=0.10\textwidth]{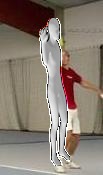}\hspace{0.54mm}
      \includegraphics[height=0.10\textwidth]{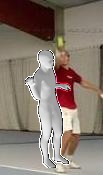}\hspace{0.54mm}
      \includegraphics[height=0.10\textwidth]{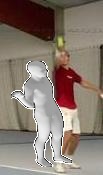}\hspace{0.54mm}
      \includegraphics[height=0.10\textwidth]{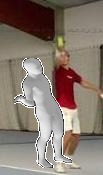}\hspace{0.54mm}
      \includegraphics[height=0.10\textwidth]{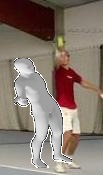}\hspace{0.54mm}
      \includegraphics[height=0.10\textwidth]{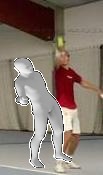}
      \\
      \hspace{-2.7mm}
      \includegraphics[height=0.10\textwidth]{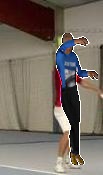}\hspace{0.54mm}
      \includegraphics[height=0.10\textwidth]{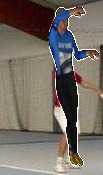}\hspace{0.54mm}
      \includegraphics[height=0.10\textwidth]{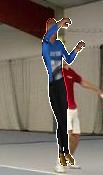}\hspace{0.54mm}
      \includegraphics[height=0.10\textwidth]{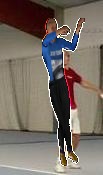}\hspace{0.54mm}
      \includegraphics[height=0.10\textwidth]{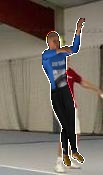}\hspace{0.54mm}
      \includegraphics[height=0.10\textwidth]{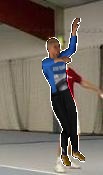}\hspace{0.54mm}
      \includegraphics[height=0.10\textwidth]{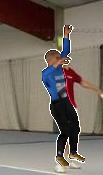}\hspace{0.54mm}
      \includegraphics[height=0.10\textwidth]{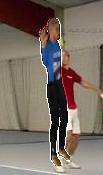}\hspace{0.54mm}
      \includegraphics[height=0.10\textwidth]{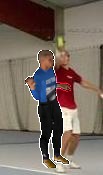}\hspace{0.54mm}
      \includegraphics[height=0.10\textwidth]{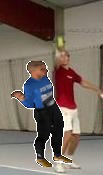}\hspace{0.54mm}
      \includegraphics[height=0.10\textwidth]{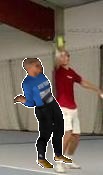}\hspace{0.54mm}
      \includegraphics[height=0.10\textwidth]{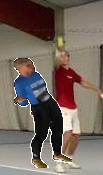}\hspace{0.54mm}
      \includegraphics[height=0.10\textwidth]{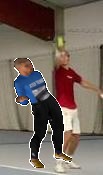}
    \end{tabular}
  \end{tabular}
  % \vspace{-2mm}
  \caption{\small Additional qualitative results of pose forecasting. The left
column shows the input images. For each input image, we show in the right
column the sequence of ground-truth frame and pose (row 1), our forecasted pose
sequence in 2D (row 2) and 3D (row 3), and the rendered human body without
texture (row 4) and with skin and cloth textures (row 5).}
  % \vspace{-2mm}
  \label{fig:additional3}
\end{figure*}

We show additional qualitative examples of the forecasted poses in
Fig.~\ref{fig:additional1}, \ref{fig:additional2}, and \ref{fig:additional3}.
Note that the rendered human model also improves the interpretability of the
output 3D poses over skeletons. The second example in
Fig.~\ref{fig:additional3} shows a failure case of 3D pose recovery. While the
forecasted motion of the tennis serve looks plausible in 2D (row 2), the
recovered 3D poses are unrealistic in their body configurations (row 4 and 5),
which may be difficult to perceive in the visualizations of 3D skeletons (row
3). All qualitative results can also be viewed as videos at
\href{http://www.umich.edu/~ywchao/image-play/}{\texttt{http://www.umich.edu/$\sim$ywchao/image-play/}}.

\end{document}